%% file: main.tex
\documentclass[dvipsnames]{thesis}
\usepackage[utf8]{inputenc}

\input{./math_commands.tex}

\usepackage{neuralnetwork}
\usetikzlibrary{positioning, calc, decorations.pathreplacing}
\usepackage{graphicx}
\usepackage{pgfplots}
\usepackage{amsmath}
\usepackage{amssymb}
\usepackage{epsfig}
\usepackage{xspace}
\usepackage{ dsfont }
\usepackage{float}
\usepackage{subcaption}
\usepackage{xcolor}
\usepackage{geometry}
\usepackage{caption}
\usepackage{booktabs}
\usepackage{multicol}
\usepackage{multirow}
\usepackage{enumitem}

\usepackage{lipsum}

\renewcommand{\1}{\mathds{1}}
\newcommand{\Acal}[0]{\mathcal{A}}

\newcommand{\Hcal}[0]{\mathcal{H}}
\newcommand{\Rcal}[0]{\mathcal{R}}
\newcommand{\Ecal}[0]{\mathcal{E}}

\newcommand{\Lcal}[0]{\mathcal{L}}
\newcommand{\Scal}[0]{\mathcal{S}}
\newcommand{\Qcal}[0]{\mathcal{Q}}
\newcommand{\Mcal}[0]{\mathcal{M}}
\newcommand{\Tcal}[0]{\mathcal{T}}

\makeatletter
\DeclareRobustCommand\onedot{\futurelet\@let@token\@onedot}
\def\@onedot{\ifx\@let@token.\else.\null\fi\xspace}
\makeatother

\newcommand{\ie}{\emph{i.e\onedot}}
\newcommand{\eg}{\emph{e.g\onedot}}
\newcommand{\wrt}{\emph{w.r.t\onedot}}
\newcommand{\cf}{\emph{c.f\onedot}}

\newcommand{\rev}[1]{#1}

\newenvironment{proof_idea}{%
	\renewcommand{\proofname}{Proof sketch}\proof}{\endproof}

\newcommand{\ddetr}{Def\onedot~DETR\xspace}

\newcommand{\faster}{Faster R-CNN\xspace}
\newcommand{\Maml}{MAML\xspace}
\newcommand{\Proto}{ProtoNet\xspace}
\newcommand{\MC}{MC\xspace}
\newcommand{\IMP}{IMP\xspace}

\crefformat{footnote}{#2\footnotemark[#1]#3}

\newcounter{savepage}

\usepackage{xurl}
\usepackage{breakcites}
\usepackage{cleveref}
\usepackage{hyperref}
\usepackage[toc,acronym]{glossaries}
\usepackage[automake]{glossaries-extra}
\usepackage{minitoc}

\title{Towards Few-Annotation Learning in Computer Vision: \\
Application to Image Classification and Object Detection tasks}
\author{Quentin Bouniot}
\date{March 2023}

\dominitoc

\makeglossaries

\newglossaryentry{scalar}
{
    name=$a$,
    description={A scalar (integer or real)}
}

\newglossaryentry{vector}
{
    name=$\rva$,
    description={A vector}
}

\newglossaryentry{matrix}
{
    name=$\rmA$,
    description={A matrix}
}

\newglossaryentry{set}
{
    name=$\Acal$,
    description={A set}
}

\newglossaryentry{space}
{
    name=$\sA$,
    description={A space}
}

\newglossaryentry{real}
{
    name=$\R$,
    description={The space of real numbers}
}

\newglossaryentry{element_vector}
{
    name=$\rva_i$,
    description={Element $i$ of vector $\rva$, with indexing starting at 1}
}

\newglossaryentry{element_matrix}
{
    name=$\rmA_{(i,j)}$,
    description={Element $i,j$ of matrix $\rmA$}
}

\newglossaryentry{set_int}
{
    name=$[[N]]$,
    description={The set of all integers from 1 to $N$: $[[N]] := \{ 1, \dots, N \}$}
}

\newglossaryentry{param}
{
    name=$\theta$,
    description={Parameters or \emph{weights} of the model}
}

\newglossaryentry{optim}
{
    name=$\theta^*$,
    description={Optimal, or ideal parameters}
}

\newglossaryentry{emp}
{
    name=$\hat{\theta}$,
    description={Empirical parameters}
}

\newglossaryentry{transpose}
{
    name={$\rmA^\top, \rva^\top$},
    description={Transpose of matrix $\rmA$ or vector $\rva$}
}

\newglossaryentry{norm_p}
{
    name=$\|\rvx \|_p$,
    description={$L_p$ norm of $\rvx$}
}

\newglossaryentry{norm_2}
{
    name=$\|\rvx \|$,
    description={$L_2$ norm of $\rvx$}
}

\newglossaryentry{norm_fro}
{
    name=$\|\rmX \|_F$,
    description={Frobenius norm of $\rmX$}
}

\newglossaryentry{prod_scal}
{
    name={$\langle\rvx,\rvy\rangle$},
    description={Scalar product between $\rvx$ and $\rvy$}
}

\newglossaryentry{proba}
{
    name=$P(\rvx)$,
    description={Probability distribution over $\rvx$}
}

\newglossaryentry{distrib}
{
    name=$\rvx \sim P$,
    description={Random variable $\rvx$ has distribution $P$}
}

\newglossaryentry{expectation}
{
    name=$\E_{\rvx \sim P}[f(\rvx)]$,
    description={Expectation of $f(\rvx)$ with $\rvx$ following distribution $P$}
}

\newglossaryentry{gauss}
{
    name={$\gN(\mu,\Sigma)$},
    description={Gaussian distribution with mean $\mu$ and covariance $\Sigma$}
}

\newglossaryentry{indic}
{
    name=$\1_{\text{condition}}$,
    description={function that is 1 if condition is true, 0 otherwise}
}

\newacronym{ml}{ML}{Machine Learning}
\newacronym{deep}{DL}{Deep Learning}
\newacronym{cv}{CV}{Computer Vision}

\newacronym{sup}{SL}{Supervised Learning}
\newacronym{semi}{SSL}{Semi-Supervised Learning}
\newacronym{self}{Self-SL}{Self-Supervised Learning}
\newacronym{unsup}{UL}{Unsupervised Learning}

\newacronym{many}{MSL}{Many-Shot Learning}
\newacronym{few}{FSL}{Few-Shot Learning}
\newacronym{fal}{FAL}{Few-Annotation Learning}

\newacronym{od}{OD}{Object Detection}
\newacronym{fsc}{FSC}{Few-Shot Classification}
\newacronym{fsod}{FSOD}{Few-Shot Object Detection}
\newacronym{ssod}{SSOD}{Semi-Supervised Object Detection}

\newacronym{mlp}{MLP}{Multi-Layer Perceptron}
\newacronym{cnn}{CNN}{Convolutional Neural Network}

\newacronym{mtr}{MTR}{Multi-Task Representation Learning}
\newacronym{ema}{EMA}{Exponential Moving Average}

\newacronym{coco}{COCO}{Microsoft Common Objects in Context dataset}
\newacronym{ilsvrc}{ILSVRC}{ImageNet Large-Scale Visual Recognition Challenge dataset}
\newacronym{pascal}{PASCAL VOC}{PASCAL Visual Object Classes challenge dataset}

\newacronym{pp}{p.p.}{Percentage point}

\newacronym{iou}{IoU}{Intersection over Union, measures the overlap of predicted bounding box to the ground truth}

\newacronym{ap}{AP}{Average Precision metric}
\newacronym{ap50}{$\text{AP}_{50}$}{AP calculated at IoU threshold of 0.5}
\newacronym{ap50-95}{$\text{AP}_{50-95}$}{AP calculated for IoU threshold from 0.5 to 0.95 with step size of 0.05}

\newacronym{map}{mAP}{mean Average Precision metric, mean over all the class APs}

\begin{document}
\pagenumbering{roman}


\input{cover_page.tex}

\renewcommand{\abstractname}{\vspace{-\baselineskip}}

\input{chapters/abstract.tex}

\tableofcontents

\newpage

\glsaddall
\renewcommand*{\arraystretch}{1.5}
\printglossary[title=Notations] \mtcaddchapter
\renewcommand*{\arraystretch}{1.5}
\printglossary[type=\acronymtype] \mtcaddchapter
\newpage

\setcounter{savepage}{\value{page}}
\pagenumbering{arabic}

\adjustmtc

\input{chapters/intro.tex}

\input{chapters/learning.tex}

\input{chapters/sota.tex}

\input{chapters/meta_mtr.tex}
\input{chapters/selfod.tex}

\input{chapters/ssod.tex}

\input{chapters/ccl.tex}

\cleardoublepage
\pagenumbering{Roman}
\appendix
\input{chapters/appendix.tex}

\bibliography{references}
\bibliographystyle{apalike}


\newpage
\listoffigures
\newpage
\listoftables

\end{document}

%% file: math_commands.tex
\usepackage{amsmath,amsfonts,bm}

\def\eqref#1{equation~\ref{#1}}

\def\1{\bm{1}}
\newcommand{\train}{\mathcal{D_{\mathrm{train}}}}

\newcommand{\test}{\mathcal{D_{\mathrm{test}}}}

\def\rva{{\mathbf{a}}}
\def\rvb{{\mathbf{b}}}
\def\rvc{{\mathbf{c}}}

\def\rvh{{\mathbf{h}}}

\def\rvq{{\mathbf{q}}}

\def\rvs{{\mathbf{s}}}

\def\rvw{{\mathbf{w}}}
\def\rvx{{\mathbf{x}}}
\def\rvy{{\mathbf{y}}}
\def\rvz{{\mathbf{z}}}

\def\rmA{{\mathbf{A}}}

\def\rmI{{\mathbf{I}}}

\def\rmM{{\mathbf{M}}}

\def\rmW{{\mathbf{W}}}
\def\rmX{{\mathbf{X}}}
\def\rmY{{\mathbf{Y}}}

\DeclareMathAlphabet{\mathsfit}{\encodingdefault}{\sfdefault}{m}{sl}
\SetMathAlphabet{\mathsfit}{bold}{\encodingdefault}{\sfdefault}{bx}{n}

\def\gC{{\mathcal{C}}}

\def\gL{{\mathcal{L}}}

\def\gN{{\mathcal{N}}}

\def\sA{{\mathbb{A}}}

\def\sR{{\mathbb{R}}}

\def\sV{{\mathbb{V}}}

\def\sX{{\mathbb{X}}}
\def\sY{{\mathbb{Y}}}

\newcommand{\etens}[1]{\mathsfit{#1}}

\def\etG{{\etens{G}}}

\newcommand{\E}{\mathbb{E}}

\newcommand{\R}{\mathbb{R}}

\newcommand{\reg}{\lambda}

\newcommand{\softmax}{\mathrm{softmax}}

\DeclareMathOperator*{\argmin}{arg\,min}

%% file: cover_page.tex
\pagestyle{plain}


\renewcommand{\maketitle}{
\thispagestyle{empty}
\newgeometry{top=0cm,bottom=2cm,left=1.8cm,right=1.8cm}

\begin{center}
\begin{tabular}{cc}
\includegraphics[height=4cm]{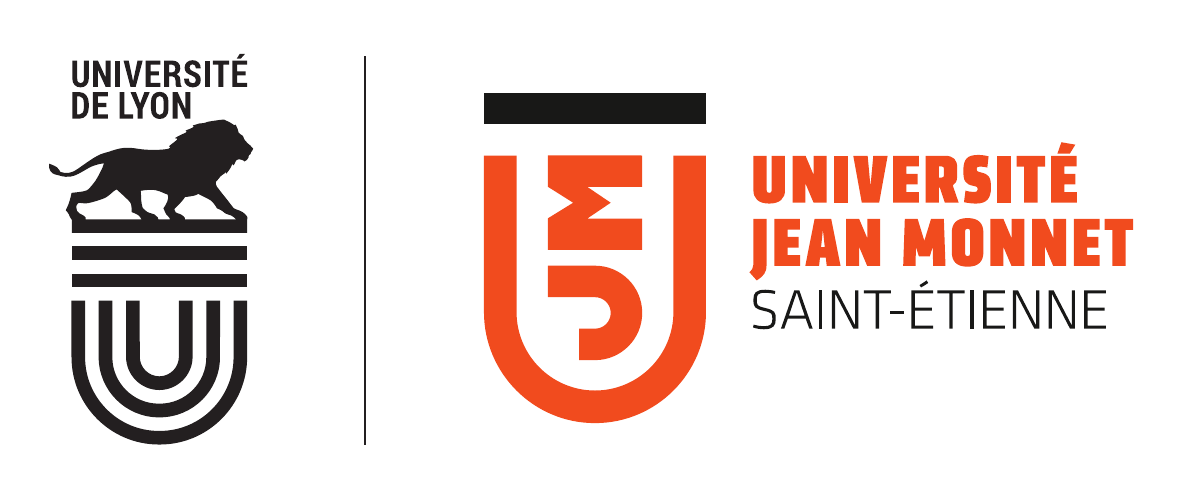}
\end{tabular}
\end{center}

\centerline{\noindent{N° d'ordre NNT : 2023STET009}}

 \begin{center}  
  \LARGE{\bf THESE de DOCTORAT DE L'UNIVERSITE \\ JEAN MONNET SAINT-ETIENNE}\\
  {membre de}\\
  \Large{ {l'}{\bf UNIVERSITE DE LYON}\\[1cm]
 
  \Large\textbf{Ecole Doctorale} N° 488\\
  \Large\textbf{Sciences Ingénierie Santé}\\[1cm]

  \Large\textbf{Spécialité de doctorat :}\\
  \Large\textbf{Discipline :} Informatique\\[1cm]

  {Soutenue publiquement le 29/03/2023, par :}\\
  \LARGE{\bf Quentin Bouniot}\\[1cm]}
    \noindent\rule{\textwidth}{1pt}
    
    \hspace{0.2cm}
    \Large{\textbf{Towards Few-Annotation Learning in Computer Vision: Application to Image Classification and Object Detection tasks}}

    \noindent\rule{\textwidth}{1pt}
    
\end{center}

{\noindent\normalsize Devant le jury composé de :}
\begin{table}[h]
    \centering
    \begin{tabular}{llll}
        Céline Hudelot & Professeure & CentraleSupélec & Rapporteuse \\
        Nicolas Thome & Professeur & Sorbonne Université & Rapporteur \\
        Diane Larlus & Chercheuse & Naver Labs Europe & Examinatrice \\
        Devis Tuia & Professeur Associé & EPFL & Examinateur \\
        David Filliat & Professeur & ENSTA & Examinateur \\
        Ievgen Redko & Chercheur & Huawei & Invité \\
        Romaric Audigier & Chercheur & CEA-List & Encadrant \\
        Angélique Loesch & Chercheuse & CEA-List & Encadrante \\
        Amaury Habrard & Professeur & Université Jean Monnet & Directeur de thèse \\
    \end{tabular}
\end{table}

\restoregeometry
}

\maketitle

\newpage

%% file: chapters/abstract.tex
\newpage

\section*{\centering Abstract}

\begin{abstract}
    In this thesis, we develop theoretical, algorithmic and experimental contributions for Machine Learning with limited labels, and more specifically for the tasks of Image Classification and Object Detection in Computer Vision. In a first contribution, we are interested in bridging the gap between theory and practice for popular Meta-Learning algorithms used in Few-Shot Classification. We make connections to Multi-Task Representation Learning, which benefits from solid theoretical foundations, to verify the best conditions for a more efficient meta-learning. Then, to leverage unlabeled data when training object detectors based on the Transformer architecture, we propose both an unsupervised pretraining and a semi-supervised learning method in two other separate contributions. For pretraining, we improve Contrastive Learning for object detectors by introducing the localization information. Finally, our semi-supervised method is the first tailored to transformer-based detectors.
\end{abstract}

\vfill

\section*{\centering Résumé}

\begin{abstract}
    Nous développons dans cette thèse des contributions théoriques, algorithmiques et expérimentales pour l’apprentissage automatique avec peu d'annota\-tions, et plus spécifiquement pour les tâches de classification d’images et de détection d’objets en vision par ordinateur. Dans une première contribution, nous nous intéressons à combler le fossé entre la théorie et la pratique concernant les algorithmes de méta-apprentissage les plus répandus, utilisés pour de la classification avec peu d’exemples. Nous établissons des liens avec l’apprentissage de représentation multi-tâches, qui bénéficie de bases théoriques solides, afin de vérifier les meilleures conditions amenant à un méta-apprentissage plus efficace. Ensuite, afin d’exploiter les données non étiquetées pour entraîner des détecteurs d’objets basés sur l’architecture Transformer, nous proposons à la fois un pré-entraînement non supervisé et une méthode d’apprentissage semi-supervisée dans deux autres contributions distinctes. Pour le pré-entraînement, nous améliorons l’apprentissage contrastif des détecteurs d’objets en introduisant les informations de localisation. Enfin, notre méthode semi-supervisée est la première adaptée aux détecteurs d’objets utilisant des modèles type Transformers.
\end{abstract}

\addstarredchapter{Abstract}

\newpage

\section*{Remerciements}

Je voudrais avant tout remercier le service du SIALV et en particulier le labo LVA pour m'avoir accueilli pendant ces dernières années. J'ai eu la chance de travailler avec des personnes remarquables et bienveillantes qui ont été d'une aide précieuse.
Je tiens également à remercier tous les autres doctorants avec qui j'ai eu la chance de partager des moments de discussions enrichissants, notamment mes co-bureaux Julien et Rémi. Vous avez été une source de motivation quotidienne.

Je voudrais exprimer ma gratitude envers mes encadrants du CEA, Romaric et Angélique, qui m'ont guidé et encadré depuis 4 ans ! On a commencé à travailler ensemble avant la thèse et je vous remercie d'avoir cru en moi. Je vais maintenant voler de mes propres ailes, mais je garderai toujours vos précieux conseils en tête. Je souhaite d'ailleurs beaucoup de courage à Angélique pour son nouveau poste de cheffe et de maman.
Je voudrais également remercier Ievgen, qui a certes quitté l'encadrement officiellement à la moitié de la thèse, mais qui est resté de bon conseil tout du long ! Même avec la distance, en France ou à l'autre bout de l'Europe, tu m'as accompagné et es resté disponible.
Un grand merci à mon directeur de thèse Amaury, qui malgré les difficultés et la distance a toujours été présent et disponible pour moi. Je n'ai pas pu venir à Saint-Étienne aussi souvent que je l'aurais voulu à cause du contexte sanitaire, mais je te suis reconnaissant de m'avoir fait confiance toi aussi.

Je voudrais aussi exprimer ma gratitude envers les membres du jury, je suis honoré d'avoir pu présenter mon travail devant vous. J'ai une pensée pour Diane et Devis qui ont pris de leur temps pour évaluer mon travail même à distance. David qui a pris la responsabilité de la présidence du jury.
Enfin, je souhaite exprimer toute ma reconnaissance envers les rapporteurs Céline et Nicolas, qui ont consacré du temps à la lecture de mon manuscrit de thèse. J'ai été touché par leurs retours pertinents et constructifs.

Je ne peux pas terminer sans remercier toute ma famille et mes amis, de Tahiti et de Métropole, qui ont été d'un soutien inconditionnel tout du long. Merci à Papa, Maman, Alexandre, Raphaël, Tiapo, Sylvie et Christophe, Mamie, Gilles et Nathalie, Alain et Cathy, sans oublier Karine et Frédéric. Je sais que vous ne comprenez pas tout ce que je fais, mais je vous remercie de votre soutien. Merci à mes amis de Tahiti, Matthias, Jonathan et Kim.
Enfin, évidemment, je voudrais exprimer ma gratitude envers ma compagne Anne-Maëlle, qui m'a soutenu et conseillé tout les jours. Ça a été essentiel et je ne serais pas arrivé au bout sans toi.

\newpage

%% file: chapters/intro.tex
\chapter{Introduction}\label{chap:intro}

\minitoc

\begin{abstract}
    \textit{
        \rev{This chapter provides an introduction to the main problems addressed in this thesis, as well as the presentation of the structure and contributions of this manuscript.}
    }
\end{abstract}

\section{Context and Background}

\subsection{Context}

Learning something new in real life does not necessarily mean going through a lot of examples in order to capture the essence of it. Even though it is said that it takes 10,000 hours to \emph{master} a new skill \cite{10000hours}, it is also true that it only takes 20 hours to \emph{learn} it \cite{20hours}. Humans are able to build upon prior experience and have the ability to adapt, allowing to combine previous observations with only little evidence for fast learning. This is particularly the case for recognition tasks, for which we are often capable of differentiating between two distinct objects after having seen only a few examples of them. 

\begin{figure}[ht]
    \centering
    \includegraphics[width=0.6\linewidth]{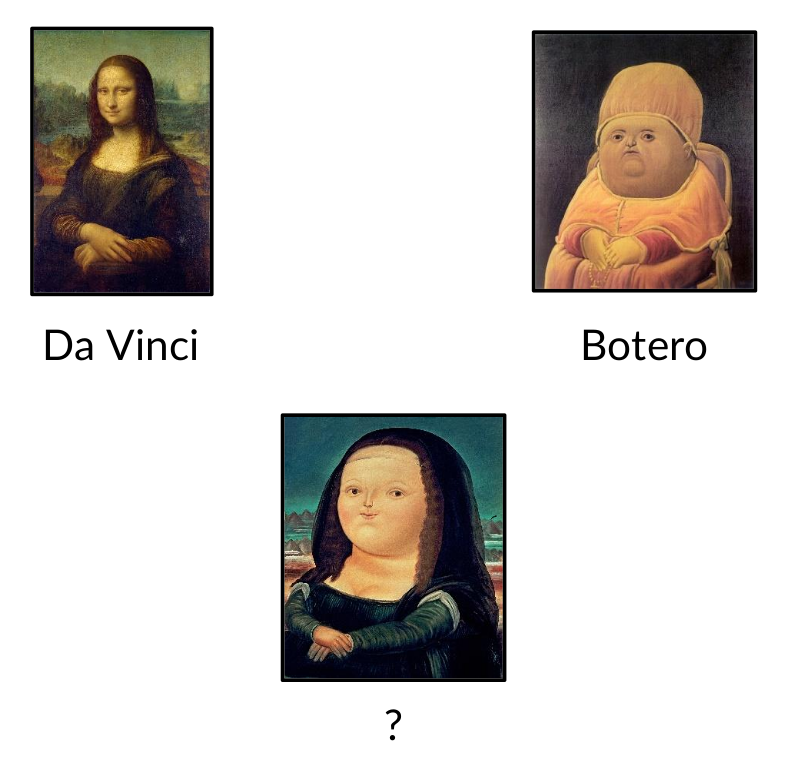}
    \caption[]{Who is the painter ? Illustration of the human capacity to \emph{learn from limited examples}. The top left painting is the \emph{Mona Lisa} by \emph{Leonardo da Vinci} and the top right one is the \emph{Pope Leo X after Raphael} by \emph{Fernando Botero}. \rev{The bottom painting is \emph{Mona Lisa, Age Twelve}, guessing the painter is left as an exercise to the reader.}}
    \label{fig:intro_botero}
\end{figure}

We give an illustrative example from paintings samples presented in \cref{fig:intro_botero}. It is quite obvious that one would easily guess and recognize the painter of the last painting below after having seen just one example of each painter's styles above. This idea has found its application in \emph{machine learning} in a more general \emph{few-shot learning} paradigm that wants to mimic the human capability to quickly learn how to solve a new problem. The above example is a direct illustration of \emph{one-shot learning}.

\subsection{Background}

Learning models, \ie{} estimating parameters from data, has long proven to be an effective approach in the literature for solving different kinds of applications, ranging from computer vision to natural language processing. 
This is what is commonly referred as \emph{Machine Learning}.

Ever since AlexNet \cite{krizhevsky2009learning}, the rise of Deep Learning (DL) \cite{lecun2015deep,Goodfellow-et-al-2016} models for \emph{Representation Learning} has led to near-human performance on many vision tasks. Current best performing architectures are based on Convolutional Neural Networks (CNN) \cite{lecun1989handwritten}, such as the popular ResNet architecture \cite{he2016deep}, and more recently on Transformers \cite{vaswani2017attention, dosovitskiy2021an, carion2020end}.

Training DL models consists in optimizing an \emph{objective}, or \emph{loss} function \rev{\wrt{} to some experience, or previously acquired \emph{information}.}
Learning algorithms considered throughout this thesis are allowed to experience an entire \emph{dataset}. A dataset is a collection of many examples, also called \emph{data samples}. These examples are usually measurements, and more specifically in the case of Computer Vision, \rev{they most commonly take the form of} \emph{images}, annotated or not.

\subsection{Learning frameworks}

Training paradigms can be broadly separated into either \emph{Supervised Learning (SL)}, \emph{Unsupervised Learning (UL)}, or \emph{Semi-supervised Learning (SSL)} depending on the amount of \rev{annotated} information available during the learning process.

\paragraph{Supervised Learning (SL)} 
The most common form of machine learning, deep or not, is \emph{supervised learning} \cite{shalev2014understanding,mohri2018foundations,lecun2015deep}. During training, SL algorithms are shown examples associated with a \emph{label}, or \emph{target}. The term \emph{supervised learning} comes from the view that an instructor, or teacher, is providing \emph{supervision} through the \emph{labels} \cite{Goodfellow-et-al-2016}. Indeed, the training optimizes supervised loss functions that take into account the additional information contained in the labels. 

\paragraph{Unsupervised learning (UL)}
Unlike SL, UL algorithms \cite{shalev2014understanding,mohri2018foundations,lecun2015deep} experience a dataset \emph{without labels}, \ie{} without additional information. The objective is to learn useful properties about the \emph{structure} of the input dataset, by, \eg{} learning the entire probability distribution that generated the dataset, either explicitly or implicitly \cite{Goodfellow-et-al-2016}. This problem is generally seen as more difficult than SL, but also potentially more impactful since it requires less information on the data provided by the user \cite{lecun2015deep}. The recent \emph{Self-Supervised learning (Self-SL)} paradigm is a particular case of UL, in which the data provides its own supervision to learn a practical representation.

\paragraph{Semi-Supervised Learning (SSL)}
In addition to data, SSL algorithms \cite{mohri2018foundations} are supervised with labels, but not for all examples. This paradigm can be seen as halfway between supervised and unsupervised learning \cite{chapelle2009semi}. The dataset can be usually divided into \emph{labeled data}, for which labels are provided, and \emph{unlabeled data}, the labels of which are not known. 


When having access to \emph{large datasets}, SSL is often a good compromise between the two other frameworks in practice. Resulting models can achieve strong performance while requiring limited supervision. 
Even though SL is the most common form of DL, it is also the most restrictive one for practical use, as it requires to have access to \emph{a large number of labels}. 

\section{Challenges and Objectives}

\subsection{The Problem of Labeling Data}

\begin{figure}
    \centering
    \includegraphics[width=\linewidth]{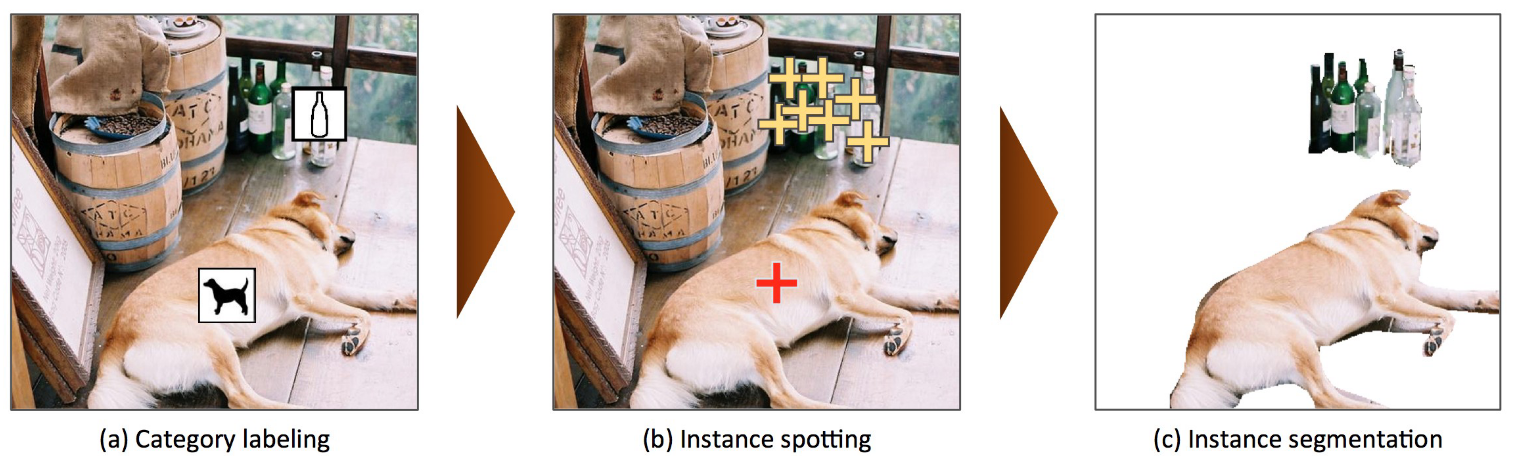}
    \caption{Illustration of the annotation pipeline from the COCO dataset \cite{lin2014microsoft}. The pipeline is split into 3 primary tasks: (a) labeling the categories (class of object) present in the image, (b) locating and marking all instances of the labeled categories, and (c) segmenting each object instance.}
    \label{fig:coco_annotation}
\end{figure}

Supervised Learning requires a lot of labeled data to be effective. The ImageNet dataset \cite{russakovskyImageNetLargeScale2015}, one of the most popular datasets for \emph{image recognition}, contains from 1.2M (\emph{ILSVRC 2012} or \emph{ImageNet-1k}) to 14M labeled images (\emph{ImageNet-21k}). However, the process of annotating data can be hard, time-consuming and thus, expensive. For a highly accurate dataset, humans must verify and manually annotate each collected example. Then, to reduce human errors, the process is repeated independently by multiple persons and the final label of each sample is taken as the convincing majority of labels given. Each year \rev{from 2010 to 2017}, ImageNet employed 20,000 to 30,000 people to annotate the dataset through the Amazon Mechanical Turk (AMT)\footnote{\url{https://www.mturk.com/}} service, paying a few cents for each image labeled\footnote{\url{https://www.nytimes.com/2012/11/20/science/for-web-images-creating-new-technology-to-seek-and-find.html}}. 

Furthermore, more \emph{dense} computer vision tasks, \ie{} problems that require a deeper understanding of images \rev{and thus more complex annotations per image}, take more time to annotate. \cref{fig:coco_annotation} is an illustration of the full annotation process for the \emph{MS COCO (COCO)} dataset \cite{lin2014microsoft}. COCO is one of the most popular large-scale dataset used for the \emph{Object Detection (OD)} task, which consists in locating and classifying every instance of object in an image. The dataset is a collection of over 200k labeled images for a total of about 2.5M object instances.
As we can see in \cref{fig:coco_annotation}, the annotation pipeline is separated in three tasks, (a) \emph{Category labeling}, (b) \emph{Instance spotting} and (c) \emph{Instance segmentation}. Here again, multiple AMT workers were asked to manually annotate and verify images, for a total of over 85,000 worker hours \cite{lin2014microsoft}, \ie{} over \emph{9 worker years}. 

\rev{The success stories of the above two examples of dataset labeling heavily rely on a \emph{collaborative annotation process} to reduce the overall cost. However, not all datasets can be annotated in the same way.}
While anyone can label images of dogs and cats, data from more complex domains might require precious time from qualified workers. For instance, labeling medical or borehole images requires insights from health specialists and geophysicists, increasing even more the cost of annotation. \rev{In addition, in some cases, sensitive data or private datasets cannot be shared with anyone.}
\rev{For all these reasons}, we often do not have access in practice to large-scale labeled datasets.



\subsection{Training data conditions in practice}

\begin{figure}
    \centering
    \includegraphics[width=0.8\linewidth]{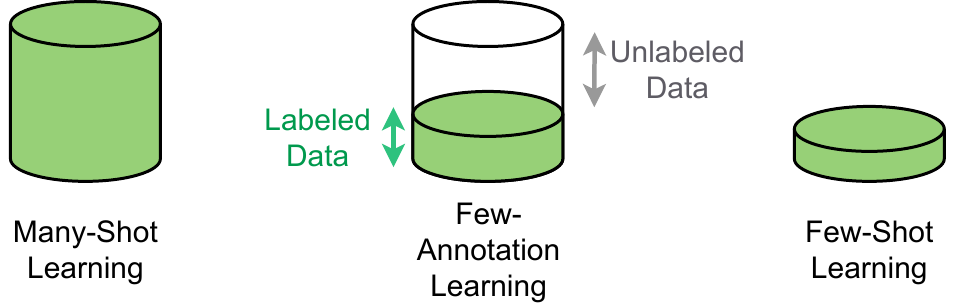}
    \caption{Illustration of the different possible \emph{training data conditions} encountered in practice, depending on the quantity of target data \emph{(the size of the cylinder)} and labels \emph{(in green)} available.}
    \label{fig:learning_pbs}
\end{figure}

We can separate learning problems in practice in three categories, depending on the amount of \emph{target data}, \ie{} data linked to the task that we want to solve, but also depending on the \emph{labels} available on this data. We give an illustration of these different categories in \cref{fig:learning_pbs}. 

\paragraph{Many-Shot Learning (MSL)} The \rev{most common} category, \emph{MSL}, corresponds to learning problems for which we have access to \emph{a lot of annotated data}. They usually arise when the data is easy to acquire \emph{and} to annotate. We will not delve into the details of this setting as it represents the easiest learning conditions, given that we can directly apply \emph{supervised algorithms}. 

\begin{figure}
    \begin{subfigure}[b]{\linewidth}
        \centering
        \includegraphics[width=\linewidth]{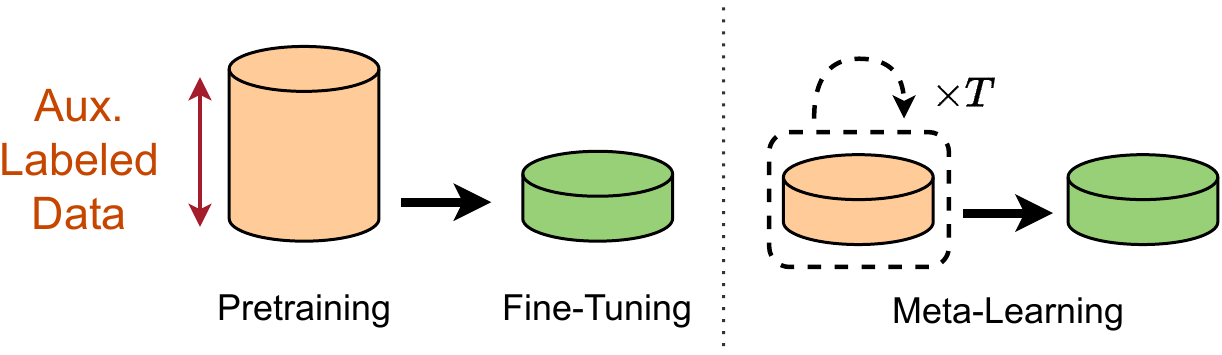}
        \caption{Few-Shot Learning (FSL)}
        \label{fig:fsl_pbs}
    \end{subfigure}
    \begin{subfigure}[b]{\linewidth}
        \centering
        \includegraphics[width=0.8\linewidth]{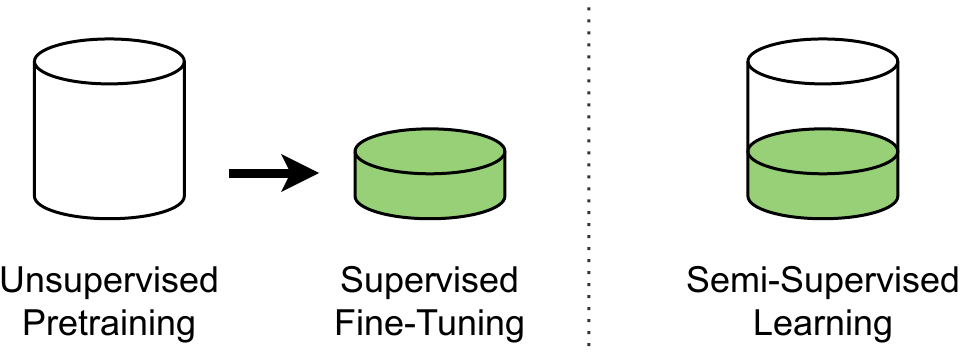}
        \caption{Few Annotation Learning (FAL)}
        \label{fig:fal_pbs}
    \end{subfigure}
    \caption{Illustration of \rev{commonly used} approaches for \textbf{(a)} \emph{FSL} and \textbf{(b)} \emph{FAL} problems. \emph{Auxiliary (Aux.) labeled data} is represented in \emph{orange}.}
\end{figure}



\paragraph{Few-Shot Learning (FSL)} The \rev{hardest} category of problems, \emph{FSL}, is the most scarce in terms of target data \emph{and} labels available. These problems have access to only few examples, or \emph{shots}, to learn from, usually because of a difficult data collection. In this setting, \emph{auxiliary data} \rev{can be used} to compensate for the lack of \emph{available target data}. We illustrate \rev{approaches commonly described in the literature} in \cref{fig:fsl_pbs}. \rev{Models} are either pretrained on the auxiliary data, \rev{with supervision or not}, and then refined on the target data \emph{(left)}, or trained with \emph{Meta-Learning} \rev{(that will be presented more in depth in \cref{chap:sota})} on auxiliary data by constructing a large number $T$ of small tasks, to be able to adapt faster to the few data of the target task \emph{(right)}.

\paragraph{Few-Annotation Learning (FAL)} An \rev{intermediate} category, \emph{FAL}, includes problems for which we have a lot of target data but \emph{few annotated ones}. While data gathering for these problems may be simple, labeling may be too costly. We give an illustration in \cref{fig:fal_pbs} of the possible approaches in this setting to leverage unlabeled data on the target task to better learn on the few labeled data available. Methods in this setting either rely on an \emph{unsupervised pretraining} phase on the unlabeled data followed by a \emph{supervised fine-tuning} phase on the labeled data \emph{(left)}, or \emph{semi-supervised learning} to leverage unlabeled data along with labeled data during training \emph{(right)}.


%



\begin{figure}
    \centering
    \includegraphics[width=0.9\linewidth]{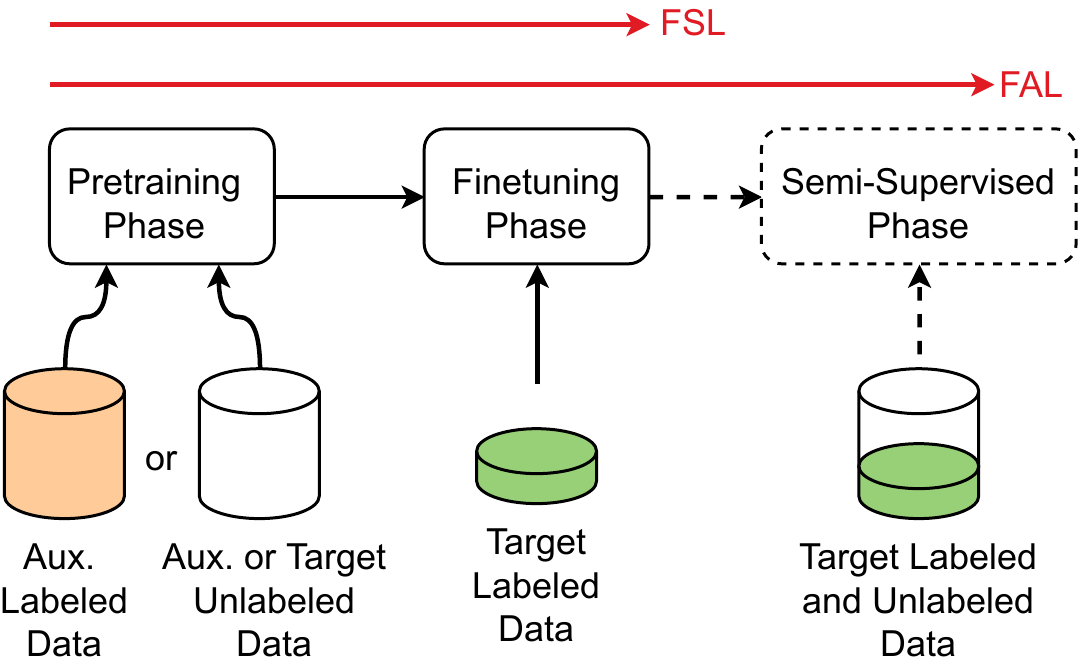}
    \caption{\rev{A tentative general pipeline encompassing approaches of FSL and FAL, coping with limited target labels}. The semi-supervised phase \emph{(dotted lines)} is possible if unlabeled target data are also available (FAL).}
    \label{fig:learning_pipeline}
\end{figure}

In this thesis, we are specifically interested in problems with limited labels available \rev{and explore both settings of FSL and FAL}.
In this case, \rev{from approaches in the literature, we can draw a general learning pipeline that distinguishes three different phases, illustrated in \cref{fig:learning_pipeline}, which are not always present}. First, a \emph{pretraining phase} either unsupervised or supervised on auxiliary labeled data, learns a general representation. Then, the \emph{finetuning phase} adapts the representation learned to the target labeled data. Additionally, a \emph{semi-supervised phase} can complete the training with both labeled and unlabeled target data. 

While there is a large literature in MSL, research in FSL \rev{and, even more, in FAL, have been much more sparsely addressed}. Yet, they represent the most common settings in industry because of the cost of data annotation, and sometimes even its collection. This discrepancy between research and industry interests motivated the work presented in this thesis.

\subsection{Issues addressed}

\rev{Throughout this thesis, we are interested in applications to \emph{Computer Vision problems}, with a focus on \emph{Image Classification} and \emph{Object Detection} tasks. As detailed above, while the easiest learning conditions and the most effective methods require having access to a large amount of labeled images, learning with scarcely annotated data remains a challenging problem even though it corresponds to the most common setting in practice. We address the issues of \emph{learning with limited labels} from different perspectives:}

\begin{itemize}
    \item \rev{On the one hand, our objective is to better understand the inner workings of learning algorithms in FSL settings from a \emph{theoretical point of view}. Conditions and guarantees for a given algorithm to learn well are given by theoretical analyses and are expressed by \emph{learning bounds}. These bounds often take the following form \cite{vapnik1999nature}:}

\begin{equation}
    \text{True Risk} - \text{Empirical Risk} \leq \frac{\text{Model Complexity}}{F(\text{Number of training data})},
\end{equation}

\noindent \rev{where the \emph{True} and \emph{Empirical Risk} correspond to the expected value of the loss function, respectively computed on the \emph{full data distribution} and the available \emph{dataset}, the \emph{Model Complexity} is a measure of the \emph{representation capacity} of the model used, and $F$ is a function of the \emph{number of training data} used. However, in FSL where the number of training data available is small, this last specific term can hinder the training process. Through connections to frameworks with solid theoretical foundations, we aim to find the best training conditions for this setting and investigate this problem in \cref{chap:contrib_eccv}, with the settings considered illustrated in \cref{fig:position_eccv}.}

\item \rev{On the other hand, we want to improve \emph{Object Detection} methods from an \emph{algorithmic point of view} in FAL settings. The specificities of Object Detection have been barely addressed in the context of FAL, even though it represents a task with many practical applications. We are specifically interested in \emph{unsupervised pretraining} and \emph{semi-supervised learning} methods, investigated respectively in \cref{chap:selfod} and \cref{chap:ssod}, with their respective settings illustrated in \cref{fig:position_iclr} and \cref{fig:position_wacv}.}
\end{itemize}


\rev{In the following section, we discuss the overall structure of the manuscript and our contributions addressing the issues introduced above.}

\section{Outline and Contributions}

We present in this section an overview of this thesis and the diverse associated publications. 

\subsection{Structure of the manuscript}

This manuscript is organized as follows. The present \cref{chap:intro} is meant as a motivation to the work presented in the following chapters. We introduced the different learning paradigms and the main settings encountered in practice. We illustrated the \emph{time-cost problem of manual data labeling} with examples of the annotation process from popular datasets. This problem is at the heart of all the questions that will interest us throughout this thesis, since it dictates the amount of target annotated data available.

In \cref{chap:learning}, we delve more in depth into the different concepts in Machine Learning that will be important to understand this work. We present the learning process and theoretical notions with simple models, then deeper models and their specificities. Finally, since we will focus on using images as our data throughout this thesis, we introduce the popular tasks of \emph{Image Classification} and \emph{Object Detection}, and their respective recent literature. 

In \cref{chap:sota}, we review related work in learning with limited labels. These works span the topics of \emph{Data Augmentation}, \emph{Meta-Learning}, \emph{Transfer Learning} and \emph{Semi-supervised Learning}. We introduce recent work in each topic, and conclude with the research directions outlined for this thesis.

\begin{figure}
    \begin{subfigure}[b]{0.95\linewidth}
        \centering
        \includegraphics[width=0.8\linewidth]{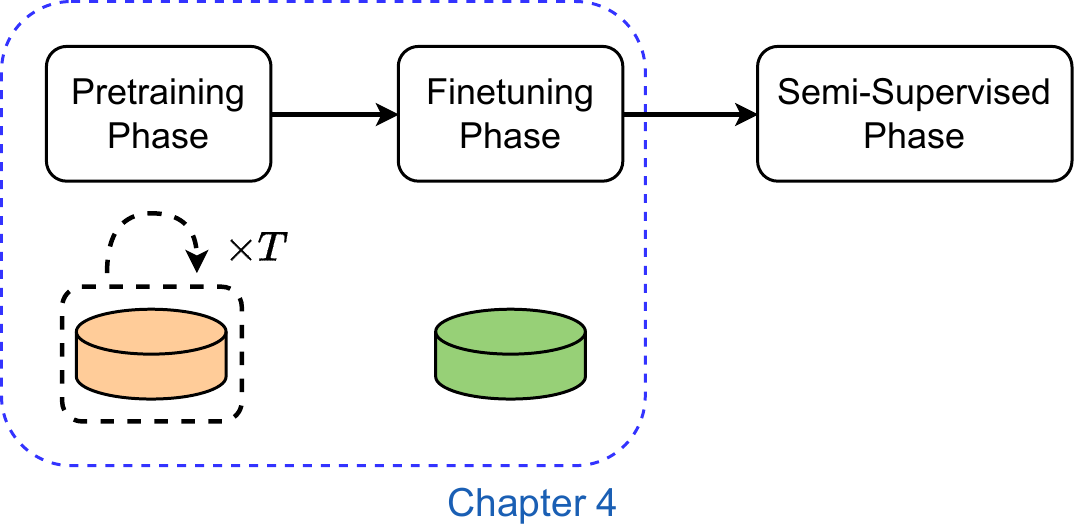}
        \caption{}
        \label{fig:position_eccv}
    \end{subfigure}
    \begin{subfigure}[b]{0.95\linewidth}
        \centering
        \includegraphics[width=0.8\linewidth]{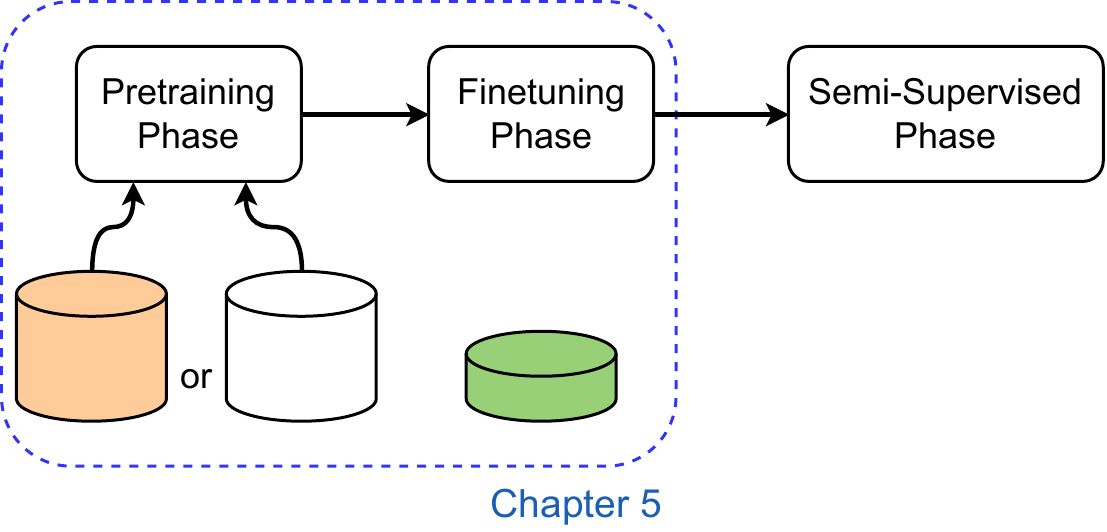}
        \caption{}
        \label{fig:position_iclr}
    \end{subfigure}
    \begin{subfigure}[b]{0.95\linewidth}
        \centering
        \includegraphics[width=0.8\linewidth]{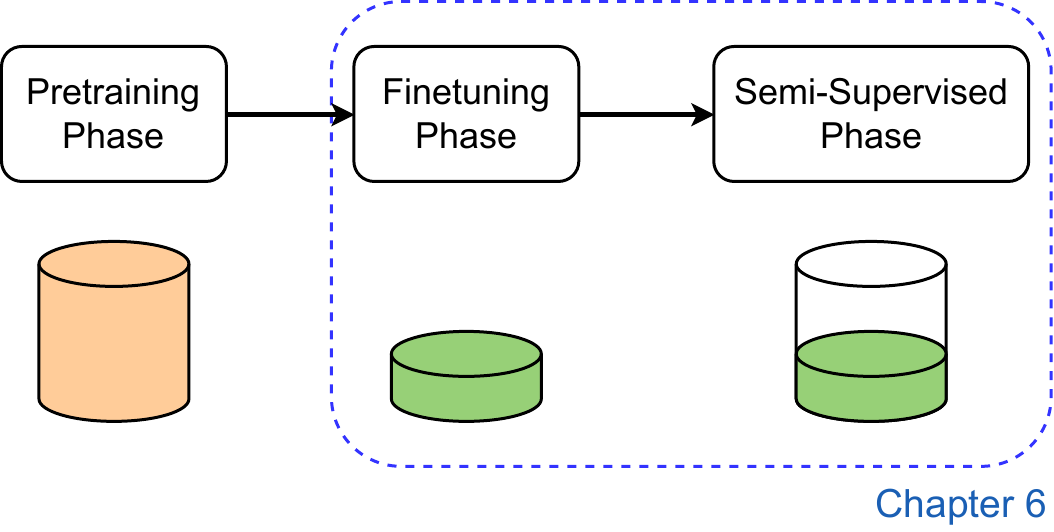}
        \caption{}
        \label{fig:position_wacv}
    \end{subfigure}
    \caption{Illustration of the position of our contributions presented in \textbf{(a)} \cref{chap:contrib_eccv}, \textbf{(b)} \cref{chap:selfod} and \textbf{(c)} \cref{chap:ssod}, in the general training pipeline from \cref{fig:learning_pipeline}.}
\end{figure}

Then, the following chapters present our contributions, each positioned \wrt{} the general training pipeline from \cref{fig:learning_pipeline}, in \cref{fig:position_eccv,fig:position_iclr,fig:position_wacv}.

\cref{chap:contrib_eccv} is intended to bridge the gap between Multi-Task Representation Learning theory and practices in Meta-Learning for FSL. We investigate the behavior and theory behind popular meta-learning algorithms through the lens of theoretical assumptions from recent learning bounds for Multi-Task Representation Learning. This leads us to better understand these methods and improve their performance by forcing them to follow the conditions \rev{for a more efficient learning in FSL}.


\cref{chap:selfod} focuses on pretraining strategies \rev{for learning Object Detector with limited labels}. In this chapter, we propose a novel pretraining method for recent transformer-based object detectors, based on a Proposal-Contrastive Learning paradigm. Throughout the chapter we advocate for consistency in the level of information learned during pretraining, which forms the foundations of our method.  


In \cref{chap:ssod}, we propose a novel semi-supervised method tailored to transformer-based object detectors. We are interested in improving performance with scarcely labeled data but with access to target unlabeled data, after fine-tuning on the target labeled data. 
\rev{We found that direct transpositions of popular practices with CNN can be counterproductive when using transformer-based detectors, and investigate ways to learn these latter models more efficiently in SSL for object detection with limited labels.}

Finally, \cref{chap:ccl} concludes this thesis with perspectives on possible future research directions and a discussion on broader impacts of this line of work.


\subsection{Scientific contributions}

The contributions presented in this thesis led to the following \rev{peer-reviewed and} published works:

\begin{itemize}[itemsep=0.3cm, label=\textbullet]
    \item International Conferences:
    \begin{enumerate}[label*={\arabic*.}, leftmargin=1cm]
        \item \textbf{Quentin Bouniot}, Ievgen Redko, Romaric Audigier, Angélique Loesch, Amaury Habrard. "Improving Few-Shot Learning through Multi-task Representation Learning Theory". In \textit{European Conference on Computer Vision (ECCV)}, Oct. 2022. \cite{bouniot2022improving}
        \item \textbf{Quentin Bouniot}, Angélique Loesch, Romaric Audigier, Amaury Habrard. "Towards Few-Annotation Learning for Object Detection:
        Are Transformer-based Models More Efficient ?". In \textit{Winter Conference on Applications of Computer Vision (WACV)}, Jan. 2023. \cite{Bouniot_2023_WACV}
        \item \textbf{Quentin Bouniot}, Romaric Audigier, Angélique Loesch, Amaury Habrard. "Proposal-Contrastive Pretraining for Object
        Detection from Fewer Data". In \textit{International Conference on Learning Representatons (ICLR)}, May 2023. \cite{bouniot2023proposalcontrastive}
    \end{enumerate}
    \item International Workshop:
    \begin{enumerate}[leftmargin=1cm]
        \item[4.] \textbf{Quentin Bouniot}, Ievgen Redko, Romaric Audigier, Angélique Loesch, Amaury Habrard. "Putting Theory to Work : From Learning Bounds to Meta-Learning Algorithms". In \textit{Neural Information Processing Conference (NeurIPS) Workshop on Meta-Learning (MetaLearn)}, Dec. 2020. \cite{bouniot2020putting}
    \end{enumerate}
    \item National Conference:
    \begin{enumerate}[leftmargin=1cm]
        \item[5.] \textbf{Quentin Bouniot}, Ievgen Redko, Romaric Audigier, Angélique Loesch, Amaury Habrard. "Vers une meilleure compréhension des méthodes de méta-apprentissage à travers la théorie de l’apprentissage de représentations multi-tâches". In \emph{Conférence sur l’Apprentissage Automatique (CAp)}, June 2021. \cite{bouniot2021cap}
    \end{enumerate}
    \item International Communication:
    \begin{enumerate}[leftmargin=1cm]
        \item[6.] \textbf{Quentin Bouniot} \& Ievgen Redko, "Understanding Few-Shot Multi-Task Representation Learning Theory". In \emph{International Conference on Learning Representations (ICLR) Blog Track}, May 2022. \cite{quentin2022understandingfewshotmultitask}
    \end{enumerate}
\end{itemize}

\cref{chap:contrib_eccv} is based on \cite{bouniot2022improving,bouniot2020putting,bouniot2021cap,quentin2022understandingfewshotmultitask}, \cref{chap:selfod} on \cite{bouniot2023proposalcontrastive} and \cref{chap:ssod} on \cite{Bouniot_2023_WACV}.

%% file: chapters/learning.tex
\chapter{The Basics of Learning from images}\label{chap:learning}

\minitoc


\begin{abstract}
    \textit{This chapter presents the preliminary knowledge to understand this thesis. The main objective is to describe basic notions and challenges in \emph{Machine Learning} and \emph{Deep Learning}, with a focus on using \emph{images} as our input data. We begin by explaining the theoretical context of Machine Learning problems in \cref{sec:learning_ml}, then we move to Deep Learning, with a justification of the necessity of deeper models as well as a presentation of the specificities and techniques in \cref{sec:learning_dl}. Finally, we give a general overview of the field of Computer Vision with a focus on the Image Classification and Object Detection tasks in \cref{sec:learning_cv}.  } 
\end{abstract}


\section{Machine Learning}\label{sec:learning_ml}

\subsection{Objective}

\begin{figure}
    \begin{minipage}{0.45\linewidth}
        \centering
        \includegraphics[width=\linewidth]{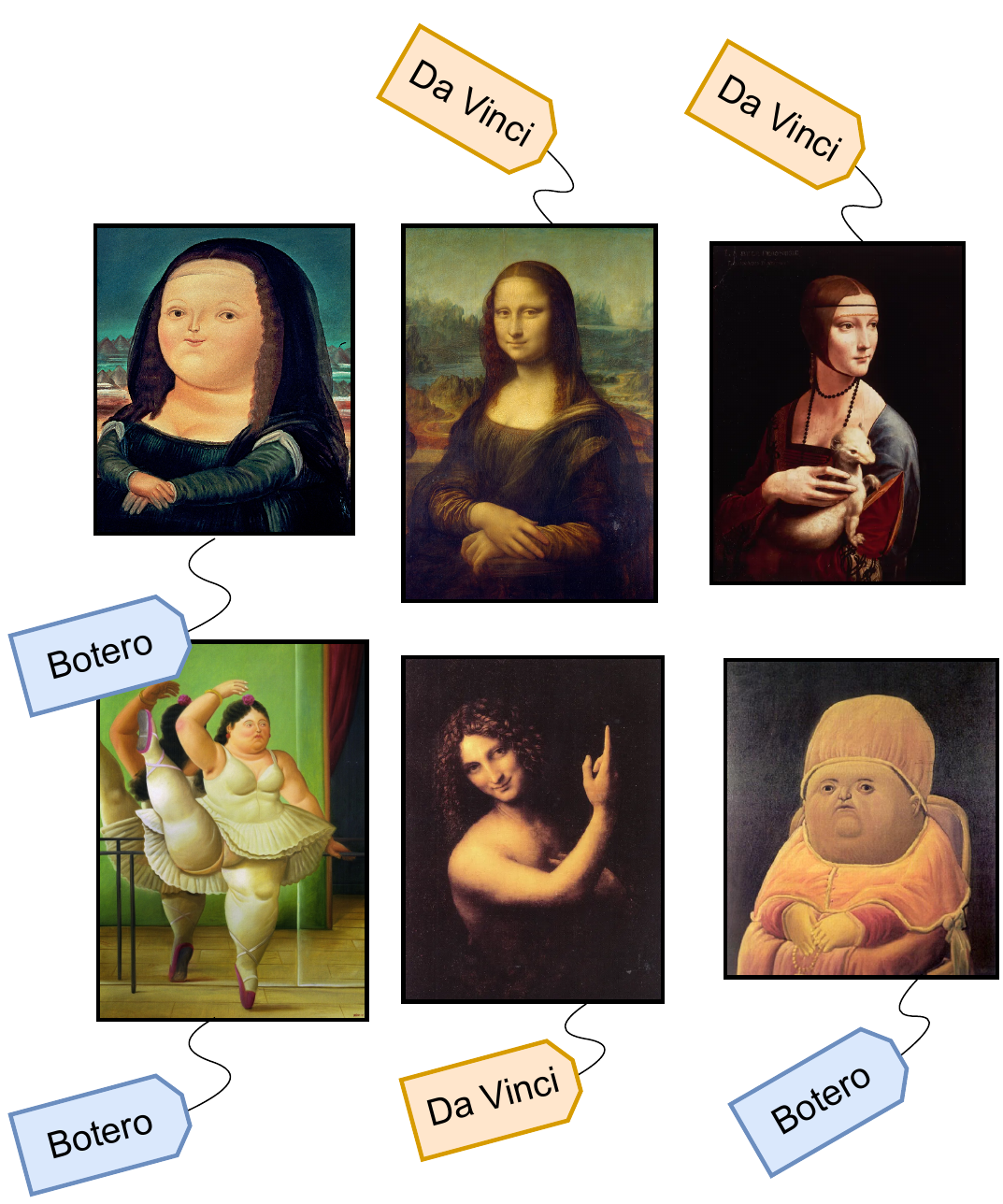}
    \end{minipage}
    \hfill
    \begin{minipage}{0.45\linewidth}
        \centering
        \includegraphics[width=\linewidth]{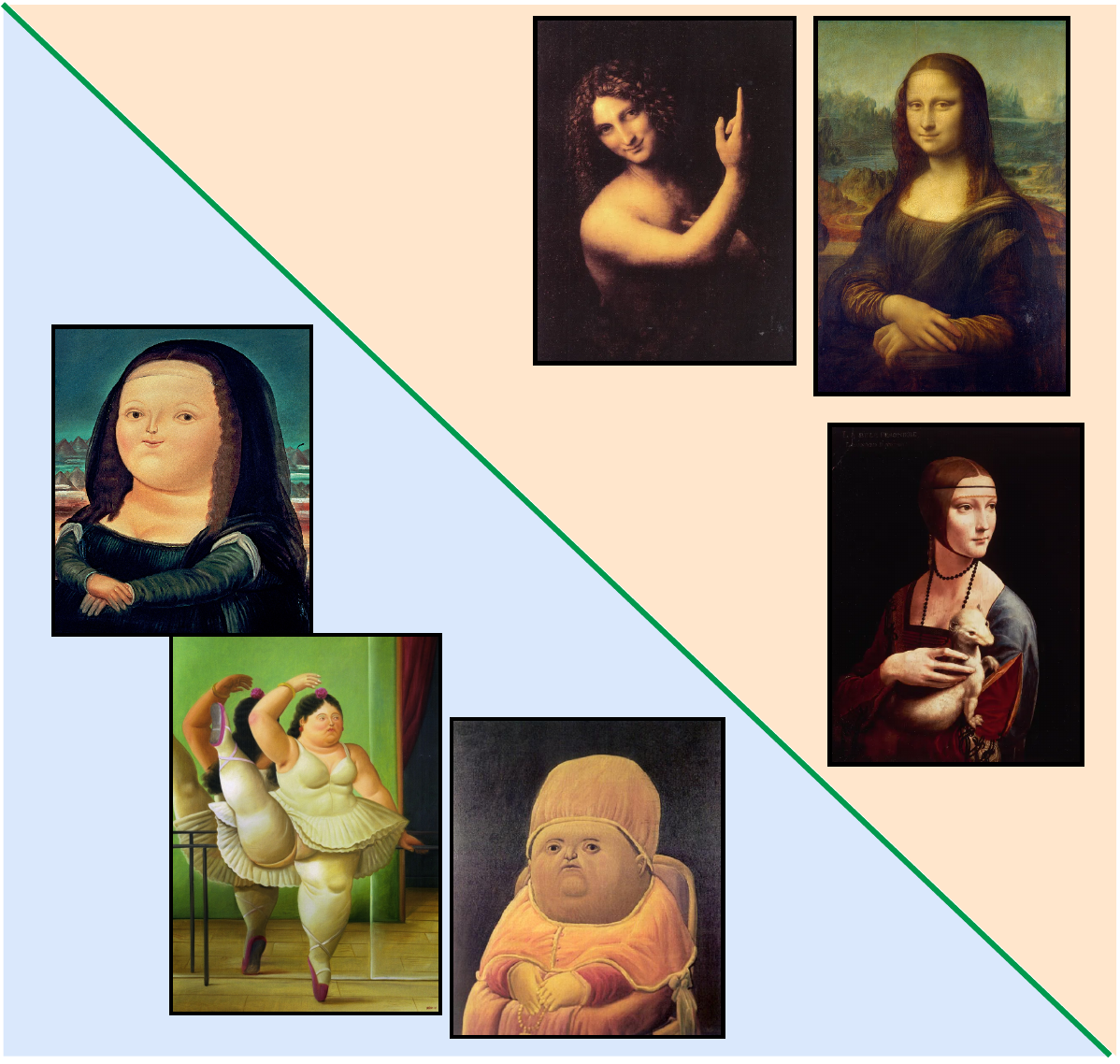}
    \end{minipage}
    
    \caption{
        \rev{We illustrate our problem using images of famous paintings, and ground-truth labels of the corresponding painters \emph{(left)}. We want to find a \emph{model} \emph{(right, the green line)} that can correctly predict the painters, which can be easily done by a human.}}
    \label{fig:learning_ml}
\end{figure}

\rev{
In this section, we want to solve a task for which we have a dataset containing a \emph{ground truth label} associated to each input samples. We illustrate the problem with images in \cref{fig:learning_ml}. Using Machine Learning (ML), the goal is to \emph{learn a model} that can correctly predict the label given an input data. This model is generally a \emph{function}, potentially complex depending on the problem to solve, and this function can be represented by the set of its parameters, that we estimate from the data during the learning process. Throughout this thesis we will be specifically interested in the case of images. However, input data could correspond to any vector of values.}

\subsection{Learning Process}

We formally describe the process of learning from data \cite{vapnik1999nature, Goodfellow-et-al-2016} in this section. 
Suppose that we have $N$ \emph{training data samples} represented as $d_1$-dimensional input vectors from an input space $\sX$: $\{ \rvx_i\}_{i=1}^N \in \sX^{N} \subseteq \R^{N \times d_1}$ with their associated \emph{ground-truths}, or \emph{labels}, represented as $d_2$-dimensional vectors from an output space $\sY$: $\{ \rvy_i\}_{i=1}^N \in \sY^{N} \subseteq \R^{N \times d_2}$, corresponding to $N$ observations of random vectors $\rvx$ and $\rvy$ that were drawn independently from fixed but unknown true data distributions $P(\rvx)$ and $P(\rvy|\rvx)$. We group these observations into matrices $\rmX \in \R^{N \times d_1}$ and $\rmY \in \R^{N \times d_2}$ to form the \emph{training set} $\train$, such that 
\begin{equation}
    \train := (\rmX, \rmY) = \left( (\rvx_1, \rvy_1), \dots, (\rvx_N, \rvy_N) \right) \sim P(\rvx, \rvy),
    \label{eq:train_set}
\end{equation}

\noindent with the joint probability distribution $P(\rvx, \rvy) = P(\rvy | \rvx) P(\rvx)$, called the \emph{data-generating distribution}. We typically make the assumption that the data samples are \emph{independent and identically distributed (i.i.d.)}. We also further assume that we have a \emph{testing set} of data $\test$, obtained by the same underlying distribution $P(\rvx, \rvy)$, that will be used to evaluate our future model and is never used during the training process that we describe in the next section.

\subsection{Risk Minimization}

The goal of Machine Learning is to find a function $f_\theta: \sX \rightarrow \sY$, or \emph{model}, that gives a good approximation of the labels associated to entries of $\sX$ \wrt{} $P(\rvy|\rvx)$. $f_\theta$ is parameterized by a set of \emph{weights} $\theta \in \Theta$. The model $f_\theta$ is selected by a \emph{learning algorithm} from a set of predefined models, called the \emph{hypothesis space}: $\Hcal \subseteq \{ f_\theta | f_\theta : \sX \rightarrow \sY, \theta \in \Theta \}$.

The learning is driven by a \emph{loss function} $\Lcal$, which provides a measure of the quality of a prediction given by the model outputs $f_\theta(\rvx)$ \wrt{} the corresponding ground-truth $\rvy$, and the \emph{true risk} $\Rcal$ given by the expected value of the loss:
\begin{equation}
    \Rcal(\theta) := \int \Lcal(\rvy, f_\theta(\rvx)) dP(\rvx, \rvy).
\end{equation}
Then, the \emph{optimal parameters} $\theta^*$ can be found by minimizing the risk functional. However, since the \emph{true} data-generating distribution is unknown, the only available information is contained in the training set \rev{$\train$ defined in \cref{eq:train_set}}. 
Thus, in practice, we compute the \emph{empirical risk} $\Ecal$ from the training set as follows:
\begin{equation}
    \Ecal(\theta) := \frac{1}{N} \sum_{i=1}^N \Lcal(\rvy_i, f_\theta(\rvx_i)).
\end{equation}
Finally, the \emph{empirical parameters} $\hat{\theta}$ are obtained by \emph{Empirical Risk Minimization (ERM)}:
\begin{equation}
    \hat{\theta} := \arg \min_{\theta \in \Theta} \Ecal(\theta).
\end{equation}
The induction principle of ERM assumes that the \emph{empirical model} $f_{\hat{\theta}}$, which minimizes $\Ecal$, results in a \emph{true risk} close to its minimum as well \cite{vapnik1999nature}. Finally, the process of minimizing $\Ecal$ is solved as an \emph{optimization problem}, and it is referred to as \emph{training a model}.

\subsection{Learning Bounds}




The soundness of the ERM principle largely depends on the \emph{uniform convergence} of the empirical risk $\Ecal$ to the true risk $\Rcal$, given $\epsilon > 0$:
\begin{equation}
    P\{ \sup_{\theta \in \Theta} | \Rcal(\theta) - \Ecal(\theta)| > \epsilon \} \rightarrow 0 \qquad \text{when} \qquad N \rightarrow \infty, 
\end{equation}

and the \emph{rate of convergence}, \ie{} how fast $f_{\hat{\theta}}$ converges towards $f_{\theta^*}$ in terms of number of samples $N$, which is also called the \emph{sample efficiency}. The more sample-efficient a learning algorithm is, the more interesting in practice it becomes. 

\emph{Rates of convergence} for a given class of function $\Hcal$ are obtained by theoretical analysis, but mainly depend on the \emph{capacity} $C(\Hcal)$ of the models. Informally, a model's capacity indicates its ability to fit a wide variety of functions, and can be quantified in different ways, with the most known ones being the \emph{Vapnik-Chervonenkis dimension} \cite{vapnik2015uniform} and the \emph{Rademacher complexity} \cite{bartlett2002rademacher}. However, these are mainly theoretical quantities useful for upper-bounding purposes. Accurately measuring the capacity of a class of model in practice is still an open problem.   
Statistical learning theory \cite{vapnik1999nature,hastie2009elements,shalev2014understanding,mohri2018foundations} gives us \emph{learning bounds} providing justification of the learning process, as well as guarantees on the rate of convergence and \emph{generalizability} of the models. We discuss the latter in the next section.

\subsection{Generalization Problem}

\begin{figure}
    \centering
    \includegraphics[width=0.8\linewidth]{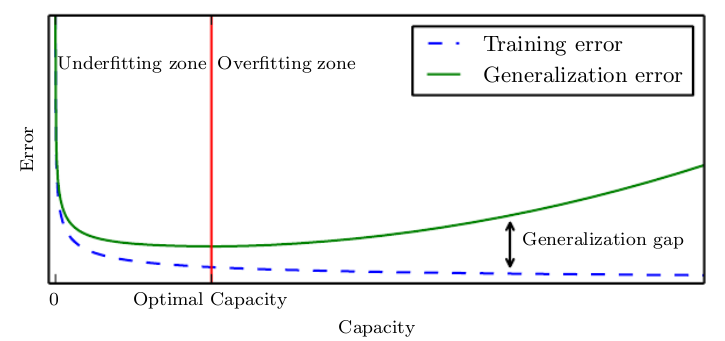}
    \caption{Illustration of the relationship between capacity and error from \cite{Goodfellow-et-al-2016}. Since we only have access to training data, training and test error behave differently. At the left end of the graph, training error and generalization error are both high. This is the underfitting regime. As we increase capacity, training error decreases, but the gap between training and generalization error increases. Eventually, the size of this gap outweighs the decrease in training error, and we enter the overfitting regime, where capacity is too large, above the optimal capacity.}
    \label{fig:generalization}
\end{figure}

The question of the \emph{generalizability} of a model arises when evaluating on the testing data $\test$. The ability of a model to perform well on unseen data is called \emph{generalization}. 

When training a model, we can compute an \emph{error measure} on the training set, that we call the \emph{training error}, and this error is usually reduced during the optimization process of the ERM. However, we want our model to be effective on the full underlying distribution of inputs $P(\rvx)$ to be useful in practice, not only on the training observations $\rmX$. Thus, we also measure the \emph{test error} on $\test$ and want it to be low as well, to estimate the \emph{generalization error} on unseen data. The need for generalizability is what separates machine learning from optimization.

The \emph{capacity} of a model has a great influence on its generalization capabilities, as can be seen in \cref{fig:generalization}. A model with low capacity will likely \emph{underfit}, \ie{} not predict accurate labels, which leads to a high training \emph{and} test error. On the other hand, a model with a too big capacity will likely \emph{overfit}, \ie{} memorize training data leading to a low training error but a high test error. 

To avoid overfitting to the training data, we can either restrict the hypothesis space to low capacity models, or more elegantly guide the optimization process towards low capacity models through \emph{regularization}, which we explain in the next section.

\subsection{Regularization}\label{sec:learning_reg}

Regularization consists in defining a \emph{penalty function}, called \emph{regularizer}, to introduce in the learning problem, that will express a preference for specific functions in the hypothesis space. 
For instance, a common regularizer used is the \emph{weight decay} \cite{kroghSimpleWeightDecay1992} $\Omega$:
\begin{equation}
    \Omega(\theta) := \| \theta \|^2_2,
\end{equation}  
defined as the squared $L_2$ norm of the parameters, that influences the optimization towards small weights and reduces overfitting. We can introduce the weight decay regularization in the training process by considering 
\begin{equation}
    \mathcal{J} := \Lcal + \reg \Omega,
\end{equation}
as our regularized loss function for the ERM, which is then called \emph{Regularized ERM (RERM)}. In the above equation, we add the parameter $\reg$ to control the strength of the regularization. 

\subsection{Types of Models}

The most simple Machine Learning algorithms are based on linear models, such that $ \Hcal = \{f(\rvx) = \rvw ^\top \rvx + b, \rvw \in \R^{d_1}, b \in \R \}$, with the most influential approach being the Support Vector Machine (SVM) \cite{boser1992training, cortes1995support}. These models are easy to learn since the training can be easily transformed into a \emph{convex optimization problem} thanks to a formulation based on a maximum margin principle:
\begin{equation}
    \hat{\rvw}, \hat{b} = \argmin_{\rvw,b} \| \rvw \|^2,  \quad \text{subject to} \quad y_i(\rvw ^\top \rvx_i - b) \geq 1, \quad \forall i \in \{ 1, \dots, N \}.
\end{equation}

To overcome the limitation of the representational capacity of linear models, which is most apparent when dealing with high dimensional data such as images, one key innovation from SVM is the \emph{kernel trick} \cite{boser1992training}. The main idea comes from observing that linear models can be rewritten as a dot product between examples: 
\begin{equation}
    \rvw ^\top \rvx + b = b + \sum_{i=1}^N \alpha_i \rvx ^ \top \rvx_i.
\end{equation}

The kernel trick consists in using a non-linear function $\phi$ to represent the data \rev{samples} $\rvx$ by $\phi(\rvx)$, and replacing the dot product by a \emph{kernel function} $k(\rvx, \rvx_i) = \langle \phi(\rvx), \phi(\rvx_i) \rangle$. The kernel trick is powerful since it allows to learn a model that is \emph{non-linear} as a function of $\rvx$ using \emph{efficient convex optimization methods}, but the mapping $\phi$ and the associated kernel are usually handcrafted or generic. The computations also become dependent on the number of training examples, which can be highly expensive when the training dataset is large. \emph{Deep Learning} represents another family of methods that allow to overcome these limitations by learning efficiently non-linear models through \emph{backpropagation} and \emph{stochastic gradient descent}.



\section{Deep Learning}\label{sec:learning_dl}

\begin{figure}
    \centering
    \includegraphics[width=0.45\linewidth]{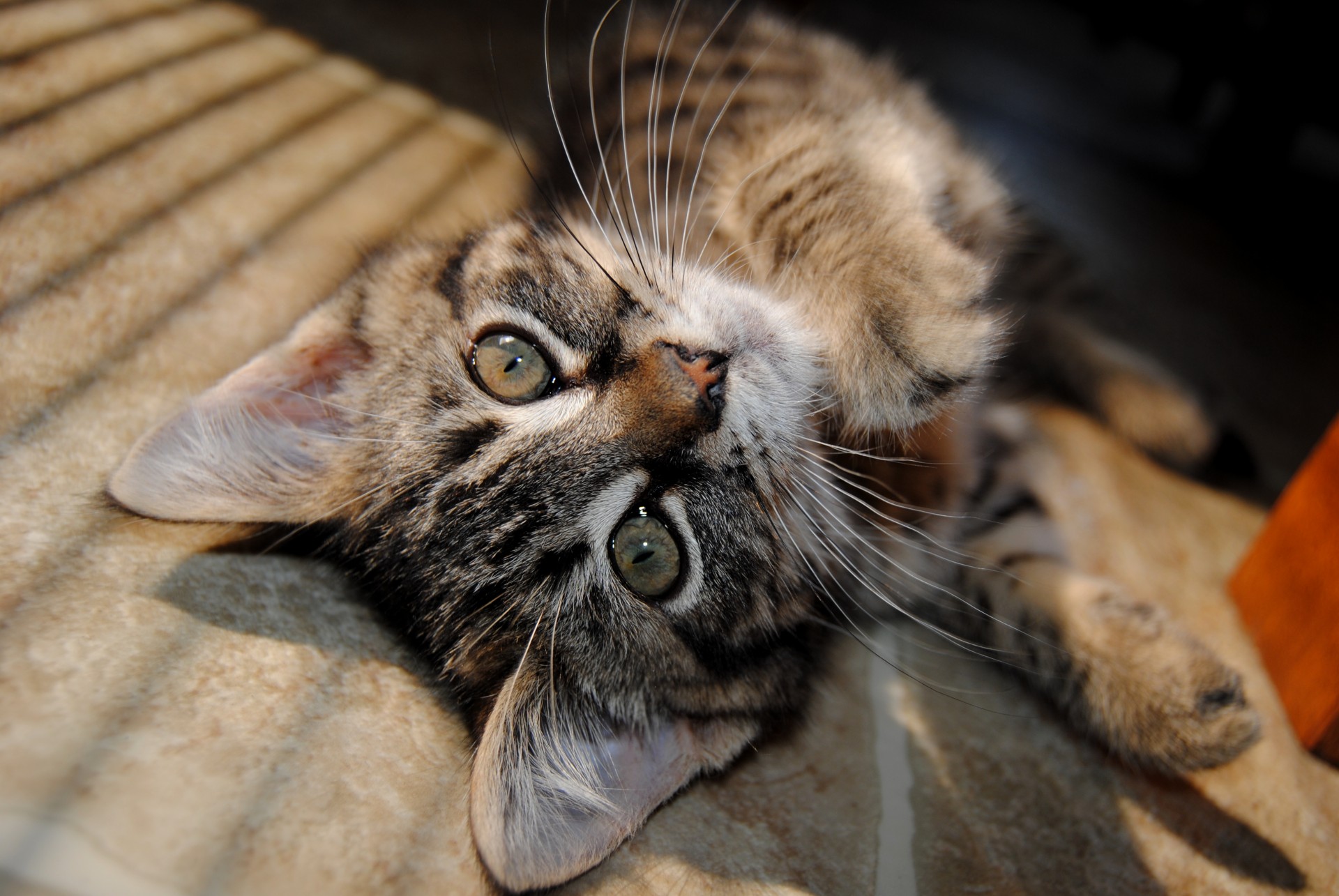}
    \includegraphics[width=0.402\linewidth]{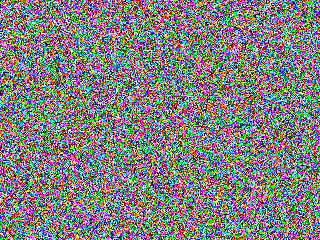}
    \caption{Although there is a non-zero probability to generate a real photo by randomly sampling values for pixels, we never actually observe this in practice. This suggests that natural images lie in specific and narrow regions of the image space, which can be difficult to represent with a simple model.}
    \label{fig:high_dim_images}
\end{figure}


Simple machine learning algorithms generally struggle with high-dimensional data, such as natural images in practice, since the number of possible configurations for the model increases exponentially with the number of dimensions. This is known as the \emph{curse of dimensionality} in Machine Learning \cite{Goodfellow-et-al-2016}. These high-dimensional data often lie in specific and complicated regions of the representation space. For instance, sampling random values from a uniform distribution for each pixel will result in an image very different from a real photo, which is illustrated in \cref{fig:high_dim_images}.  

All these problems motivated the design of more complex models to generalize to central applications such as computer vision. The strategy of Deep Learning is to \emph{learn} a non-linear mapping $\phi$ providing a representation of the data, instead of using generic or handcrafted functions with the kernel trick.

\subsection{Simple Neural Networks}

Deep Learning architectures are composed of several layers of \emph{artificial neurons}. By stacking layers and adding more units within layers, deep neural networks can represent complex, \emph{non-linear} functions. 

\subsubsection{Artificial Neurons}

A neuron is represented as a single \emph{parametric function} $f_\rvw: \sX \rightarrow \R$ which takes inputs $\rvx \in \sX$ and applies an \emph{activation function} $\varphi: \R \rightarrow \R$:

\begin{equation}
    f_\rvw(\rvx) = \varphi \left( \rvw ^\top \rvx \right).
\end{equation}

Artificial neurons were popularized by the \emph{Perceptron algorithm} \cite{rosenblatt1958perceptron} for supervised binary classification, which used a \emph{step function} as activation: $\varphi = \1_{\R_+}$. They are at the basis of most deep learning architectures.

\subsubsection{Multilayer Perceptrons}

\begin{figure}
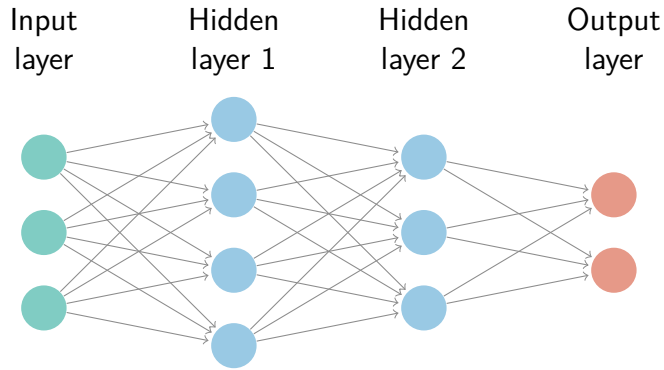

    \centering
    \begin{neuralnetwork}[height=4]
        \newcommand{\x}[2]{$x_#2$}
        \newcommand{\y}[2]{$y_#2$}
        \newcommand{\hfirst}[2]{\small $h^{(1)}_#2$}
        \newcommand{\hsecond}[2]{\small $h^{(2)}_#2$}
        \inputlayer[count=3, bias=false, title=Input\\layer]
        \hiddenlayer[count=4, bias=false, title=Hidden\\layer 1] \linklayers
        \hiddenlayer[count=3, bias=false, title=Hidden\\layer 2] \linklayers
        \outputlayer[count=2, title=Output\\layer] \linklayers
    \end{neuralnetwork}

    \caption{Illustration of an MLP with two hidden layers such that input and output dimensions are respectively $d_1 = 3$ and $d_2 = 2$.}
    \label{fig:learning_mlp}
\end{figure}

\emph{Multilayer Perceptrons (MLP)} are the simplest deep learning models, composed of multiple \emph{layers} of neurons. The number of neurons in a layer determine its \emph{width}, \ie{} its dimensionality. The first and last layers are respectively the \emph{input layer} and \emph{output layer}, and the dimensionality of the input and output data respectively define their widths. All the other layers are called \emph{hidden layers}, and their respective widths are hyperparameters specified by design. Each layer describes a function, and the full model $f_\theta$ can be seen as the composition of all the functions:

\begin{equation}
    f_\theta = f^{(d)}_{\theta^{(d)}} \circ \dots \circ f^{(1)}_{\theta^{(1)}},
\end{equation}

where $f^{(i)}_{\theta^{(i)}}$ corresponds to the function describing the $i$-th layer. The resulting parameters $\theta$ of the whole model are defined as the combination of parameters $\theta^{(i)}$ of each layer. The number of layers gives the \emph{depth} $d$ of the model. An illustration of an MLP with two hidden layers is given in \cref{fig:learning_mlp}.
A specificity of this architecture is that for a given layer, each neuron is connected to every node of the next layer, leading to many connections, and these layers are also called \emph{fully connected}. Non-linearities can be introduced between layers through \emph{activation functions}, but they make the optimization of most loss functions a non-convex problem. Popular choice of activation functions for hidden layers are the Rectified Linear Unit (ReLU) \cite{jarrett2009best} functions. 

\subsection{Gradient-Based Learning}

\subsubsection{Loss Functions and Predictive Modeling Problems}

\rev{
An important aspect of the design of neural networks is the choice of the loss function, which is tightly coupled with the choice of the activation function of the output layer. Suppose that the network provides a set of \emph{hidden features} $\rvh$, from the last hidden layer. 
Depending on the \emph{prediction problem} that the network must perform, the role of the output layer is to provide an additional transformation to the features $\rvh$ to follow the correct framing of the problem. The loss function then computes the error depending on the predictions and labels. }

\rev{
\textbf{Regression Problems} In a \emph{regression problem}, the network must predict an $n$-dimensional vector $\rvy \in \R^n$, which can be seen as the mean of a Gaussian distribution. \emph{Affine functions} are often used in this case:
\begin{equation}
    \hat{\rvy} = \theta^{(d) \top} \rvh + \rvb.
\end{equation}
The error between the output and ground-truth can then be measured using the \emph{Mean Squared Error (MSE)} loss function:
\begin{equation}
    \Lcal_{MSE}(\rvy,\hat{\rvy}) = \| \rvy - \hat{\rvy} \|_2^2.
\end{equation}
}

\rev{
\textbf{Binary Classification Problems} In a \emph{binary classification problem}, the model has to classify samples as belonging to one of two possible classes. The two classes are often labeled $0$ and $1$ for simplicity, such that $y \in \{0,1\}$. These problems can be cast as predicting a Bernoulli distribution: $\hat{y} = P(y=1)$. A popular choice is then to use a \emph{logistic sigmoid activation function} defined by:
\begin{align}
    \sigma &: \R \rightarrow \R \\
    \sigma(x) &= \frac{1}{1 + e^{-x}},
\end{align}
and the output $\hat{y}$ is then computed as:
\begin{equation}
    \hat{y} = \sigma \left(\theta^{(d) \top} \rvh + b \right).
\end{equation}
It can be seen as the combination of a \emph{linear layer} to compute $z = \theta^{(d) \top} \rvh + b$, also called \emph{logit}, with a \emph{sigmoid activation} to convert $z$ into a probability. The usual loss function is then the \emph{Binary Cross Entropy (BCE)}:
\begin{equation}
    \Lcal_{BCE}(y, \hat{y}) = - (y \cdot \log \hat{y} + (1 - y) \cdot \log \hat{y}).
\end{equation}
}

\rev{
\textbf{Multi-Class Classification Problems} In a \emph{multi-class classification problem}, the model must predict a class for an input sample over $C$ possible classes. For a sample with ground-truth class $c$, the target label is usually either the class number $y = c$ or a vector $\rvy \in \{0,1\}^C$ such that $\rvy_c = 1$, also called \emph{one-hot label vector}. These problems can be represented as predicting a probability distribution over a discrete variable with $C$ possible values, also called a Multinoulli distribution. The output is a vector $\hat{\rvy}$ such that $\hat{\rvy}_c$ represents the predicted probability of class $c$: $\hat{\rvy}_c = P(y = c)$. A \emph{softmax activation function} is often used for these problems:
\begin{align}
    \softmax &: \R^C \rightarrow \R^C \\
    \softmax(\rvx)_i &= \frac{\exp(\rvx_i)}{\sum_{j=1}^C \exp(\rvx_j)}.
\end{align}
The output $\hat{\rvy}$ is obtained such that:
\begin{equation}
    \forall i \in [1, \dots,C], \quad \hat{\rvy}_i = \softmax\left(\theta^{(d) \top} \rvh + \rvb \right)_i.
\end{equation} 
Similarly to binary classification, the output layer is the combination of a linear layer to compute $\rvz = \theta^{(d) \top} \rvh + \rvb$ with a \emph{softmax} activation. The \emph{Cross Entropy (CE)} loss function is used for these problems:
\begin{equation}
    \Lcal_{CE}(\rvy, \hat{\rvy}) = - \sum_{i=1}^C \rvy_i \cdot \log \hat{\rvy}_i.
\end{equation}
}

\rev{
While the \emph{affine}, \emph{sigmoid} and \emph{softmax} activation functions described above are the most common, neural networks can generalize to almost any kind of output layer. The total cost function used to train a neural network will often combine one of a loss function described above with additional regularization terms, such as the \emph{weight decay} discussed in \cref{sec:learning_reg}.}

\subsubsection{Optimization by Backpropagation}

Deep Learning models are usually trained iteratively using \emph{gradient-based} optimizers \cite{bottou1998online,kingma2015adam,loshchilov2018decoupled} on the training data. The loss functions being non-convex because of non-linear activation functions, the optimization problems can have several local and global minima. The optimization process aims to drive the loss function to a very low value, but has no convergence guarantees and is very sensitive to initialization. 

The optimization algorithms typically compute the updates of the parameters based on an expected value of the loss function computed on randomly sampled subsets, or \emph{minibatch}, of the training data. We usually call \emph{batch size} the size of each minibatch in the training process. 
In practice, the choice of the batch size is important and is driven by several factors. On the one hand, large minibatch leads to more accurate gradients, and more different data seen for the same number of training iterations. On the other hand, a small batch size induces a noise that has also a regularization effect during training. However, the batch size is ultimately limited by the hardware capacities.

Stochastic Gradient Descent (SGD) \cite{bottou1998online} and its variants, with the popular Adam \cite{kingma2015adam}, are the most used optimization algorithms in machine learning in general and deep learning in particular. A crucial parameter in these algorithms is the \emph{learning rate}, \ie{} the step size in the direction given by the gradient. In practice, it is necessary to define a function, called \emph{scheduler}, to decay the learning rate over the course of the training process. Popular choices of scheduler include linear, cosine or step functions.

Computing an analytical expression for the gradient of the model can be straightforward, but numerically evaluating it can be computationally expensive. The \emph{back-propagation} algorithm \cite{rumelhart1986learning} is a simple procedure to efficiently compute an approximation of the gradients. The general idea of the algorithm is to recursively apply the \emph{chain rule of calculus} to compute the derivatives of the model as a function formed by the composition of other functions. In practice, back-propagation is integrated and computed by the deep learning frameworks implementing \emph{automatic differentiation}, such as Pytorch\footnote{\url{https://pytorch.org/}} or Tensorflow\footnote{\url{https://www.tensorflow.org/}}.

\rev{
However, gradient-based learning methods with backpropagation also suffers from instabilities in the training process. During each training iteration, the updates of the weights are scaled depending on the current weights. With deep architectures, the scaling accumulates between layers because of backpropagation, and updates can become either very small \emph{(vanishing gradients)} or very large \emph{(exploding gradients)} \cite{hochreiter2001gradient}. Vanishing gradients effectively prevent the weights from changing their values, while exploding gradients can result in an unstable training.}

\subsection{Deeper Architectures}


\subsubsection{Convolutional Neural Networks}

\begin{figure}
    \centering
    \includegraphics[width=0.8\linewidth]{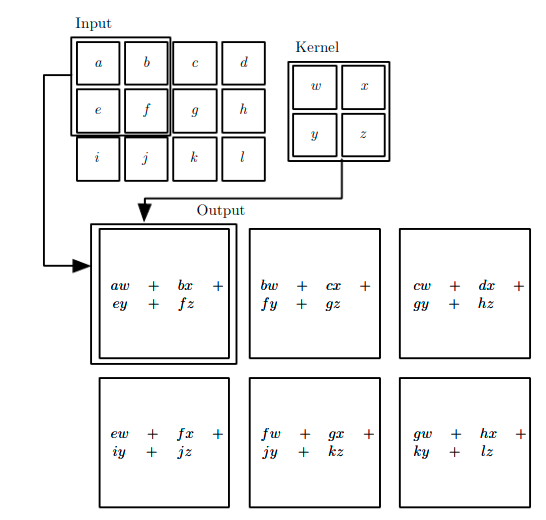}
    \caption{An illustration of the 2D convolution operation used in CNNs from \cite{Goodfellow-et-al-2016}. The input and output are also called \emph{feature maps}. A \emph{kernel} matrix slides across the input feature map to compute the output feature map.}
    \label{fig:conv}
\end{figure}

Convolutional Neural Networks \cite{lecun1989handwritten}, or \emph{CNNs}, are a specialized kind of MLP for data with a grid-like topology and local information. CNNs are composed of \emph{convolutional layers} which apply \emph{convolution} on the input using different filters, or \emph{kernels}, and pass the result to the next layer. For instance, with two-dimensional input $I$ and kernel $K$, the convolution operation is defined as:
\begin{equation}
    (I * K)(i,j) = \sum_m \sum_n I(m,n)K(i-m,j-n).
\end{equation}
We usually assume that these kernels are 0 everywhere but a finite set of points.
See \cref{fig:conv} for an illustration of the convolution operation. CNNs have the advantage of working with inputs of variable sizes and implement by design three important concepts: 
\begin{description}
    \item[Sparse connectivity:] with a kernel smaller than the input, there are fewer parameters and connections than in MLP layers.
    \item[Parameter sharing:] the same parameters in the kernel are used for different locations in the input feature map.
    \item[Equivariance to translation:] the convolution operation is invariant to a translation in the inputs since the kernel is translated at different location in the input feature map.      
\end{description}

\begin{figure}
    \centering
    \includegraphics[width=\linewidth]{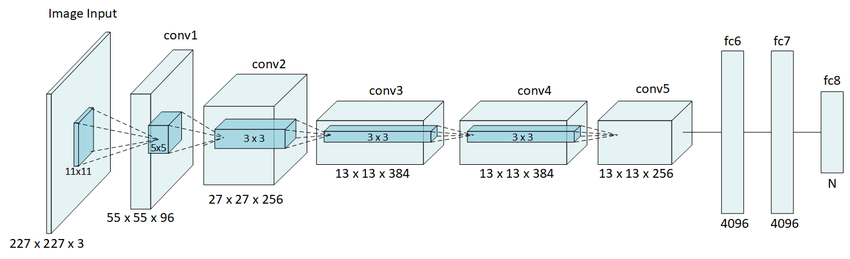}
    \caption{Illustration of the AlexNet architecture, from \cite{krizhevsky2009learning}. Dimensions of the resulting features are written below each layers. The architecture is composed of 5 convolutional layers \emph{(conv)} followed by 3 fully-connected layers \emph{(fc)}.}
    \label{fig:learning_alexnet}
\end{figure}

\begin{figure}
    \centering
    \begin{tikzpicture}

        \node[fill=orange!50] (l1) {layer 1};
        \node[blue!50!black, right=of l1, label={below:activation}] (act1) {$a(\rvx)$};
        \node[fill=teal!50, right=of act1] (l2) {layer 2};
        \node[right=of l2, font=\Large, label={below:add}, inner sep=0, pin={60:$f(\rvx) + \rvx$}] (add) {$\oplus$};
        \node[blue!50!black, right=of add, label={below:activation}] (act2) {$a(\rvx)$};
      
        \draw[->] (l1) -- (act1);
        \draw[->] (act1) -- (l2);
        \draw[<-] (l1) -- ++(-2,0) node[below, pos=0.8] {$\rvx$};
        \draw[->] (l2) -- (act2) node[above, pos=0.8] {};
        \draw[->] ($(l1)-(1.5,0)$) to[out=90, in=90] node[below=1ex, midway, align=center] {skip connection\\(identity)} node[above, midway] {$\rvx$} (add);
        \draw[decorate, decoration={brace, amplitude=1ex, raise=1cm}] (l2.east) -- node[midway, below=1.2cm] {$f(\rvx)$} (l1.west);
      
      \end{tikzpicture}
      \caption{Illustration of residual connection, or skip connection, inspired by \cite{he2016deep}. This is an important building block of the ResNet architecture.}
      \label{fig:residual}
\end{figure}
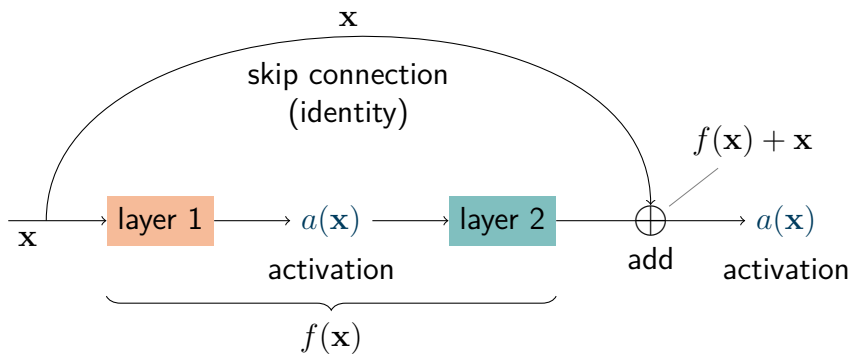

The CNN architectures have been at the heart of many advances in deep learning, their widespread application leading to deep CNNs reaching state of the art in many vision problems starting with AlexNet \cite{krizhevsky2017imagenet}, with the architecture illustrated in \cref{fig:learning_alexnet}. Among the most popular ones are the Residual Networks or \emph{ResNet} \cite{he2016deep}, that implement \emph{residual connections} illustrated in \cref{fig:residual}, allowing to go deeper in the architectures design and represent even more complex spaces, also called \emph{embeddings}. \rev{Notably, residual connections allow to limit the problem of \emph{vanishing gradients} \cite{he2016deep}.} They are now an important building block used in almost all recent architectures, notably the Transformers \cite{vaswani2017attention} detailed hereafter.

\subsubsection{Transformers}

\begin{figure}
    \centering
    \includegraphics[width=0.5\linewidth]{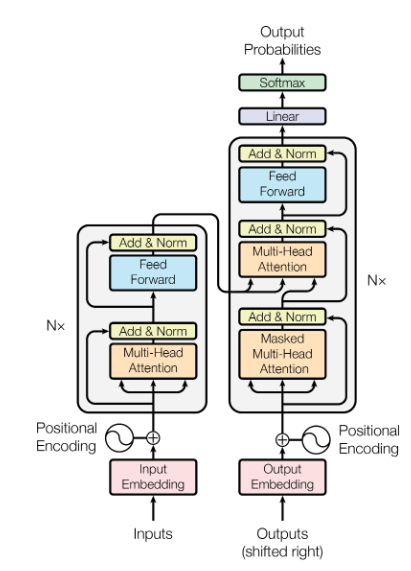}
    \caption{The Transformer encoder-decoder architecture from \cite{vaswani2017attention} used for NLP tasks. The left and right halves respectively represent the \emph{Transformer encoder} and \emph{Transformer decoder} architectures.}
    \label{fig:transformer}
\end{figure}

The \emph{Transformer} architecture \cite{vaswani2017attention}, illustrated in \cref{fig:transformer}, was originally introduced for Natural Language Processing (NLP), but has been recently used and adapted to Computer Vision tasks as well \cite{carion2020end,dosovitskiy2021an}. Transformer are designed to represent \emph{set-to-set functions}, \ie{} they take a \emph{fixed number of tokens} as input and outputs \emph{the same number of processed tokens}. However, there is no particular association between the input tokens and the output tokens. 

\rev{
Before being used by the encoder and decoder, input and output tokens are converted to \emph{embedding vectors} through simple \emph{representation functions} such as a linear projection \cite{dosovitskiy2021an}. 
For the model to make use of the order of the input sequence, information about the position of each token are injected through \emph{positional encoding}. There are many choices of positional encodings, learned or fixed \cite{gehring2017convolutional}.}
The \emph{original} Transformer architecture \cite{vaswani2017attention} is decomposed into two parts, an \emph{encoder} and a \emph{decoder}. 

\begin{figure}
    \begin{subfigure}{0.45\linewidth}
        \centering
        \includegraphics[width=0.4\linewidth]{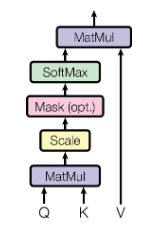}
        \caption{Scaled Dot-product Attention}
        \label{fig:attention}
    \end{subfigure}
    \begin{subfigure}{0.45\linewidth}
        \centering
        \includegraphics[width=0.527\linewidth]{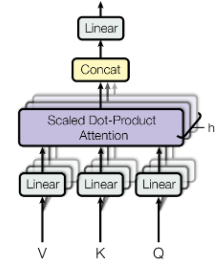}
        \caption{Multi-head Attention}
        \label{fig:multi-head}
    \end{subfigure}
    \caption{Illustration of the sequencing of \textbf{(a)} the Scaled Dot-product Attention and \textbf{(b)} the Multi-head Attention, from \cite{vaswani2017attention}. These operations are computed between the keys $K$, queries $Q$ and values $V$ matrices.}
\end{figure}

The \textbf{encoder} is made of a stack of several ($N$) identical layers, with each layer split into two sub-layers: a \emph{multi-head self-attention}, \rev{illustrated in \cref{fig:multi-head}}, and a \rev{\emph{position-wise MLP}}. \rev{Each sub-layers are surrounded by residual connections \cite{he2016deep}, followed by layer normalization \cite{ba2016layer}.}
The inputs of the layer are projected into three matrices called \emph{Key} ($K$), \emph{Value} ($V$), and \emph{Query} ($Q$).
The \emph{attention operation} corresponds to a \emph{scaled dot-product} between these three matrices:

\begin{equation}
    Attention(Q,K,V) = \softmax(\frac{QK^\top}{\sqrt{d}}) V,
\end{equation}
with $d$ the dimension of the keys $K$.
\rev{The main objective of \emph{attention} is to focus on the important elements of the task by allowing for an adequate weighting of the inputs. An illustration of the sequencing of operations in attention can be found in \cref{fig:attention}.}

The projection and computation of the attention are repeated on several sub-layers \emph{in parallel}, hence the name \emph{multi-head}. Using different $K$, $Q$ and $V$ matrices for each head, allows considering different possible representations for the same inputs. The term \emph{self-attention} is also used in the literature, since the attention function is computed between the input vectors. The outputs of the Multi-head attention layer is then sent to the \emph{position-wise MLP}.
\rev{In each layers of the encoder and decoder, the \emph{position-wise MLP} is applied to each position separately but identically. While the architectures of the MLP are the same across different positions, they use different parameters.}

The \textbf{decoder} part is also made of a stack of $N$ identical layers. In addition to the two sub-layers in each encoder layer, the decoder inserts a third sub-layer, which performs multi-head attention over the output of the encoder stack. \rev{The multi-head attention is also modified by a \emph{masking}, to prevent positions from attending to subsequent positions.}

For Computer Vision, Transformers have been successfully adapted for Image Classification \cite{dosovitskiy2021an} but also for Object Detection \cite{carion2020end}, and are now strong contenders to CNNs. We present these computer vision tasks in the next section.

\section{Deep Computer Vision}\label{sec:learning_cv}

\begin{figure}
    \centering
    \begin{subfigure}{0.45\linewidth}
        \includegraphics[width=\linewidth]{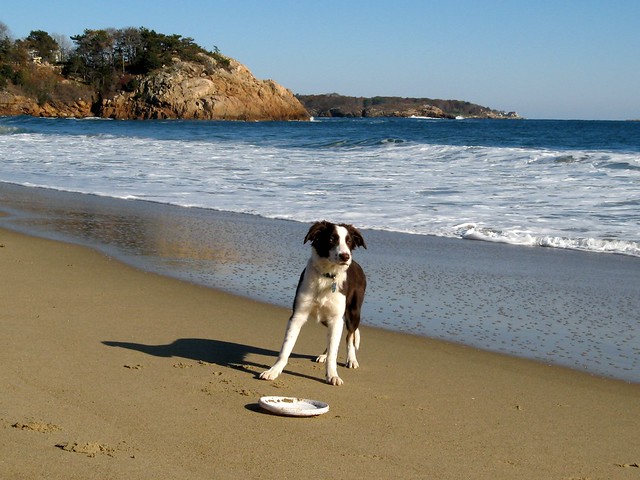}
        \caption{Class: Dog}
    \end{subfigure}
    \begin{subfigure}{0.45\linewidth}
        \includegraphics[width=\linewidth]{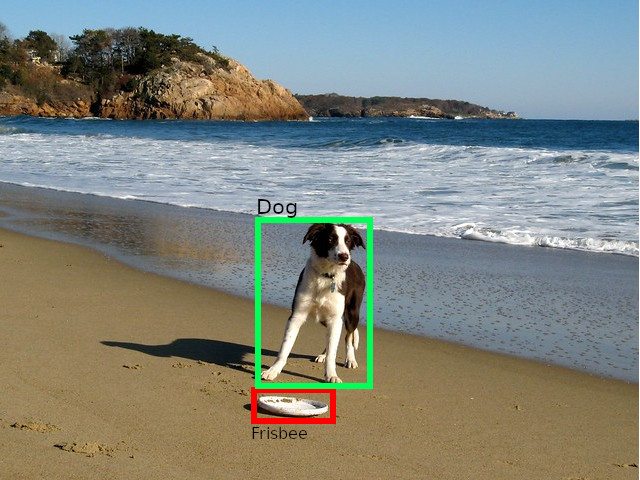}
        \caption{Object 1: Dog / Object 2: Frisbee}
    \end{subfigure}
    \caption{An illustration of the difference between \textbf{(a)} Image Classification and \textbf{(b)} Object Detection tasks. In Image Classification, the most probable class among a set of predefined class is predicted for the whole image, ignoring the fact that the image could represent multiple different objects. In Object Detection, all objects among all known class of objects appearing in an image must be identified and located through a bounding box.}
    \label{fig:ic_od}
\end{figure}

\subsection{Image Classification}

In Computer Vision, the task of \emph{Image Classification} consists in predicting a class for a whole image, among multiple predefined classes. Classification is one of the go-to problems for evaluating and comparing Deep Learning models, thanks to several benchmarks from black and white handwritten digits in \emph{MNIST} \cite{lecun1998gradient} in the early days of Deep Learning, to tiny natural images in \emph{CIFAR} \cite{krizhevsky2009learning}, and now large-scale image recognition challenges with \emph{ImageNet} \cite{russakovskyImageNetLargeScale2015}. These benchmarks have allowed to quantitatively compare results and improvements between different methods and relatively to baselines. 
With the advent of large-scale datasets such as the full ImageNet dataset \cite{deng2009imagenet}, \rev{containing 14M images divided into 21k classes} \cite{NEURIPS_BENCHMARKS2021_98f13708}, the Image Classification task on ImageNet serves as the main initialization of models before training on other downstream tasks \cite{ren2015faster,He_2019_ICCV}. Improving performance on the classification tasks often serves as a proxy metric for general applicability \cite{kornblith2019better}. 

However, this task has its limitations. It does not take into account multiple different objects in a single image, nor their locations, illustrated in \cref{fig:ic_od}. The model only learns general information from the images, rather than local information as it is done in more \emph{dense tasks}, \ie{} tasks that require a deeper understanding of the image, such as Object Detection.





\subsection{Object Detection}\label{sec:learning_od}



\begin{figure}
    \includegraphics[width=1.0\linewidth]{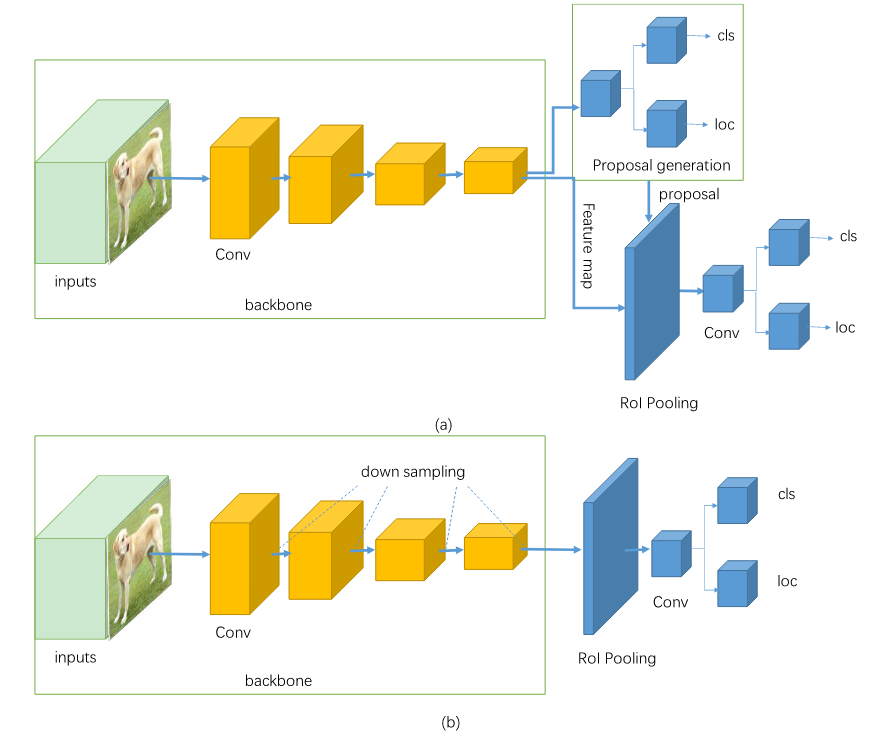}
    \caption{Illustration from \cite{jiao2019survey} of the basic architectures of (a) \emph{two-stage detectors} and (b) \emph{one-stage detectors}, based on deep convolutional networks. The \emph{cls} and \emph{loc} outputs correspond respectively to the class and location predictions of objects. The detectors are composed of a \emph{backbone network (in yellow)}, with additional \emph{detection-specific layers} \emph{(in blue)}.}
    \label{fig:conv_od}
\end{figure}

\begin{figure}
    \includegraphics[width=1.0\linewidth]{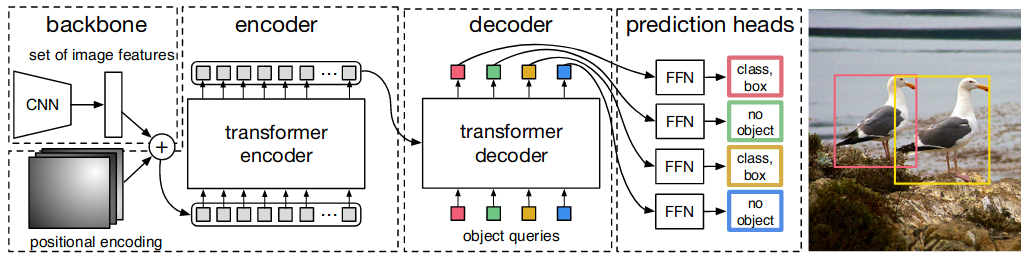}
    \caption{Illustration from \cite{carion2020end} of DETR, the original architecture based on Transformers for end-to-end object detection. The model is also composed of a \emph{backbone network (top left)}, followed by detection-specific \emph{transformer encoders and decoders}.}
    \label{fig:detr}
\end{figure}

The goal of \emph{Object Detection} is to locate and classify every single object appearing in an image, from a pool of known classes objects. The \emph{locating task} is done by predicting a \emph{Bounding Box} around the object. Early \emph{Object Detectors} were based on the computation of local features such as SIFT \cite{lowe2004distinctive} or HOG \cite{dalal2005histograms} along with more classic ML paradigms, but they are now largely outperformed by Deep Learning models. These more recent object detectors can be separated into three categories, of which we give a brief review below. We refer the reader to \cite{jiao2019survey,zaidi2022survey} for more thorough surveys of modern deep-learning based object detection methods. They all use a \emph{backbone} network as a \emph{feature extractor}, typically a ResNet \cite{he2016deep}, then inserts specific additional networks. The backbone is usually pretrained on Image Classification or an unsupervised task \cite{He_2019_ICCV}, before \emph{finetuning} the full network, \ie{} updating the parameters through training, on the Object Detection downstream task. \rev{Illustrations of typical object detection models are given in \cref{fig:conv_od,fig:detr}.}

Similarly to Image Classification, Object Detection has several large scale benchmark datasets to compare improvements between methods. The most notorious are the \emph{PASCAL VOC} \cite{everingham2010pascal} and \emph{MS COCO} \cite{lin2014microsoft} benchmarks.

\paragraph{Two-stage Detectors}

These networks \cite{girshick2014rich,girshick2015fast,ren2015faster,he2015spatial,dai2016r,lin2017feature}, illustrated in \cref{fig:conv_od} (a), have a separate module to generate \emph{Region Proposals} \cite{uijlings2013selective}. They find an arbitrary number of objects proposals in the input image during the first stage, and then classify and refine their location in the second. The most representative detector from this category is \faster{} \cite{ren2015faster}, in which the first stage proposes candidate object bounding boxes using a \emph{Region Proposal Network (RPN)}, and the second stage extracts features from each candidate box by the RoI (Region of Interest) pooling operation for the following classification and regression tasks. As these methods have two separate steps, they generally take more time to generate and process the region proposals, but they have strong detection performance.

\paragraph{One-stage Detectors}

These detectors \cite{redmon2016you, redmon2017yolo9000, liu2016ssd,lin2017focal, law2018cornernet,zhou2019objects,tian2019fcos} classify and locate in a single shot using a dense sampling. They either predefine boxes, or \emph{anchors}, of various scales and aspect ratios in the feature maps, that are then refined to locate and classify objects \cite{redmon2016you, redmon2017yolo9000, liu2016ssd}, or predict bounding boxes according to reference points in the image \cite{law2018cornernet,zhou2019objects,tian2019fcos}. One-stage methods have a much simpler design than two-stage ones, and have the benefit of real-time inference speed at the cost of lower detection performance.  

\paragraph{Transformer-based Detectors}

Since the introduction of \emph{Detection Transformers} (DETR) \cite{carion2020end}, transformer-based detectors \cite{carion2020end,zhu2020deformable,dai2021dynamic, meng2021conditional, wang2021anchor, liu2022dabdetr, yang2022querydet, li2022dn} have been an increasingly popular architecture. They allow end-to-end detection with a simpler overall architecture and without the need for hand-crafted heuristics such as the Non-Maximal Suppression (NMS) used in \faster{}, or the predefined boxes locations in one-stage detectors. However, Transformers being set-to-set functions, these detectors rely on the Hungarian algorithm \cite{munkres1957algorithms} to find the optimal matching between predicted object proposals and the ground-truths, which can be costly to compute at each iterations.

\rev{In the next chapter, we now focus on the main problem of interest of this thesis, \emph{learning from few labels}. Some methods and problems presented in this chapter will be revisited according to this setting.}


%% file: chapters/sota.tex
\chapter{Learning with few labels: an Overview}\label{chap:sota}

\minitoc


\begin{abstract}
    \textit{This chapter aims at discussing related work in the context of learning with limited labels. \rev{The main goal is to give a general overview of the recent background related to this topic while allowing us to present our contributions in the following chapters}. 
    The organization of the chapter is as follows. First, we present previous work on \emph{data augmentation} in \cref{sec:related_work_data_augm}, to alleviate label scarcity by increasing diversity. Then we focus on learning frameworks adapted to the limited data setting, \emph{Meta-Learning} in \cref{sec:related_work_meta_learning}, specifically designed for FSL, \emph{Transfer Learning \rev{by Pretraining}} in \cref{sec:related_work_transfer_learning} which includes discussion on both \emph{pretraining} and \emph{fine-tuning} strategies, and finally \emph{Semi-Supervised Learning \rev{with few annotations}} in \cref{sec:related_work_semi_sup}, that leverages unlabeled data along with few labeled data. In each section, we discuss applications for image classification and object detection separately.}
\end{abstract}

\section{Data Augmentation with few images}\label{sec:related_work_data_augm}

\begin{figure}
    \centering
    \includegraphics[width=0.9\linewidth]{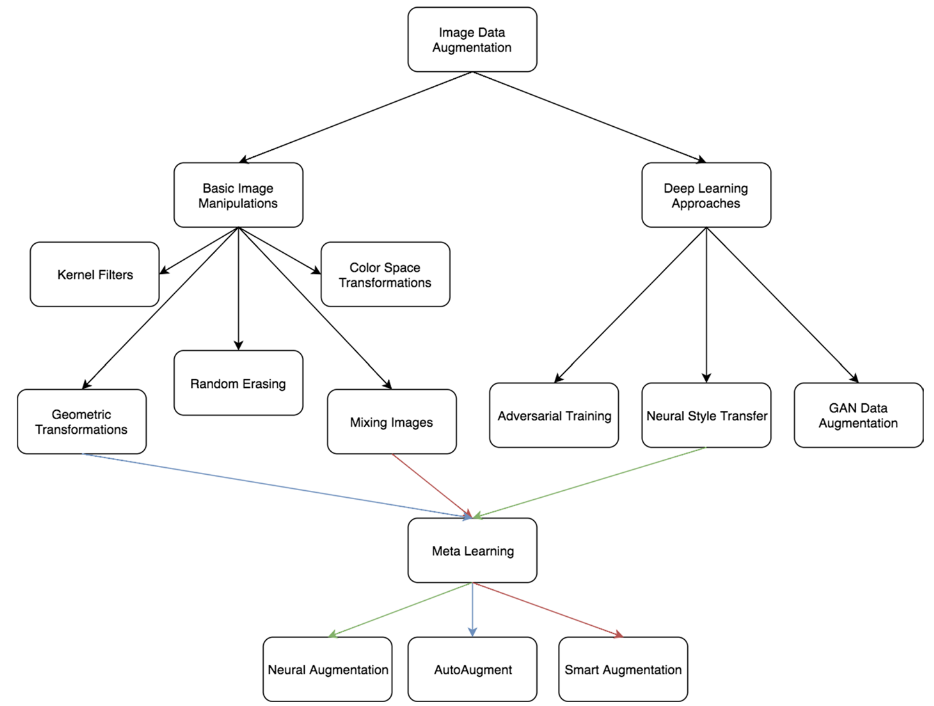}
    \caption{A general taxonomy of data augmentations techniques from \cite{shorten2019survey}.}
    \label{fig:data_augm_taxo}
\end{figure}

The most direct way to deal with the problem of having too few samples, is to artificially increase the diversity in the training dataset through \emph{data augmentation procedures}. Data augmentation has been a staple since the early days of deep learning \cite{yaeger1996effective}. It helps to guide the model into learning invariance properties or have a better estimate of the true data distribution. For a comprehensive overview of data augmentation techniques for computer vision applications, we refer the reader to \cite{shorten2019survey}, and to \cite{wang2020generalizing} for more details in the specific case of few-shot learning. A taxonomy of the different techniques is given in \cref{fig:data_augm_taxo}.

\subsection{Hand-crafted augmentations}

\begin{figure}
    \centering
    \includegraphics[width=\linewidth]{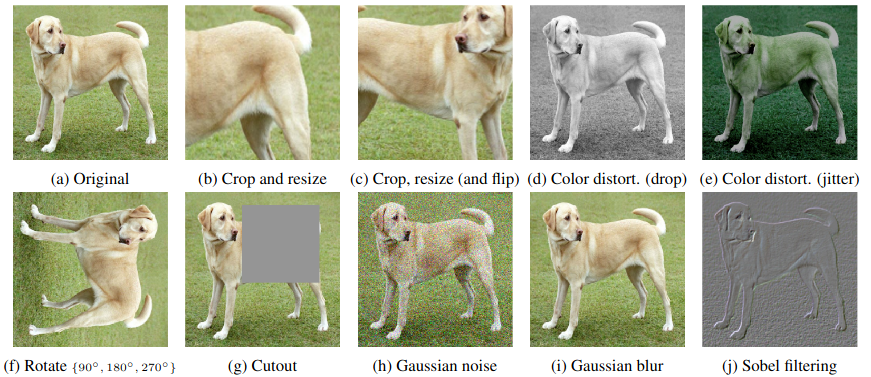}
    \caption{\rev{Illustration from \cite{chen2020simple} of common hand-crafted augmentations.}}
    \label{fig:augm_simclr}
\end{figure}

\begin{figure}
    \begin{subfigure}[b]{0.45\linewidth}
        \centering
        \includegraphics[width=\linewidth]{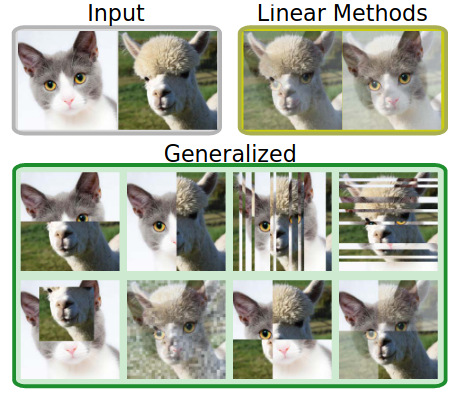}
    \end{subfigure}
    \begin{subfigure}[b]{0.49\linewidth}
        \centering
        \includegraphics[width=\linewidth]{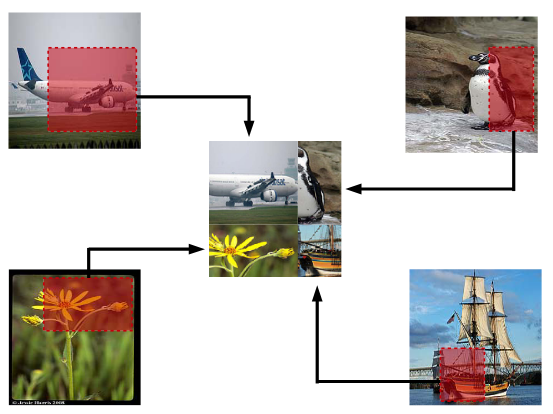}
    \end{subfigure}
    \caption{Different ways of mixing images. \textbf{(left)} Non-linear vs linear mixing from \cite{summers2019improved}. \textbf{(right)} Mixing with random patch and cropping \cite{takahashi2019data}.}
    \label{fig:image_mixing}
\end{figure}

Choosing the best augmentations for a given problem can require guidance from domain experts. Indeed, depending on the data and tasks to solve, some augmentations can be important while others can become inadequate \cite{cubuk2019autoaugment}. Furthermore, the magnitude of the transformation must be chosen carefully. Indeed, every image processing function can result in a label changing transformation at some distortion magnitude \cite{shorten2019survey}. 
Common hand-crafted augmentations are based on basic image transformations, \rev{illustrated in \cref{fig:augm_simclr}}, that can be grouped into \emph{geometric transformations} such as flipping, cropping, rotation, translation, and \emph{photometric transformations} with color jittering, gaussian blur or grayscale conversion \cite{taylor2018improving}.
A counterintuitive but effective augmentation technique is \emph{mixing images} together, illustrated in \cref{fig:image_mixing}. The combination can be done linearly \cite{zhang2017mixup,tokozume2018between,inoue2018data}, with non-linear functions \cite{summers2019improved,takahashi2019data}, or through random cropping, patching and pasting training images together \cite{yun2019cutmix,takahashi2019data,bochkovskiy2020yolov4,li2021cutpaste}.

\begin{figure}
    \centering
    \includegraphics[width=0.8\linewidth]{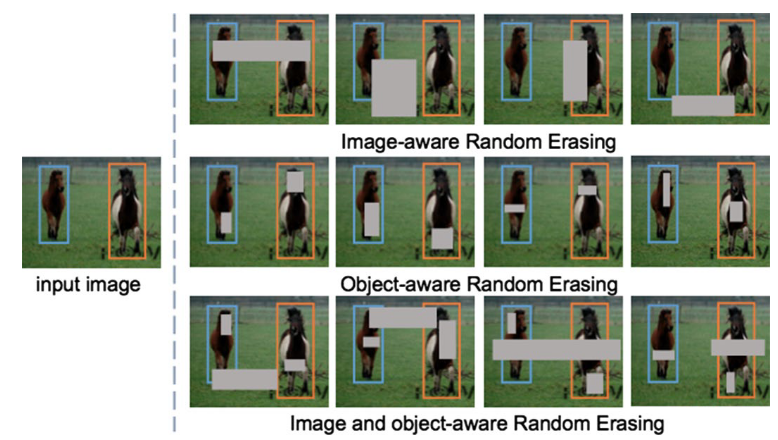}
    \caption{Example of random erasing on object detection task \cite{zhong2020random}}
    \label{fig:random_erasing}
\end{figure}

Another interesting procedure is the \emph{Random erasing}, also called \emph{CutOut}, augmentation \cite{zhong2020random,devries2017improved} and illustrated in \cref{fig:augm_simclr,fig:random_erasing}. It works by randomly masking a patch in an image with random pixel values, preventing the model from overfitting on particular visual features in images.

\subsection{Learning augmentations}

\begin{figure}
    \centering
    \includegraphics[width=\linewidth]{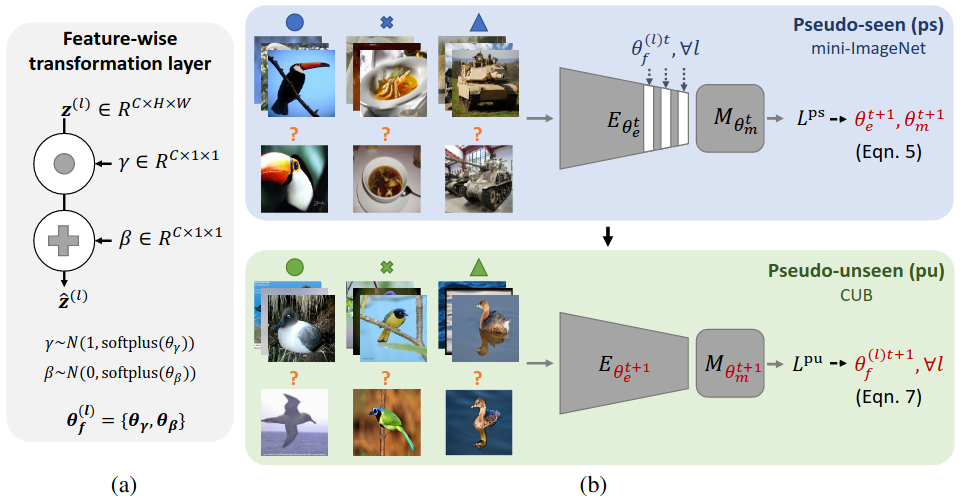}
    \caption{\rev{Illustration of an augmentation learned in the feature space, from \cite{Tseng2020Cross-Domain}. (a) Learned feature-wise transformation layer. (b) The layers are inserted and learned during training, to simulate feature distributions extracted from tasks in various domains.}}
    \label{fig:augm_feat}
\end{figure}

In contrast to the above-mentioned transformations that are applied in the input space, augmentations can also be applied in the feature space \cite{chawla2002smote,devries2017dataset}. To do so, recent methods learn several layers of neural networks for the transformation, using additional unlabeled data \cite{fu2015transductive}, or online during training \cite{hariharanLowShotVisualRecognition2017,wangLowShotLearningImaginary2018a,schwartz2018delta,Tseng2020Cross-Domain}, \rev{with one of such method illustrated in \cref{fig:augm_feat}}. However, it is very difficult to interpret the augmented data after feature space augmentations. Finally, if transformations are plausible in the image space, they provide greater benefit for improving performance and reducing overfitting than generic augmentations in the feature space \cite{wong2016understanding}.


\subsection{Limitations of augmentation procedures}

\begin{figure}
    \centering
    \includegraphics[width=\linewidth]{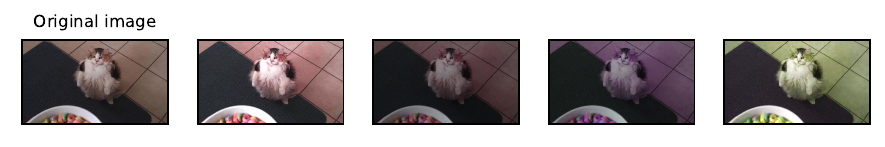}
    \caption{In this example, augmenting the original image of a cat with only color transformations overly biases towards spatial characteristics of the cat. Learning a deep model on limited data with these color transformations will result in more overfitting than learning without augmentations.}
    \label{fig:augm_biases_cat}
\end{figure}

One important consideration when choosing augmentation procedures is the intrinsic bias in the initial dataset, since augmented samples are always based on the original samples. While augmentations are designed to alleviate specific biases and prevent overfitting by modifying limited datasets with characteristics of larger data, they cannot overcome all biases present in small datasets and will always reproduce some of them. An illustration of this problem is given in \cref{fig:augm_biases_cat}. Over-extensive use of augmentations can even lead to more overfitting by emphasizing specific features \cite{shorten2019survey}. Therefore, one cannot only rely on data augmentations for few-shot learning.

\section{Meta-Learning}\label{sec:related_work_meta_learning}


\begin{figure}
    \centering
    \includegraphics[width=\linewidth]{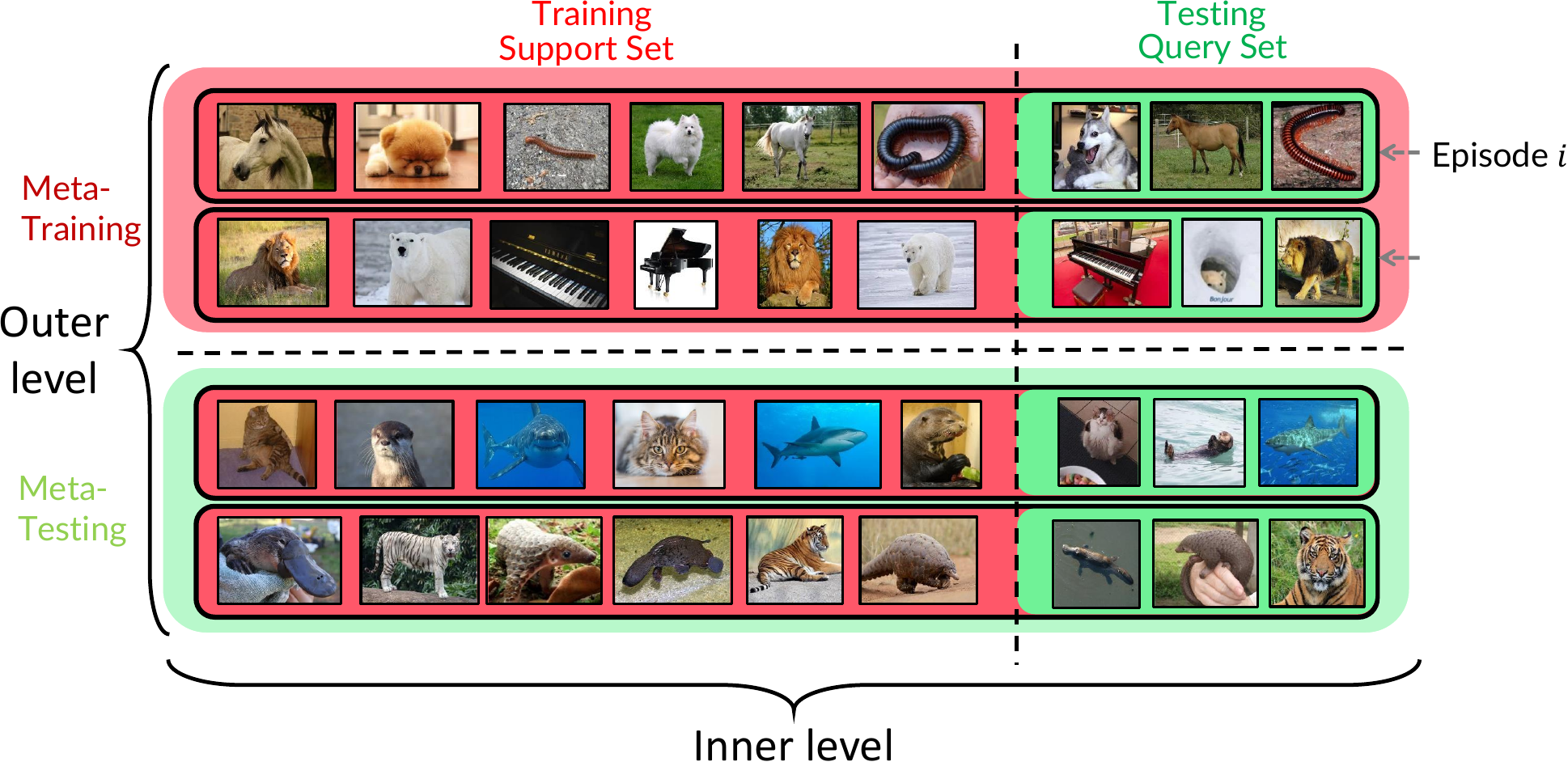}
    \caption{A general illustration of meta-learning for FSC. The training dataset is separated into \emph{episodes}, \ie{} small learning tasks. In FSC, an episode is an $N$-way $K$-shot classification problem. We illustrate here 3-way 2-shot episodes. The meta-learning algorithm learns each task at the \emph{inner-level} and between tasks at the \emph{outer-level}.}
    \label{fig:meta_episodes}
\end{figure}

Meta-learning, or \emph{learning to learn} \cite{thrun1998learning}, refers to the process of improving a learning algorithm over multiple learning \emph{episodes}, \ie{} a distribution of related tasks, as opposed to multiple data samples in conventional learning frameworks. Meta-learning is separated into an \emph{inner-level} and \emph{outer-level} learning phase. In the inner-level, a learning algorithm solves a \emph{task}, such as image classification, defined by a small dataset and an objective function. In the outer-level, the \emph{inner algorithm} is updated by an \emph{outer algorithm} to improve an \emph{outer objective}, such as generalization performance or learning speed, \rev{\ie{} the number of iterations and samples required for convergence to a good solution}.

\rev{More formally, we consider that the model learns over a distribution of \emph{tasks} $P(\Tcal)$, also called \emph{episodes} in practice, where each task $\Tcal_t$ corresponds to $K$ i.i.d. training observations called \emph{support set} $\Scal_t = \{(\rvx_{(t,j)}, \rvy_{(t,j)} \}_{j=1}^K \in \sX^K \times \sY^K$ and $q$ i.i.d. testing observations called \emph{query set} $\mathcal{Q}_t = \{(\rvx^\prime_{(t,j)}, \rvy^\prime_{(t,j)}) \}_{j=1}^q \in \sX^q \times \sY^q $, all generated by the corresponding distribution $\mu_t$ over the joint data space $\sX \times \sY$. During \emph{meta-training}, a task $\Tcal_t$ is sampled from $P(\Tcal)$ and the model is trained \emph{at the inner level} to learn the new task from the $K$ support samples. The model is then improved \emph{at the outer-level} by considering the test error on the unseen \emph{query samples} from $\Tcal_t$. In effect, the test error on episode $\Tcal_t$ serves as the meta-training error of the meta-learning process. For \emph{meta-testing}, new unseen episodes are sampled from $P(\Tcal)$ and the model's performance is measured on the query samples \emph{after} learning from the $K$ support samples. In practice meta-learning is usually done through a \emph{bi-level optimization} procedure. We give an illustration of meta-learning episodes for \emph{Few-Shot Classification (FSC)} in \cref{fig:meta_episodes}.
}

\begin{figure}
    \centering
    \includegraphics[width=\linewidth]{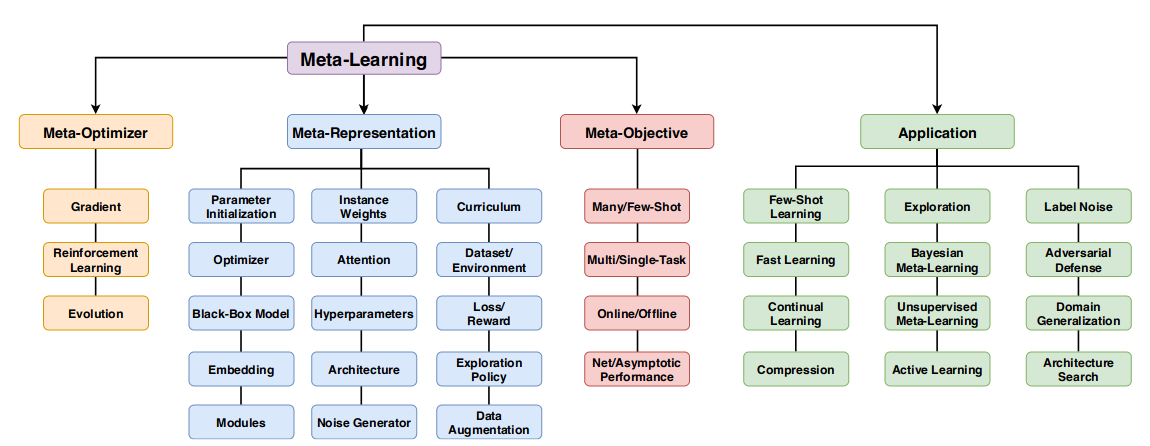}
    \caption{A taxonomy of Meta-Learning and its applications going beyond FSL, from \cite{hospedales2021meta}.}
    \label{fig:meta_taxo}
\end{figure}

While meta-learning has become a popular paradigm that successfully applied in areas spanning reinforcement learning \cite{Alet2020Meta-learning}, hyperparameter optimization \cite{franceschi2018bilevel} and neural architecture search \cite{liu2018darts}, we will focus on the few-shot image recognition applications \cite{finnModelAgnosticMetaLearningFast2017,snellPrototypicalNetworksFewshot2017}.
Meta-learning-based approaches are increasingly powerful to train CNNs on small datasets for many vision problems. By learning to solve many tasks with only few training samples, the meta-learned model can more easily adapt to a novel unseen task with few examples as well.  
We refer the reader to \cite{hospedales2021meta} for a more general presentation of deep meta-learning and its applications, with its respective taxonomy given in \cref{fig:meta_taxo}, and to \cite{wang2020generalizing} for a detailed review of the FSL applications. \rev{Note more details will be provided in \cref{chap:contrib_eccv} describing our first contribution on FSC. In the following sections, we detail related work applying meta-learning for Few-Shot Image Recognition and Object Detection.}

\subsection{Meta-Learning for Few-Shot Image Recognition}

\begin{figure}
    \begin{subfigure}[b]{0.5\linewidth}
        \centering
        \includegraphics[width=\linewidth]{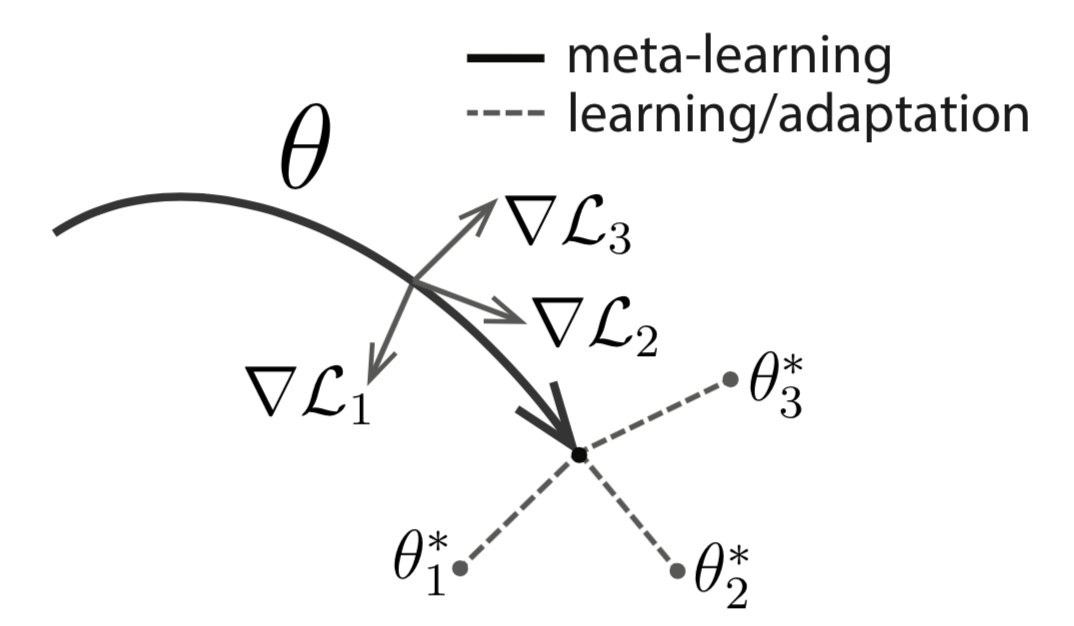}
        \caption{}
        \label{fig:maml_fig}
    \end{subfigure}
    \begin{subfigure}[b]{0.5\linewidth}
        \centering
        \includegraphics[width=0.85\linewidth]{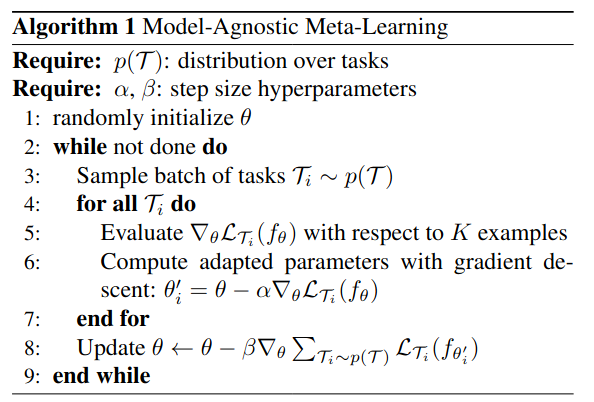}
        \caption{}
        \label{fig:maml_pseudo}
    \end{subfigure}
    \caption{\rev{Illustrations from \cite{finnModelAgnosticMetaLearningFast2017} of \textbf{(a)} the Optimization-based method \Maml{}, \textbf{(b)} the corresponding \emph{pseudo-code}.}}
    \label{fig:maml}
\end{figure}

\begin{figure}
    \centering
    \includegraphics[width=0.6\linewidth]{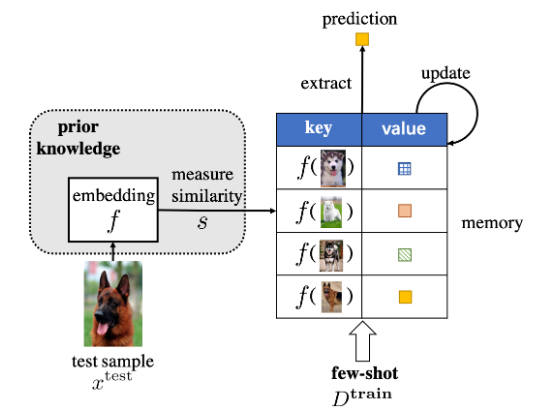}
    \caption{Illustration of memory-augmented neural network from \cite{wang2020generalizing}}
    \label{fig:meta_model}
\end{figure}

\begin{figure}
    \begin{subfigure}[b]{\linewidth}
        \centering
        \includegraphics[width=0.4\linewidth]{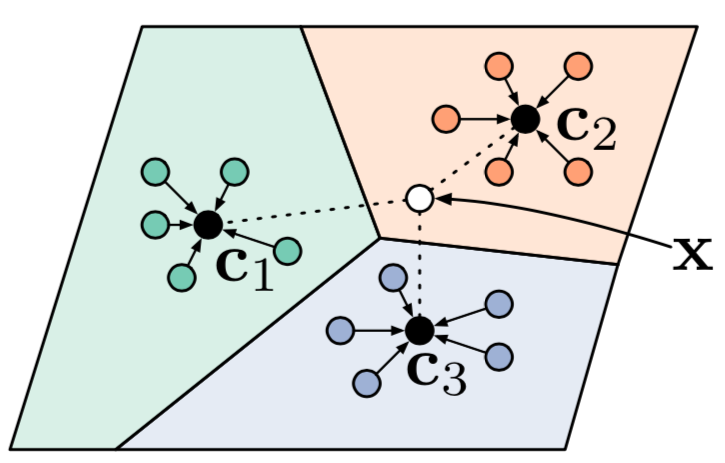}
        \caption{}
        \label{fig:proto_fig}
    \end{subfigure}
    \begin{subfigure}[b]{\linewidth}
        \centering
        \includegraphics[width=0.85\linewidth]{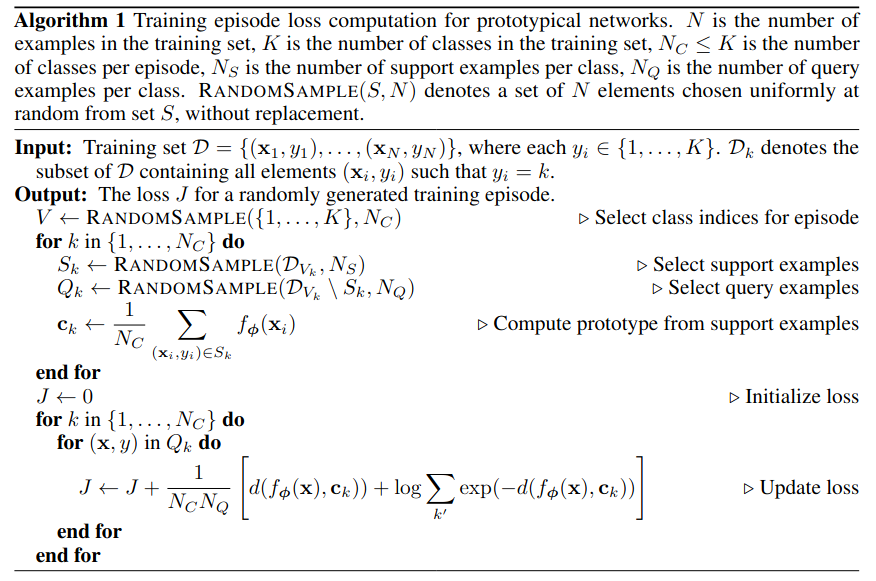}
        \caption{}
        \label{fig:proto_pseudo}
    \end{subfigure}
    \caption{\rev{Illustrations from \cite{snellPrototypicalNetworksFewshot2017} of \textbf{(a)} the Metric-based method \Proto{}, \textbf{(b)} the corresponding \emph{pseudo-code}.}}
    \label{fig:proto}
\end{figure}


The most common application of meta-learning in computer vision is multi-class FSC, using cross-entropy as the inner and outer loss functions \cite{snellPrototypicalNetworksFewshot2017,raviOPTIMIZATIONMODELFEWSHOT2017,liMetaSGDLearningLearn2017,antoniou2018how,yeHowTrainYour2021a,vinyalsMatchingNetworksOne2016,garcia2018fewshot,rusu2018meta,leeMetaLearningDifferentiableConvex2019,Yin2020Meta-Learning,bertinettoMetalearningDifferentiableClosedform2019,Raghu2020Rapid}. Meta-learning methods are usually separated in \emph{optimization-based} \cite{finnModelAgnosticMetaLearningFast2017,Raghu2020Rapid,DBLP:conf/iclr/OhYKY21}, \emph{model-based} \cite{pmlr-v48-santoro16,qiao2018few} or \emph{metric-based} \cite{snellPrototypicalNetworksFewshot2017,pmlr-v97-allen19b} categories \rev{that are briefly presented below}. 

\subsubsection{Optimization-based} \emph{Optimization-based} methods are literally solving the meta-learning objective as an optimization problem. The most representative example is \emph{Model Agnostic Meta-Learning (\Maml{})} \cite{finnModelAgnosticMetaLearningFast2017}, illustrated in \cref{fig:maml}, which aims to learn initialization parameters for novel tasks such that a small number of inner steps produces a classifier that performs well. In this case, the optimization is done end-to-end by gradient descent, but other alternatives also considers SVM \cite{leeMetaLearningDifferentiableConvex2019}, Ridge Regression \cite{bertinettoMetalearningDifferentiableClosedform2019} or recurrent networks \cite{raviOPTIMIZATIONMODELFEWSHOT2017}. However, optimization through multiple inner-steps leads to several computation and memory challenges which make the models difficult to train \cite{antoniou2018how} and to analyze theoretically \cite{DBLP:conf/iclr/FinnL18,sha_maml_2021}.

\subsubsection{Model-based} \emph{Model-based} methods, also called \emph{black-box} methods, wrap the inner learning step in a forward pass of a model designed specifically for learning with few training steps. Typical architectures include convolutional networks \cite{DBLP:conf/iclr/MishraR0A18}, hyper-networks \cite{qiao2018few,gidarisDynamicFewShotVisual2018} or memory-augmented neural networks \cite{DBLP:journals/corr/GravesWD14,pmlr-v48-santoro16,kaiserLearningRememberRare2017,munkhdalai2017meta,caiMemoryMatchingNetworks} which are illustrated in \cref{fig:meta_model}. While these methods enjoy simpler optimization without second-order gradients compared to optimization-based methods, they are usually less able to generalize to out-of-distributions tasks and asymptotically weaker as they struggle to embed large training set \cite{DBLP:conf/iclr/FinnL18}. 

\subsubsection{Metric-based} \emph{Metric-based} methods \cite{snellPrototypicalNetworksFewshot2017,pmlr-v97-allen19b,vinyalsMatchingNetworksOne2016,kochSiameseNeuralNetworks,sungLearningCompareRelation2018,DBLP:conf/cvpr/BateniGMWS20} perform non-parametric learning at the inner-level by comparing query points with support samples through similarity functions, and metric learning at the outer-level to find a good embedding space suited for comparison. The most popular representative is \emph{Prototypical Networks (\Proto{})} \cite{snellPrototypicalNetworksFewshot2017}, illustrated in \cref{fig:proto}, which uses Euclidean distance to compare query samples with class prototypes, represented by the average features of support examples for each class. The non-parametric inner-level computations make these methods simpler and faster to train than other methods, but they are also limited to FSL applications compared to other meta-learning methods that could be potentially applied for other settings.

\subsubsection{Datasets used}

\begin{figure}
    \centering
    \includegraphics[width=\linewidth]{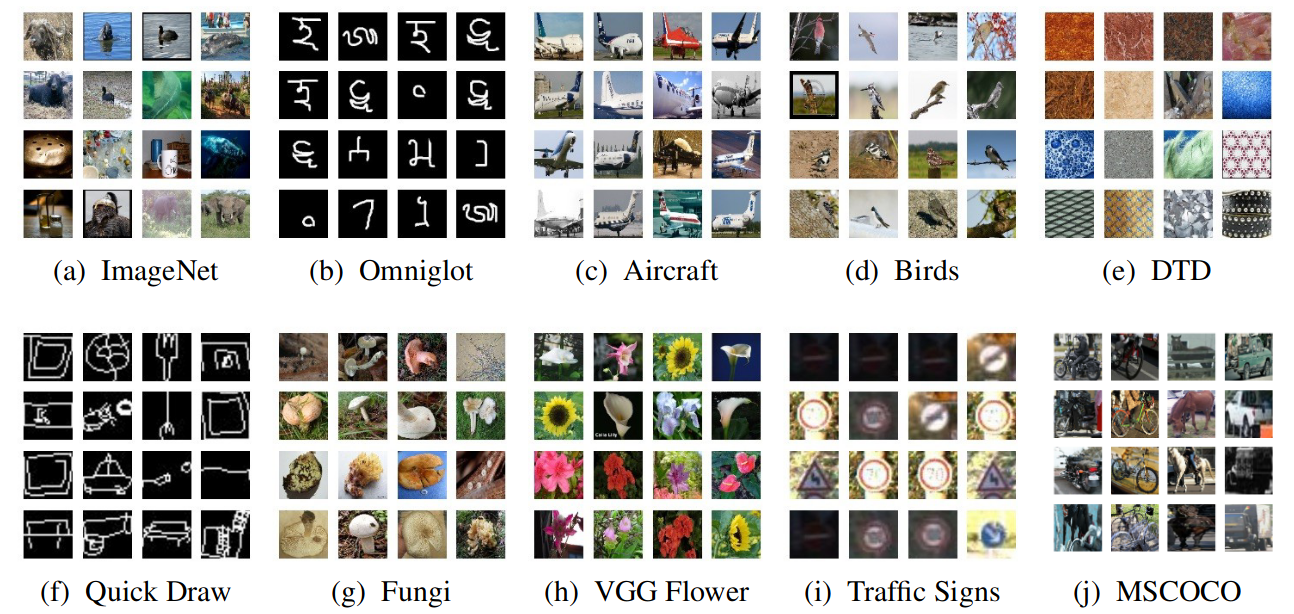}
    \caption{Training examples taken from the various datasets forming Meta-Dataset from \cite{DBLP:conf/iclr/TriantafillouZD20}.}
    \label{fig:meta_dataset}
\end{figure}

We present here common datasets used for FSC problems. The first dataset introduced for this problem is \emph{Omniglot} \cite{lakeHumanlevelConceptLearning2015}, a dataset of 20 instances of 1623 characters from 50 different alphabets. Each image was hand-drawn by different people using Amazon Mechanical Turk. This dataset is still popular in the literature for benchmarks and has the advantage of being lightweight, but it is now considered too simple for modern few-shot learning methods, with state-of-the-art performance of about 99\% of accuracy.
For more complicated datasets, recent methods are comparing performance on \emph{miniImageNet} \cite{raviOPTIMIZATIONMODELFEWSHOT2017}, \emph{tiered-ImageNet} \cite{renMetaLearningSemiSupervisedFewShot2018} or \emph{CIFAR-FS} \cite{bertinettoMetalearningDifferentiableClosedform2019} datasets. Both miniImageNet and tiered-ImageNet are constructed by taking subsets of the ImageNet dataset \cite{russakovskyImageNetLargeScale2015}. The former consists of 100 randomly chosen classes and 600 images for each class, while the latter is chosen such that training classes are semantically unrelated to testing classes to be more complicated, and is composed of 779 165 images divided into 608 classes.
\emph{CIFAR-FS}, which stands for \emph{CIFAR100 few-shot}, corresponds to classes from CIFAR-100 \cite{krizhevsky2009learning} randomly separated \rev{and with 600 images for each class}, to obtain a dataset harder than Omniglot but lighter than miniImageNet.
Recent work introduced \emph{Meta-Dataset} \cite{DBLP:conf/iclr/TriantafillouZD20} for a large-scale benchmark consisting of data from 10 different image datasets. Examples from each dataset are shown in \cref{fig:meta_dataset}. This benchmark aims to introduce realistic class imbalance with varying number of classes and training set sizes, to test the robustness across the spectrum of low-shot learning and domain discrepancy.


\vspace{\baselineskip}

Meta-learning approaches and the episodic formulation of the problem have steadily improved performance in FSC compared to early methods \cite{finnModelAgnosticMetaLearningFast2017,kochSiameseNeuralNetworks,vinyalsMatchingNetworksOne2016}, leading to the recent interest in this line of work. However, the benefit over simpler methods based on transfer learning is also being challenged \cite{chenCloserLookTraining2020,tian2020rethinking,DBLP:conf/iccv/Chen00D021}, as we will see in \cref{sec:related_work_transfer_learning}.

\subsection{Meta-Learning for Few-Shot Object Detection}\label{sec:fsod_meta}

\begin{figure}
    \centering
    \includegraphics[width=\linewidth]{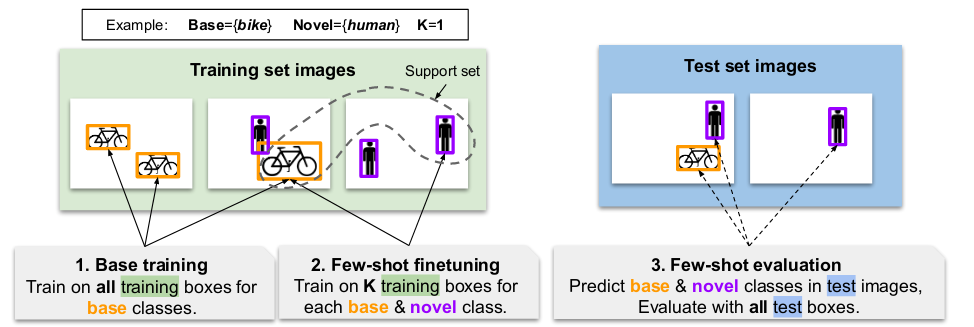}
    \caption{The Few-shot object detection protocol proposed by \cite{DBLP:conf/iccv/KangLWYFD19}, as illustrated by \cite{DBLP:journals/corr/abs-2110-14711}. The model is trained on \emph{base classes}, then finetuned on a combination of \emph{base and novel classes} from a limited \emph{support set}. During testing, the model is evaluated on \emph{base and novel class detection}.}
    \label{fig:fsod_protocol}
\end{figure}

\begin{figure}
    \centering
    \includegraphics[width=0.7\linewidth]{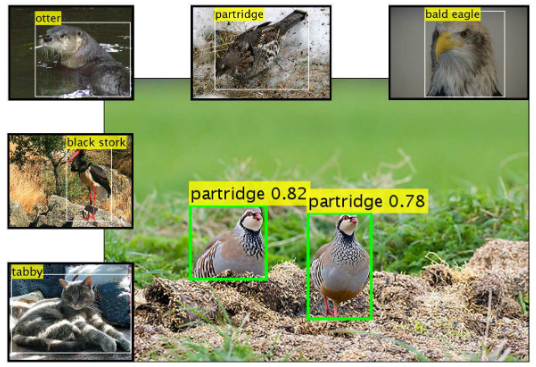}
    \caption{One-shot Object Detection example from \cite{DBLP:conf/cvpr/KarlinskySHSAFG19}. \emph{Support examples} representing novel classes to detect are represented in the surrounding images. The \emph{query image} is in the center, with detection results for the \emph{partridge} category, one of the novel class.}
    \label{fig:fsod_example}
\end{figure}

\emph{Few-Shot Object Detection (FSOD)} is the task of learning to detect new categories of objects in an image using only few training examples per class. The categories of objects are separated into two disjoint sets: the \emph{base classes}, for which we have access to many training examples, and \emph{novel classes}, for which we only have few examples, or shots, per class. We give an illustration of the framework introduced by \cite{DBLP:conf/iccv/KangLWYFD19} in \cref{fig:fsod_protocol}. With Meta-Learning, methods are trained on \emph{query images}, \ie{} the image to detect objects from, using annotated \emph{support images}, \ie{} reference examples of each class to detect. Support images generally have a single bounding-box around the object to detect and are randomly sampled from all base classes during training. Then, a predefined set of base and novel images are used during a \emph{finetuning phase} and in the \emph{evaluation phase}. An example of evaluation in one-shot object detection with support and query images is given in \cref{fig:fsod_example}. We refer the reader to \cite{DBLP:journals/corr/abs-2110-14711} for a more detailed survey on FSOD.

\subsubsection{Metric-based methods}

Similarly to FSC, meta-learning methods for FSOD are based on \emph{metric learning}. They compute prototypes for each object categories using support examples, and then classifies object proposals according to a similarity function to each representative. The prototypes can be obtained through linear operations \cite{DBLP:conf/mm/WuSH20,DBLP:conf/cvpr/FanYLOGXL20,DBLP:conf/aaai/HanHMHC22, DBLP:conf/eccv/XiaoM20,DBLP:conf/iccv/QiaoZLQWZ21} or learned with a separated module \cite{DBLP:conf/cvpr/KarlinskySHSAFG19,DBLP:conf/iccv/KangLWYFD19,DBLP:conf/cvpr/LiL21a,DBLP:conf/iccv/YanCXWLL19}, and the similarity can be computed through an explicit distance function \cite{DBLP:conf/cvpr/KarlinskySHSAFG19,DBLP:conf/mm/WuSH20,DBLP:conf/iccv/QiaoZLQWZ21} or from an attention operation \cite{DBLP:conf/iccv/KangLWYFD19,DBLP:conf/cvpr/LiL21a,DBLP:conf/iccv/YanCXWLL19,fan2020few,DBLP:conf/cvpr/FanYLOGXL20,DBLP:conf/aaai/HanHMHC22,DBLP:conf/eccv/XiaoM20}. Even though the large majority of the methods presented could technically be used \emph{without finetuning}, by directly conditioning on the support examples at few-shot evaluation time, in practice, most works find it beneficial to finetune, and in fact many do not even report numbers without finetuning. \rev{It shows the necessity found in finetuning, and it is thus difficult to really assess the benefit of the methods without finetuning on target data.}

\subsubsection{Datasets used}

\begin{table}
    \centering
    \begin{tabular}{@{}lcc@{}}
        \toprule
        Statistics & Train & Test \\
        \midrule
        No. Class & 800 & 200 \\
        No. Image & 52350 & 14152 \\
        No. Box & 147489 & 35102 \\
        Avg No. Box / Img & 2.82 & 2.48 \\
        Min No. Img / Cls & 22 & 30 \\
        Max No. Img / Cls & 208 & 199 \\
        Avg No. Img / Cls & 75.65 & 74.31 \\
        Box Size & [6, 6828] & [13, 4605] \\
        Box Area Ratio & [0.0009, 1] & [0.0009, 1] \\
        Box W/H Ratio & [0.0216, 89] & [0.0199, 51.5] \\
        \bottomrule
    \end{tabular} 
    \caption{\emph{FSOD-dataset} summary from \cite{fan2020few}.}
    \label{tab:fsod_dataset_summary}     
\end{table}   

We describe here the dominant FSOD benchmarks used in the literature. PASCAL VOC \cite{everingham2010pascal} is one of the most popular smaller benchmarks for traditional object detection, and has naturally been adapted for FSOD \cite{DBLP:conf/iccv/KangLWYFD19}. The object categories are separated into \emph{15 base classes} and \emph{5 novel classes}, and three different splits between classes are usually considered, denoted \emph{splits 1, 2 and 3}. For \emph{base training}, training and validation images of the \emph{base classes} from (PASCAL) VOC2007 and (PASCAL) VOC2012 are used (\emph{trainval} sets). For \emph{finetuning}, a fixed subset of the VOC2007 and VOC2012 \emph{trainval} sets is taken as the support set. The benchmarks consider between 1 and 10 shots which correspond to between 15 and 150 base bounding boxes and between 5 and 50 novel bounding boxes. For evaluation, about 5k images from VOC2007 test set are used.
Another popular dataset for object detection, \emph{Microsoft Common Object in COntext (COCO)} \cite{lin2014microsoft}, has also been adapted to FSOD \cite{DBLP:conf/iccv/KangLWYFD19}. The object categories are split between \emph{20 novel classes}, shared with PASCAL VOC, and \emph{60 base classes}. A subset of 5k images from COCO validation set are used for evaluation, denoted \emph{val5k}, and the remaining training and validation images of COCO are used for training and finetuning, denoted \emph{trainvalno5k}.
A novel dataset, \emph{FSOD} \cite{fan2020few}, that we will refer to \emph{FSOD-dataset} to avoid confusion, has been introduced specifically for the FSOD problem. The dataset contains 1000 categories with 800 training classes and 200 testing classes, unambiguously separated. The total number of images is around 66k with about 182k bounding boxes. Detailed statistics are given in \cref{tab:fsod_dataset_summary}.

\vspace{\baselineskip}

Similarly to FSC, recent work \cite{DBLP:conf/icml/WangH0DY20,bar2022detreg} has shown that simply finetuning a classic detection model through supervised learning on the few-shot dataset after a pretraining phase on a bigger dataset achieves competitive results, compared with more complicated methods tailored for few-shot learning. Furthermore, even with meta-learning, the models learned are mainly doing \emph{feature reuse} between tasks rather than a \emph{rapid learning} of each task \cite{Raghu2020Rapid}. These papers reignited research on \emph{Transfer Learning} with limited labeled data on the \emph{target dataset}, both for FSC and FSOD.

\section{Transfer Learning by Pretraining}\label{sec:related_work_transfer_learning}


Transfer Learning is one of the oldest practice in deep learning to deal with novel unseen tasks with limited labeled data \cite{pan2009survey}. \rev{We focus in this section more specifically on pretraining-based methods,} a classical learning paradigm which aims to reuse features learned on an \emph{extensive source dataset} to a \emph{potentially limited target dataset}. 

\begin{figure}
    \centering
    \includegraphics[width=0.95\linewidth]{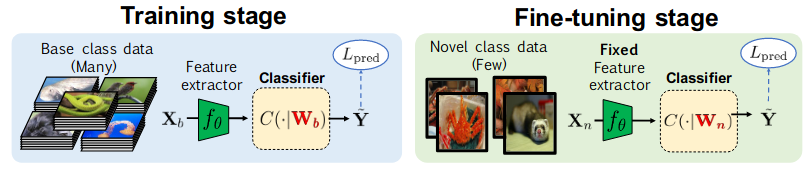}
    \caption{Illustration of the general process of Transfer Learning from \cite{chenCLOSERLOOKFEWSHOT2019}. The training can be separated in \emph{pretraining} and \emph{fine-tuning} stages, in combination with other techniques.}
    \label{fig:transfer_fsl}
\end{figure}

\subsection{Pretraining and Fine-tuning for Few-Shot Learning}

Transfer learning can be divided into a first \emph{pretraining stage}, in which we aim to learn a general representation of the data encoded in a \emph{feature extractor}, and then a \emph{fine-tuning stage}, during which the model is fully or partially updated using the few labeled data available for the target task. The general process in the case of FSC is illustrated in \cref{fig:transfer_fsl}. Even though early works on FSC had shown large improvements against this simple baseline \cite{vinyalsMatchingNetworksOne2016,raviOPTIMIZATIONMODELFEWSHOT2017,finnModelAgnosticMetaLearningFast2017}, later works have found that they achieve competitive performance when compared on common grounds, with same training details and also with deeper networks \cite{chenCLOSERLOOKFEWSHOT2019,DBLP:journals/corr/abs-1910-00216,wangSimpleShotRevisitingNearestNeighbor2019}, while others have refined the fine-tuning stage specifically for FSL \cite{gidarisDynamicFewShotVisual2018,DBLP:conf/cvpr/LifchitzAPB19,dhillonBaselineFewshotImage2020,tian2020rethinking,DBLP:conf/aaai/ShenLQSC21}. Several recent works are initializing meta-learning with a pretraining stage, showing that they also benefit from strong pretrained feature extractors \cite{rusu2018meta,ye2020fewshot,yeHowTrainYour2021a}.
In \cite{tian2020rethinking}, authors have also investigated using different type of pretrained models, from classical supervised pretraining, to early \emph{self-supervised pretraining} methods \cite{he2020momentum,DBLP:conf/eccv/TianKI20}, and obtained similar results between them, showing the potential of strong pretraining for FSL. More recent works in self-supervised learning are evaluating the fine-tuning performance on limited labeled data \cite{chen2020big}. Self-supervised learning have also been used in parallel during the fine-tuning stage to further boost FSL \cite{DBLP:conf/iccv/GidarisBKPC19}. Using an \emph{unsupervised pretraining stage} would also alleviate even more the need of labels.

\subsection{Self-supervised Pretraining}


Self-Supervised Learning (Self-SL) is an unsupervised learning procedure in which the data provides its own supervision to learn a good representation. The representation learned can then be transferred and fine-tuned on various downstream tasks such as classification or object detection. Several works have shown that a self-supervised pretrained model can outperform its supervised counterpart \cite{caron2020unsupervised,grill2020bootstrap,caron2021emerging}. Self-SL is based on \emph{pretext tasks}, \ie{} unsupervised tasks that apply \emph{transformations} to the data, or part of it, in order to provide its supervision.

\subsubsection{Early pretext tasks}

Early works proposed different pretext tasks from different set of augmentations to learn good data representations, such as \emph{instance discrimination} \cite{DBLP:journals/pami/DosovitskiyFSRB16}, \emph{patch localization} \cite{DBLP:conf/iccv/DoerschGE15}, \emph{colorization} \cite{DBLP:conf/eccv/ZhangIE16}, \emph{solving jigsaw puzzle} \cite{noroozi2016unsupervised}, \emph{counting} \cite{DBLP:conf/iccv/NorooziPF17} or \emph{angle rotation prediction} \cite{gidaris2018unsupervised}. These tasks are currently outperformed by \emph{Contrastive Learning}, which represents now the majority of recent work in Self-SL.

\subsubsection{Contrastive Learning}

Contrastive Learning is a learning paradigm relying on \emph{instance discrimination} using a pair of  \emph{positive views} from the same input contrasted with all other negative instances in the batch of images, called \emph{negatives} \cite{oord2018representation, wu2018unsupervised, he2020momentum, chen2020simple, misra2020self, grill2020bootstrap, caron2020unsupervised, chen2020big, chen2021empirical, denize2021similarity}. The popular InfoNCE loss function \cite{oord2018representation} is widely used for contrastive learning:

\begin{equation}
    \Lcal_{\text{InfoNCE}} = - \frac{1}{N} \sum_{i=1}^N \log \left( \frac{\exp(\rvz_i \cdot \rvz^\prime_i / \tau)}{\sum^N_{j=1} \exp (\rvz_i \cdot \rvz^\prime_j / \tau)} \right).
\end{equation}

\rev{This loss function computes the pairwise similarity scaled by a temperature parameter $\tau$ from a batch of feature vectors $\mathcal{B} = \{ (\rvz_i,\rvz^\prime_i), i \in N\}$, and maximizes agreement between positive pairs of features $(\rvz_i, \rvz^\prime_i)$ while pushing negative pairs $\{(\rvz_i, \rvz^\prime_j), j \neq i\}$ away.}
However, InfoNCE \cite{oord2018representation} requires a large amount of negative instances to be effective \cite{pmlr-v119-wang20k}. While most methods use a huge batch size \cite{chen2020simple,chen2020big,chen2021empirical,DBLP:conf/nips/Tian0PKSI20}, the popular MoCo \cite{he2020momentum,DBLP:journals/corr/abs-2003-04297} maintains a memory buffer of representations to keep a high number of negatives along with a small batch size. 
\emph{Hard negatives}, \ie{} negative samples that are difficult for the model to separate from positive ones, are the most important for contrastive learning \cite{cai2020all}, but they are also potentially harmful to the training because of the \emph{class collision} problem \cite{cai2020all,DBLP:conf/iclr/0005W0u21,DBLP:conf/nips/ChuangRL0J20}, where semantically close instances can be used as negatives in the loss computation, which damages the quality of the representation learned. Several samplers of negatives have been proposed to alleviate this problem \cite{DBLP:conf/iccv/DwibediATSZ21,DBLP:conf/iccv/WangWWTL21,DBLP:conf/cvpr/Hu00Q21}. Other popular methods are based on a siamese architecture, \ie{} using two copies of the model that are jointly updated, to perform contrastive learning without the use of negatives by ensuring consistency between the outputs of the two models \cite{grill2020bootstrap, caron2020unsupervised,caron2021emerging,DBLP:conf/icml/ZbontarJMLD21,DBLP:conf/iclr/BardesPL22}. Recent works \cite{zheng2021ressl,denize2021similarity} have also tackled the class collision problem by introducing the \emph{relational aspect} between instances.


\subsection{Pretraining and Fine-tuning for Object Detection}

\begin{figure}
    \centering
    \includegraphics[width=\linewidth]{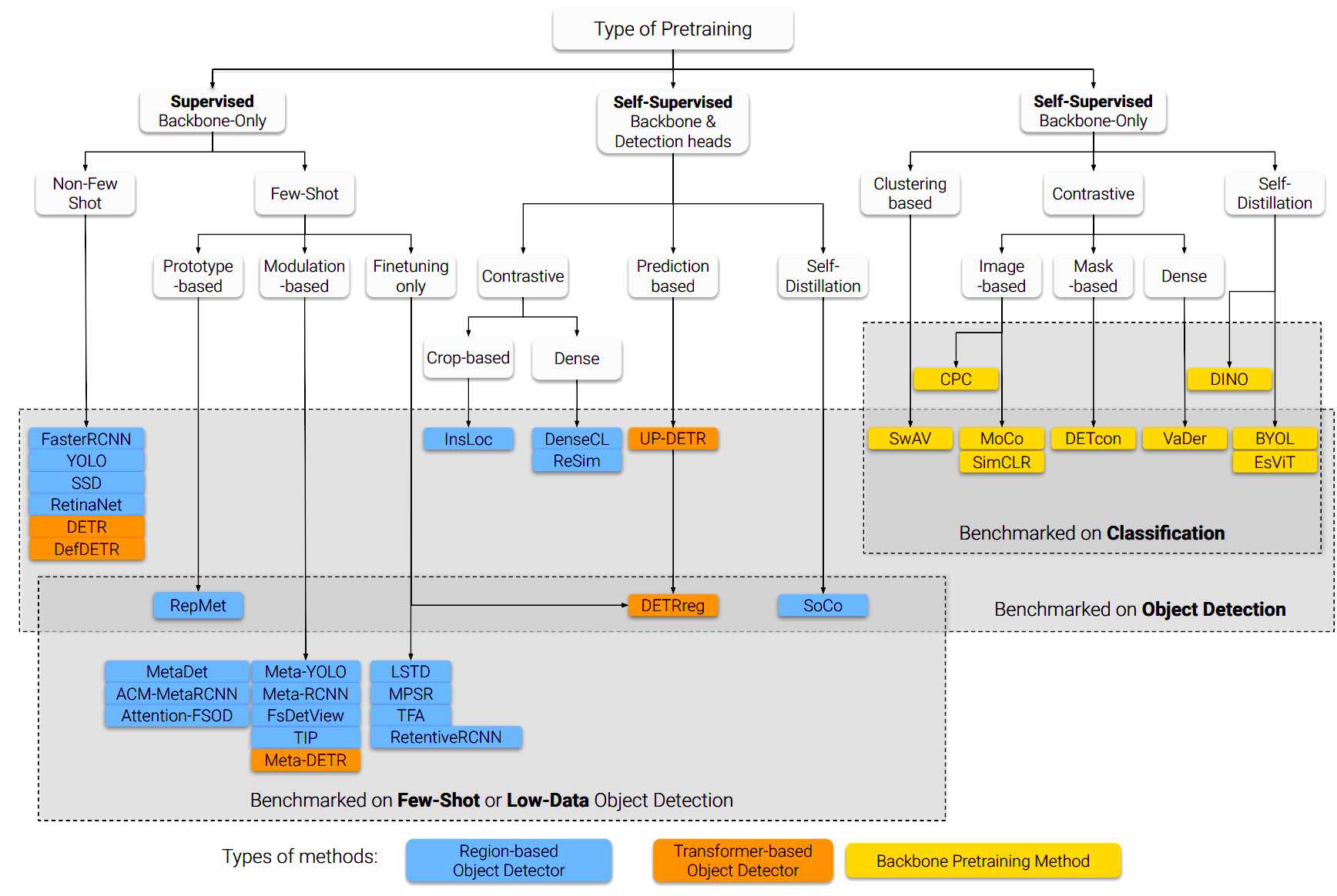}
    \caption{A taxonomy of Few-shot Object Detection methods through the lens of pretraining, with Self-supervised learning, Meta-learning and Transfer learning methods \cite{DBLP:journals/corr/abs-2110-14711}.}
    \label{fig:taxo_fsod}
\end{figure}

\rev{Using pretrained models then finetuning is a popular choice to deal with scarce data settings in Object Detection, since the models can be difficult to train from a random initialization. \cref{fig:taxo_fsod} represents a taxonomy of pretraining methods for Few-Shot Object Detection. }

\subsubsection{The use of pretraining in Object Detection}\label{sec:pretrain_od}

\paragraph{Backbone pretraining}
Object detectors trained in practice are rarely starting from a random initialization. The backbone of the models are usually pretrained on ImageNet \cite{russakovskyImageNetLargeScale2015}, before fine-tuning on the target dataset and detection task \cite{ren2015faster,He_2019_ICCV}. Until recently, the standard approach was to pretrain the backbone in fully supervised learning. 
While pretraining from a \emph{large general dataset} allows for faster convergence during fine-tuning on a similar dataset, both in terms of number of samples and iterations, it is less helpful for the localization task \cite{He_2019_ICCV}. 
Recent \emph{unsupervised} approaches have started investigating pretraining tasks tailored for object detection by imposing local consistency, either at the \emph{pixel} or \emph{region-level}, with the goal of helping the model to locate. These methods respectively propose to match in the representation space the features corresponding to the same location in the input space \cite{o2020unsupervised, xie2021propagate, wang2021dense}, or ensure local consistency between features from regions in the image \cite{roh2021spatially, yang2021instance, xiao2021region}. 

\begin{figure}
    \centering
    \includegraphics[width=\linewidth]{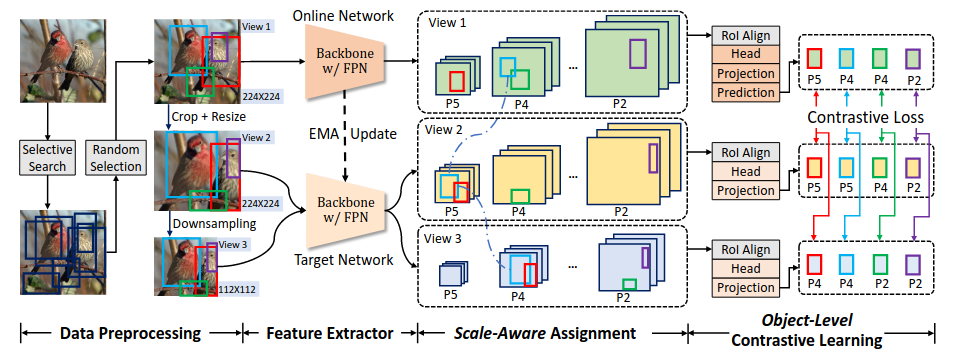}
    \caption{\rev{Illustration of SoCo \cite{wei2021aligning}. The pretraining is specialized to two-stage architectures and use a contrastive learning objective.}}
    \label{fig:soco}
\end{figure}

\begin{figure}
    \centering
    \includegraphics[width=\linewidth]{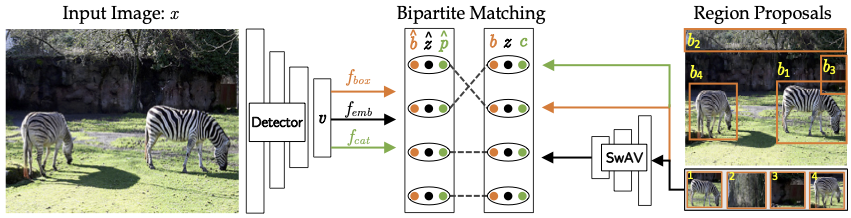}
    \caption{\rev{Illustration of DETReg \cite{bar2022detreg}. The detector must predict region proposals ($f_{box}(v)=\hat{b}$) given by an unsupervised algorithm \cite{uijlings2013selective}, associated object embeddings ($f_{emb}(v) = \hat{z}$) obtained by an existing pretrained self-supervised backbone \cite{caron2020unsupervised}, and object scores ($f_{cat}(v) = \hat{p}$), fixed to $1$.}}
    \label{fig:detreg}
\end{figure}

\paragraph{Pretraining the overall detection model}
Due to architecture mismatch between detection and classification models, only the backbone feature extractor can be initialized with pretraining while the detection-specific parts of the models must be learned from scratch. Few approaches in the literature have tackled this problem, by pretraining detection-specific parts along with the backbone \cite{wei2021aligning}, or independently \cite{dai2021up,bar2022detreg}. SoCo \cite{wei2021aligning}, illustrated in \cref{fig:soco}, proposes a contrastive pretraining strategy inspired by BYOL \cite{grill2020bootstrap}, for the \emph{overall} detection model but specialized to convolutional two-stage detectors. UP-DETR \cite{dai2021up} and DETReg \cite{bar2022detreg}, \rev{the latter more recent method illustrated in \cref{fig:detreg},} use a fixed pretrained backbone to extract features from patches and pretrain the detector \emph{independently} by locating and reconstructing the features. Initializing detectors from an overall pretraining leads to even faster convergence and stronger performance with limited labels \cite{wei2021aligning,bar2022detreg}.

\subsubsection{Fine-Tuning Detectors}

\begin{figure}
    \centering
    \includegraphics[width=\linewidth]{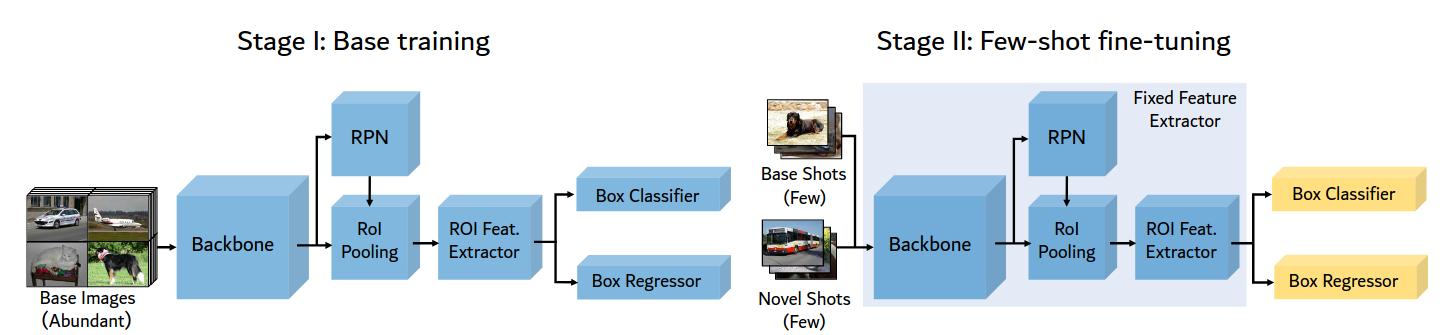}
    \caption{Fine-tuning for FSOD in two-stage, from \cite{DBLP:conf/icml/WangH0DY20}. Pretraining and fine-tuning detection models represents a strong baseline for FSOD, compared to more complex approaches presented in \cref{sec:fsod_meta}.}
    \label{fig:fsod_tfa}
\end{figure}

\paragraph{Fine-tuning for Few-Shot Object Detection} Similarly to FSC, recent work have shown that simply fine-tuning traditional object detectors with only minor architecture modifications achieves strong results on FSOD \cite{DBLP:conf/icml/WangH0DY20}. This fine-tuning approach is illustrated in \cref{fig:fsod_tfa}. Following works have then further refined the fine-tuning stage for FSOD \cite{DBLP:conf/eccv/WuL0W20,DBLP:conf/cvpr/FanMLS21}.

\paragraph{Fine-tuning for Low-shot Object Detection} Instead of having a distinction between \emph{base classes} with many examples, and \emph{novel classes} with few shots, \emph{Low-shot Object Detection} (LSOD) assumes that the number of examples for \emph{all} classes is limited. \emph{Data scarcity} is generally simulated by considering a fraction of the labels of traditional object detection datasets. The most prominent benchmark is \emph{mini}COCO, which uses random subsets of 1\%, 5\% or 10\% of the full COCO dataset. Several pretraining methods for object detection report numbers in this setting to compare performance, since comparison of transfer learning on the full COCO dataset may be of limited significance due to the large-scale supervision available \cite{yang2021instance,wang2021dense,wei2021aligning,bar2022detreg}.

\vspace{\baselineskip}

Unsupervised pretraining methods presented in this section have the advantage of leveraging large unlabeled data available by providing an initialization of the models more adapted to the target task \cite{reed2022self}. One can further leverage these unlabeled data after fine-tuning, through \emph{semi-supervised learning}.

\section{Semi-supervised learning \rev{with few annotations}}\label{sec:related_work_semi_sup}

\begin{figure}
    \centering
    \includegraphics[width=\linewidth]{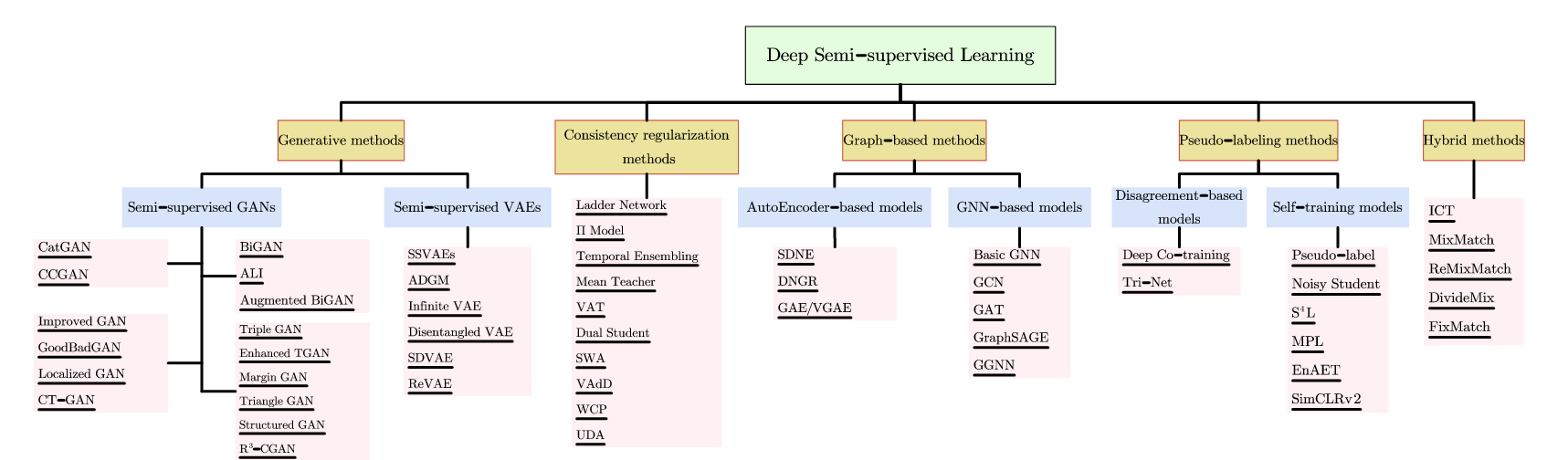}
    \caption{A general taxonomy of semi-supervised methods with deep learning, from \cite{DBLP:journals/corr/abs-2103-00550}.}
    \label{fig:dssl_taxo}
\end{figure}

Semi-supervised learning is a learning paradigm aiming to train models by using both labeled and unlabeled data. It can improve performance by using additional unlabeled samples compared to supervised learning algorithms which can only use labeled data, alleviating the need for numerous labels \cite{DBLP:conf/nips/OliverORCG18}. One can easily extend fully supervised or unsupervised algorithms to obtain semi-supervised ones. In this section, we will focus on presenting \emph{few-annotation learning} applications for \emph{image classification} and \emph{object detection}, and limit the discussion to \emph{consistency regularization} and \emph{pseudo-labeling} as they represent the most successful and promising approaches in semi-supervised learning. We refer the reader to \cite{DBLP:journals/ml/EngelenH20,DBLP:journals/corr/abs-2103-00550} for a general overview of the semi-supervised learning paradigm, with a taxonomy of deep learning-based methods given in \cref{fig:dssl_taxo}.

\subsection{Few-Annotation Classification}

In order to present and explain the key concepts in semi-supervised learning, we begin by the image classification problem. We are interested specifically in Few-Annotation Classification (FAC), in which the model has only access to few labeled examples along with a large number of unlabeled data.

\subsubsection{Consistency Regularization}

\emph{Consistency regularization methods} are based on a prior assumption that realistic perturbations of data should not change the output of the model \cite{DBLP:conf/nips/OliverORCG18}. These methods follow the general structure of \emph{Teacher-Student}: the model learns as the \emph{student}, and generates targets as the \emph{teacher} to compute a \emph{consistency regularization term} in the final loss function.
\rev{The consistency constraint can be formally written as:}

\begin{equation}
    \E_{\rvx \in \rmX} \left[ \Rcal \left(f_\theta(\rvx), f^\prime_{\theta^\prime}(t(\rvx))\right) \right],
\end{equation}

\noindent \rev{where $f_\theta(\rvx)$ are the predictions of the student model for input $\rvx$, $f^\prime_{\theta^\prime}(t(\rvx))$ are the consistency target of the teacher model $f^\prime$ with parameters $\theta^\prime$ for the same input transformed through $t$, and $\Rcal$ measures the distance between the two vectors which is usually MSE or KL-divergence.}

Consistency constraints can be considered at the input-level, by adding perturbations to input examples \cite{DBLP:conf/nips/RasmusBHVR15,DBLP:conf/icml/PezeshkiFBCB16,DBLP:conf/nips/SajjadiJT16,miyato2018virtual,DBLP:conf/nips/XieDHL020}, or at the network-level, by averaging the weights \cite{tarvainen2017mean,DBLP:conf/uai/IzmailovPGVW18,DBLP:conf/iccv/KeWYRL19}, the predictions \cite{LaineA17} or even by dropping connections \cite{DBLP:conf/aaai/ParkPSM18,DBLP:conf/cvpr/ZhangQ20}.  

\subsubsection{Self-Training}

\emph{Self-Training} differs from consistency regularization in that the former relies on \emph{high confident predictions} and the latter on consistency between \emph{rich transformations}. Self-Training methods can be based on \emph{disagreement} \cite{DBLP:journals/kais/ZhouL10} of predictions on unlabeled instances, from different views of the data \cite{DBLP:conf/eccv/QiaoSZWY18} or between different models \cite{DBLP:conf/ijcai/ChenWGZ18}, or based on \emph{Pseudo-labeling} by leveraging the model's predictions on unlabeled data \cite{DBLP:conf/nips/GrandvaletB04,lee2013pseudo,DBLP:conf/cvpr/XieLHL20,DBLP:conf/iccv/BeyerZOK19,chen2020big,DBLP:journals/tip/WangKLQ21,DBLP:conf/cvpr/PhamDXL21}. Self-training has recently been at the heart of a lot of improvements in unsupervised training with the advent of large-scale training and architectures \cite{chen2020big,caron2021emerging}.

\subsubsection{Hybrid Methods}

\begin{figure}
    \centering
    \includegraphics[width=0.9\linewidth]{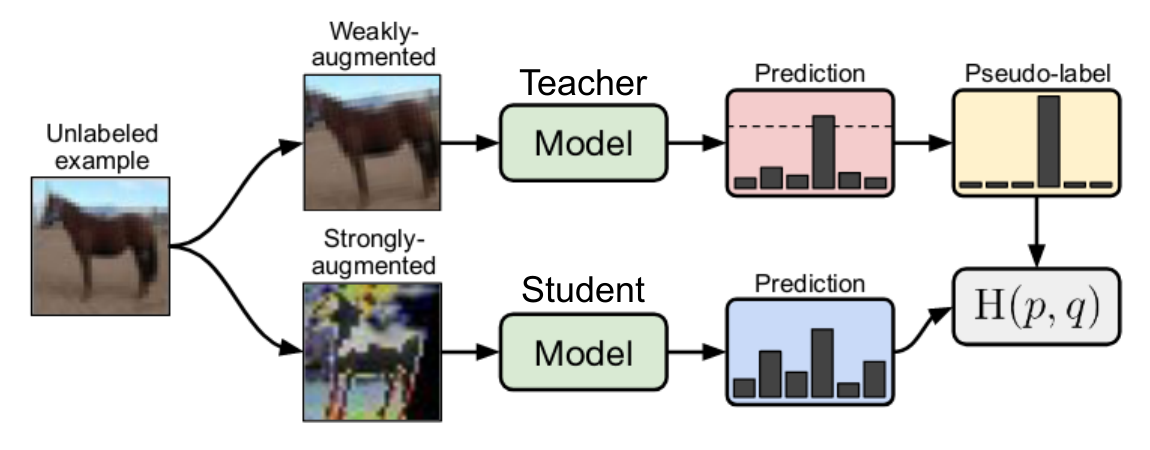}
    \caption{Illustration from FixMatch \cite{sohn2020fixmatch}, a popular hybrid method taking inspiration from both \emph{pseudo-labeling} and \emph{consistency regularization}. Two differently augmented views are created for each unlabeled example, with one view going to the model as the teacher to create a pseudo-label that will be used to supervise the model as the student with the other view.}
    \label{fig:fixmatch}
\end{figure}

\emph{Hybrid methods} \cite{DBLP:conf/ijcai/VermaLKBL19,berthelot2019mixmatch,DBLP:conf/iclr/LiSH20,DBLP:conf/iclr/BerthelotCCKSZR20,sohn2020fixmatch} combine ideas from both self-training and consistency regularization above-mentioned methods. They use input and network transformations along with entropy minimization \cite{berthelot2019mixmatch}, distribution alignment \cite{DBLP:conf/iclr/LiSH20,DBLP:conf/iclr/BerthelotCCKSZR20,DBLP:conf/ijcai/VermaLKBL19} or pseudo-labeling \cite{sohn2020fixmatch}. These methods achieve currently state-of-the-art performance on various benchmarks in FAC, using as low as \emph{a single} labeled example for each class with FixMatch \cite{sohn2020fixmatch} for which an illustration is given in \cref{fig:fixmatch}.


\subsection{Semi-Supervised Object Detection for Few-Annotation Learning}

\begin{figure}
    \centering
    \includegraphics[width=0.9\linewidth]{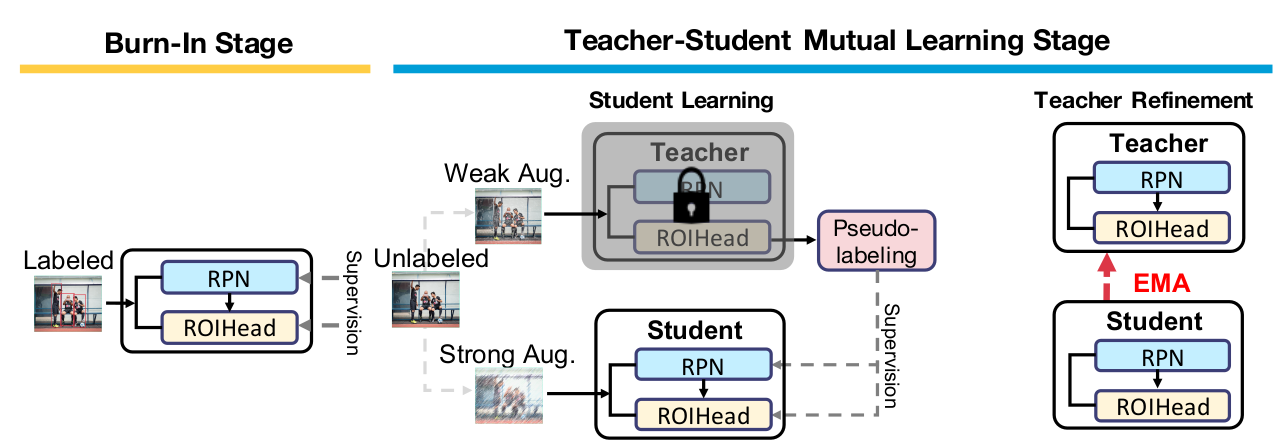}
    \caption{Illustration of \emph{Unbiased Teacher} \cite{liu2021unbiased}, a popular approach in SSOD inspired by hybrid methods. Semi-supervised learning is initialized after a fully supervised training (\emph{Burn-In stage}). The model is based on a \emph{Teacher-Student} architecture, using two \emph{differently} augmented views of the data, and \emph{pseudo-labeling}. The teacher model is an independent copy of the student model, updated throughout training by the \emph{Exponential Moving Average (EMA)} \cite{tarvainen2017mean} of the student's weights.}
    \label{fig:ubt}
\end{figure}

First \emph{Semi-Supervised Object Detection} (SSOD) methods were mostly based on \emph{self-training} \cite{DBLP:conf/cvpr/TangWGDGC16,DBLP:conf/cvpr/WangYZ0L18,DBLP:conf/iccv/GaoWDLN19}, or \emph{consistency regularization} \cite{jeong2019consistency}, before combining both.
More recent methods \cite{sohn2020simple,liu2021unbiased,tang2021humble,xu2021end} are mainly taking inspiration from \emph{hybrid methods} presented above, and focusing on \emph{two-stage convolutional detectors}, presented in \cref{sec:learning_od}. They achieve strong performance in FAL for object detection by leveraging additional unlabeled examples along with LSOD training data. A popular recent approach is illustrated in \cref{fig:ubt}. 

\section{\rev{Towards the contributions of this thesis}}\label{sec:related_work_discussion}

As can be seen throughout this chapter, the problem of learning with limited labels in Deep Learning spans various topics, and can be tackled from different ways. The most obvious option is to work on the data itself to artificially increase its diversity, but it has its limits. The main research direction is to refine training schemes to deal with limited data or labels. \emph{Meta-learning} with \emph{episodic training}, is a popular framework for FSL and originally designed for this setting. Besides, it has found applications for other problems such as Reinforcement Learning \cite{finnModelAgnosticMetaLearningFast2017} or Continual Learning \cite{gupta2020look}. However, a strong theoretical justification of meta-learning algorithms matching with empirical results is still lacking. Inspired by recent theoretical results in \emph{multitask representation learning} \cite{duFewShotLearningLearning2020,tripuraneniProvableMetaLearningLinear2020,pmlr-v139-wang21ad} published at the beginning of this thesis, our first research direction investigate the link between meta-learning and multitask representation learning algorithms to leverage strong learning bounds in this setting. Then as novel works challenging meta-learning in favor of fine-tuning approaches was published \cite{chenCLOSERLOOKFEWSHOT2019,Raghu2020Rapid,DBLP:conf/icml/WangH0DY20}, we focus on leveraging unlabeled data along with few annotated data in object detection through pretraining and semi-supervised learning \rev{to develop the other contributions of this thesis}.    









%% file: chapters/meta_mtr.tex
\chapter{Improving Few-Shot Classification with Meta-Learning through Multi-Task Learning}\label{chap:contrib_eccv}

\minitoc

\begin{abstract}
    \textit{
        In this chapter, we \rev{introduce our first contribution which focuses on} Meta-Learning algorithms for Few-Shot Learning (FSL), and more specifically their link with Multi-Task Representation Learning (MTR) algorithms. We recall the positioning in the general training pipeline in \cref{fig:position_eccv_chap}. As recent theoretical works in MTR theory have achieved efficient learning bounds for FSL, our goal is to leverage their theoretical settings to improve and better understand popular Meta-learning algorithms in practice. The organization of the chapter is as follows. We present the existing theoretical results for the MTR problem with their corresponding assumptions, and introduce considered meta-learning algorithms in \cref{sec:prelim}. In \cref{sec:understand}, we investigate how metric-based and gradient-based algorithms behave in practice with respect to the identified assumptions and provide theoretical explanation to the observed behavior. We further show that one can force meta-learning algorithms to satisfy such assumptions through adding an appropriate spectral regularization term to their objective function.  
        In \cref{sec:expe}, we provide an experimental evaluation of several state-of-the-art meta-learning methods and highlight the different advantages brought by the proposed regularization technique in practice for FSC tasks. 
        Finally, we conclude the chapter in \cref{sec:conclusion}.
    }
\end{abstract}

\begin{figure}
    \centering
    \includegraphics[width=0.8\linewidth]{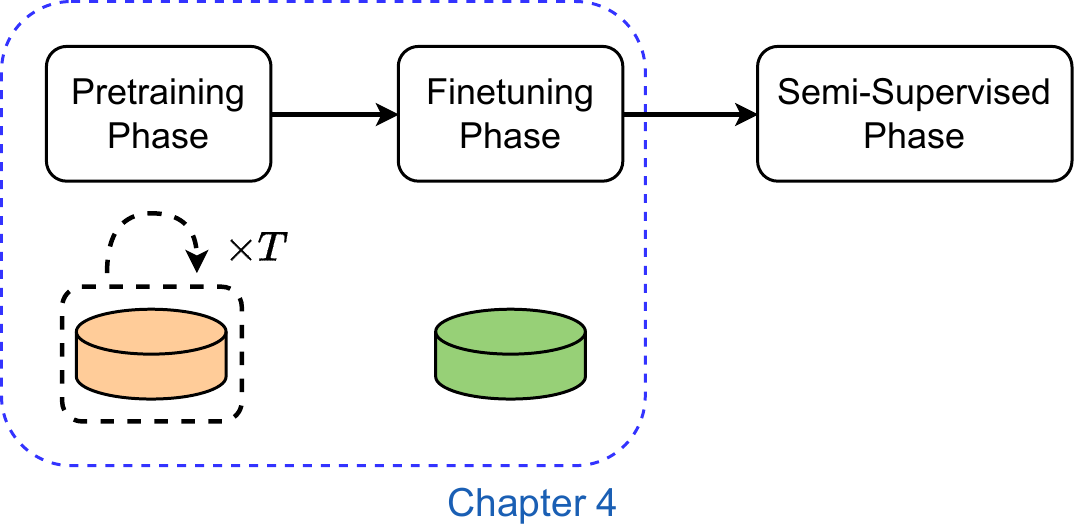}
    \caption{Illustration of the position of our contributions from this chapter in the full training pipeline presented in \cref{chap:intro}. We are interested on Meta-Learning for Pretraining then Finetuning.}
    \label{fig:position_eccv_chap}
\end{figure}

\section{Introduction}\label{sec:introduction}


\rev{As discussed in the previous chapter, meta-learning is an appealing solution for learning efficient models in the scarce data regime. Indeed, meta-learning aims at producing a model on data coming from a set of (meta-train) source tasks to use it as a starting point for learning successfully a new previously unseen (meta-test) target task. Each of these tasks can be composed of small sets of data samples.}

The success of many meta-learning approaches is directly related to their capacity of learning a good representation \cite{Raghu2020Rapid} from a set of tasks making it closely related to Multi-Task Representation learning (MTR).
For this latter, several theoretical studies \cite{baxterModelInductiveBias2000,pentinaPACBayesianBoundLifelong,maurerBenefitMultitaskRepresentation2016,AmitM18,Yin2020Meta-Learning}\footnote{We omit the results on meta-learning in the context of online convex optimization \cite{FinnRKL19,BalcanKT19,KhodakBT19,DeneviCGP19} as they concern a different learning setup.} provided probabilistic learning bounds that require the amount of data in the meta-train source task \textit{and} the number of meta-train tasks to tend to infinity for it to be efficient. While capturing the underlying general intuition, these bounds do not suggest that all the source data is useful in such learning setup due to the additive relationship between the two terms mentioned above and thus, for instance, cannot explain the empirical success of MTR in few-shot classification (FSC) task. To tackle this drawback, two very recent studies \cite{duFewShotLearningLearning2020,tripuraneniProvableMetaLearningLinear2020} aimed at finding deterministic assumptions that lead to faster learning rates allowing MTR algorithms to benefit from all the source data. Contrary to probabilistic bounds that have been used to derive novel learning strategies for meta-learning algorithms \cite{AmitM18,Yin2020Meta-Learning}, there has been no attempt to verify the validity of the assumptions leading to the fastest known learning rates in practice or to enforce them through an appropriate optimization procedure.

In this \rev{chapter}, we aim to use the recent advances in MTR theory \cite{tripuraneniProvableMetaLearningLinear2020,duFewShotLearningLearning2020} to explore the inner workings of these popular meta-learning methods. Our rationale for such an approach stems from a recent work \cite{pmlr-v139-wang21ad} proving that the optimization problem behind the majority of meta-learning algorithms can be written as an MTR problem.
Thus, we believe that looking at meta-learning algorithms through the recent MTR theory lens, could lead to better understanding the capacity to work well in few-shot regime.
In particular, we take a closer look at two families of meta-learning algorithms, notably: gradient-based algorithms \cite{raviOPTIMIZATIONMODELFEWSHOT2017,nicholFirstOrderMetaLearningAlgorithms2018,liMetaSGDLearningLearn2017, NEURIPS2018_66808e32,leeMetaLearningDifferentiableConvex2019,bertinettoMetalearningDifferentiableClosedform2019,ParkO19, NEURIPS2020_8989e07f, Raghu2020Rapid} including \Maml{} \cite{finnModelAgnosticMetaLearningFast2017} and metric-based algorithms \cite{kochSiameseNeuralNetworks,vinyalsMatchingNetworksOne2016,snellPrototypicalNetworksFewshot2017,sungLearningCompareRelation2018,pmlr-v97-allen19b,liFindingTaskRelevantFeatures2019, simonAdaptiveSubspacesFewShot2020} with its most prominent example given by \Proto{} \cite{snellPrototypicalNetworksFewshot2017}, which we presented in the previous chapter. 

Our main contributions \rev{in this chapter} are then two-fold: 
\begin{enumerate}
    \item We empirically show that tracking the validity of assumptions on optimal predictors used in \cite{tripuraneniProvableMetaLearningLinear2020,duFewShotLearningLearning2020} reveals a striking difference between the behavior of gradient-based and metric-based methods in how they learn their optimal feature representations. We provide elements of theoretical analysis that explain this behavior and explain the implications of it in practice. Our work is thus complementary to Wang et al. \cite{pmlr-v139-wang21ad} and connects MTR, FSC and Meta-Learning from both theoretical and empirical points of view.
    \item We show that theoretical assumptions mentioned above can be forced during the training of meta-learning algorithms for both families of considered methods, and that enforcing them leads to better generalization of the considered algorithms for FSC baselines. We further show that our analysis can be extended to a Multi-Task Learning algorithm (MTL \cite{pmlr-v139-wang21ad}), and that enforcing the same theoretical assumptions leads to improved results as well.
\end{enumerate}

\rev{Before presenting our contributions, we give in the next section some necessary knowledge to introduce the concepts that will be presented in this chapter.}

\section{Preliminary Knowledge}
\label{sec:prelim}

\rev{We start by recalling the framework of Multi-Task Representation Learning (MTR). Then we provide review of the existing learning bounds and assumptions allowing one to obtain guarantees in this setting. Finally, we describe the meta-learning frameworks considered in the context of this contribution.}

\begin{figure}[t]
    \centering
    \begin{tabular}{c|c}
        \textbf{Violated assumptions} & \textbf{Satisfied assumptions}\\
        \includegraphics[trim= 50 50 50 45, clip, width = .49\linewidth]{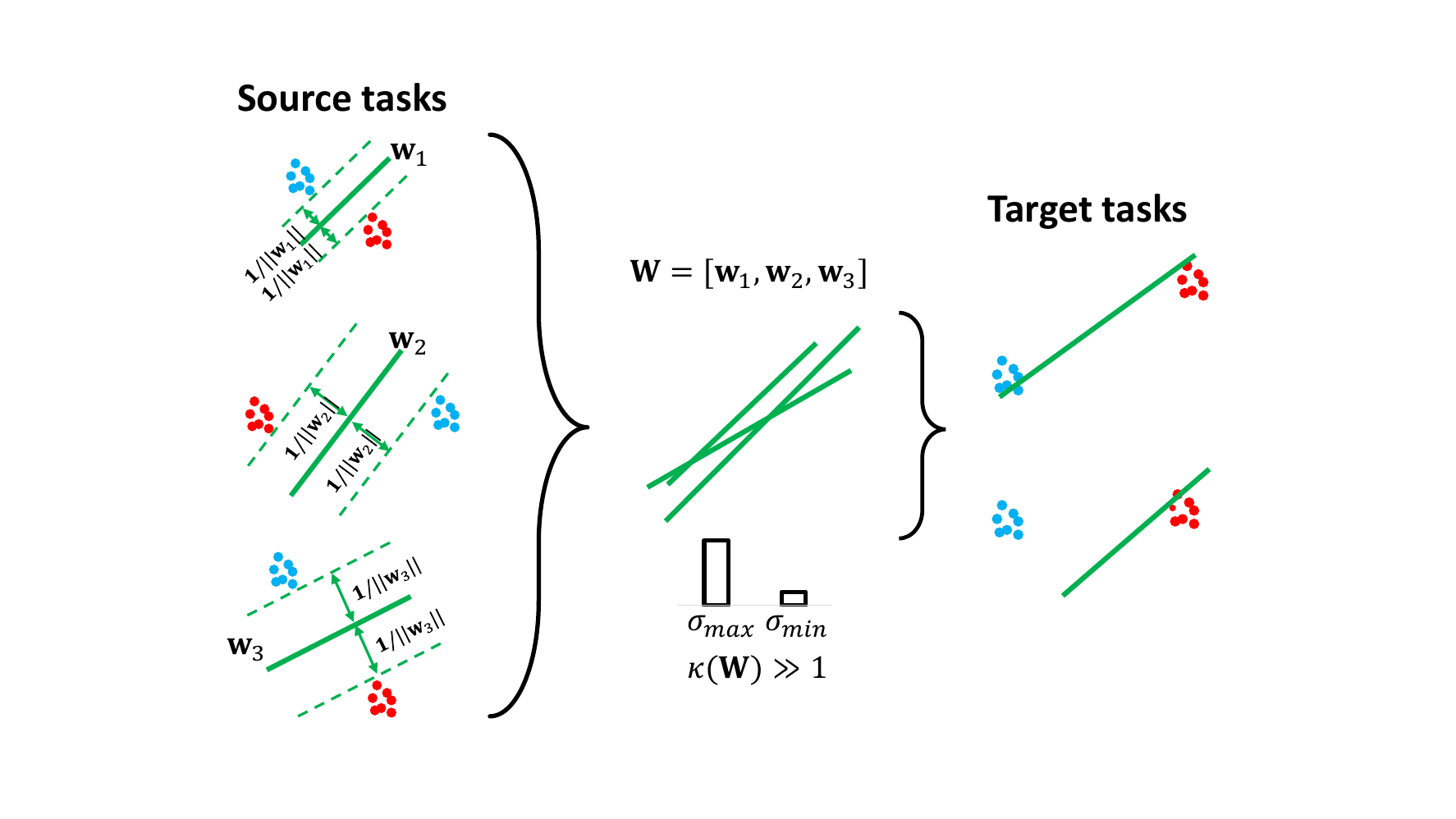}
        &
        \includegraphics[trim= 30 40 30 30, clip, width = .49\linewidth]{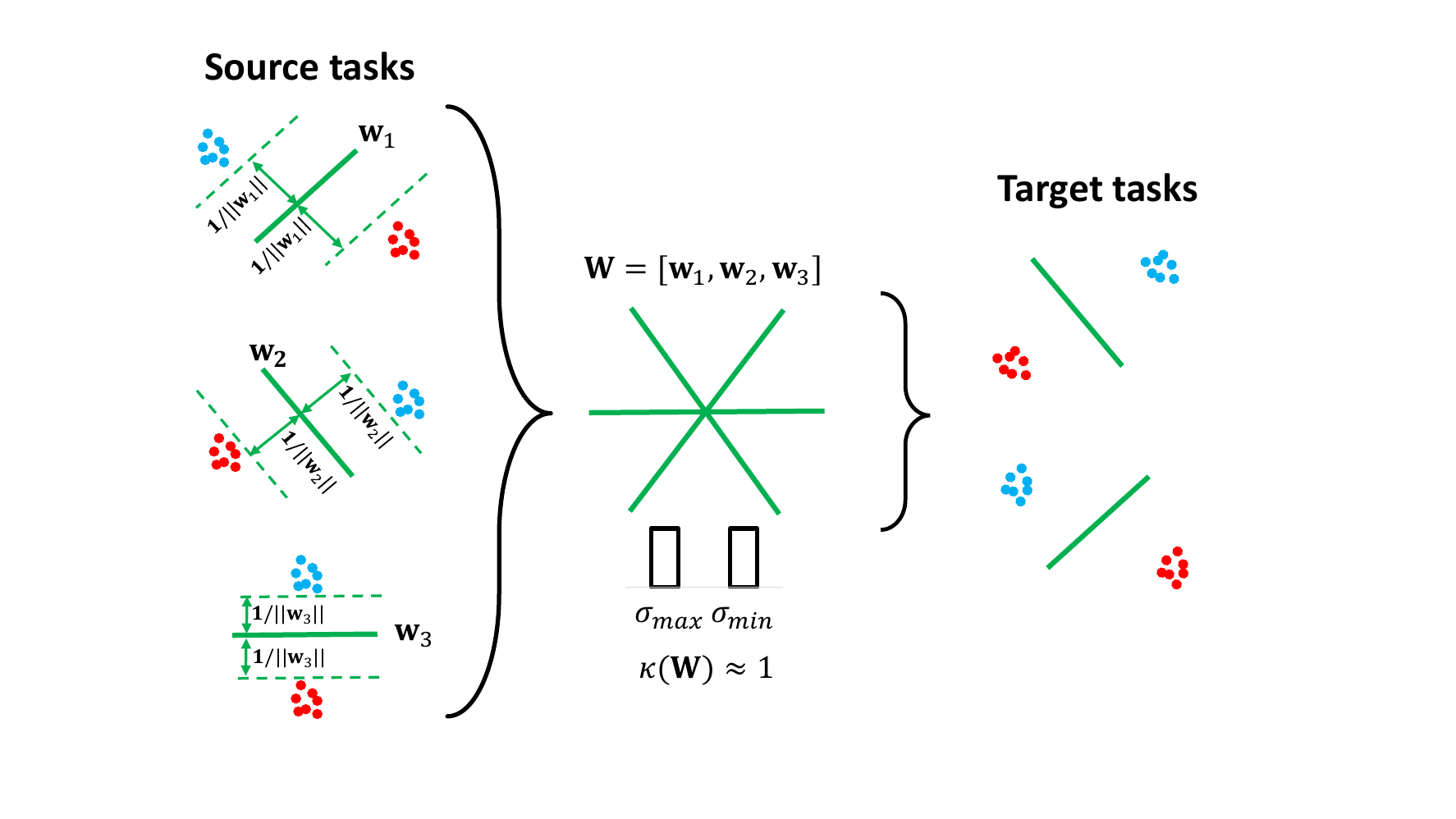}
    \end{tabular}
    \caption{Illustration of the intuition behind the assumptions derived from the MTR learning theory. \textbf{(left)} Lack of diversity and increasing norm of the linear predictors restrict them from being useful on the target task. \textbf{(right)} When the assumptions are satisfied, the linear predictors cover the embedding space evenly and their norm remains roughly constant on source tasks making them useful for a previously unseen task.}
    \label{fig:intuition0}
\end{figure}

\subsection{Multi-Task Representation Learning Setup}
Given a set of $T$ source tasks observed through finite size samples of size $n_1$ grouped into matrices $\rmX_t = (\rvx_{t,1}, \dots, \rvx_{t,n_1}) \in \sR^{n_1\times d}$ and vectors of outputs $\rvy_t = (y_{t,1}, \dots, y_{t,n_1}) \in \sR^{n_1},\ \forall t \in [[T]]:= \{1, \dots, T\}$ generated by their respective distributions $\mu_{t}$, the goal of Multi-Task Representation Learning (MTR) is to learn a shared representation $\phi$ belonging to a certain class of functions $\Phi := \{\phi\ |\ \phi: \sX \rightarrow \sV,\ \sX \subseteq \sR^d, \ \sV \subseteq \sR^k\}$, generally represented as (deep) neural networks, and linear predictors $\rvw_t \in \sR^{k},\ \forall t \in [[T]]$ grouped in a matrix $\rmW \in \sR^{T\times k}$. More formally, this is done by solving the following joint optimization problem:
\begin{equation}
    \hat{\phi}, \widehat{\rmW} \!=\! \argmin_{\phi \in \Phi, \rmW \in \sR^{T\times k}} \frac{1}{Tn_1}\sum_{t=1}^T\sum_{i=1}^{n_1} \ell(y_{t,i},\langle \rvw_t, \phi(\rvx_{t,i})\rangle),
    \label{eq:mtr_pb}
\end{equation}
where $\ell:\sY \times \sY \rightarrow \sR_+$, with $\sY \subseteq \sR$, is a loss function. Once such a representation $\hat{\phi}$ is learned, we want to apply it to a new previously unseen target task observed through a pair $(\rmX_{T+1} \in \sR^{n_2\times d}, \rvy_{T+1} \in \sR^{n_2})$ containing $n_2$ samples generated by the distribution $\mu_{T+1}$. We expect that a linear classifier $\rvw$ learned on top of the obtained representation leads to a low true risk over the whole distribution $\mu_{T+1}$. For this, we first use $\hat{\phi}$ to solve the following problem:
\begin{equation}
    \hat{\rvw}_{T+1} = \argmin_{\rvw \in \sR^{k}} \frac{1}{n_2}\sum_{i=1}^{n_2} \ell(y_{T+1,i},\langle \rvw, \hat{\phi}(\rvx_{T+1,i})\rangle).
\end{equation}
Then, we define the true target risk of the learned linear classifier $\hat{\rvw}_{T+1}$ as:
$\gL(\hat{\phi}, \hat{\rvw}_{T+1}) = \mathop{\E}_{(\rvx,y)\sim \mu_{T+1}}[\ell(y,\langle \hat{\rvw}_{T+1}, \hat{\phi}(\rvx)\rangle)]$
and want it to be as close as possible to the ideal true risk $\gL(\phi^*, \rvw_{T+1}^*)$ where $\rvw_{T+1}^*$ and $\phi^*$ satisfy:
\begin{equation}
        \forall t \in [[T+1]] \text{  and  } (\rvx,y) \sim \mu_t, \quad y = \langle \rvw_t^*, \phi^*(\rvx)\rangle + \varepsilon, \quad \varepsilon \sim \gN(0,\sigma^2).
    \label{eq:data_gen_model}
\end{equation}

\noindent Equivalently, most of the works found in the literature seek to upper-bound the \textit{excess risk} defined as $\text{ER}(\hat{\phi}, \hat{\rvw}_{T+1}) := \gL(\hat{\phi}, \hat{\rvw}_{T+1}) - \gL(\phi^*, \rvw_{T+1}^*)$. 

\subsection{Learning Bounds and Assumptions}\label{sec:theoretical_results}
First studies in the context of MTR relied on the probabilistic assumption \cite{baxterModelInductiveBias2000,pentinaPACBayesianBoundLifelong,maurerBenefitMultitaskRepresentation2016,AmitM18,Yin2020Meta-Learning} stating that meta-train and meta-test tasks distributions are all sampled i.i.d. from the same random distribution. 
\rev{This assumption, however, is considered unrealistic as in many learning settings, such as FSC, source and target tasks' data are often given by different draws (without replacement) from the same dataset. In this setup, the above-mentioned works obtained the bounds having the following form:
$$\text{ER}(\hat{\phi}, \hat{\rvw}_{T+1}) \leq O\left(\frac{1}{\sqrt{n_1}}+\frac{1}{\sqrt{T}}\right).$$
Such a guarantee implies that even with the increasing number of source data, one would still have to increase the number of tasks as well, in order to draw the second term to $0$.} 
A natural improvement to this bound was then proposed by \cite{duFewShotLearningLearning2020} and \cite{tripuraneniProvableMetaLearningLinear2020} that obtained the bounds on the excess risk behaving as 
\begin{equation}
    \text{ER}(\hat{\phi}, \hat{\rvw}_{T+1}) \leq O\left(\frac{C(\Phi)}{n_1T}+\frac{k}{n_2}\right),
\end{equation}
where $C(\Phi)$ is a measure of the complexity of $\Phi$.
Both these results show that all the source and target samples are useful in minimizing the excess risk. Thus, in the FSC regime where target data is scarce, all source data helps to learn well. From a set of assumptions made by the authors in both of these works, we note the following two:

\textbf{Assumption 1: Diversity of the source tasks}\phantom{.} The matrix of optimal predictors $\rmW^*$ should cover all the directions in $\sR^k$ evenly. More formally, this can be stated as
\begin{align}
    \kappa(\rmW^*) := \frac{\sigma_1(\rmW^*)}{\sigma_k(\rmW^*)} = O(1),
\end{align}
where $\sigma_i(\cdot)$ denotes the $i\textsuperscript{th}$ singular value of $\rmW^*$. As pointed out by the authors, such an assumption can be seen as a measure of diversity between the source tasks that are expected to be complementary to each other to provide a useful representation for a previously unseen target task. In the following, we will refer to $\kappa(\rmW)$ as the \emph{condition number} for matrix $\rmW$. 

\textbf{Assumption 2: Consistency of the classification margin}\ The norm of the optimal predictors $\rvw^*$ should not increase with the number of tasks seen during meta-training\footnote{While not stated separately, this assumption is used in \cite{duFewShotLearningLearning2020} to derive the final result on p.5 after the discussion of Assumption 4.3.}. This assumption says that the classification margin of linear predictors should remain constant thus avoiding over- or under-specialization to the seen tasks. 

\rev{While being highly insightful, the authors did not provide any experimental evidence suggesting that verifying these assumptions in practice helps to learn more efficiently in the considered learning setting. 
An intuition behind these assumptions can be seen in \cref{fig:intuition0}. When the assumptions do not hold, the linear predictors can be biased towards a single part of the space and over-specialized to the tasks. The representation learned will not generalize well to unseen tasks. If the assumptions are respected, the linear predictors are complementary and will not under- or over-specialize to the tasks seen. The representation learned can more easily adapt to the target tasks and achieve better generalization.}


\subsection{Meta-Learning Algorithms}

\rev{We precise now the meta-learning frameworks that are considered in this chapter.}
Meta-learning algorithms considered below learn an optimal representation sequentially via the so-called episodic training strategy introduced by~\cite{vinyalsMatchingNetworksOne2016}, instead of jointly minimizing the training error on a set of source tasks as done in MTR. \rev{As introduced in \cref{chap:sota},} episodic training mimics the training process at the task scale with each task data being decomposed into a training set \emph{(support set $\Scal$)} and a testing set \emph{(query set $\mathcal{Q}$)}. 
Recently, \cite{chenCloserLookTraining2020} showed that the episodic training setup used in meta-learning leads to a generalization bounds of $O(\frac{1}{\sqrt{T}})$. This bound is independent of the task sample size $n_1$, which could explain the success of this training strategy for FSC in the asymptotic limit. 
However, unlike the results obtained by \cite{duFewShotLearningLearning2020} studied in this \rev{chapter}, the lack of dependence on $n_1$ makes such a result uninsightful in practice as we are in a finite-sample size setting. This bound does not give information on other parameters to leverage when the task number cannot increase. 
We now \rev{recall} two major families of meta-learning approaches below. 

\noindent \textbf{Metric-based methods}\phantom{.} These methods learn an embedding space in which feature vectors can be compared using a similarity function (usually an $L_2$ distance or cosine similarity) \cite{kochSiameseNeuralNetworks,vinyalsMatchingNetworksOne2016,snellPrototypicalNetworksFewshot2017,sungLearningCompareRelation2018,pmlr-v97-allen19b,liFindingTaskRelevantFeatures2019, simonAdaptiveSubspacesFewShot2020}. They typically use a form of contrastive loss as their objective function, similarly to Neighborhood Component Analysis (NCA)~\cite{NIPS2004_NCA_goldberger} or Triplet Loss~\cite{hofferDeepMetricLearning2015}. In this \rev{chapter}, we focus our analysis on the popular Prototypical Networks~\cite{snellPrototypicalNetworksFewshot2017} (\Proto{}) that computes prototypes as the mean vector of support points belonging to the same class: 
$\rvc_i = \frac{1}{|\Scal_i|} \sum_{\rvs \in \Scal_i} \phi(\rvs),$
with $\Scal_i$ the subset of support points belonging to class $i$.

\Proto{} minimizes the negative log-probability of the true class $i$ computed as the softmax over distances to prototypes $\rvc_i$: 
\begin{equation}
    \Lcal_{proto}(\Scal,\Qcal,\phi):= \E_{\rvq \sim \Qcal} \left[ - \log \frac{\exp(-d(\phi(\rvq), \rvc_i))}{\sum_{j} \exp{(- d(\phi(\rvq), \rvc_j))}} \right]
\end{equation}
with $d$ being a distance function used to measure similarity between points in the embedding space.


\noindent \textbf{Gradient-based methods}\phantom{.} These methods learn through end-to-end or two-step optimization  \cite{raviOPTIMIZATIONMODELFEWSHOT2017,nicholFirstOrderMetaLearningAlgorithms2018,liMetaSGDLearningLearn2017, NEURIPS2018_66808e32,leeMetaLearningDifferentiableConvex2019,bertinettoMetalearningDifferentiableClosedform2019,ParkO19, NEURIPS2020_8989e07f, Raghu2020Rapid} where given a new task, the goal is to learn a model from the task's training data specifically adapted for this task.
\Maml{}~\cite{finnModelAgnosticMetaLearningFast2017} updates its parameters $\theta$ using an end-to-end optimization process to find the best initialization such that a new task can be learned quickly, \ie{} with few examples. More formally, given the loss $\ell_t$ for each task $t \in [[T]]$, \Maml{} minimizes the expected task loss after an \emph{inner loop} or \emph{adaptation} phase, computed by a few steps of gradient descent initialized at the model's current parameters:
\begin{equation}
    \Lcal_{\Maml{}}(\theta) := \E_{t \sim \eta} [\ell_t(\theta - \alpha \nabla \ell_t(\theta))],
\end{equation}
with $\eta$ the distribution of the meta-training tasks and $\alpha$ the learning rate for the adaptation phase. For simplicity, we take a single step of gradient update in this equation.

In what follows, we establish our theoretical analysis for the popular methods \Proto{} and \Maml{}. We add their improved variations respectively called Infinite Mixture Prototypes~\cite{pmlr-v97-allen19b} (\IMP{}) and Meta-Curvature~\cite{ParkO19} (\MC{}) in the experiments to validate our findings. \rev{The interested reader can find an introduction to these two more advanced algorithms in \cref{ax:intro_imp_mc}.}


\section[Understanding Meta-learning Algorithms through MTR Theory]{Understanding Meta-learning Algorithms \\ through MTR Theory}\label{sec:understand}

\rev{In this section, we study the behavior of gradient- and metric-based meta-learning algorithms with respect to the theoretical insights from MTR theory.}  

\subsection{Link between MTR and Meta-learning}

Recently, \cite{pmlr-v139-wang21ad} has shown that meta-learning algorithms that only optimize the last layer in the inner-loop, solve the same underlying optimization procedure as multi-task learning. In particular, their contributions have the following implications:
\begin{enumerate}
    \item For \emph{Metric-based algorithms}, the majority of methods can be seen as MTR problems. This is true, in particular, for \Proto{}{}\ and \IMP{}\ algorithms considered in this work.
    \item In the case of \emph{Gradient-based algorithms}, such methods as ANIL~\cite{Raghu2020Rapid} and MetaOptNet~\cite{leeMetaLearningDifferentiableConvex2019} that do not update the embeddings during the inner-loop, can be also seen as multi-task learning. However, \Maml{}\ and \MC{}\ in principle do update the embeddings even though there exists strong evidence suggesting that the changes in the weights during their inner-loop are mainly affecting the last layer \cite{Raghu2020Rapid}. Consequently, we follow \cite{pmlr-v139-wang21ad} and use this assumption to analyze \Maml{}\ and \MC{}\ in MTR framework as well.
    \item In practice, \cite{pmlr-v139-wang21ad} showed that the mismatch between the multi-task and the actual episodic training setup leads to a negligible difference.
\end{enumerate}


\noindent In the following section, we start by empirically verifying that
the behavior of meta-learning methods reveals very distinct features when looked at through the prism of the considered MTR theoretical assumptions. 
\rev{We then set on a quest of explaining the differences in their behavior, leading to novel insights into meta-learning algorithms and interesting open problems for future research.}

\subsection{What happens in practice?}\label{sec:practice}
To verify whether theoretical results from MTR setting are also insightful for episodic training used by popular meta-learning algorithms, we first investigate the natural behavior of \Maml{} and \Proto{} when solving FSC tasks on the popular \emph{miniImageNet}~\cite{raviOPTIMIZATIONMODELFEWSHOT2017}, \emph{tieredImageNet}~\cite{renMetaLearningSemiSupervisedFewShot2018} and \emph{Omniglot} \cite{lakeHumanlevelConceptLearning2015} datasets. The full experimental setup is detailed in \cref{sec:expe_setup}.

To verify Assumption 1 from MTR theory, we want to compute singular values of $\rmW$ during the meta-training stage and to follow their evolution. In practice, as $T$ is typically quite large, we propose a more computationally efficient solution that is to calculate the condition number only for the last batch of $N$ predictors (with $N \ll T$) grouped in the matrix $\rmW_N \in \sR^{N\times k}$ that capture the latest dynamics in the learning process. We further note that $\sigma_i(\rmW_N \rmW_N^{\top}) = \sigma_i^2(\rmW_N),\ \forall i \in [[N]]$ implying that we can calculate the SVD of $\rmW_N \rmW_N^{\top}$ (or $\rmW_N^{\top} \rmW_N$ for $k \leq N$) and retrieve the singular values from it afterwards. 
We now want to verify whether $\rvw_t$ cover all directions in the embedding space and track the evolution of the ratio of singular values $\kappa(\rmW_N)$ during training. 

For the first assumption to be satisfied, we expect $\kappa(\rmW_N)$ to decrease gradually during the training thus improving the generalization capacity of the learned predictors and preparing them for the target task. To verify the second assumption, the norm of the linear predictors should not increase with the number of tasks seen during training, \textit{i.e.}, $\|\rvw\|_2 = O(1)$ or, equivalently, $\|\rmW\|_F^2 = O(T)$ and $\|\rmW_N\|_F = O(1)$.

\begin{figure}[!t]
    \centering
    \includegraphics[width=0.32\linewidth]{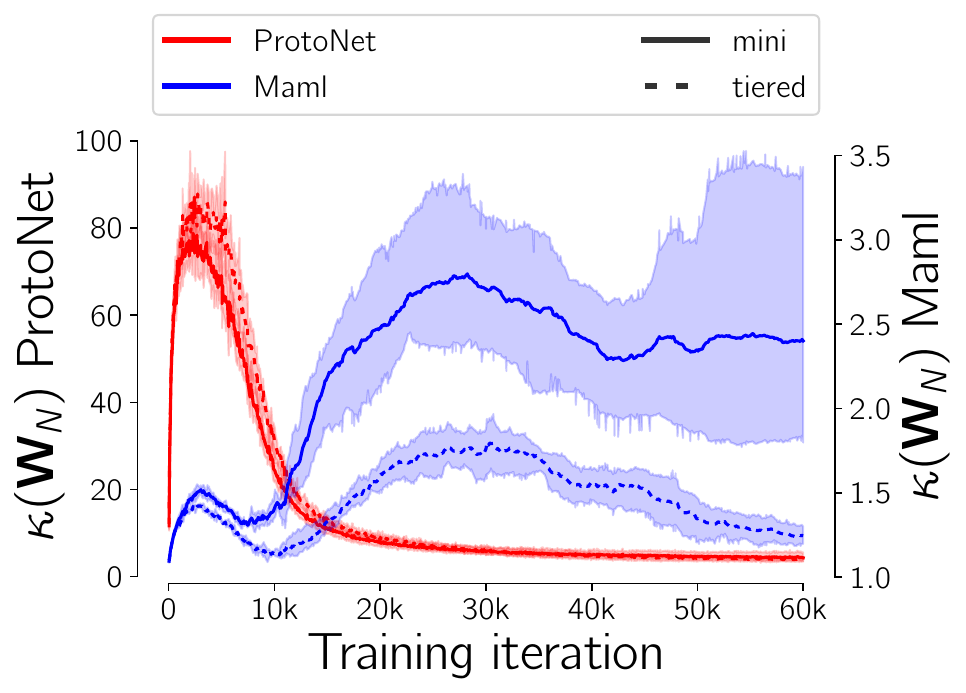}
    \includegraphics[width=0.32\linewidth]{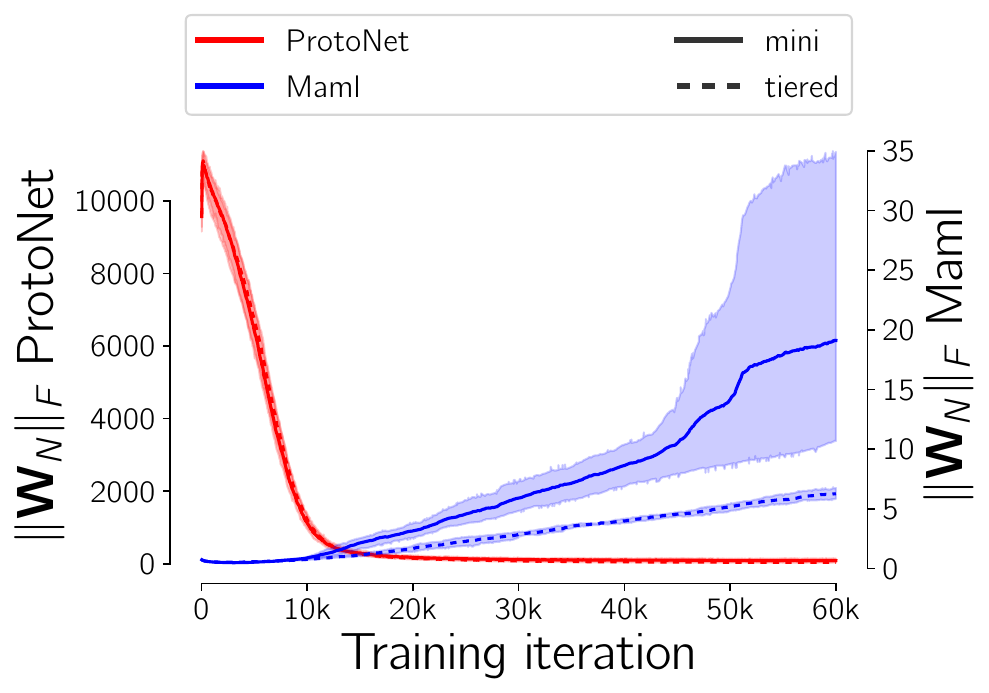}
    \includegraphics[width=0.28\linewidth]{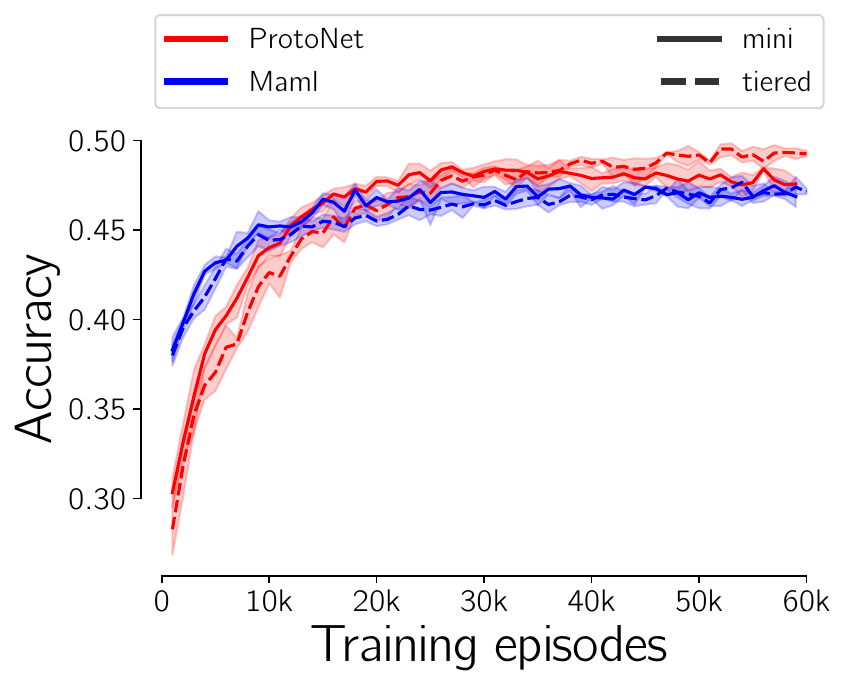}
    \\
    \includegraphics[width=0.32\linewidth]{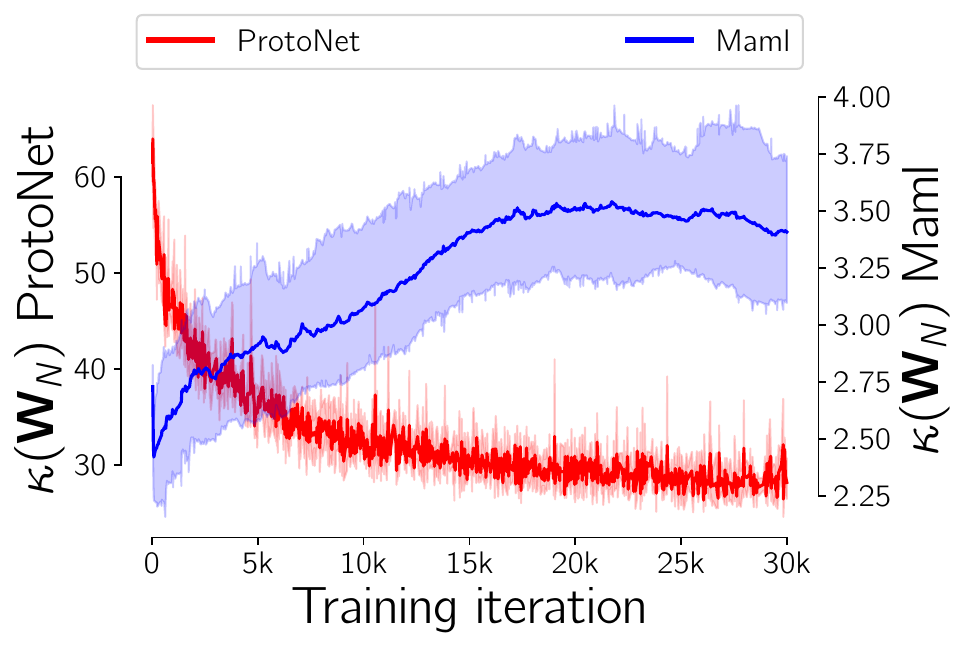}
    \includegraphics[width=0.32\linewidth]{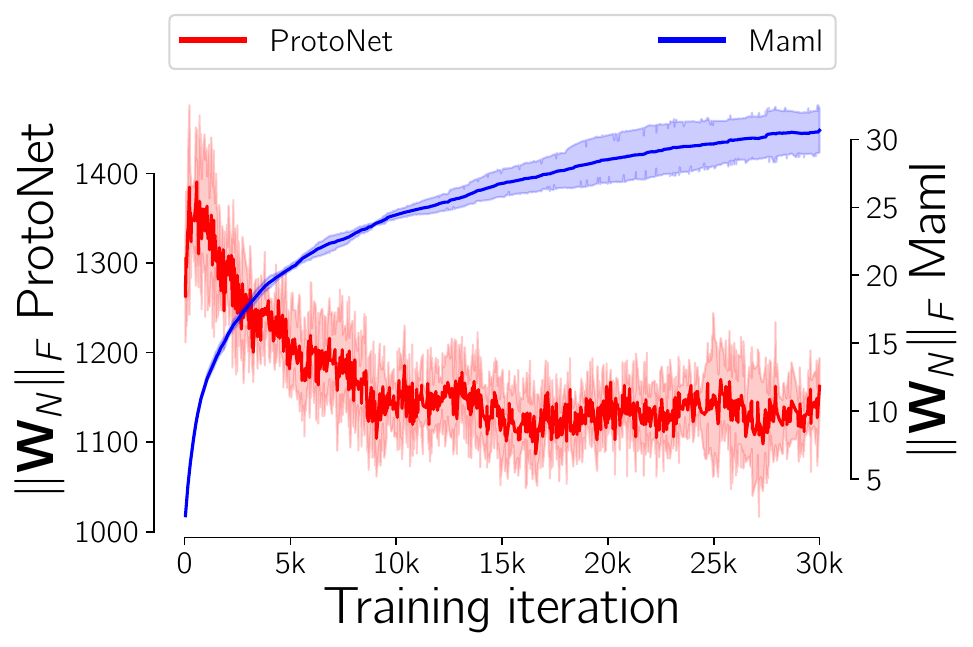}
    \includegraphics[width=0.28\linewidth]{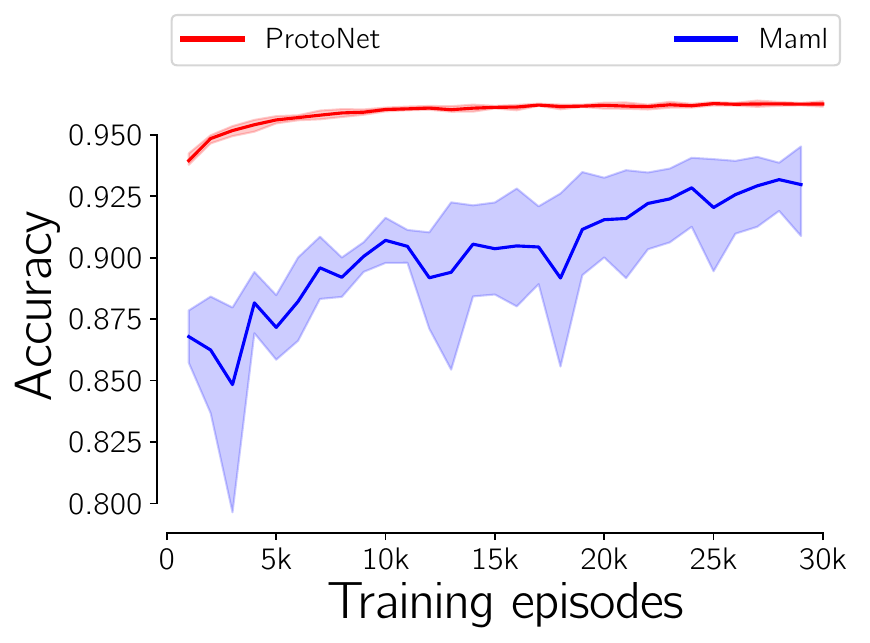}
    \caption{Evolution of $\kappa(\rmW_N)$ (left), $\|\rmW_N\|_F$ (middle) and accuracy (right) during the training of \Proto{}\ (\emph{red, left axes}) and \Maml{}\ (\emph{blue, right axes}) on miniImageNet (\emph{mini, solid lines}) and tieredImageNet (\emph{tiered, dashed lines}) with 5-way 1-shot episodes (\emph{top}), \rev{and on the Omniglot dataset with 20-way 1-shot episodes (\emph{bottom})}. 
    }
    \label{fig:training_curves}
\end{figure}

\begin{figure}
    \centering
    \includegraphics[width=0.6\linewidth]{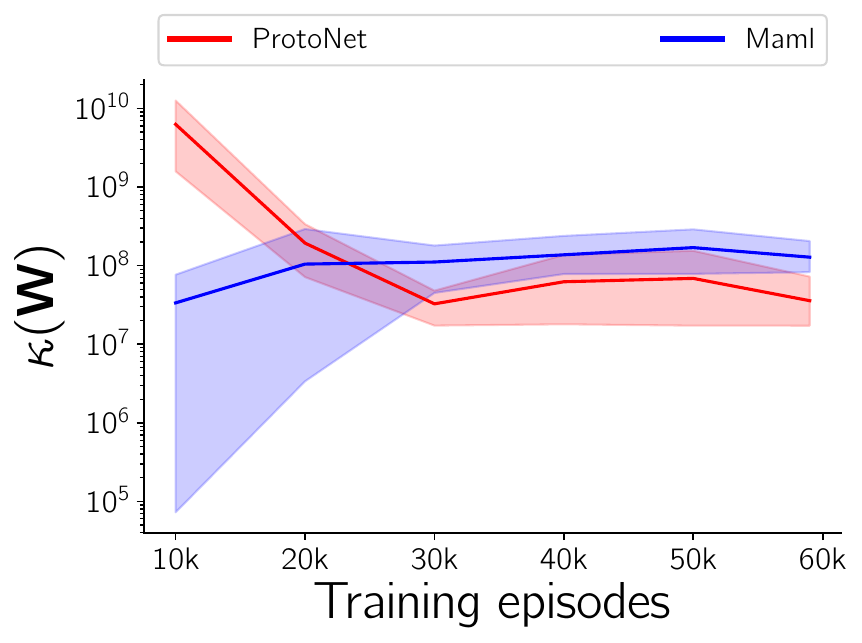}
    \caption{Evolution of $\kappa(\rmW)$ (in log scale) during the training of \Proto{} (in red) and \Maml{} (in blue) on miniImageNet 5-way 1-shot episodes.}
    \label{fig:meta_mtr_kappa_full}
\end{figure}


For gradient-based methods, linear predictors are directly the weights of the last layer of the model. Indeed, for each task, the model learns a batch of linear predictors and we can directly take the weights for $\rmW_N$.
Meanwhile, metric-based methods do not use linear predictors but compute a similarity between features. In the case of \Proto{}, the similarity is computed with respect to class prototypes that are mean features of the instances of each class. For the Euclidean distance, \emph{this is equivalent to a linear model} with the prototype of a class acting as the linear predictor of this class \cite{snellPrototypicalNetworksFewshot2017}. This means that we can apply our analysis directly to the \emph{prototypes} computed by \Proto{}. In this case, the matrix $\rmW^*$ will be the matrix of the optimal prototypes and we can then take the \emph{prototypes} computed for each task as our matrix $\rmW_N$.

From \cref{fig:training_curves}, we can see that for \Maml{}\ (\emph{blue}), both $\|\rmW_N\|_F$ (\emph{left}) and $\kappa(\rmW_N)$ (\emph{middle}) increase with the number of tasks seen during training, whereas \Proto{}\ (\emph{red}) naturally learns prototypes with a good coverage of the embedding space, and minimizes their norm.
Since we compute the singular values of the last $N$ predictors in $\kappa(\rmW_N)$, we can only compare the \emph{overall behavior} throughout training between methods. For the sake of completeness, we also compute $\kappa(\rmW)$ at different stages in the training, provided in \cref{fig:meta_mtr_kappa_full}. To do so, we fix the encoder $\phi_T$ learned after $T$ episodes and recalculate the linear predictors of the $T$ past training episodes with this fixed encoder. We can see that $\kappa(\rmW)$ of \Proto{}\ also decreases during training and reach a lower final value than $\kappa(\rmW)$ of \Maml{}. This confirms that the dynamics of $\kappa(\rmW_N)$ and $\kappa(\rmW)$ are similar whereas the values $\kappa(\rmW_N)$ between methods should not be directly compared.
The behavior of $\kappa(\rmW)$ also validate our finding that \Proto{}\ learns to cover the embedding space with prototypes.
This behavior is rather peculiar as neither of the two methods explicitly controls the theoretical quantities of interest, and still, \Proto{}\ manages to do it implicitly. 

\rev{Before confirming this claim through extensive empirical evaluations involving more baseline methods and benchmark datasets, we first prove several results that provide an explanation to the difference of behavior of these two families of methods.} 

\subsection{The case of Meta-Learning algorithms}
\label{sec:grad_vs_metric}

The differences observed above for the two methods call for a deeper analysis of their behavior. 
\rev{To this end, we provide an explanation of why \Proto{}\ naturally leads to small condition number of the obtained predictors and a consistent behavior of their norm, while for \Maml{}\ we consider a common simplified learning model leaving the general result as an open problem for future works.} 

\noindent \textbf{\Proto{}}\phantom{.} We start by first explaining why \Proto{}\ learns prototypes that cover the embedding space efficiently. This result is given by the following theorem (cf. \cref{ax:meta_mtr_proofs} for the full proof).
\begin{theorem}\label{th:norm_proto}\emph{(Normalized \Proto{})} \\
If $\forall i\ \|\rvc_i\| = 1$, then $\forall \hat{\phi} \in \argmin_{\phi} \Lcal_{proto}(S,Q,\phi)$, the matrix of the optimal prototypes $\rmW^*$ is well-conditioned, \ie{} $\kappa(\rmW^*) = O(1)$.
\end{theorem}

\begin{proof_idea}
\rev{We prove this result by recalling that, similarly to other metric-based methods, $\Lcal_{proto}$ is a form of contrastive loss studied in~\cite{pmlr-v119-wang20k}. By rewriting $\Lcal_{proto}$ and using a part of their Theorem 1 proof on the convergence of the loss, we get the desired result.} 
\end{proof_idea}
This theorem explains the empirical behavior of \Proto{}\ in FSC task: the minimization of its objective function naturally minimizes the condition number when the norm of the prototypes is low. 
In particular, it implies that norm minimization seems to initiate the minimization of the condition number seen afterwards due to the contrastive nature of the loss function minimized by \Proto{}. We confirm this latter implication through experiments in \cref{sec:expe} showing that norm minimization is enough for considered metric-based methods to obtain the well-behaved condition number and that minimizing both seems redundant. 

\textbf{\Maml{}}\phantom{.} Unfortunately, the analysis of \Maml{}\ in the most general case is notoriously harder, as even expressing its loss function and gradients in the case of an overparametrized linear regression model with only 2 parameters requires using a symbolic toolbox for derivations \cite{sha_maml_2021}. 

To this end, we resort to the linear regression model considered in this latter paper and defined as follows. We assume for all $t \in [[T]]$ that the task parameters $\boldsymbol{\theta}_t$ are normally distributed with $\boldsymbol{\theta}_t \sim \mathcal{N}(\bm{0}_d,\bm{I}_d)$, the inputs $\rvx_t \sim \mathcal{N}(\bm{0}_d,\bm{I}_d)$ and the output $y_t \sim \mathcal{N}(\langle\boldsymbol{\theta}_t, \rvx_t\rangle, 1)$. For each $t$, we consider the following learning model and its associated square loss:
\begin{equation}
\hat{y}_t = \langle \rvw_t, \rvx_t\rangle, \quad \ell_t = \mathbb{E}_{p(\rvx_t,y_t|\boldsymbol{\theta}_t)} (y_t-\langle \rvw_t, \rvx_t\rangle)^2.   
\label{eq:setup_maml}
\end{equation}
We can now state the following result (the full proof can be found in \cref{ax:meta_mtr_proofs}).
\begin{proposition}\label{prop:maml_kappa}
Let $\forall t \in [[T]]$, $\boldsymbol{\theta}_t \sim \mathcal{N}(\bm{0}_d,\bm{I}_d)$, $\rvx_t \sim \mathcal{N}(\bm{0}_d,\bm{I}_d)$ and $y_t \sim \mathcal{N}(\langle \boldsymbol{\theta}_t, \rvx_t\rangle, 1)$. Consider the learning model from \cref{eq:setup_maml}, let $\boldsymbol{\Theta}_i := [\boldsymbol{\theta}_i, \boldsymbol{\theta}_{i+1}]^T$, and denote by $\widehat{\rmW}_2^i$ the matrix of last two predictors learned by \Maml{}\ at iteration $i$ starting from $\widehat{\rvw}_0 = \bm{0}_d$. Then, we have that: \\
\begin{equation}
    \forall i, \quad \kappa(\widehat{\rmW}_2^{i+1})\geq \kappa(\widehat{\rmW}_2^i),\ \text{ if } \sigma_\text{min}(\boldsymbol{\Theta}_i) = 0.
\end{equation}
\end{proposition}

\begin{proof_idea}
    \rev{We use the analytical expression for the $t$\textsuperscript{th} task gradient given in \cite{sha_maml_2021} for this case. We then show that when starting the optimization process from $\widehat{\rvw}_0 = \bm{0}_d$, the sequence of the obtained linear predictors are interdependent such that $\widehat{\rmW}_2^{i+1} \propto \widehat{\rmW}_2^i + \boldsymbol{\Theta}_i$. We then use upper and lower bounds for singular values of a sum of matrices to derive the desired result.} 
\end{proof_idea}

This proposition provides an explanation of why \Maml{}\ may tend to increase the ratio of singular values during the iterations. Indeed, the condition when this happens indicates that the optimal predictors forming matrix $\boldsymbol{\Theta}_i$ are linearly dependent implying that its smallest singular values becomes equal to 0. While this is not expected to be the case for all iterations, we note, however, that in FSC task the draws from the dataset are in general not i.i.d. and thus may correspond to co-linear optimal predictors. In every such case, the condition number is expected to remain non-decreasing, as illustrated in \cref{fig:training_curves} (\emph{left}) where for \Maml{}, contrary to \Proto{}, $\kappa(\rmW_N)$ exhibits plateaus but also intervals where it is increasing. 

This highlights a major difference between the two approaches: \Maml{}\ does not specifically seek to diversify the learned predictors, while \Proto{}\ does.
\rev{Even though the condition number and the classifier norm both increase for \Maml{}, it still achieves good performance. \textbf{This means that both assumptions are not necessary to meta-learning. However, verifying them leads to a more favorable framework that allows for better performance as confirmed by our experimental evaluation below.}} 
\cref{prop:maml_kappa} gives a hint to the essential difference between the methods studied. On the one hand, \Proto{}\ constructs its classifiers directly from the data of the current task and they are independent of the other tasks. On the other hand, for \Maml{}, the weights of the classifiers are reused between tasks and only slightly adapted to be specific to each task. This limits the generalization capabilities of the linear predictors learned by \Maml{}\ since they are based on predictors from previous tasks.

\rev{In the following section, we advance this insight further by showing that for some data generating distributions one may learn two data representations that achieve a perfect fidelity to the data generating process but differ drastically in terms of the coverage of the embedding space provided by their linear predictors.}

\subsection{Enforcing the assumptions}\label{sec:toy_problem}
\rev{So far, we have gathered evidence for the fact that MTR theory seems to be insightful for meta-learning algorithms as well. As satisfying the assumptions from MTR theory is expected to come in hand with better generalization performance, we now study what impact forcing these assumptions may have on the learning process when the optimal predictors involved in the data generating process do not naturally satisfy them.}

\textbf{Why should we force the assumptions ?}\phantom{.} From the results obtain by \cite{duFewShotLearningLearning2020}, and with the same assumptions, we can easily make appear $\kappa(\rmW^*)$ to obtain a more explicit bound:

\begin{proposition}\label{prop:new_bound}
If $\forall t \in [[T]], \|\rvw_t^*\| = O(1)$ and $\kappa(\rmW^*) = O(1)$, and $\rvw_{T+1}$ follows a distribution $\nu$ such that $\| \E_{\rvw \sim \nu}[\rvw\rvw^\top]\| \leq O \left(\frac{1}{k} \right)$, then
\begin{equation}
    \text{ER}(\hat{\phi}, \hat{\rvw}_{T+1}) \leq O\left(\frac{C(\Phi)}{n_1T}\cdot \kappa(\rmW^*) +\frac{k}{n_2}\right).
\end{equation}
\end{proposition}

\cref{prop:new_bound} suggests that the terms $\|\rvw_t^*\|$ and $\kappa(\rmW^*)$ underlying the assumptions directly impact the tightness of the established bound on the excess risk. The full proof can be found in \cref{ax:meta_mtr_proofs}.

\begin{figure}[!t]
    \centering
    \includegraphics[width = .85\linewidth]{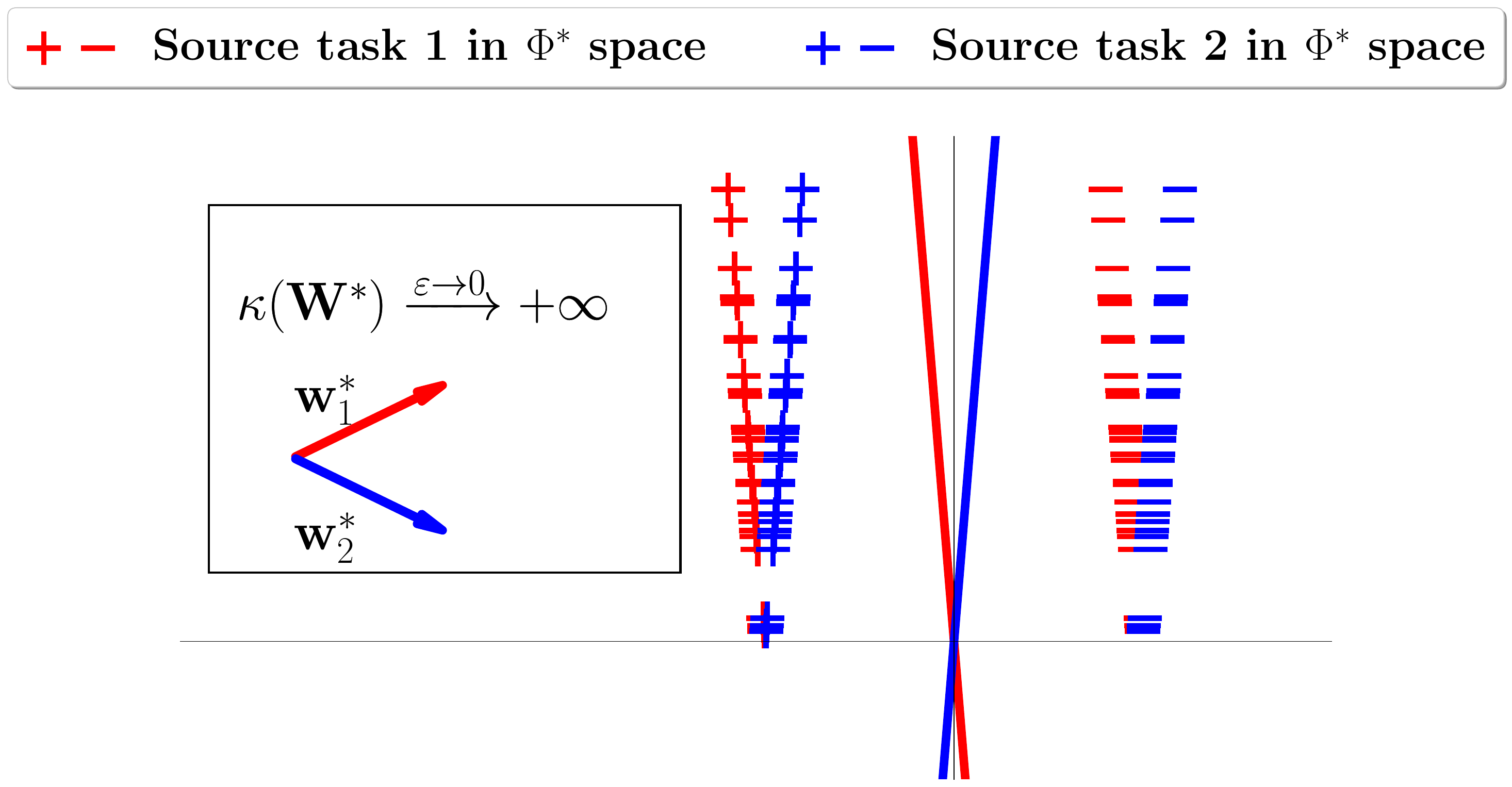}
    \includegraphics[width = .85\linewidth]{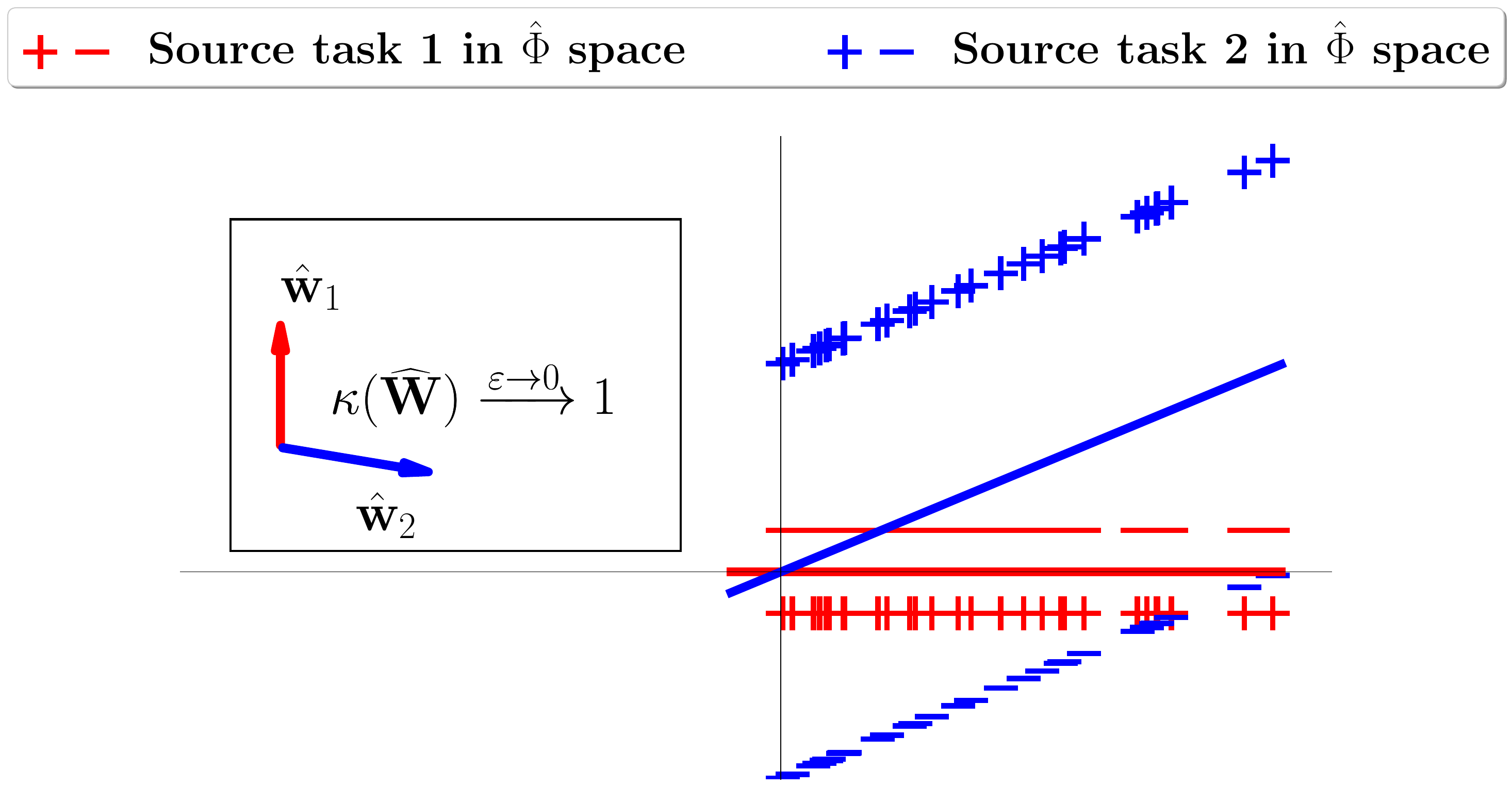}
    \caption{Visualization of the distributions used in the constructive example for the proof of \cref{prop:example}, with $\epsilon = 0.02$. In this example, $\kappa(\widehat{\rmW})$ is closer to 1 than $\kappa(\rmW^*)$. It shows that we can search for a representation $\widehat{\phi}$ such that optimal predictors in this space are fulfilling the assumptions, while solving the underlying problem equally well.}
    \label{fig:intuition}
\end{figure}

\noindent \textbf{Can we force the assumptions ?}\phantom{.} According to the previous result, satisfying the assumptions from MTR theory is expected to come in hand with better performance. However all the terms involved refer to optimal predictors, that we cannot act upon. Thus, we aim to answer the following question: \\
\vspace{-15pt}
\begin{center}
\begin{minipage}{.95\linewidth}
\begin{center}
    \textit{Given $\rmW^*$ such that $\kappa(\rmW^*)\gg 1$, can we learn $\widehat{\rmW}$ with $\kappa(\widehat{\rmW})\approx 1$ while solving the underlying classification problems equally well?}
\end{center}
\end{minipage}
\end{center}
\vspace{-2pt}
\noindent While obtaining such a result for any distribution seems to be very hard in the considered learning setup, we provide a constructive proof for the existence of a distribution for which the answer to the above-mentioned question is positive in the case of two tasks. The latter restriction comes out of the necessity to analytically calculate the singular values of $\rmW$ but we expect our example to generalize to more general setups and a larger number of tasks as well.
\begin{proposition}
Let $T=2$, $\sX \subseteq \sR^d$ be the input space and $\sY = \{-1,1\}$ be the output space. Then, there exist distributions $\mu_1$ and $\mu_2$ over $\sX \times \sY$, representations $\widehat{\phi}\neq \phi^*$ and matrices of predictors $\widehat{\rmW} \neq \rmW^*$ that satisfy the data generating model (\cref{eq:data_gen_model}) with $\kappa(\widehat{\rmW})\approx 1$ and $\kappa(\rmW^*) \gg 1$.
\label{prop:example}
\end{proposition}
\begin{proof_idea}
\rev{The construction used in this proof is illustrated in \cref{fig:intuition}. We define two uniform distributions $\mu_1$ and $\mu_2$ parametrized by a scalar $\varepsilon>0$ and find a tuple $(\rmW^*,\phi^*)$ that satisfies the data generating process from \cref{eq:data_gen_model}. To this end, $\phi^*$ is picked as a linear operator that projects the data generated by $\mu_i$ to a two-dimensional space by discarding its last $d-2$ dimensions and ensuring that $\kappa(\rmW^*) = \frac{1}{\varepsilon}$. The dependence of the ratio on $\varepsilon$ allows us to make it arbitrary large. Then, we construct another operator $\widehat{\phi}$ that projects the data generated by $\mu_i$ to a two-dimensional space by discarding the first and last $d-1$ dimensions. We then show that in this space the matrix of optimal predictors $\widehat{\rmW}$ has a ratio $\kappa(\widehat{\rmW}) \xrightarrow[]{\varepsilon \rightarrow 0} 1$.
}
\end{proof_idea}

The established results show that in some cases even when $\rmW^*$ does not satisfy Assumptions 1-2 in the $\phi^*$ space, it may still be possible to learn a new representation $\widehat{\phi}$ such that the optimal predictors in this space do satisfy them. The full proof of this result can be found in \cref{ax:meta_mtr_proofs}.

\noindent \textbf{How to force the assumptions ?}\phantom{.}  This can be done either by considering the constrained problem:
\begin{align}
    \widehat{\phi}, \widehat{\rmW} &= \argmin_{\phi \in \Phi, \rmW \in \sR^{T\times k}} \frac{1}{Tn_1}\sum_{t=1}^T\sum_{i=1}^{n_1} \ell(y_{t,i},\langle \rvw_t, \phi(\rvx_{t,i})\rangle) \notag \\
    \text{ s.t. } &\kappa(\rmW) = O(1), \|\rvw_t\| = 1,
    \label{eq:meta_train_hat}
\end{align}
or by using a more common strategy that consists in adding $\kappa(\rmW)$ and $\|\rmW\|_F^2$ directly as regularization terms:
\begin{equation}
    \widehat{\phi}, \widehat{\rmW} = \argmin_{\phi \in \Phi, \rmW \in \sR^{T\times k}} \frac{1}{Tn_1}\sum_{t=1}^T\sum_{i=1}^{n_1} \ell(y_{t,i},\langle \rvw_t, \phi(\rvx_{t,i})\rangle) + \lambda_1 \kappa(\rmW) + \lambda_2 \|\rmW\|_F^2.
    \label{eq:meta_train_reg}
\end{equation}

\rev{By adding $\|\rmW\|_F^2$ in the loss, we force the model to have a low norm on the weights. Since it cannot be put to 0 or below, the model will keep the norm relatively constant instead of increasing it. The second regularizer term is a softer way to apply the constraint on the norm rather than normalizing the weights as in \cref{eq:meta_train_hat}.}

\rev{According to Theorem 7.1 from \cite{lewisNonsmoothAnalysisSingular2005}, subgradients of singular values function are well-defined for absolutely symmetric functions. In our case, we are computing the squared singular values $\sigma^2(\rmW)$ and we retrieve the singular values by taking the square root. This means that effectively, we are computing $\kappa(\rmW) = \max(|\sigma(\rmW)|) / \min(|\sigma(\rmW)|)$, which is an absolutely symmetric function. Consequently, subgradients of the spectral regularization term $\kappa(\rmW)$ are well-defined and can be optimized efficiently when used in the objective function.}

To the best of our knowledge, such regularization terms based on insights from the advances in MTR theory have never been used in the literature before. 
\rev{We also further use the basic quantities involved in the proposed regularization terms as indicators of whether a given meta-learning algorithm naturally satisfies the assumptions ensuring an efficient meta-learning in practice or not.}


\subsection{Positioning with respect to Previous Work}\label{sec:related_work}

\rev{Below, we discuss several related studies aiming at improving the general understanding of meta-learning, and mention other regularization terms specifically designed for meta-learning.}\\

\textbf{Understanding meta-learning}\phantom{.} While a complete theory for meta-learning is still lacking, several recent works aim to shed light on phenomena commonly observed in meta-learning by evaluating different intuitive heuristics. For instance, \cite{Raghu2020Rapid} investigate whether \Maml{} 
works well due to rapid learning with significant changes in the representations when deployed on target task, or due to feature reuse where the learned representation remains almost intact. 
They establish that the latter factor is dominant and propose a new variation of \Maml{} that freezes all but task-specific layers of the neural network when learning new tasks. 
In \cite{goldblumUnravelingMetaLearningUnderstanding2020}, the authors explain the success of meta-learning approaches by their capability to either cluster classes more tightly in feature space (task-specific adaptation approach), or to search for meta-parameters that lie close in weight space to many task-specific minima (full fine-tuning approach). Finally, the effect of the number of shots on the FSC
accuracy was studied in \cite{CaoLF20} for \Proto{}. More recently, \cite{yeHowTrainYour2021a} studied the impact of the permutation of labels when training \Maml{}. 
They show that this key part of meta-training, severely impacts \Maml{}\ as it does not enjoy permutation invariance.
Our \rev{work} investigates a new aspect of meta-learning that has never been studied before and, unlike \cite{Raghu2020Rapid}, \cite{CaoLF20}, \cite{goldblumUnravelingMetaLearningUnderstanding2020} and \cite{yeHowTrainYour2021a}, provides a more complete experimental evaluation with the two different approaches of meta-learning. \\
\textbf{Normalization}\phantom{.} Multiple methods in the literature introduce a normalization of their features either to measure cosine similarity instead of Euclidean distance \cite{gidarisDynamicFewShotVisual2018, qiLowShotLearningImprinted2018, chenCLOSERLOOKFEWSHOT2019} or because of the noticed improvement in their performance \cite{wangSimpleShotRevisitingNearestNeighbor2019, tian2020rethinking, pmlr-v139-wang21ad}. 
In this chapter, we proved in \cref{sec:grad_vs_metric} above that for \Proto{} prototypes normalization is enough to achieve a good coverage of the embedding space, and we empirically show in \cref{sec:expe} below that it leads to better performance. Since we only normalize the prototypes and not all the features, we do not measure cosine similarity. Moreover, with our \cref{th:norm_proto}, we give explanations through MTR theory regarding the link between the coverage of the representation space and performance. \\
\textbf{Common regularization strategies}\phantom{.}  
In general, we note that regularization in meta-learning (i) is applied to either the weights of the whole neural network \cite{balajiMetaRegDomainGeneralization,Yin2020Meta-Learning}, or (ii) the predictions \cite{jamalTaskAgnosticMetaLearning2019, goldblumUnravelingMetaLearningUnderstanding2020} or (iii) is introduced via a prior hypothesis biased regularized empirical risk minimization \cite{pentinaPACBayesianBoundLifelong,kuzborskij2017fast,denevi2018incremental,denevi2018learning,DeneviCGP19}.
Contrary to the first group of methods and to the weight decay approach \cite{kroghSimpleWeightDecay1992}, we do not regularize the whole weight matrix learned by the neural network but the linear predictors of its last layer. While weight decay is used to avoid overfitting by penalizing large magnitudes of weights, our goal is to keep the classification margin unchanged during the training to avoid over-/under-specialization to source tasks. Similarly, spectral normalization proposed by \cite{MiyatoKKY18} does not affect the condition number $\kappa$. Second, we regularize the singular values of the matrix of linear predictors obtained in the last batch of tasks instead of the predictions used by the methods of the second group (\textit{e.g.}, using the theoretic-information quantities in \cite{jamalTaskAgnosticMetaLearning2019}). Finally, the works of the last group are related to the online setting with convex loss functions only and do not specifically target the spectral properties of the learned predictors.

\begin{figure}[tb]
    \centering
    \includegraphics[width=0.30\linewidth]{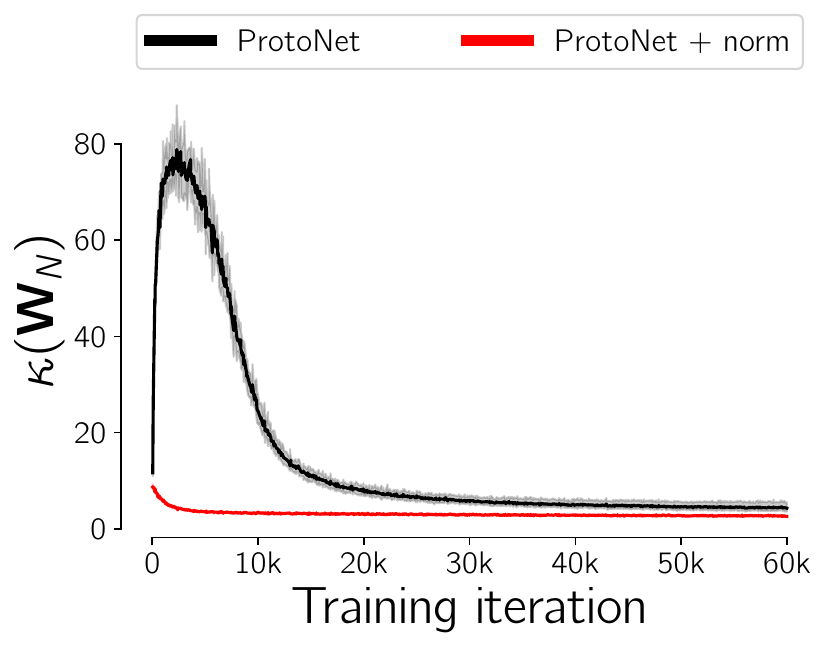}
    \includegraphics[width=0.30\linewidth]{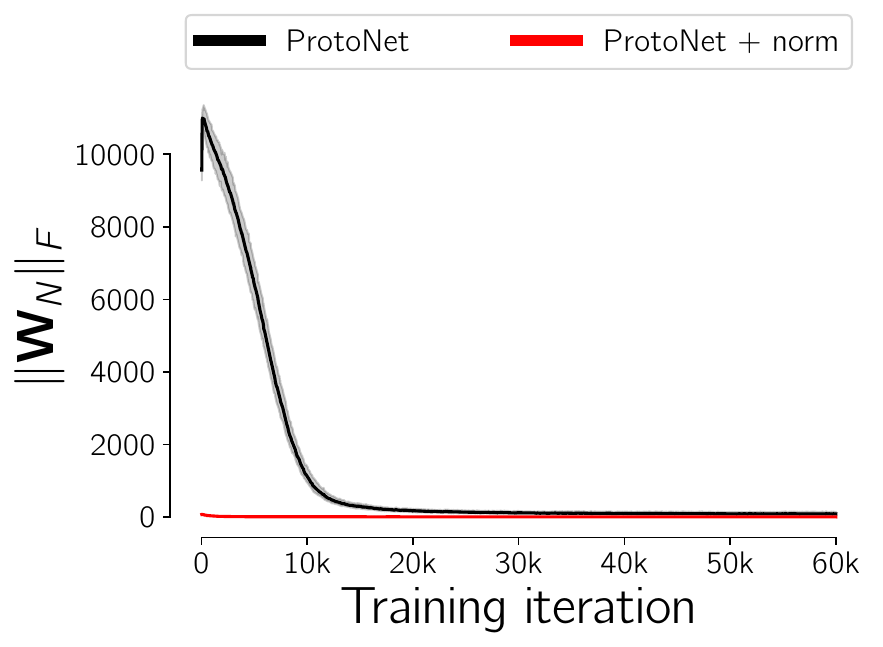} 
    \includegraphics[width =.30\linewidth]{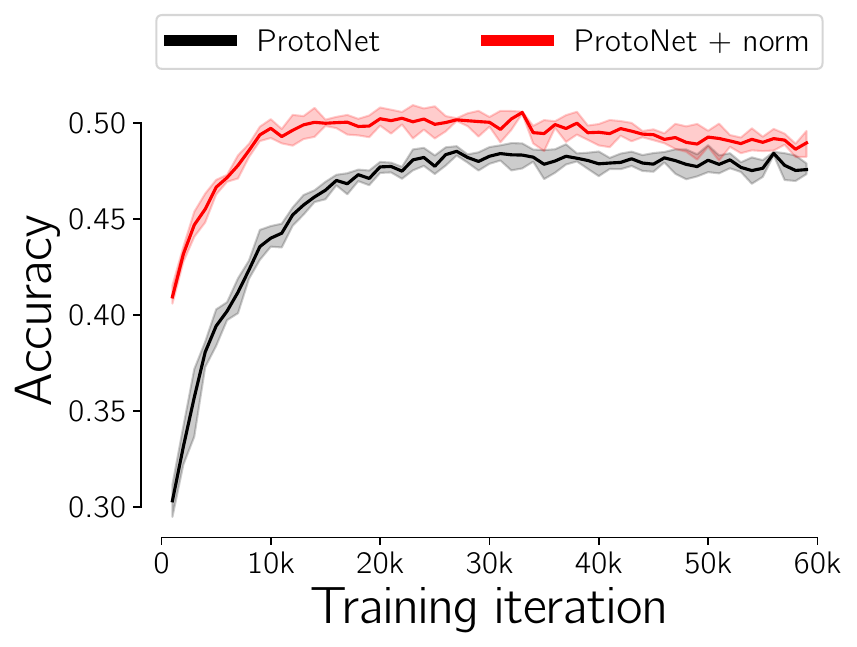}
    \includegraphics[width=0.30\linewidth]{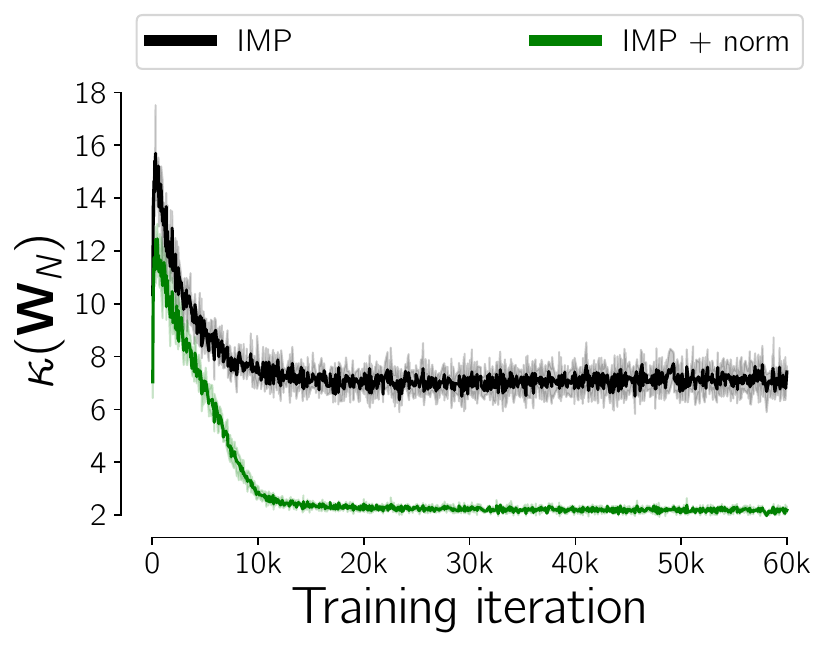}
    \includegraphics[width=0.30\linewidth]{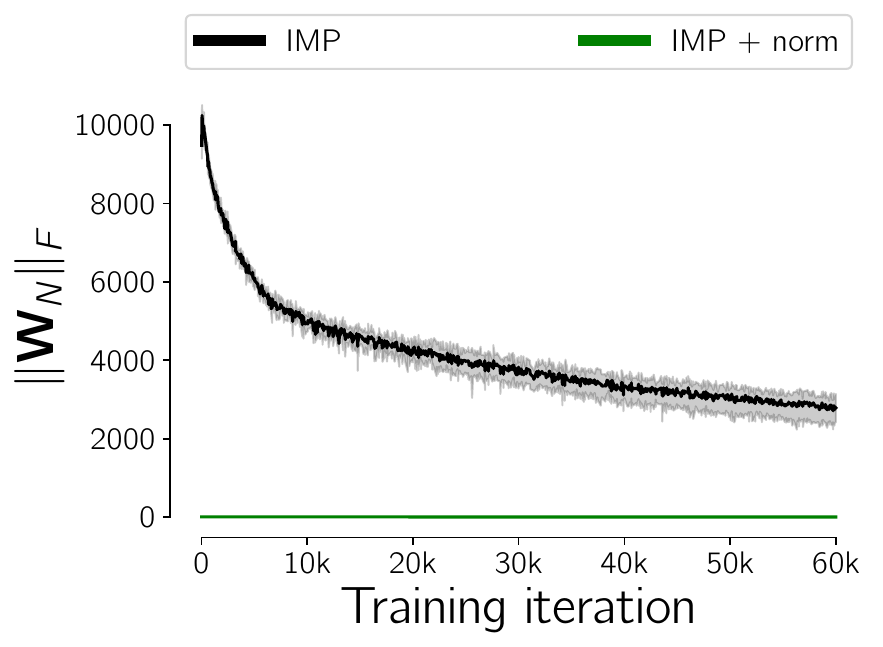}
    \includegraphics[width =.30\linewidth]{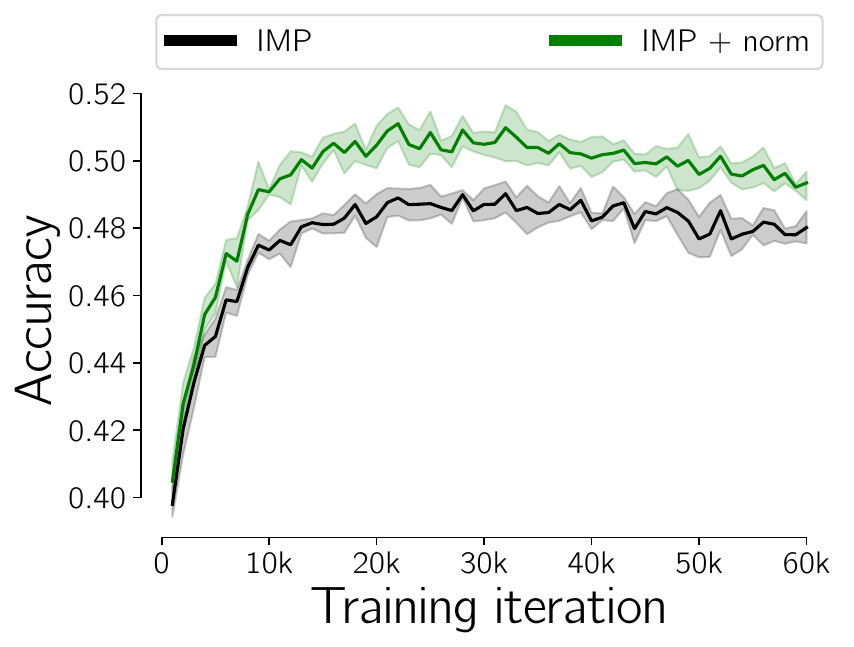} 
    \includegraphics[width=0.30\linewidth]{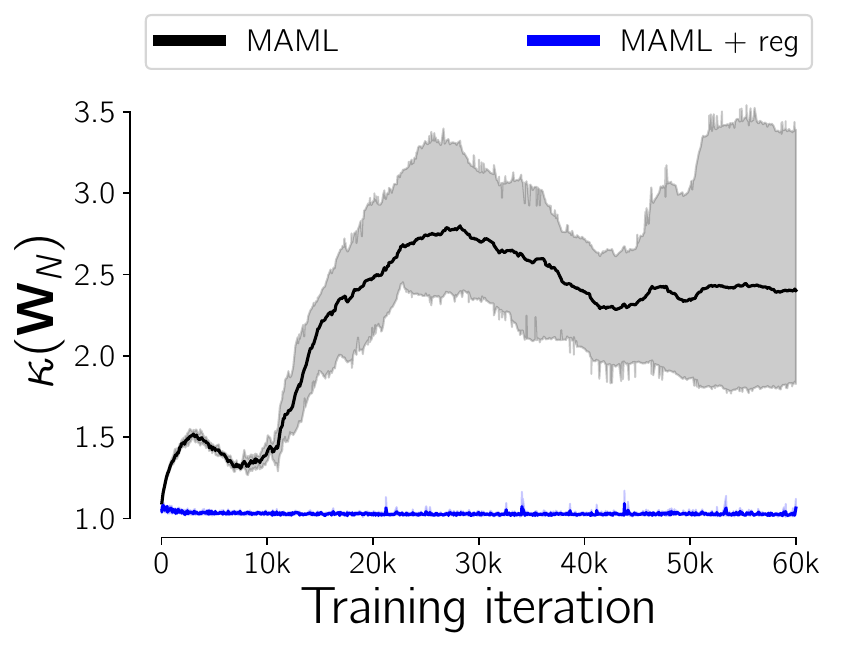}
    \includegraphics[width=0.30\linewidth]{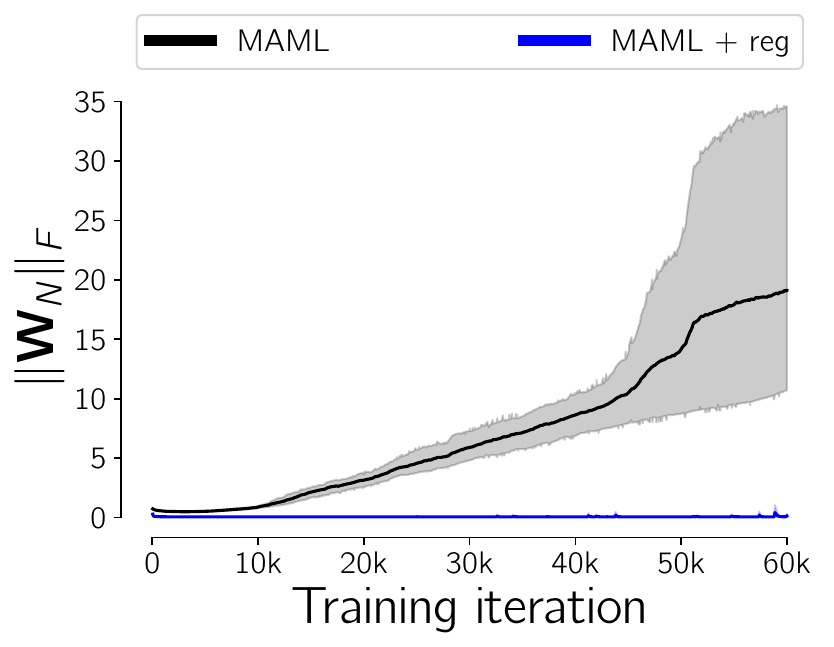}
    \includegraphics[width =.30\linewidth]{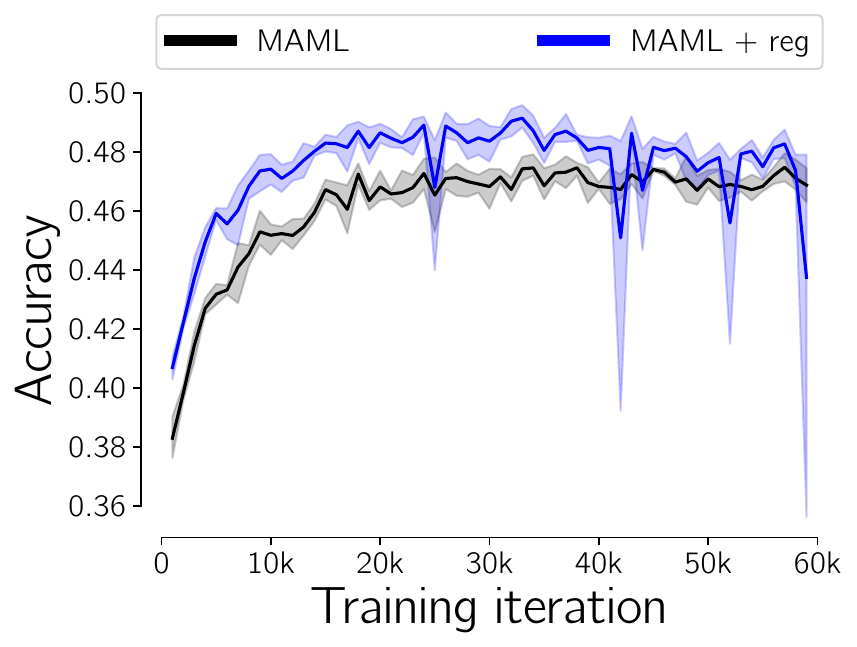}
    \includegraphics[width=0.30\linewidth]{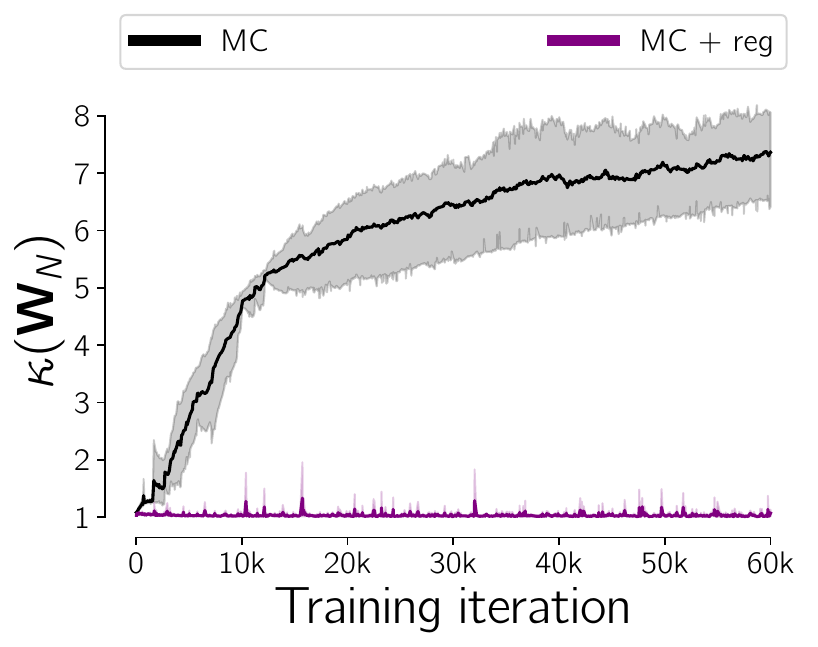}
    \includegraphics[width=0.30\linewidth]{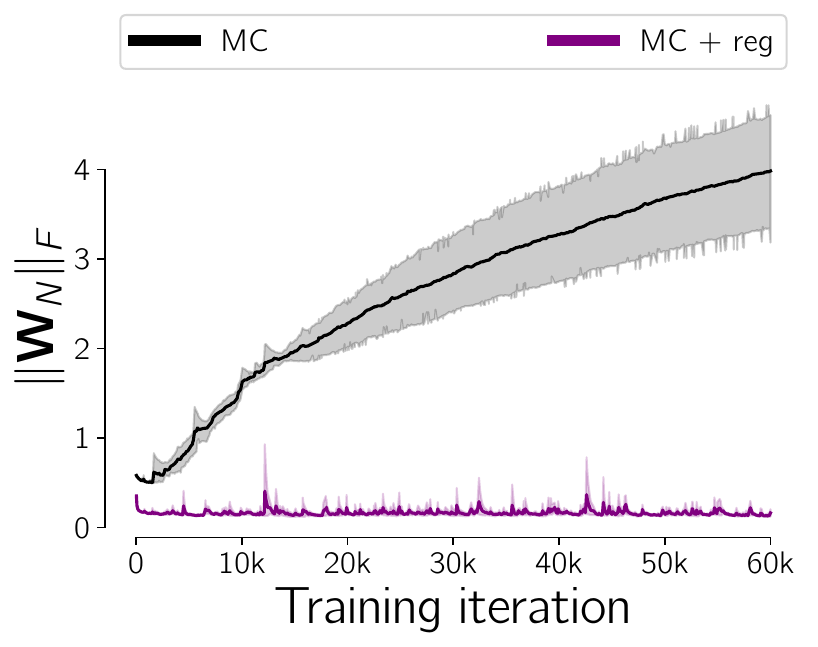}
    \includegraphics[width =.30\linewidth]{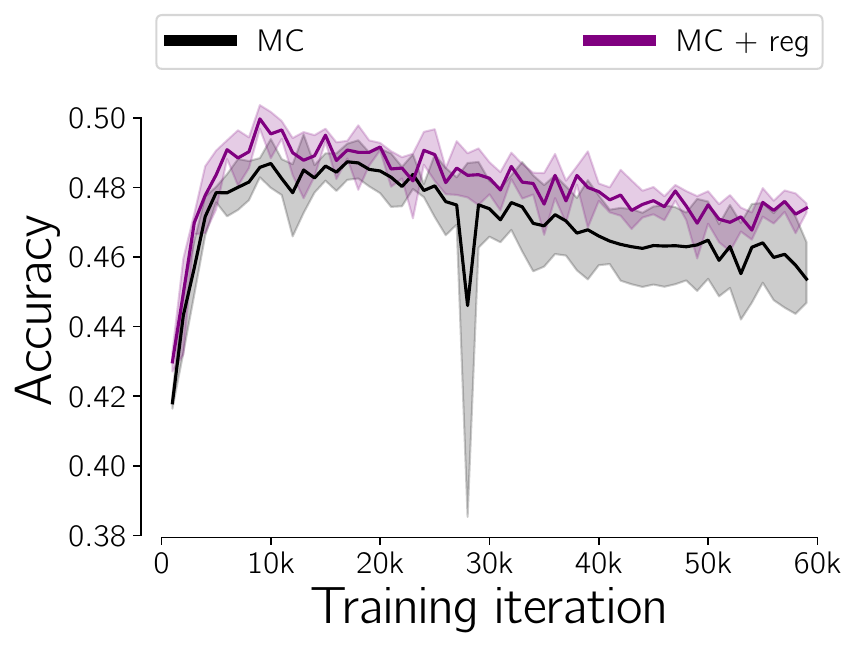}
   \caption{
    Evolution of $\kappa(\rmW_N)$ (\emph{left}), $\|\rmW_N\|_F$ (\emph{middle}) and the accuracy (\emph{right}) on 5-way 1-shot episodes from \emph{miniImageNet},
    for the metric-based methods \Proto{} (\emph{first row}) and \IMP{} (\emph{second row}), as well as the gradient-based methods \Maml{} (\emph{third row}) and \MC{} (\emph{last row}) and their regularized or normalized counterparts. 
    }
    \label{fig:acc_curves}
\end{figure}

\begin{figure}
    \centering
    \includegraphics[width=0.6\linewidth]{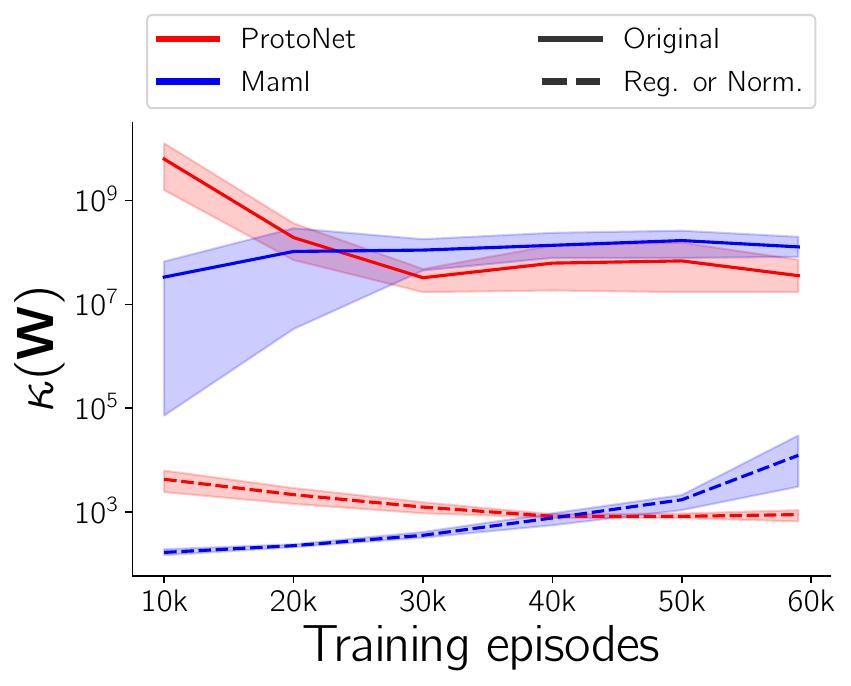}
    \caption{Evolution of $\kappa(\rmW)$ on 5-way 1-shot episodes from \emph{miniImageNet}, for \Proto{} (\emph{red, straight lines}), \Maml{} (\emph{blue, straight lines}) along with their regularized and normalized counterparts (\emph{respectively the same colors, in dashed lines}).}
    \label{fig:meta_mtr_kappa_full_reg}
\end{figure}

\section{Impact of enforcing theoretical assumptions}\label{sec:expe}

\rev{In this section, we investigate the impact of enforcing the aforementioned theoretical assumptions for meta-learning algorithms in practice.} 



\subsection{Experimental Setup}\label{sec:expe_setup}

\rev{
    We consider the FSC problem on the three benchmarks introduced in \cref{chap:sota} that we recall briefly here:
\begin{enumerate}
    \item \textbf{Omniglot} \cite{lakeHumanlevelConceptLearning2015} is a dataset of 20 instances of 1623 characters from 50 different alphabets. Each image was hand-drawn by different people. The images are resized to $28 \times 28$ pixels and the classes are augmented with rotations by multiples of $90$ degrees. 
    \item \textbf{miniImageNet} \cite{raviOPTIMIZATIONMODELFEWSHOT2017} is a dataset made from randomly chosen classes and images taken from the ILSVRC-12 dataset \cite{russakovskyImageNetLargeScale2015}. The dataset consists of $100$ classes and $600$ images for each class. The images are resized to $84 \times 84$ pixels and normalized.
    \item \textbf{tieredImageNet} \cite{renMetaLearningSemiSupervisedFewShot2018} is also a subset of ILSVRC-12 dataset. However, unlike miniImageNet, training classes are semantically unrelated to testing classes. The dataset consists of $779,165$ images divided into $608$ classes. Here again, the images are resized to $84 \times 84$ pixels and normalized.
\end{enumerate}
 }



\rev{
For each dataset, we follow a common experimental protocol used in \cite{finnModelAgnosticMetaLearningFast2017,chenCLOSERLOOKFEWSHOT2019}. \rev{More specifically, we} use a four-layer convolution backbone with 64 filters (C64) as done by \cite{chenCLOSERLOOKFEWSHOT2019} optimized with Adam~\cite{kingma2015adam} and a learning rate of $0.001$. On \emph{miniImageNet} and \emph{tieredImageNet}, models are trained on $60000$ 5-way 1-shot or 5-shot episodes and on $30000$ 20-way 1-shot or 5-shot episodes for \emph{Omniglot}. We use a batch size of $4$ and evaluate on the validation set every $1000$ episodes. We keep the best performing model on the validation set to evaluate on the test set. The seeds used for all experiments are $1$, $10$, $100$ and $1000$. For \Maml{}\ and \MC{}, we use an inner learning rate of $0.01$ for \emph{miniImageNet} and \emph{tieredImageNet}, and $0.1$ for \emph{Omniglot}. During training, we perform $5$ inner gradient step and $10$ step during testing. For all FSC experiments, unless explicitly stated, we use the regularization parameters $\lambda_1 = \lambda_2 = 1$ in \cref{eq:meta_train_reg}. We measure the performance using the top-1 accuracy with $95\%$ confidence intervals, reproduce the experiments with 4 different random seeds using a single NVIDIA V100 GPU, and average the results over $2400$ test tasks.}

\rev{We also provide experiments with the ResNet-12 architecture backbone \cite{leeMetaLearningDifferentiableConvex2019}. In this case, we follow the recent practice and initialize the models with the weights pretrained on the entire meta-training set \cite{ye2020fewshot, rusu2018meta, qiao2018few}. Like in their protocol, this initialization is updated by meta-training with \Proto{} or \Maml{} on at most $20 000$ episodes, grouping every $100$ episodes into an \emph{epoch}. Then, the best performing model on the validation set, evaluated at every epoch, is kept and the performance on $10 000$ test tasks is measured. For all experiments with the ResNet-12 architecture, the SGD optimizer with a weight decay of $0.0005$ and momentum of $0.9$ and a batch of episodes of size $1$ are used.
For \Proto{}, following the protocol of Ye et al. \cite{ye2020fewshot}, an initial learning rate of $0.0002$, decayed by a factor $0.5$ every $40$ epochs, is used. For \Maml{}, following Ye et Chao \cite{yeHowTrainYour2021a}, the initial learning rate is set to $0.001$, decayed by a factor $0.1$ every 20 epochs. The number of inner loop updates are respectively set to 15 and 20 with a step size of $0.05$ and $0.1$ for 1-shot and 5-shot episodes on the miniImageNet dataset, and respectively 20 and 15 with a step size of 0.001 and 0.05 on the tieredImageNet dataset.}


\begin{figure}[t]
    \centering
    \includegraphics[width=0.3\linewidth]{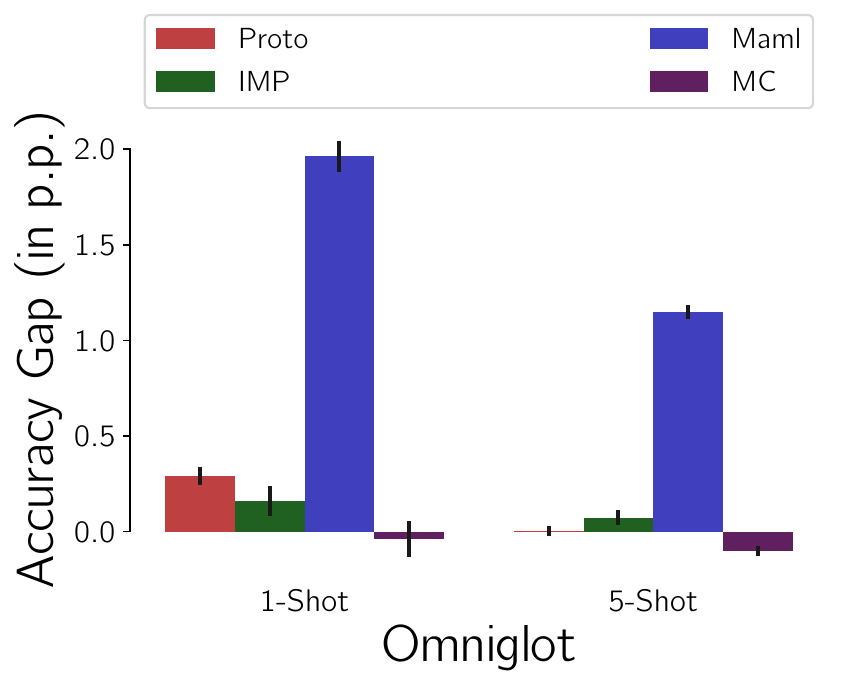}
    \includegraphics[width=0.3\linewidth]{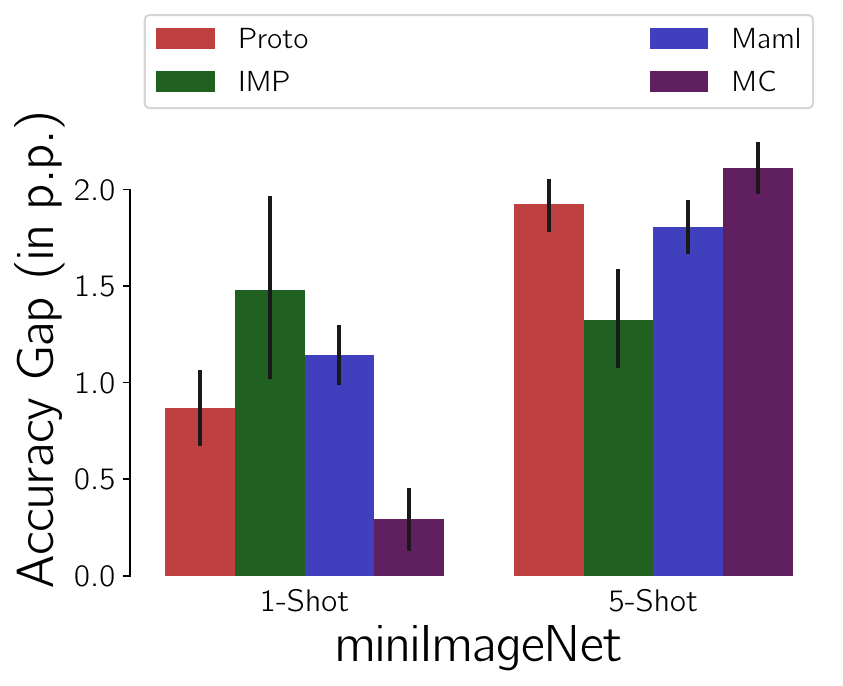}
    \includegraphics[width=0.3\linewidth]{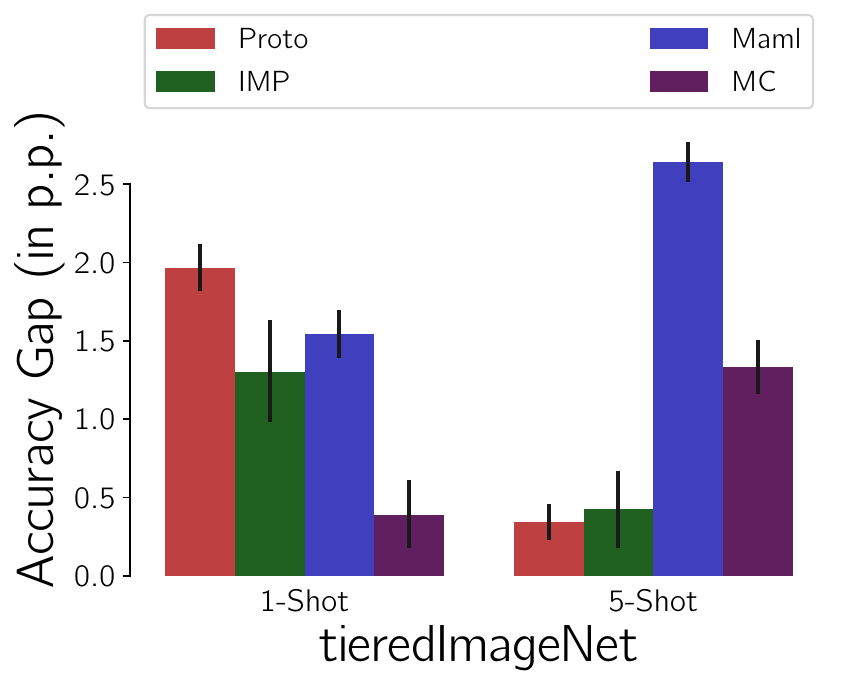}

    \caption{\small{Accuracy gap (in p.p.) when adding the normalization of prototypes for \Proto{}\ (\emph{red}) and \IMP{}\ (\emph{green}), and both spectral and norm regularization for \Maml{}\ (\emph{blue}) and \MC{}\ (\emph{purple}) enforcing the theoretical assumptions on Omniglot (\emph{left}), miniImagenet (\emph{middle}) and tieredImagenet (\emph{right}) datasets. \rev{Exact performance values are reported in \cref{ax:detailed_perf}.}
    }}
    \label{fig:barplots}
\end{figure}

\begin{table}[!htb]
    \centering
\resizebox{0.92\linewidth}{!}{%
    \begin{tabular}{@{}llcccc@{}} \toprule
    & & \multicolumn{2}{c}{miniImageNet 5-way} & \multicolumn{2}{c}{tieredImageNet 5-way} \\
    \cmidrule(lr){3-4}
    \cmidrule(lr){5-6}
    Model & Arch. & 1-shot & 5-shot & 1-shot & 5-shot \\
    \midrule
    \multicolumn{6}{l}{Gradient-based algorithms} \\
    \midrule
    \Maml{} \cite{finnModelAgnosticMetaLearningFast2017} & C64 & 48.70 $\pm$ 1.84 & 63.11 $\pm$ 0.92 &  - & - \\
    ANIL \cite{Raghu2020Rapid} & C64 & 48.0 $\pm$ 0.7 & 62.2 $\pm$ 0.5 & - & - \\
    Meta-SGD \cite{liMetaSGDLearningLearn2017} & C64 & \textbf{50.47 $\pm$ 1.87} & 64.03 $\pm$ 0.94 & - & - \\
    TADAM \cite{NEURIPS2018_66808e32} & R12 & 58.50 $\pm$ 0.30 & 76.70 $\pm$ 0.30 & - & - \\
    MC \cite{ParkO19} & R12 & 61.22 $\pm$ 0.10 & 75.92 $\pm$ 0.17 & \textbf{66.20 $\pm$ 0.10} & \textbf{82.21 $\pm$ 0.08} \\
    MetaOptNet \cite{leeMetaLearningDifferentiableConvex2019} & R12 & 62.64 $\pm$ 0.61 & 78.63 $\pm$ 0.46 & \underline{65.99 $\pm$ 0.72} & { \underline{81.56 $\pm$ 0.53}} \\
    MATE \cite{NEURIPS2020_8989e07f} & R12 & 62.08 $\pm$ 0.64 & 78.64 $\pm$ 0.46 & - & - \\
    \\
    \Maml{}\ (Ours) & C64 & 47.93 $\pm$ 0.83 & 64.47 $\pm$ 0.69 & 50.08 $\pm$ 0.91 & 67.5 $\pm$ 0.79 \\
    \Maml{}\ + reg. (Ours) & C64 & 49.15 $\pm$ 0.85 & \textbf{66.43 $\pm$ 0.69} & 51.5 $\pm$ 0.9 & 70.16 $\pm$ 0.76 \\
    \MC{}\ (Ours) & C64 & 49.28 $\pm$ 0.83 & 63.74 $\pm$ 0.69 & \underline{55.16 $\pm$ 0.94} & 71.95 $\pm$ 0.77 \\
    \MC{}\ + reg. (Ours) & C64 & \underline{49.64 $\pm$ 0.83} & \underline{65.67 $\pm$ 0.70} & \textbf{55.85 $\pm$ 0.94} & \textbf{73.34 $\pm$ 0.76} \\
    \Maml{}\ (Ours) & R12 & 63.52 $\pm$ 0.20 & 81.24 $\pm$ 0.14 & 63.96 $\pm$ 0.23 & \underline{81.79 $\pm$ 0.16} \\
    \Maml{}\ + reg. (Ours) & R12 & \textbf{64.04 $\pm$ 0.22} & \textbf{82.45 $\pm$ 0.14} & 64.32 $\pm$ 0.23 & \underline{81.28 $\pm$ 0.11} \\ 
    
    \midrule
    \multicolumn{6}{l}{Metric-based algorithms} \\
    \midrule
    \Proto{}\ \cite{snellPrototypicalNetworksFewshot2017} & C64 & 46.61 $\pm$ 0.78 & 65.77 $\pm$ 0.70 & -- & -- \\
    \IMP{}\ \cite{pmlr-v97-allen19b} & C64 & 49.6 $\pm$ 0.8 & \textbf{68.1 $\pm$ 0.8} & -- & -- \\
    SimpleShot \cite{wangSimpleShotRevisitingNearestNeighbor2019} & C64 & \underline{49.69 $\pm$ 0.19} & 66.92 $\pm$ 0.17 & 51.02 $\pm$ 0.20 & 68.98 $\pm$ 0.18 \\
    Relation Nets \cite{sungLearningCompareRelation2018} & C64 & \underline{50.44 $\pm$ 0.82} & 65.32 $\pm$ 0.70 & -- & -- \\
    SimpleShot \cite{wangSimpleShotRevisitingNearestNeighbor2019} & R18 & \textbf{62.85 $\pm$ 0.20} & \textbf{80.02 $\pm$ 0.14} & \textbf{69.09 $\pm$ 0.22} & \textbf{84.58 $\pm$ 0.16} \\
    CTM \cite{liFindingTaskRelevantFeatures2019} & R18 & \underline{62.05 $\pm$ 0.55} & 78.63 $\pm$ 0.06 & 64.78 $\pm$ 0.11 & 81.05 $\pm$ 0.52 \\
    DSN \cite{simonAdaptiveSubspacesFewShot2020} & R12 & \underline{62.64 $\pm$ 0.66} & 78.83 $\pm$ 0.45 & 66.22 $\pm$ 0.75 & 82.79 $\pm$ 0.48 \\
    \\
    \Proto{}\ (Ours) & C64 & 49.53 $\pm$ 0.41 & 65.1 $\pm$ 0.35 & 51.95 $\pm$ 0.45 & 71.61 $\pm$ 0.38 \\
    \Proto{} + norm. (Ours) & C64 & \underline{50.29 $\pm$ 0.41} & \underline{67.13 $\pm$ 0.34} & \textbf{54.05 $\pm$ 0.45} & \underline{71.84 $\pm$ 0.38} \\
    \IMP{}\ (Ours) & C64 & 48.85 $\pm$ 0.81 & 66.43 $\pm$ 0.71 & 52.16 $\pm$ 0.89 & 71.79 $\pm$ 0.75 \\
    \IMP{}\ + norm. (Ours) & C64 & \textbf{50.69 $\pm$ 0.8} & \underline{67.29 $\pm$ 0.68} & \underline{53.46 $\pm$ 0.89} & \textbf{72.38 $\pm$ 0.75} \\
    \Proto{}\ (Ours) & R12 & 59.25 $\pm$ 0.20 & 77.92 $\pm$ 0.14 & 41.39 $\pm$ 0.21 & 83.06 $\pm$ 0.16 \\
    \Proto{} + norm. (Ours) & R12 & \textbf{62.69 $\pm$ 0.20} & \textbf{80.95 $\pm$ 0.14} & \textbf{68.44 $\pm$ 0.23} & \textbf{84.20 $\pm$ 0.16} \\
 
    \bottomrule
    \end{tabular}%
    }
    \caption{Performance comparison of FSC models on FSC benchmarks. \emph{For a given architecture}, \textbf{bold} values are the highest accuracy and \underline{underlined values} are near-highest accuracies (at less than 1-point lower than the highest one).}
    \label{tab:sota_perfs}
\end{table}


\subsection{Metric-based Methods}\label{sec:metric}

\subsubsection{Using normalized class prototypes}

\begin{table}[ht]
    \centering
    \resizebox{0.95\textwidth}{!}{
    \begin{tabular}{@{}lllll@{}} \toprule
    \multirow{2}{*}{Dataset} & \multirow{2}{*}{Episodes} & without Norm., & with Norm.,  & with Norm.,  \\
     & & $\lambda_1 = 0$ & $\lambda_1 = 0$ & $\lambda_1 = 1$ \\
    \midrule
    \multirow{2}{*}{Omniglot} & 20-way 1-shot & $95.56 \pm 0.10\%$ & $\mathbf{95.89 \pm 0.10\%}$ & $91.90 \pm 0.14\%$ \\
     & 20-way 5-shot & $\mathbf{98.80 \pm 0.04\%}$ & $\mathbf{98.80 \pm 0.04\%}$ & $96.40 \pm 0.07\%$ \\
    \midrule
    \multirow{2}{*}{miniImageNet} & 5-way 1-shot & $49.53 \pm 0.41\%$ & $\mathbf{50.29 \pm 0.41\%}$ & $49.43 \pm 0.40\%$ \\
     & 5-way 5-shot & $65.10 \pm 0.35\%$ & $\mathbf{67.13 \pm 0.34\%}$ & $65.71 \pm 0.35\%$ \\
    \midrule
    \multirow{2}{*}{tieredImageNet} & 5-way 1-shot & $51.95 \pm 0.45\%$ & $\mathbf{54.05 \pm 0.45\%}$ & $53.54 \pm 0.44\%$ \\
     & 5-way 5-shot & $\mathbf{71.61 \pm 0.38\%}$ & $\mathbf{71.84 \pm 0.38\%}$ & $70.30 \pm 0.40\%$ \\
    \bottomrule
    \end{tabular}}
    \begin{center}
        \caption{Performance of \Proto{} with and without our regularization on the entropy and/or normalization. All accuracy results (in \%) are averaged over 2400 test episodes and 4 different random seeds and are reported with $95\%$ confidence interval. Further enforcing regularization on the singular values can be detrimental to performance.}
        \label{tab:proto_perf}
    \end{center}

\end{table}

\begin{table}[ht]
    \begin{center}
    \resizebox{0.95\textwidth}{!}{
    \begin{tabular}{@{}lllllllll@{}} \toprule
    Dataset & Episodes & Original & $\lambda_1 = 0$ & $\lambda_1 = 1$ & $\lambda_1 = 0.1$ & $\lambda_1 = 0.01$ & $\lambda_1 = 0.001$ & $\lambda_1 = 0.0001$ \\
    \midrule
    \multirow{2}{*}{miniImageNet} & 5-way 1-shot & $49.53 \pm 0.41\%$ & $\mathbf{50.29 \pm 0.41\%}$ & $49.43 \pm 0.40\%$ & $\mathbf{50.19 \pm 0.41\%}$ & $\mathbf{50.44 \pm 0.42\%}$ & $\mathbf{50.46 \pm 0.42\%}$ & $\mathbf{50.45 \pm 0.42\%}$ \\
    & 5-way 5-shot & $65.10 \pm 0.35\%$ & $\mathbf{67.13 \pm 0.34\%}$ & $65.71 \pm 0.35\%$ & $66.69 \pm 0.36\%$ & $66.69 \pm 0.34\%$ & $\mathbf{67.2 \pm 0.35\%}$ & $\mathbf{67.12 \pm 0.35\%}$ \\
    \midrule
    \multirow{2}{*}{Omniglot} & 20-way 1-shot & $95.56 \pm 0.10\%$ & $\mathbf{95.89 \pm 0.10\%}$ & $91.90 \pm 0.14\%$ & $94.38 \pm 0.12\%$  & $95.60 \pm 0.10\%$ & $95.7 \pm 0.10\%$ & $95.77 \pm 0.10\%$ \\
     & 20-way 5-shot & $98.80 \pm 0.04\%$ & $98.80 \pm 0.04\%$ & $96.40 \pm 0.07\%$ & $97.93 \pm 0.05\%$ & $98.62 \pm 0.04\%$ & $98.76 \pm 0.04\%$ & $\mathbf{98.91 \pm 0.03\%}$ \\
    \bottomrule
    \end{tabular}
    }
    \end{center}
    \caption{Ablative study on the strength of the regularization with normalized \Proto{}. All accuracy results (in \%) are averaged over 2400 test episodes and 4 random seeds and are reported with $95\%$ confidence interval.}
    \label{tab:hparam_proto}
\end{table}



\cref{th:norm_proto} tells us that with normalized class prototypes that act as linear predictors, \Proto{}\ naturally decreases the condition number of their matrix. Furthermore, since the prototypes are directly the image features, adding a regularization term on the norm of the prototypes makes the model collapse to the trivial solution which maps all images to 0.
To this end, we choose to ensure the theoretical assumptions for metric-based methods (\Proto{}\ and \IMP{}) only with the prototype normalization: \\
\begin{align}
    \widehat{\phi}, \widehat{\rmW} = \argmin_{\phi \in \Phi, \rmW \in \sR^{T\times k}} \frac{1}{Tn_1}\sum_{t=1}^T\sum_{i=1}^{n_1} \ell(y_{t,i},\langle \tilde{\rvw}_t, \phi(\rvx_{t,i})\rangle),
    \label{eq:meta_train_metric}
\end{align}    

where $\tilde{\rvw} = \frac{\rvw}{\|\rvw\|}$ are the normalized prototypes. 
According to \cref{th:norm_proto}, the normalization of the prototypes makes the problem similar to the constrained problem given in \cref{eq:meta_train_hat}.

As can be seen in \cref{fig:acc_curves,fig:meta_mtr_kappa_full_reg}, the normalization of the prototypes has the intended effect on the condition number of the matrix of predictors. Indeed, $\kappa(\rmW_N)$ (\emph{left}) stay constant and low during training, and we achieve a much lower $\kappa(\rmW)$ (\emph{right}) than without normalization.

From \cref{fig:barplots}, we note that normalizing the prototypes from the very beginning of the training process has an overall positive effect on the obtained performance, and this gain is statistically significant in most of the cases according to the \emph{Wilcoxon signed-rank} test ($p < 0.05$) \cite{wilcoxon, David1956NonparametricSF}.
In \cref{tab:sota_perfs}, we compare the performance obtained against state-of-the-art algorithms behaving similarly to Instance Embedding algorithms~\cite{ye2020fewshot} such as \Proto{}, depending on the architecture used. Even with a ResNet-12 architecture, the proposed normalization still improves the performance to reach competitive results with the state-of-the-art. On the miniImageNet 5-way 5-shot benchmark, our normalized \Proto{}\ achieves 80.95\%, better than DSN (78.83\%), CTM (78.63\%) and SimpleShot (80.02\%).

\subsubsection{Further enforcing a low condition number on Metric-based methods}\label{sec:entropy}

\rev{To guide the model into learning an encoder with the lowest condition number, we consider adding $\kappa(\rmW_N)$ as a regularization term in \cref{eq:meta_train_metric}. In addition to the normalization of the prototypes, this should further enforce the assumption on the condition number. Unfortunately, this latter strategy hinders the convergence of the network and leads to numerical instabilities. It is most likely explained by prototypes being computed from image features which suffer from rapid changes across batches, making the smallest singular value $\sigma_N(\rmW_N)$ close to $0$.
Consequently, we propose to replace the condition number as a regularization term by the \emph{negative entropy of the vector of singular values} as follows:
\begin{align}
    & H_\sigma(\rmW_N):= \sum_{i=1}^N \softmax(\mathbf{\sigma}(\rmW_N))_i \cdot \log \softmax(\mathbf{\sigma}(\rmW_N))_i,
\end{align}
where $\softmax(\cdot)_i$ is the $i^{th}$ output of the \emph{softmax} function. Since uniform distribution has the highest entropy, regularizing with $\kappa(\rmW_N)$ or $H_\sigma(\rmW_N)$ leads to a better coverage of $\sR^k$ by ensuring a nearly identical importance regardless of the direction.}

\noindent
\rev{We obtain the following regularized optimization problem:
\begin{align}
    \widehat{\phi}, \widehat{\rmW} &= \argmin_{\phi \in \Phi, \rmW \in \sR^{T\times k}} \frac{1}{Tn_1}\sum_{t=1}^T\sum_{i=1}^{n_1} \ell(y_{t,i},\langle \tilde{\rvw}_t, \phi(\rvx_{t,i})\rangle) + \lambda_1 H_\sigma(\rmW),
    \label{eq:meta_train_reg_metric}
\end{align}
where $\tilde{\rvw} = \frac{\rvw}{\|\rvw\|}$ are the normalized prototypes.}

\rev{In \cref{tab:proto_perf}, we report the performance of \Proto{}\ without normalization, with normalization and with both normalization and regularization on the entropy. Finally, we can see that further enforcing a regularization on the singular values through the entropy does not help the training since \Proto{}\ naturally learns to minimize the singular values of the prototypes. In \cref{tab:hparam_proto}, we show that reducing the \emph{strength} of the regularization with the entropy can help to improve the performance.}

\subsection{Gradient-based Methods}\label{sec:gradient}

\begin{table}[ht]
    \begin{subtable}{0.45\textwidth}
    \centering
    \resizebox{0.95\textwidth}{!}{
    \begin{tabular}{@{}lll@{}} \toprule
     & $\lambda_1 = 0$ & $\lambda_1 = 1$ \\
    \midrule
    \multirow{2}{*}{$\lambda_2 = 0$} & $91.72 \pm 0.29\%$ & $89.86 \pm 0.31 \%$ \\
    & $97.07 \pm 0.14\%$ & $72.47 \pm 0.17 \%$ \\
    \midrule
    \multirow{2}{*}{$\lambda_2 = 1$} & $92.80 \pm 0.26 \%$ & $\mathbf{95.67 \pm 0.20 \%}$ \\
    & $96.99 \pm 0.14 \%$ & $\mathbf{98.24 \pm 0.10\%}$ \\
    \bottomrule
    \end{tabular}}
    \subcaption{}
    \end{subtable}
    \hfill
    \begin{subtable}{0.45\textwidth}
    \centering
    \resizebox{0.95\textwidth}{!}{
    \begin{tabular}{@{}lll@{}} \toprule
     & $\lambda_1 = 0$ & $\lambda_1 = 1$ \\
    \midrule
    \multirow{2}{*}{$\lambda_2 = 0$} & $47.93 \pm 0.83\%$ & $47.76 \pm 0.84\%$ \\
    & $64.47 \pm 0.69\%$ & $64.44 \pm 0.68\%$ \\
    \midrule
    \multirow{2}{*}{$\lambda_2 = 1$} & $48.27 \pm 0.81\%$ & $\mathbf{49.16 \pm 0.85\%}$ \\
    & $64.16 \pm 0.72\%$ & $\mathbf{66.43 \pm 0.69\%}$ \\
    \bottomrule
    \end{tabular}}
    \subcaption{}
    \end{subtable}
    \caption{
    \rev{Ablative study of the regularization parameter for \Maml, on Omniglot \textbf{(a)} with 20-way 1-shot (\emph{top values}) and 20-way 5-shot (\emph{bottom values}) episodes, and miniImageNet \textbf{(b)} with 5-way 1-shot (\emph{top values}) and 5-way 5-shot (\emph{bottom values}) episodes. All accuracy results (in \%) are averaged over 2400 test episodes and 4 different random seeds and are reported with $95\%$ confidence interval. We can see that in all cases, using both regularization terms is important.}}
    \label{tab:maml_ablative}
\end{table}

Gradient-based methods learn a batch of linear predictors for each task, and we can directly take them as $\rmW_N$ to compute its SVD. 
In the following experiments, we consider the regularized problem of \cref{eq:meta_train_reg} for \Maml{} as well as Meta-Curvature (\MC{}). 
As expected, the dynamics of $\|\rmW_N\|_F$ and $\kappa(\rmW_N)$ during the training of the regularized methods remain bounded and the effect of the regularization is confirmed with the lower value of $\kappa(\rmW)$ achieved (cf. \cref{fig:acc_curves}).

The impact of our regularization on the results is quantified in \cref{fig:barplots} where a statistically significant accuracy gain is achieved in most cases, according to the \emph{Wilcoxon signed-rank} test ($p < 0.05$) \cite{wilcoxon, David1956NonparametricSF}.
In \cref{tab:sota_perfs}, we compare the performance obtained to state-of-the-art gradient-based algorithms. We can see that our proposed regularization is globally improving the results, even with a bigger architecture such as ResNet-12 and with an additional pretraining. On the miniImageNet 5-way 5-shot benchmark, with our regularization \Maml{} achieves 82.45\%, better than TADAM (76.70\%), MetaOptNet (78.63\%) and MATE with MetaOptNet (78.64\%).
\rev{In Table~\ref{tab:maml_ablative}, we compare the performance of \Maml{} without regularization ($\lambda_1 = \lambda_2 = 0$), with a regularization on the condition number $\kappa(\rmW_N)$ ($\lambda_1 = 1$ and $\lambda_2 = 0$), on the norm of the linear predictors ($\lambda_1 = 0$ and $\lambda_2 = 1$), and with both regularization terms ($\lambda_1 = \lambda_2 = 1$) on Omniglot and miniImageNet. We can see that both regularization terms are important in the training and that using only a single term can be detrimental to the performance.}


\subsection{Multi-Task learning Methods}\label{sec:mtl}

\begin{table}
    \centering
\resizebox{0.90\linewidth}{!}{
\begin{tabular}{@{}llcccc@{}} \toprule
    & & \multicolumn{2}{c}{miniImageNet 5-way} & \multicolumn{2}{c}{tieredImageNet 5-way} \\
    \cmidrule(lr){3-4}
    \cmidrule(lr){5-6}
    Model & Arch. & 1-shot & 5-shot & 1-shot & 5-shot \\
    \midrule
    MTL & R12 & 55.73 $\pm$ 0.18 & 76.27 $\pm$ 0.13 & 62.49 $\pm$ 0.21 & 81.31 $\pm$ 0.15 \\
    MTL + norm. & R12 & 59.49 $\pm$ 0.18 & \textbf{77.3 $\pm$ 0.13} & \textbf{66.66 $\pm$ 0.21} & \textbf{83.59 $\pm$ 0.14} \\
    MTL + reg. (Ours) & R12 & \textbf{61.12 $\pm$ 0.19} & \underline{76.62 $\pm$ 0.13} & \underline{66.28 $\pm$ 0.22} & 81.68 $\pm$ 0.15 \\

    \bottomrule
    \end{tabular}%
    }
    \caption{Performance comparison of MTL models \cite{pmlr-v139-wang21ad} on FSC benchmarks with the ResNet-12 backbone architecture. \textbf{Bold} values are the highest accuracy and \underline{underlined values} are near-highest accuracies (at less than 1-point lower than the highest one).}
    \label{tab:mtl_perfs}
\end{table}

\begin{table}[!ht]
    \begin{center}
    \resizebox{0.95\linewidth}{!}{
    \begin{tabular}{@{}llllllllll@{}} \toprule
    $\lambda_1$ & 1 & 0.8 & 0.6 & 0.4 & 0.2 & 0.1 & 0.05 & 0.01 & 0 \\
    Accuracy & \multirow{2}{*}{75.84} & \multirow{2}{*}{75.85} & \multirow{2}{*}{76.02} & \multirow{2}{*}{76.11} & \multirow{2}{*}{76.15} & \multirow{2}{*}{75.99} & \multirow{2}{*}{75.65} & \multirow{2}{*}{75.08} & \multirow{2}{*}{74.64} \\
    ($\lambda_2 = 1$) & & & & & & & & & \\
    \midrule
    $\lambda_2$ & 1 & 0.8 & 0.6 & 0.4 & 0.2 & 0.1 & 0.05 & 0.01 & 0 \\
    Accuracy & \multirow{2}{*}{75.84} & \multirow{2}{*}{76.09} & \multirow{2}{*}{75.81} & \multirow{2}{*}{76.28} & \multirow{2}{*}{76.23} & \multirow{2}{*}{76.1} & \multirow{2}{*}{76.25} & \multirow{2}{*}{\textbf{76.42}} & \multirow{2}{*}{76.06} \\
    ($\lambda_1 = 1$) & & & & & & & & & \\
    \bottomrule
    \end{tabular}
    }
    \end{center}
    \caption{\rev{Performance of MTL~\cite{pmlr-v139-wang21ad} when varying either $\lambda_1$ or $\lambda_2$, the other being fixed to 1, on the miniImageNet 5-way 5-shot benchmark. All accuracy results (in \%) are averaged over 2000 test episodes on a single random seed.}
    }
    \label{tab:hparam_mtl}
\end{table}

We implement our regularization on a recent Multi-Task Learning (MTL) method \cite{pmlr-v139-wang21ad}, following the same experimental protocol.
The objective is to empirically validate our analysis on a method using the MTR framework.
As mentioned in \cref{sec:related_work}, the authors introduce feature normalization in their method, speculating that it improves coverage of the representation space \cite{pmlr-v119-wang20k}. Using their code, we reproduce their experiments on three different settings compared in \cref{tab:mtl_perfs}: the vanilla MTL, the MTL with feature normalization, and MTL with our proposed regularization on the condition number and the norm of the linear predictors. We use $\lambda_1 = 1$ in all the settings, and $\lambda_2 = 1$ in the 1-shot setting and $\lambda_2 = 0.01$ in the 5-shot settings. Our regularization, as well as the normalization, globally improve the performance over the non-normalized models. Notably, our regularization is the most effective when there is the less data which is well-aligned with the MTR theory in few-shot setting. We can also note that in most of the cases, the normalized models and the regularized ones achieve similar results, hinting that they may have a similar effect.

\rev{\cref{tab:hparam_mtl} presents the effect of varying independently either parameter $\lambda_1$ or $\lambda_2$ in the regularization, the other being fixed to 1. From these results, we can see that performance is much more impacted by the condition number regularization (parameter $\lambda_1$) than by the normalization (parameter $\lambda_2$). Indeed, varying the regularization weight can lead from the lowest accuracy (74.64\%, for $\lambda_1=0$) to one of the highest accuracies (76.15\% for $\lambda_1=0.2$).}

All of these results show that our analysis and our proposed regularization are also valid in the MTL framework. 

\subsection{\rev{Out-of-domain Analysis}}\label{sec:ood}

\begin{figure}
    \begin{center}
        \includegraphics[width = .24\linewidth]{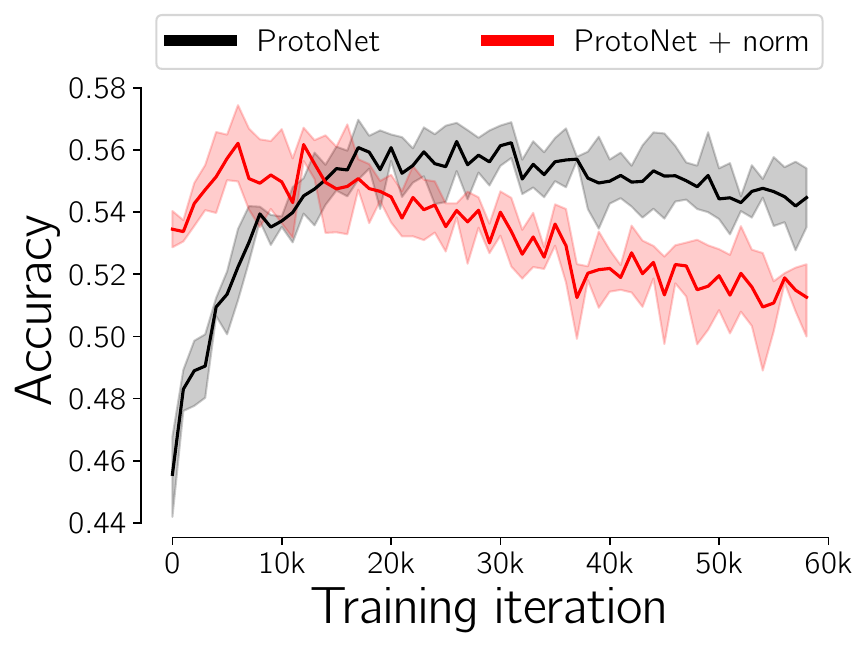}
        \includegraphics[width = .24\linewidth]{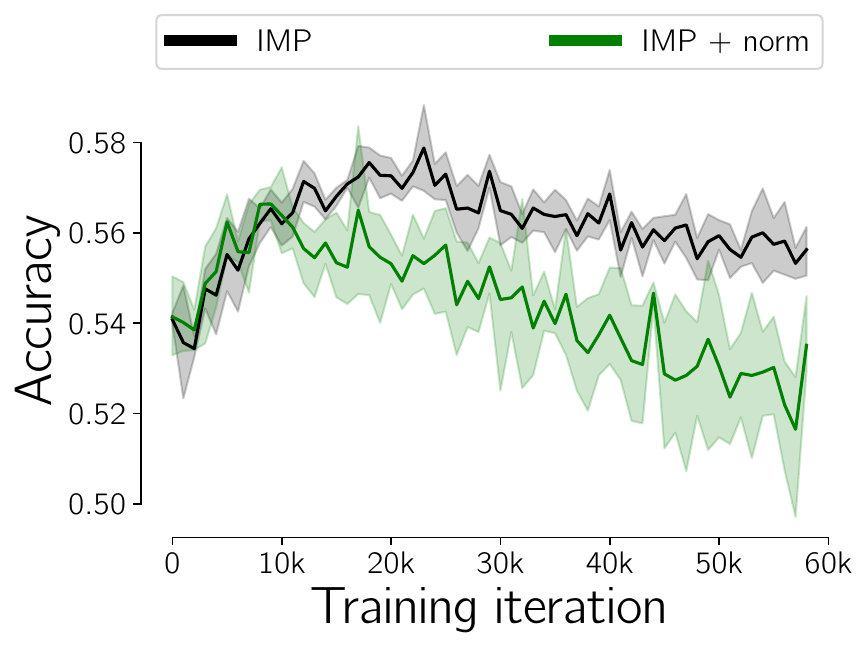}
        \includegraphics[width = .24\linewidth]{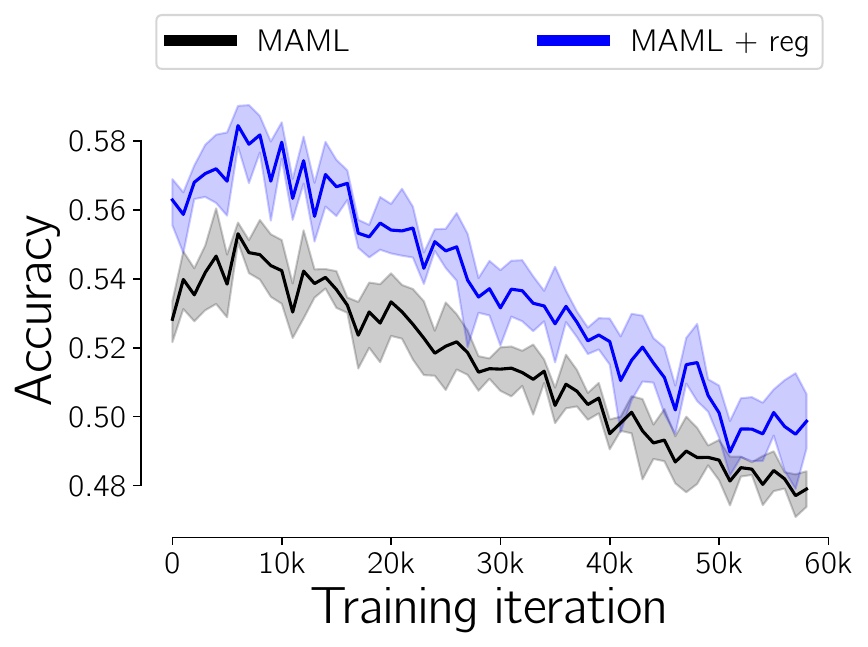}
        \includegraphics[width = .24\linewidth]{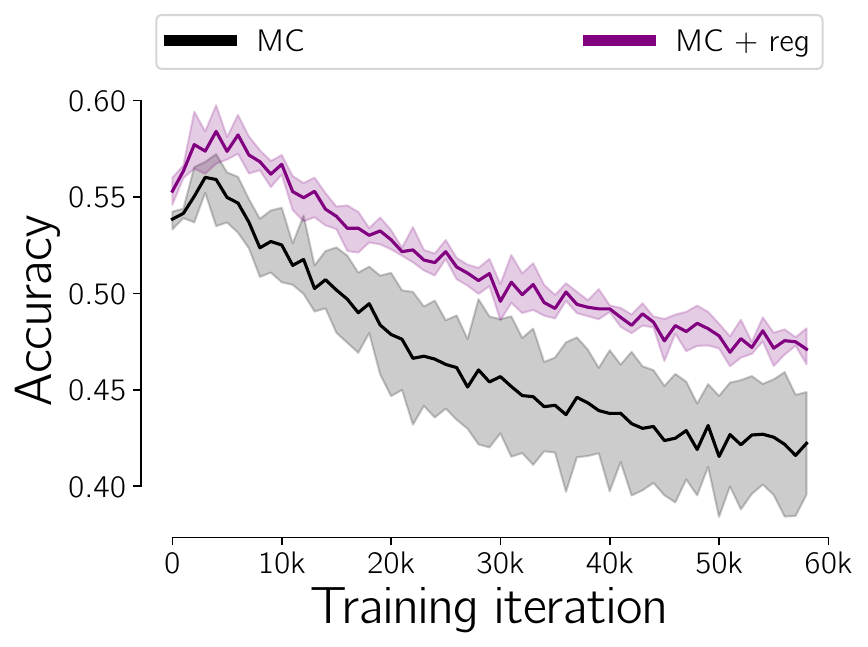}
        
        \includegraphics[width = .24\linewidth]{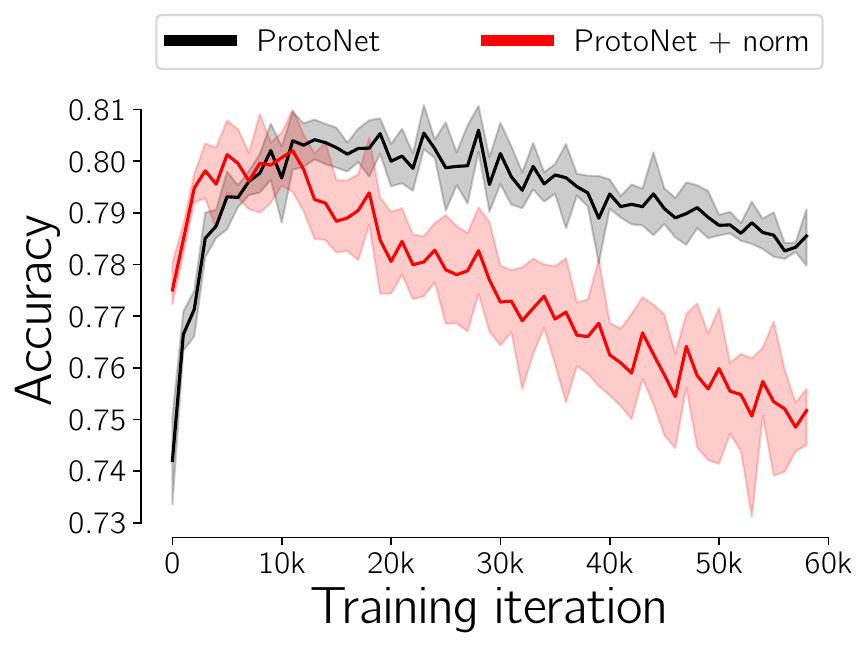}
        \includegraphics[width = .24\linewidth]{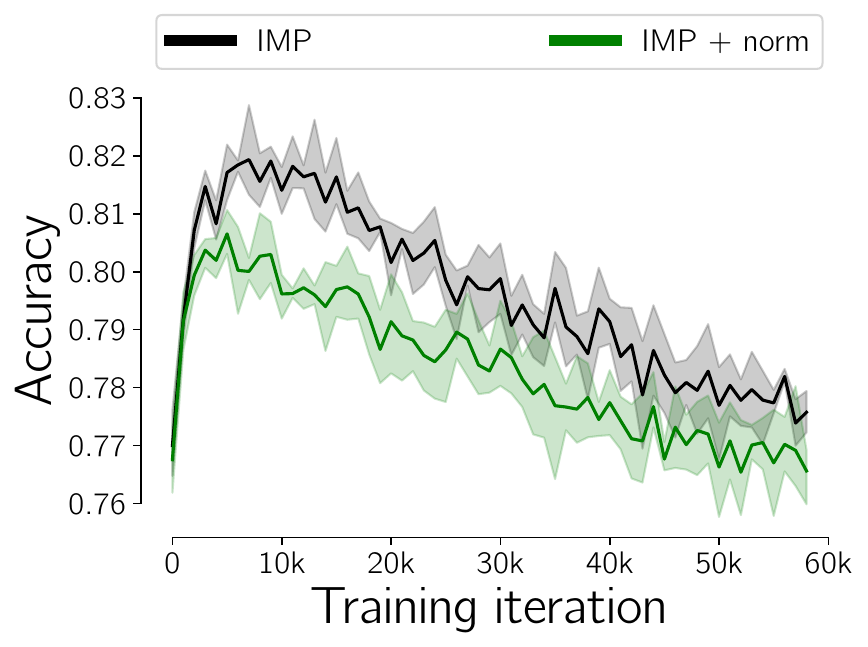}
        \includegraphics[width = .24\linewidth]{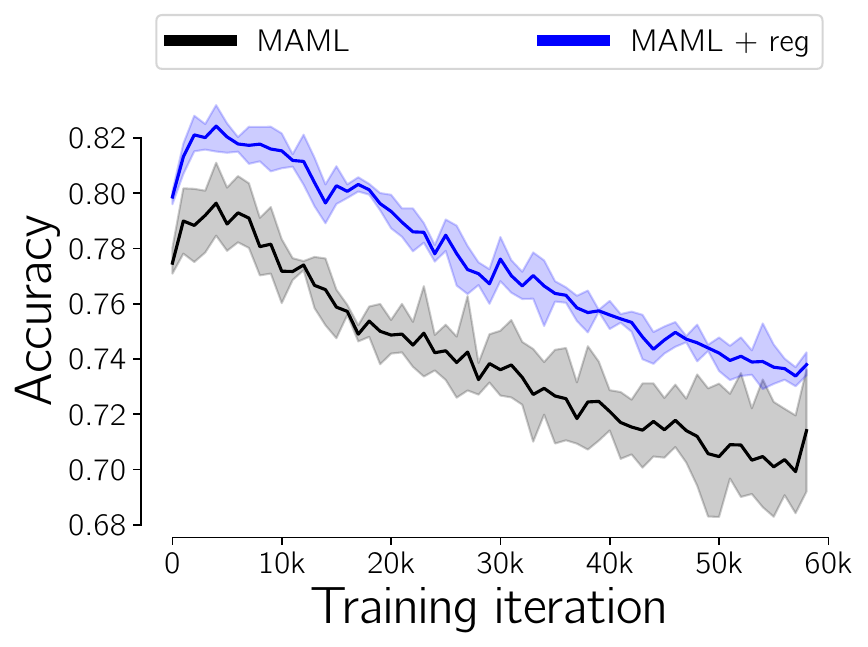}
        \includegraphics[width = .24\linewidth]{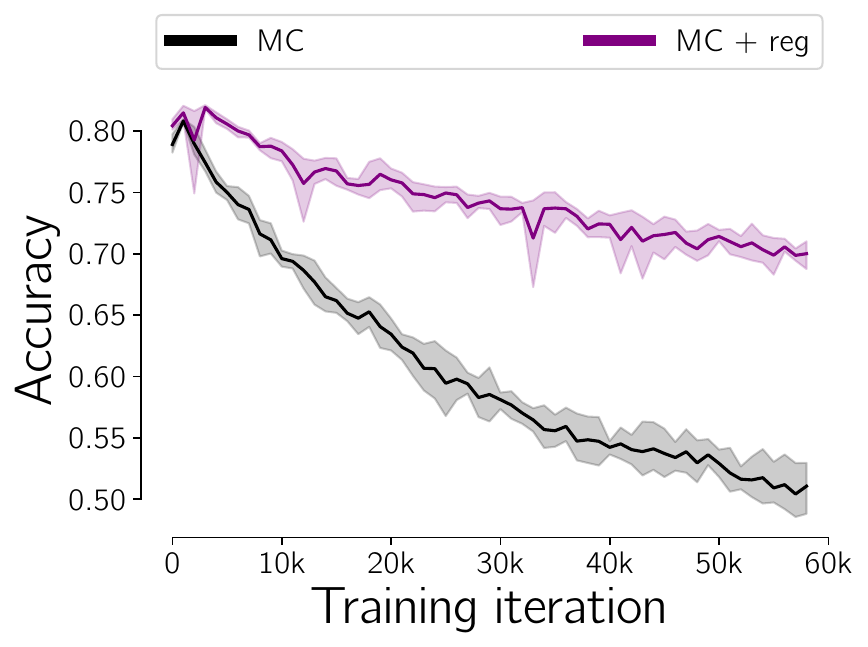}

        \captionof{figure}{
        Evolution of accuracy of 5-way 1-shot (\emph{top}, resp. 5-way 5-shot, \emph{bottom}) meta-test episodes from \emph{CropDisease} during meta-training on 5-way 1-shot (\emph{top}, resp. 5-way 5-shot, \emph{bottom}) episodes from \emph{miniImageNet},
        for \Proto, \IMP, \Maml, \MC\ (\emph{from left to right}) and their regularized or normalized counterparts (\emph{in red, green, blue and purple, respectively}).
        All results are averaged over 4 different random seeds. The shaded areas show $95\%$ confidence intervals.}
        \label{fig:acc_curves_ood}
        \end{center}
    \end{figure}
\rev{
In the theoretical MTR framework, one additional critical assumption made is that 
the task data marginal distributions 
are similar\footnote{We refer the reader to \cref{ax:MTR} for more information on this assumption}, which does not hold in a \emph{cross-domain} setting, where we evaluate a model on a 
dataset 
different from the training dataset. In this setting, we do not have the same guarantees that our regularization or normalization schemes will be as effective as in \emph{same-domain}.
To verify this, we measure out-of-domain performance on the \textbf{CropDiseases} dataset~\cite{mohantyUsingDeepLearning2016} adopted by \cite{guoBroaderStudyCrossDomain2020a}. Following their protocol, this dataset is used only for testing purposes. In this specific experiment, evaluated models are trained on \emph{miniImageNet}.
}
 
\rev{
On the one hand, for metric-based methods, the improvement in the \emph{same-domain} setting does not translate to the \emph{cross-domain} setting. From \cref{fig:acc_curves_ood}, we can see that even though the low condition number in the beginning of training leads to improved early generalization capabilities of \Proto{}, this is not the case for \IMP{}. We attribute this discrepancy between \Proto{} and \IMP{} to a difference 
in cluster radius parameters of
\IMP{} and normalized \IMP{}, making the encoder less adapted to \emph{out-of-domain} features. 
On the other hand, we found that gradient-based models keep their accuracy gains when evaluated 
in cross-domain setting with improved generalization capabilities due to our regularization. This can be seen on \cref{fig:acc_curves_ood}, where we achieve an improvement of about 2 percentage points \emph{(p.p.)} for both \Maml{} and \MC{} models on both 1-shot and 5-shot settings.
}    

\rev{
These results confirm that minimizing the norm and condition number of the linear predictors learned improves the generalization capabilities of meta-learning models. As opposed to metric-based methods which are already implicitly doing so, the addition of the regularization terms for gradient-based methods leads to a more significant improvement of performance in \emph{cross-domain}.
}

\section{Conclusion}\label{sec:conclusion}
In this \rev{chapter}, we studied the validity of the theoretical assumptions made in recent papers of Multi-Task Representation Learning theory when applied to popular metric- and gradient-based meta-learning algorithms. We found a striking difference in their behavior and provided both theoretical and experimental arguments explaining that metric-based methods satisfy the considered assumptions, while gradient-based
don't. We further used this as a starting point to implement a regularization strategy ensuring these assumptions and observed that it leads to faster learning and better generalization. 

\rev{While this work proposes an initial approach to bridging the gap between theory and practice for Meta-Learning, some questions remain open on the inner workings of these algorithms. Considering settings different from the episodic one represents another important line of research for the understanding of few-shot learning \cite{tian2020rethinking, shuoYang2021freelunchfew-shot}. In the remaining chapters, we are interested in the more challenging Object Detection task that requires to take into account the \emph{location} of objects along with their classification. For object detectors in FSL, recent work have found that pretraining and fine-tuning are more effective than episodic approaches \cite{DBLP:conf/icml/WangH0DY20,bar2022detreg}, which motivated the development presented in the subsequent chapters.}

%% file: chapters/selfod.tex
\chapter{Unsupervised Pretraining for Object Detection with Fewer Annotation}\label{chap:selfod}

\minitoc

\begin{abstract}
    \textit{
        The present chapter investigates unsupervised pretraining specifically designed for Object Detection with Transformers. We recall the positioning in the general training pipeline in \cref{fig:position_iclr_chap}. The objective is to propose a strong pretraining strategy for a more efficient fine-tuning of these detectors with few labeled data. The chapter is organized as follows. In \cref{sec:related_work_self}, we review previous work on unsupervised pretraining with a focus on application for object detection, and discuss the positioning of our method. Then, in \cref{sec:method}, we introduce our proposed \emph{ProSeCo}, and detail the localization-aware contrastive loss used in \cref{sec:loc_sce}. In \cref{sec:exps}, we present results on different benchmarks with a focus on learning with fewer data with several ablation studies. We conclude this chapter in \cref{sec:conclusion_self}.
    }
\end{abstract}

\begin{figure}
    \centering
    \includegraphics[width=0.8\linewidth]{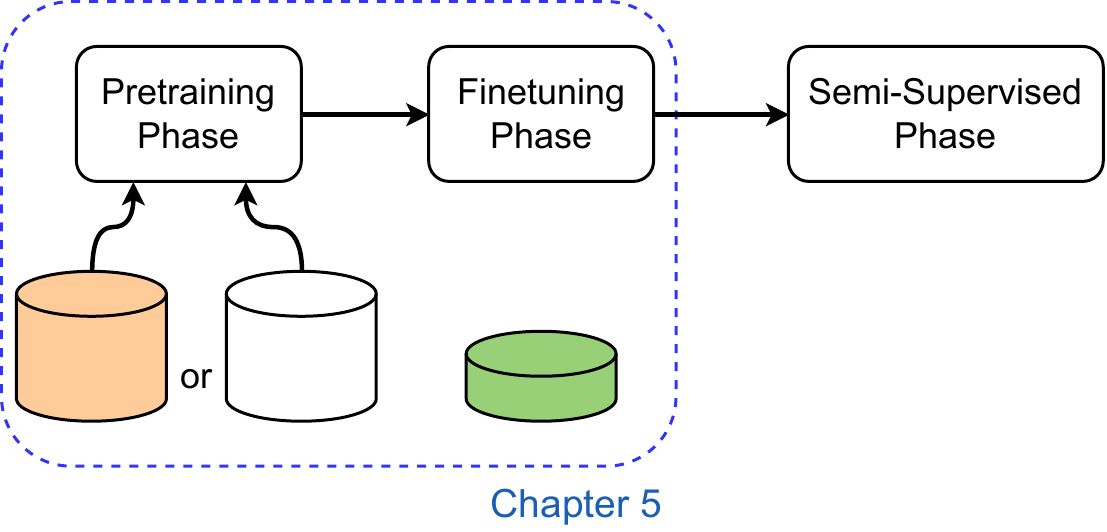}
    \caption{Illustration of the position of our contributions from this chapter in the full training pipeline presented in \cref{chap:intro}. We are interested on Unsupervised Pretraining for Object Detection.}
    \label{fig:position_iclr_chap}
\end{figure}

\section{Introduction}



\rev{In this chapter, we address the object detection task in the scarce data regime. As shown in the recent literature, unsupervised pretraining which consists in initializing a model with a pretrained backbone, helps train big architectures more efficiently \cite{chen2020big, caron2020unsupervised, he2020momentum}. While gathering data is not difficult in most cases, its labeling is always time-consuming and costly. Pretraining leverages huge amounts of unlabeled data and helps achieve better performance with fewer data and fewer iterations, when finetuning the pretrained models afterwards.}

\begin{figure}[t]
    \centering
    \includegraphics[width=1.\linewidth]{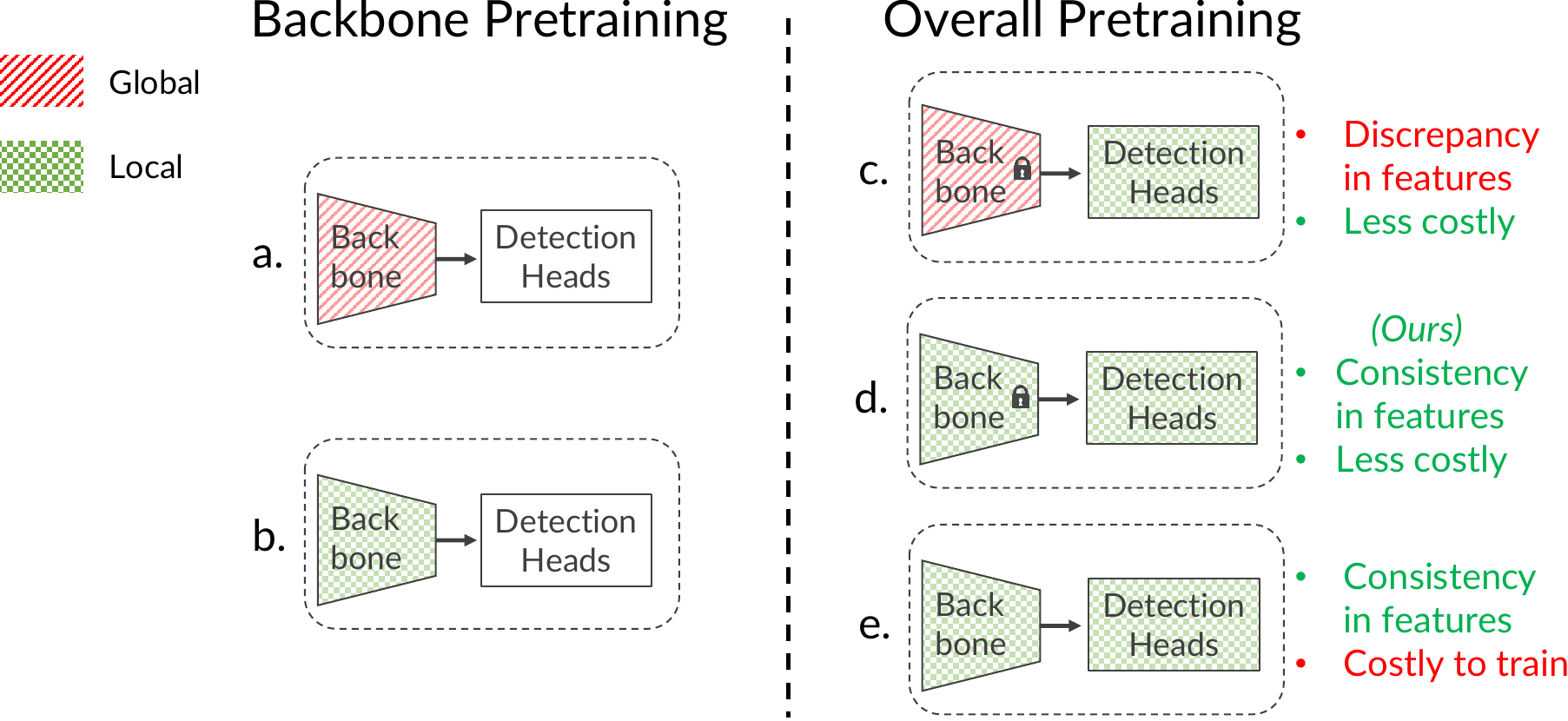}
    \caption{Illustration of the different pretraining possibilities for Object Detection. The pretraining can be either limited to the backbone (\emph{left}), or overall including the detection heads (\emph{right}). The few previous overall approaches either suffer from a discrepancy in the features between the backbone, that is trained at the image-level (\emph{global}), and the detection heads, trained to encode object-level (\emph{local}) information (\emph{c.}), or from the cost of training lots of parameters with a large batch size (\emph{e.}).}
    \label{fig:od_pt}
\end{figure}

The design of pretraining tasks for dense problems such as Object Detection has to take into account the fine-grained information in the image. Furthermore, complex object detectors contain different specific parts that can be either pretrained \emph{independently} \cite{xiao2021region, xie2021propagate, wang2021dense, henaff2021efficient, dai2021up, bar2022detreg} or \emph{jointly} \cite{wei2021aligning}. The different pretraining possibilities for Object Detection in the literature are illustrated in \cref{fig:od_pt}. A pretraining of the backbone tailored to dense tasks has been the subject of many recent efforts \cite{xiao2021region, xie2021propagate, wang2021dense, henaff2021efficient} (\emph{Backbone Pretraining}), but few have been interested in incorporating the detection-specific components of the architectures during pretraining \cite{dai2021up, bar2022detreg, wei2021aligning} (\emph{Overall Pretraining}). Among them, SoCo~\cite{wei2021aligning} focuses on convolutional architectures and pretrains the whole detector, \ie{} the backbone along with the detection heads (approach \emph{e.} in \cref{fig:od_pt}), whereas UP-DETR~\cite{dai2021up} and DETReg~\cite{bar2022detreg} pretrain only the transformers~\cite{vaswani2017attention} in transformer-based object detectors~\cite{carion2020end, zhu2020deformable} and keep the backbone fixed (approach \emph{c.} in \cref{fig:od_pt}).

Due to the numerous parameters that must be learned and the huge number of iterations needed because of random initialization, pretraining the entire detection model is expensive (\cref{fig:od_pt}, \emph{e.}). 
On the other hand, pretraining only the detection-specific parts with a fixed backbone leads to fewer parameters and allows leveraging strong pretrained backbones already available. However, fully relying on aligning embeddings given by the fixed backbone during pretraining and those given by the detection head, as done in DETReg or UP-DETR, introduces a discrepancy in the information contained in the features (\cref{fig:od_pt}, \emph{c.}). Indeed, while the pretrained backbone has been trained to learn image-level features, the object detector must understand object-level information in the image. \rev{Aligning inconsistent features hinders the pretraining quality.}

In this \rev{chapter, we propose our second contribution,} \emph{Proposal Selection Contrast} (ProSeCo), an unsupervised pretraining method using transformer-based detectors with a fixed pretrained backbone. 
ProSeCo makes use of two models. The first one aims to alleviate the discrepancy in the features by \rev{maintaining a copy of} the whole detection model. This model is referred to as a \emph{teacher} in charge of the \emph{object proposals} embeddings, and is updated through an Exponential Moving Average (EMA) of another \emph{student} network making the object predictions and using a similar architecture. This latter network is trained by a contrastive learning approach that leverages the high number of object proposals that can be obtained from the detectors. This methodology, in addition to the absence of batch normalization in the architectures, reduces the need for a large batch size.
We further adapt the contrastive loss commonly used in pretraining to take into account the locations of the object proposals in the image, which is crucial in object detection. In addition, the localization task is independently learned through region proposals generated by Selective Search \cite{uijlings2013selective}.
Our contributions are summarized as:
\begin{itemize}
    
\item We propose \emph{Proposal Selection Contrast (ProSeCo)}, a contrastive learning method tailored for pretraining transformer-based object detectors. 
\item We introduce the information of the localization of object proposals for the selection of positive examples in the contrastive loss to improve its efficiency for pretraining.
\item We show that our proposed ProSeCo outperforms previous pretraining methods for transformer-based object detectors on standard benchmarks as well as novel benchmarks.
\end{itemize}

\rev{Before detailing these contributions, we first provide a review of specific related work.}

\section{Related Work}\label{sec:related_work_self}


\paragraph{Supervised Object Detection with transformer-based architectures}

\rev{As presented in \cref{chap:learning,chap:sota}}, Object Detection is an important and extensively researched problem in computer vision~\cite{girshick2014rich,girshick2015fast,ren2015faster, redmon2016you, liu2016ssd, lin2017feature, tian2019fcos}. It essentially combines object localization and classification tasks. 
\rev{We are particularly interested in this chapter and the following in the recent detector family based on an encoder-decoder architecture using Transformers \cite{vaswani2017attention}, the DETR \cite{carion2020end} family of transformer-based architectures introduced in \cref{chap:learning}.}
The training complexity of this architecture was subsequently improved in Deformable DETR (\ddetr)~\cite{zhu2020deformable}, by changing the attention operations into deformable attention, resulting in a faster convergence speed. Several other follow-up works \cite{dai2021dynamic, meng2021conditional, wang2021anchor, liu2022dabdetr, yang2022querydet, li2022dn} have also focused on increasing the training efficiency of DETR. 
Transformer-based architectures now represent a strong alternative to traditional convolutional object detectors, reaching better performance for a similar training cost. Furthermore, recent work have shown strong results of transformer-based detectors in a data-scarce setting \cite{bar2022detreg}, compared to convolutional architectures \cite{liu2021unbiased}, \rev{which we also observe and discuss in \cref{chap:ssod}.}

\paragraph{Unsupervised Pretraining for detection models}

\rev{We introduced in \cref{chap:sota} recent pretraining methods tailored for OD, but few of them have tackled the problem of pretraining detection-specific parts of the architectures. We discuss their limitations here.}
SoCo \cite{wei2021aligning} proposes a contrastive pretraining strategy for convolutional detectors inspired by BYOL \cite{grill2020bootstrap}. Object locations are generated using Selective Search \cite{uijlings2013selective}, and then object-level features are extracted and contrasted with each other using a \emph{teacher-student} strategy. The small amount of object proposals and object features generated requires using a large batch size for the contrastive loss to be effective. Pretraining the backbone along with the detection modules this way makes the method difficult and costly to train due to the high amount of parameters to learn.
For transformer-based architectures, UP-DETR \cite{dai2021up} and DETReg \cite{bar2022detreg} use a fixed pretrained backbone to extract features respectively from random patch, or from object locations given by Selective Search \cite{uijlings2013selective}, then pretrain the detector by locating and reconstructing the features of the patch extracted from the input images. However, since the features to reconstruct are obtained by a backbone which was trained to encode image-level information, there is a discrepancy in the information between the features to match.

Our proposed ProSeCo is designed specifically for transformer-based detectors, and use a fixed backbone pretrained for \emph{local information}. In this work, we leverage the high amount of object proposals generated by transformer-based detectors as instances for contrastive learning. Target object-level features and localizations are provided by a \emph{teacher} detection model updated through EMA, inspired by recent advances in self-supervised and semi-supervised learning \cite{liu2021unbiased, grill2020bootstrap, denize2021similarity, wei2021aligning}. The \emph{student} detection model is pretrained by computing the contrastive loss between \emph{object proposals} given by the student and teacher detectors. The large number of proposals generated by transformer-based detectors alleviates the need for a large batch size for the contrastive loss. The contrastive loss function used is further improved by introducing the location of objects for selecting positive proposals. 



\section{Overview of the approach}\label{sec:method}

\begin{figure}[t]
    \centering
    \includegraphics[width=\linewidth]{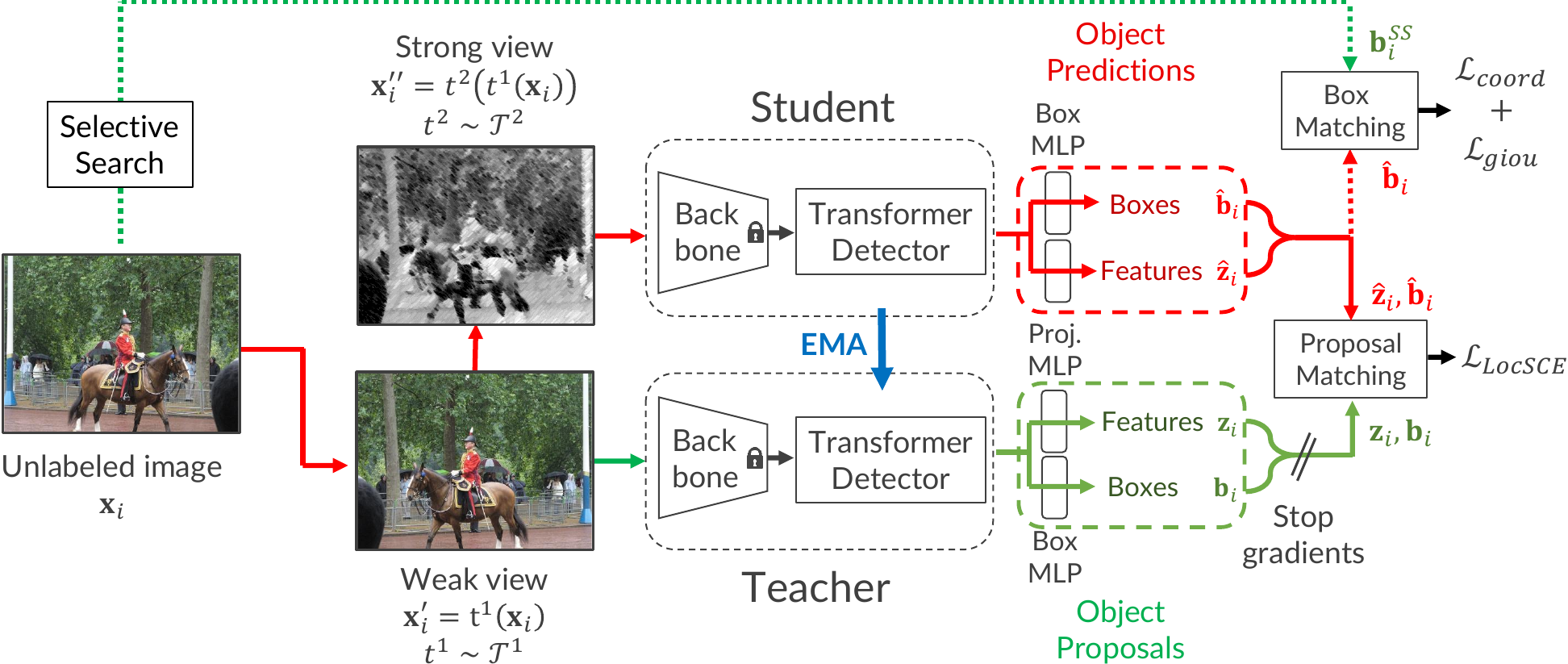}
    \caption{Overview of our proposed \emph{ProSeCo} for unsupervised pretraining. The method follows a \emph{student-teacher architecture}, with the teacher updated through an Exponential Moving Average (EMA) of the student. For each input image, $K$ random boxes are computed using the \emph{Selective Search} algorithm, and two different views are generated through an asymmetric set of \emph{weak augmentations} $\mathcal{T}^1$ and \emph{strong augmentations} $\mathcal{T}^2$. Then, \emph{object predictions} are obtained from the student model for the strongly augmented view, and \emph{object proposals} from the teacher model with the weakly augmented view. Finally, the boxes predicted by the student are matched with the boxes sampled from Selective Search to compute the localization losses $\Lcal_\text{coord}$ and $\Lcal_\text{giou}$, and the full predictions are matched with the object proposals to compute \emph{our novel contrastive loss} $\Lcal_\text{LocSCE}$ between them.}
    \label{fig:ProSeCo}
\end{figure}

We present in this section our proposed unsupervised pretraining approach, illustrated in \cref{fig:ProSeCo}, beginning with the data processing pipeline. Then, we detail the contrastive loss used with the localization-aware positive object proposal selection. The transformer-based detectors are built on a general architecture that consists of a backbone encoder (\eg{} a ResNet-50), followed by several transformers encoder and decoder layers, and finally two prediction heads for the bounding boxes coordinates and class logits \cite{carion2020end, zhu2020deformable}.

\subsection{Data processing pipeline}\label{sec:data}

Throughout the section, we assume to have sampled a batch of unlabeled images $\rvx = \{\rvx_i\}_{i=1}^{N_b}$, with $\rvx_i$ the $i$\textsuperscript{th} image and $N_b$ the batch size.

For each input image $\rvx_i$, we compute two different views with two asymmetric distributions of augmentations\footnote{The exact sets of augmentations chosen are given in \cref{tab:augmentations_self}.} $\mathcal{T}^1$ and $\mathcal{T}^2$: a \emph{weakly augmented view} ${\rvx_i}^\prime = t^1(\rvx_i)$, with $t^1 \sim \mathcal{T}^1$, and a \emph{strongly augmented view} obtained from the weakly augmented one ${\rvx_i}'' = t^2(t^1(\rvx_i))$, with $t^2 \sim \mathcal{T}^2$. 

To guide the model into discovering locations of objects in unlabeled images, we use bounding boxes obtained from the \emph{Selective Search} algorithm \cite{uijlings2013selective}, similarly to previous work \cite{wei2021aligning, bar2022detreg}. Since Selective Search is deterministic and the generated boxes are not ordered, we compute the boxes for all images offline and, at training time, randomly sample $K$ boxes $\rvb^{SS}_i = \{ \rvb^{SS}_{(i,j)} \in \mathbb{R}^4\}_{j=1}^K$ for each image in the batch. 
Then, the two views and the corresponding boxes sampled are used as input for our method.

\subsection{Pretraining method}\label{sec:proseco}

As shown in \cref{fig:ProSeCo}, our approach is composed of a \emph{student-teacher architecture}~\cite{grill2020bootstrap, denize2021similarity, wei2021aligning}. With ProSeCo, we extend the student-teacher pretraining for transformer-based detectors to tackle the discrepancy in information-level when aligning features, and introduce a \emph{dual unsupervised bipartite matching} presented below.

First, the backbone and the detection heads are respectively initialized from pretrained weights and randomly for both the student and teacher models. The teacher model is updated through an \emph{Exponential Moving Average (EMA)} of the student's weights at every training iteration:

\begin{equation}
    \theta_t \leftarrow \alpha \theta_t + (1-\alpha) \theta_s.
    \end{equation}
\noindent \rev{where $\theta_t$ and $\theta_s$ are respectively the teacher's and student's weights, and $\alpha \in [0,1]$ is the \emph{keep rate parameter}. This way, the teacher model represents a slightly delayed but more stable version of the student. When $\alpha=1$, the teacher is not updated and remains constant. When $\alpha=0$, all weights are updated to be equal to those of the student. Naturally, there is a trade-off between a too high and too low keep rate parameter.}

For both models, the classification heads in the detectors are replaced by an MLP, called \emph{projector}, to obtain latent representations of the objects. 

From the weakly augmented view ${\rvx_i}^\prime$, the teacher model provides \emph{object proposals} $\rvy_i = \{ \rvy_{(i,j)} \}_{j=1}^N = \{(\rvz_{(i,j)}, \rvb_{(i,j)} \}_{j=1}^N$, with $\rvz_{(i,j)}$ the latent embedding and $\rvb_{(i,j)}$ the coordinates of the $j$\textsuperscript{th} object found. The student model infers predictions $\hat{\rvy}_i = \{ \hat{\rvy}_{(i,j)} \}_{j=1}^N = \{(\hat{\rvz}_{(i,j)}, \hat{\rvb}_{(i,j)} \}_{j=1}^N$ from the corresponding strongly augmented view ${\rvx_i}''$.

Then, we apply an \emph{unsupervised} Hungarian algorithm \cite{munkres1957algorithms} for \emph{proposal matching} to find from all the permutations of $N$ elements $\mathfrak{S}_N$, the optimal bipartite matching $\hat{\sigma}_i^\text{prop}$ between the predictions $\hat{\rvy}_i$ of the student and the object proposals $\rvy_i$ of the teacher: 

\begin{equation}
\hat{\sigma}_i^\text{prop} = \arg \min_{\sigma \in \mathfrak{S}_N} \sum_{j=1}^N \Lcal_{\text{prop\_match}}(\rvy_{(i,j)}, \hat{\rvy}_{(i,\sigma(j))}).
\end{equation}

Therefore, for each image $\rvx_i$, the $j$\textsuperscript{th} proposal $\rvy_{(i,j)}$ found by the teacher is associated to the $\hat{\sigma}_i^\text{prop}(j)$\textsuperscript{th} prediction of the student $\hat{\rvy}_{(i, \hat{\sigma}_i^\text{prop}(j))}$. Our matching cost $\Lcal_{\text{prop\_match}}$ for the Hungarian algorithm takes into account both features and bounding box predictions through a linear combination of features similarity $\Lcal_\text{sim}$, the $\ell_1$ loss of the box coordinates $\Lcal_{\text{coord}}$, and the generalized IoU loss $\Lcal_{\text{giou}}$ from \cite{rezatofighi2019generalized}:

\begin{align}
    \Lcal_\text{sim}(\rvz_{(i,j)}, \hat{\rvz}_{(i,\sigma(j))}) &= \frac{ \langle \rvz_{(i,j)}, \hat{\rvz}_{(i,\sigma(j))} \rangle}{\| \rvz_{(i,j)} \|_2 \cdot \| \hat{\rvz}_{(i,\sigma(j))} \|_2} \\
    \Lcal_{\text{coord}}(\rvb_{(i,j)}, \hat{\rvb}_{(i,\hat{\sigma}_i(j))}) &= \| \rvb_{(i,j)} - \hat{\rvb}_{(i,\hat{\sigma}_i(j))} \|_1 \\
\begin{split}
\Lcal_\text{prop\_match}(\rvy_{(i,j)}, \hat{\rvy}_{(i,\sigma(j))}) &= \Bigl[ \lambda_\text{sim} \Lcal_\text{sim}\left(\rvz_{(i,j)}, \hat{\rvz}_{(i,\sigma(j))}\right) \\
& + \lambda_{\text{coord}} \Lcal_\text{coord}\left(\rvb_{(i,j)}, \hat{\rvb}_{(i,\sigma(j))}\right) + \lambda_\text{giou} \Lcal_\text{giou}\left(\rvb_{(i,j)}, \hat{\rvb}_{(i,\sigma(j))}\right)\Bigr].
\end{split}
\end{align}

Similarly, we also use an \emph{unsupervised} Hungarian algorithm  for \emph{box matching}, to find the optimal bipartite matching $\hat{\sigma}_i^\text{box} \in \mathfrak{S}_N$ between the predicted boxes $\hat{\rvb}_i$ of the student and the sampled boxes $\rvb^{SS}_i$ from Selective Search, using the matching cost $\Lcal_{\text{box\_match}}$:

\begin{align}
\hat{\sigma}_i^\text{box} &= \arg \min_{\sigma \in \mathfrak{S}_N} \sum_{j=1}^N \Lcal_{\text{box\_match}}(\rvy_{(i,j)}, \hat{\rvy}_{(i,\sigma(j))}), \\
\Lcal_\text{box\_match}(\rvb^{SS}_{(i,j)}, \hat{\rvb}_{(i,\sigma(j))}) &= \lambda_{\text{coord}} \Lcal_\text{coord}\left(\rvb^{SS}_{(i,j)}, \hat{\rvb}_{(i,\sigma(j))}\right) + \lambda_\text{giou} \Lcal_\text{giou}\left(\rvb^{SS}_{(i,j)}, \hat{\rvb}_{(i,\sigma(j))}\right).
\end{align}

Finally, the global unsupervised loss $\Lcal_u$ used for training is a combination of a loss function between the object latent embeddings of the teacher and student models, and between the object localization of the student predictions and Selective Search boxes. 
More formally, it is computed as:
\begin{align}
\begin{split}
    \Lcal_u(\rvx) &= \lambda_\text{contrast} \Lcal_\text{LocSCE}\left(\rvy, \hat{\rvy},\hat{\sigma}^\text{prop}\right) + \\
    & \frac{1}{N_b K} \sum_{i=1}^{N_b} \Bigl[ \sum_{j=1}^K \lambda_{\text{coord}} \Lcal_{\text{coord}}\left(\rvb^{SS}_{(i,j)}, \hat{\rvb}_{(i,\hat{\sigma}^\text{box}_i(j))}\right) + \sum_{j=1}^K \lambda_\text{giou} \Lcal_{\text{giou}}\left(\rvb^{SS}_{(i,j)}, \hat{\rvb}_{(i,\hat{\sigma}^\text{box}_i(j))}\right) \Bigr]. 
\end{split}
\end{align}

In the above equations, we define $\lambda_\text{sim}, \lambda_{\text{coord}}, \lambda_\text{giou}, \lambda_\text{contrast} \in \R^+$ as the coefficients for the different losses. For the consistency in the latent embeddings of the objects, we introduce the object locations information in our contrastive loss $\Lcal_\text{LocSCE}$. This loss is used to contrast the predictions $\hat{\rvy} = \{ \hat{\rvy}_i \}_{i=1}^{N_b}$ of the student with object proposals $\rvy = \{ \rvy_i \}_{i=1}^{N_b}$ found by the teacher, matched according to the proposal matching $\hat{\sigma}^\text{prop} = \{ \hat{\sigma}_i^\text{prop} \}_{i=1}^{N_b}$ over the batch. We detail the computations behind this loss in the next section. 

\subsection{Localization-aware contrastive loss}\label{sec:loc_sce}

\begin{figure}[t]
    \centering
    \includegraphics[width=1.\linewidth]{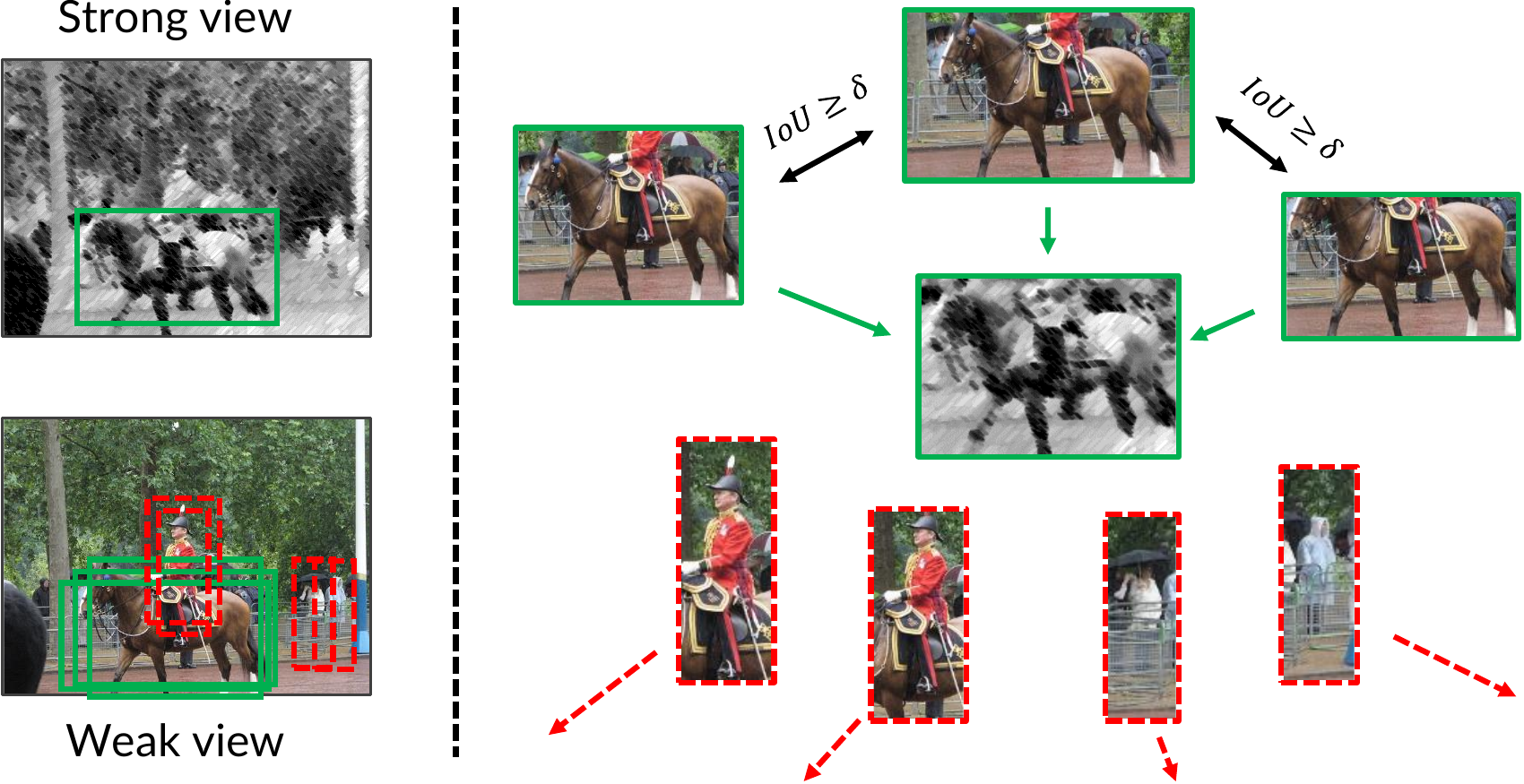}
    \caption{Illustration of the effect of the \emph{localized contrastive loss} used. Predictions of the student and teacher models are contrasted with each other to leverage the large number of object proposals obtained from transformer-based detectors. To introduce the \emph{object locations} information, overlapping proposals \emph{(in green)} in each \rev{weak view}, according to an \emph{IoU threshold} $\delta$, are also considered as positive along with the matched proposal. \rev{Proposals that neither match nor overlap the matched proposal, are considered as negative \emph{(in red)} in the contrastive loss.}}
    \label{fig:loc_sce}
\end{figure}


Inspired by advances in self-supervised learning, we propose a contrastive objective function \cite{oord2018representation, chen2020simple, he2020momentum} between object proposals that also maintains relations \cite{zheng2021ressl, denize2021similarity} among these proposals.
This objective function extends the latest SCE \cite{denize2021similarity}, \rev{as well as the original InfoNCE \cite{oord2018representation},} for \emph{instance discrimination between object proposals} and is illustrated in \cref{fig:loc_sce}. We compute the contrastive loss between all the object latent embeddings from a batch of image. The \emph{positive pair} of objects in an image $\rvx_i$ is given by the proposal matching $\sigma_i^\text{prop}$.
First, we define the distributions of similarity between objects embeddings :

\begin{align}
    p_{(in,jm)}^\prime &= \frac{ \1_{i \neq n} \1_{j \neq m} \exp(\rvz_{(i,j)} \cdot \rvz_{(n,m)} / \tau_t) }{ \sum_{k=1}^{N_b} \sum_{l=1}^{N} \1_{i \neq k} \1_{j \neq l} \exp(\rvz_{(i,j)} \cdot \rvz_{(k,l)} / \tau_t) }, \\
    p_{(in,jm)}^{\prime\prime} &= \frac{ \exp( \rvz_{(i,j)} \cdot \hat{\rvz}_{(n,m)} / \tau )}{ \sum_{k=1}^{N_b} \sum_{l=1}^{N} \exp( \rvz_{(i,j)} \cdot \hat{\rvz}_{(k,l)} / \tau )}.
\end{align}

The distribution $p_{(in,jm)}^\prime$ represents the \emph{relations} between weakly augmented object embeddings scaled by the temperature $\tau_t$, and $p_{(in,jm)}^{\prime\prime}$ the similarity between the strongly augmented embeddings and the weakly augmented ones, scaled by $\tau$.
Then, the \emph{target similarity distribution} for the objective function is a weighted combination of one-hot label and the teacher's embeddings relations:

\begin{equation}
    w_{(in,jm)} = \lambda_\text{SCE} \cdot \1_{i=n} \1_{j=m} + (1-\lambda_\text{SCE}) \cdot p_{(in,jm)}^\prime.
\end{equation}


To introduce the location information in our objective, we compute the pairwise \emph{Intersection over Union} (IoU) between object proposals of the same image and consider overlapping objects as other positives when computing the target similarity distribution:
\begin{equation}
    w_{(in,jm)}^\text{Loc} = \lambda_\text{SCE} \cdot \1_{i=n} \1_{IoU_i(j,m) \geq \delta} + (1-\lambda_\text{SCE}) \cdot p_{(in,jm)}^\prime,
\end{equation}

\noindent where $IoU_i(j,m)$ corresponds to the IoU between teacher proposals $\rvy_{(i,j)}$ and $\rvy_{(i,m)}$ found in the same image $\rvx_i$, and $\delta$ is the \emph{IoU threshold} to consider the proposal as a positive example. 
Finally, we use this tailored target similarity distribution in our \emph{Localized SCE} (LocSCE) loss:

\begin{equation}
    \Lcal_\text{LocSCE}(\rvy, \hat{\rvy}, \hat{\sigma}^\text{prop}) = - \frac{1}{N_b N} \sum_{i=1}^{N_b} \sum_{n=1}^{N_b} \sum_{j=1}^N \sum_{m=1}^N w^\text{Loc}_{(in,jm)} \log ( p_{(in,j\hat{\sigma}_n^\text{prop}(m))}^{\prime\prime}).
\end{equation} 

Note that we do not match the proposals according to $\hat{\sigma}^\text{prop}$ in the target similarity, as we compare proposals obtained by the teacher model. We also require the full proposal as input of the loss to compute the pairwise IoU using the box coordinates.

\rev{We can also extend InfoNCE the same way:}

\begin{equation}
    \Lcal_\text{InfoNCE}(\rvz, \hat{\rvz}, \hat{\sigma}^\text{prop}) = - \frac{1}{N_b N} \sum_{i=1}^{N_b} \sum_{j=1}^N \log \left( \frac{\exp(\rvz_{(i,j)} \cdot \hat{\rvz}_{(i, \hat{\sigma}_i^\text{prop}(j))} / \tau )}{ \sum_{k=1}^{N_b} \sum_{l=1}^{N} \exp(\rvz_{(i,j)} \cdot \hat{\rvz}_{(k,l)} / \tau) } \right).
\end{equation} 

\rev{Similarly to our \emph{LocSCE}, the localization of the objects can be introduced in the InfoNCE loss function to obtain the \emph{LocNCE} objective as follows:} 
\begin{equation}
    \Lcal_\text{LocNCE}(\rvy, \hat{\rvz}, \hat{\sigma}^\text{prop}) = - \frac{1}{N_b N} \sum_{i=1}^{N_b} \sum_{j=1}^N \sum_{m=1}^N \log \left( \frac{ \1_{IoU_i(j,m) \geq \delta} \exp(\rvz_{(i,j)} \cdot \hat{\rvz}_{(i, \hat{\sigma}_i^\text{prop}(m))} / \tau )}{ \sum_{k=1}^{N_b} \sum_{l=1}^{N} \exp(\rvz_{(i,j)} \cdot \hat{\rvz}_{(k,l)} / \tau) } \right).
\end{equation} 

We recover the original formulations of both SCE \rev{and InfoNCE} when $\delta =1$.
These formulations lead to an effective batch size of $N_b \cdot N$.
The localization-aware contrastive loss functions aim to pull together the objects embeddings that overlap subsequently with each others, as they should correspond to the same object in the image. 

\section{Experiments}\label{sec:exps}

In this section, we present a comparative study of the results of our proposed method on standard and novel benchmarks for learning with fewer data, as well as an ablative study on the most relevant parts. First, we recall the datasets \rev{introduced in \cref{chap:sota} with the corresponding} evaluation and training settings.

\subsection{Implementation details}\label{sec:exp_details_self}

\paragraph{Datasets and evaluation}

We use different datasets throughout this work that we present here.
\begin{itemize}
    \item For pretraining purposes, we use the standard \emph{ImageNet ILSVRC 2012 (IN)} \cite{russakovskyImageNetLargeScale2015} dataset, which contains 1.2M training images separated in 1000 class categories.
    \item We also use the \emph{MS-COCO (COCO)} \cite{lin2014microsoft} dataset for pretraining in ablation studies, but mainly for finetuning and evaluation purposes. This dataset contains 80 classes of objects and about 118k training images \emph{(train2017 subset)}. Performance is evaluated on the \emph{val2017} subset.
    \item For finetuning, we also use the \emph{PASCAL VOC 2007 and 2012 (VOC 07-12)} \cite{everingham2010pascal} datasets. This dataset contains 20 classes of objects, and we use the combination of the \emph{trainval} subsets from both VOC2007 and VOC2012 for training, corresponding to about 16k training images in total. Performance is evaluated on the \emph{test} subset from VOC2007.
    \item For a more complicated dataset, we use the \emph{Few-Shot Object Detection} dataset \cite{fan2020few}. Since the dataset is designed as \emph{open-set}, \ie{} with different classes between training and testing, we \rev{separately} use the \rev{\emph{train} and} \emph{test} sets for benchmarking. We separate the test set into a training and testing subset, by randomly taking 80\% of images for training and the 20\% remaining for testing, \rev{and do the same for the train set.} We make sure that all classes appears at least once in both training and testing subsets. The images selected for training and testing will be made available for reproducibility. This separation leads to about 11k training images and 3k testing images \rev{for the \emph{test} set and about 42k training images and 10k testing images for the \emph{train} set}.
\end{itemize}

To evaluate the performance in learning with fewer data, following previous work \cite{wei2021aligning, bar2022detreg}, we consider the \emph{Mini-COCO} benchmarks, where we randomly sample 1\%, 5\% or 10\% of the training data. Similarly, we also introduce the novel \emph{Mini-VOC} benchmark, in which we randomly sample 5\% or 10\% of the training data. 
We also use the Few-Shot Object Detection (FSOD) dataset \cite{fan2020few} in novel \emph{FSOD-test} \rev{and \emph{FSOD-train}} benchmarks. 
In all benchmarks, the image ids selected for training and testing will be made available for reproducibility. 

\paragraph{Pretraining}
We initialize the backbone with the publicly available pretrained SCRL \cite{roh2021spatially} checkpoint and pretrain ProSeCo for 10 epochs on IN. The hyperparameters are set as follows: the EMA keep rate parameter to 0.999, the IoU threshold $\delta = 0.5$, a batch size of $N_b = 64$ images over 8 A100 GPUs, and the coefficients in the different losses $\lambda_\text{sim} = \lambda_\text{contrast} = 2$ which is the same value used for the coefficient governing the class cross-entropy in the supervised loss. The projector is defined as a 2-layer MLP with a hidden layer of 4096 and a last layer of 256, without batch normalization. Following SCE \cite{denize2021similarity}, we set the temperatures $\tau = 0.1, \tau_t = 0.07$ and the coefficient $\lambda_\text{SCE} = 0.5$. Similarly to DETReg \cite{bar2022detreg}, we sample $K=30$ boxes from the outputs of Selective Search for each image at every iteration. Other training and architecture hyperparameters are defined as in \ddetr \cite{zhu2020deformable} with, specifically, the coefficients $\lambda_\text{coord} = 5$ and $\lambda_\text{giou} = 2$, the number of object proposals (queries) $N=300$, and the learning rate is set to $lr = 2\cdot 10^{-4}$. For weak augmentations $\mathcal{T}^1$, we use a random combination of flip, resize and crop, and for strong augmentations $\mathcal{T}^2$, we use a random combination of color jittering, grayscale and Gaussian blur. In $\mathcal{T}^1$, we resize images with the same range of scales as the supervised training protocol on COCO (\emph{Large-scale}).

\paragraph{\rev{Augmentations used}}
\rev{We detail in \cref{tab:augmentations_self} the distributions of augmentations used to create the weak view and the strong view. The \emph{Weak Augmentations} follow standard supervised training for transformer-based detectors~\cite{carion2020end,zhu2020deformable}. The \emph{Strong Augmentations} follow typical contrastive learning augmentations \cite{chen2020simple, bar2022detreg}.}

\paragraph{Finetuning protocols}
For finetuning the pretrained models, we follow the standard supervised learning hyperparameters of \ddetr \cite{zhu2020deformable}. In all experiments, we train the models with a batch size of 32 images over 8 A100 GPUs until the validation performance stops increasing, \ie{} for Mini-COCO, up to 2000 epochs for 1\%, 500 epochs for 5\%, 400 epochs for 10\%, for Mini-VOC, up to 2000 epochs for both 5\% and 10\%, up to 100 epochs for both FSOD-test and PASCAL VOC, \rev{and up to 50 epochs for FSOD-train}. We always decay the learning rates by a factor of 0.1 after about 80\% of the number of training epochs.
To compare our method to DETReg \cite{bar2022detreg} on our novel benchmarks, we use their publicly available checkpoints from GitHub.

\begin{table}[ht]
    \centering
    \resizebox{0.95\linewidth}{!}{%
    \begin{tabular}{@{}lcc@{}}
    \toprule
    \midrule
    \multicolumn{3}{c}{\textbf{Weak Augmentations ($\mathcal{T}^1$)}} \\
    \midrule
    Augmentations & Probability & Parameters \\
    \midrule
    Horizontal Flip & 0.5 & -- \\
    \midrule
    \multirow{2}{*}{Resize} & \multirow{2}{*}{0.5} & \emph{Mid-scale:} short edge = range(320,481,16) \\
    & & \emph{Large-scale:} short edge = range(480,801,32) \\
    \midrule
    Resize & \multirow{4}{*}{0.5} & short edge $\in \{400,500,600\}$ \\
        \cmidrule(r){1-1}
        \cmidrule(l){3-3}
    Random Size Crop & & min size = 384 ; max size = 600 \\
        \cmidrule(r){1-1}
        \cmidrule(l){3-3}
    \multirow{2}{*}{Resize} & & \emph{Mid-scale:} short edge = range(320,481,16) \\
    & & \emph{Large-scale:} short edge = range(480,801,32) \\
    \bottomrule
    \toprule
    \multicolumn{3}{c}{\textbf{Strong Augmentations ($\mathcal{T}^2$)}} \\
    \midrule
    Color Jitter & 0.8 & (brightness, contrast, saturation, hue) = (0.4, 0.4, 0.4, 0.1) \\
    \midrule
    GrayScale & 0.2 & -- \\
    \midrule
    Gaussian Blur & 0.5 & (sigma x, sigma y) = (0.1, 2.0) \\
    \bottomrule
    \end{tabular}%
    }
    \begin{center}
        \caption{The different sets of augmentations used for each branch (\emph{weak} or \emph{strong}). \emph{Probability} indicates the probability of applying the corresponding augmentation.}
        \label{tab:augmentations_self}
        \end{center}
\end{table}

\subsection{Finetuning and transfer learning}

\paragraph{Different datasets} We evaluate the transfer learning ability of our pretrained model on several datasets.
\cref{tab:mini_coco,tab:voc_fsod} present the results obtained compared to previous methods in the literature when learning from fewer labeled data. We can see that our method outperforms state-of-the-art results in unsupervised pretraining on all benchmarks and datasets, and obtain even more strong results when training data is scarce. The improvement is even more significant as the overall performance with few training data is low.
When using 5\% of the COCO training data (\ie{} Mini-COCO 5\% in \cref{tab:mini_coco}), corresponding to about 5.9k images, ProSeCo achieves 28.8 mAP, which represents an improvement of +5.2 \emph{percentage point} (p.p.) over the supervised pretraining baseline and +2 p.p. over both state-of-the-art overall pretraining methods. 

\paragraph{\rev{Different models}} \rev{Then, we evaluate pretraining with other transformer-based detectors in different settings. In \cref{tab:dab_def_detr}, we fine-tune with DAB-\ddetr{} \cite{liu2022dabdetr} after pretraining with \ddetr{}. We observe that ProSeCo still performs strongly when transferring the pretrained weights to other architectures. We can see that using a more consistent pretrained backbone is important for pretraining the overall detection model.
In \cref{tab:cond_detr}, we pretrain \emph{and} fine-tune with Conditional DETR \cite{meng2021conditional}. We can see that in the most scarce setting, pretraining the detection-specific parts with our ProSeCo is less effective than only pretraining the backbone with SCRL \cite{roh2021spatially}. However, with more labels pretraining the overall detection models improves the performance. We can see that Conditional DETR has globally lower performance in FSL than \ddetr, which might hinders overall pretraining in the scarcest setting.  
}

\begin{table}[ht]
    \begin{subtable}{\linewidth}
    \centering
    \resizebox{0.98\linewidth}{!}{%
    \begin{tabular}{@{}lllccc@{}}
    \toprule
    \multirow{2}{*}{Method} & \multirow{2}{*}{Detector} & Pretrain. & \multicolumn{3}{c}{Mini-COCO} \\
    \cmidrule(lr){4-6}
    & & Dataset & 1\% (1.2k) & 5\% (5.9k) & 10\% (11.8k)\\
    \midrule
        Supervised & \ddetr & IN & 13.0 & 23.6 & 28.6 \\
        SwAV \cite{caron2020unsupervised} & \ddetr & IN & 13.3 & 24.5 & 29.5 \\ 
        SCRL \cite{roh2021spatially} & \ddetr & IN & 16.4 & 26.2 & 30.6 \\ 
        DETReg \cite{bar2022detreg} & \ddetr & COCO & 15.8 & 26.7 & 30.7 \\
        DETReg \cite{bar2022detreg} & \ddetr & IN & 15.9 & 26.1 & 30.9 \\
    \midrule
        Supervised \cite{wei2021aligning} & Mask R-CNN & IN & -- & 19.4 & 24.7 \\
        SoCo$^*$ \cite{wei2021aligning} & Mask R-CNN & IN & -- & 26.8 & 31.1 \\ 
    \midrule
        \textit{ProSeCo (Ours)} & \ddetr & IN & \textbf{18.0} & \textbf{28.8} & \textbf{32.8} \\
    \bottomrule
    \end{tabular}%
    }
    \subcaption{ }
    \label{tab:mini_coco}
    \end{subtable}
    \begin{subtable}{\linewidth}
    \centering
    \resizebox{0.98\linewidth}{!}{%
    \begin{tabular}{@{}lccccc@{}}
    \toprule
    \multirow{2}{*}{Method} & FSOD-test & \rev{FSOD-train} & PASCAL VOC & \multicolumn{2}{c}{Mini-VOC} \\
    \cmidrule(lr){5-6}
     & 100\% (11k) & \rev{100\% (42k)} & 100\% (16k) & 5\% (0.8k) & 10\% (1.6k) \\
    \midrule
        Supervised & 39.3 & \rev{42.6} & 59.5 & 33.9 & 40.8 \\
        DETReg \cite{bar2022detreg} & 43.2 & \rev{43.3} & 63.5 & 43.1 & 48.2 \\
        \textit{ProSeCo (Ours)} & \textbf{46.6} & \rev{\textbf{47.2}} & \textbf{65.1} & \textbf{46.1} & \textbf{51.3} \\
    \bottomrule
    \end{tabular}%
    }
    \subcaption{ }
    \label{tab:voc_fsod}
    \end{subtable}
    \begin{center}
        \caption{Performance (mAP in \%) of our proposed pretraining approach after fine-tuning using different percentage of training data (with the corresponding number of images reported). We show that our ProSeCo outperforms previous pretraining methods in all benchmarks.}
        \end{center}
\end{table}

\begin{table}
    \centering
        \begin{tabular}{@{}lccc@{}}
        \toprule
        \multirow{2}{*}{Pretraining} & \multicolumn{3}{c}{Mini-COCO} \\
        \cmidrule(lr){2-4}
        & 1\% (1.2k) & 5\% (5.9k) & 10\% (11.8k) \\
        \midrule
            Supervised & 16.7 & 27.0 & 31.0 \\
            DETReg & 17.7 & 28.6 & 33.0 \\
            ProSeCo w/ SwAV & 16.2 & 28.7 & 33.9 \\
            ProSeCo w/ SCRL & \textbf{19.4} & \textbf{29.9} & \textbf{34.1} \\ 
        \bottomrule
        \end{tabular}
        \caption{\rev{Performance comparison when transferring pretrained weights to DAB-Deformable DETR \cite{liu2022dabdetr}, after fine-tuning on different fractions of labeled data.}}
        \label{tab:dab_def_detr}
    \end{table}

\begin{table}
    \centering
        \begin{tabular}{@{}lccc@{}}
        \toprule
        \multirow{2}{*}{Pretraining} & \multicolumn{3}{c}{Mini-COCO} \\
        \cmidrule(lr){2-4}
        & 1\% (1.2k) & 5\% (5.9k) & 10\% (11.8k) \\
        \midrule
            Supervised & 8.7 & 19.3 & 24.6 \\
            SCRL & \textbf{11.7} & 20.9 & 26.2 \\
            ProSeCo & 9.6 & \textbf{22.4} & \textbf{27.5} \\ 
        \bottomrule
        \end{tabular}
        \caption{\rev{Performance comparison when pretraining and fine-tuning with Conditional DETR \cite{meng2021conditional} on different fractions of labeled data.}}
        \label{tab:cond_detr}
    \end{table}

\subsection{Ablation Studies}

In the following, we provide several ablation studies for our proposed approach. All experiments and results \rev{for the different ablations} are compared on the Mini-COCO 5\% benchmark \rev{with the pretrained SwAV backbone unless explicitly stated}. 

\paragraph{Pretraining dataset and backbone} In \cref{tab:abl_bb}, we show the effect of changing the pretraining dataset or the fixed backbone used in our pretraining method. We can see that using a backbone more adapted to dense tasks that learned local information (\eg{} SCRL) helps the model by having consistent features (+1 p.p.), compared to a backbone pretrained for global features (\eg{} SwAV). Furthermore, even with a less adapted backbone, our ProSeCo initialized with the SwAV backbone outperforms DETReg (+1.7 p.p.). To compare with IN, we also pretrain ProSeCo on COCO for 120 epochs. We obtain better results when pretraining the model on IN than using COCO thanks to the large number of different images in IN (about 10 times the number of images of COCO, leading to +0.4 p.p.), which is consistent with previous findings \cite{wei2021aligning}. The difference in the number of images might be slightly compensated in COCO by the closeness to the downstream task and the diversity of objects.

\paragraph{Location information in contrastive loss} In \cref{tab:abl_loss_iou}, we show the effect of the location information in the contrastive loss SCE. We can see that when introducing multiple positive examples for each image based on the IoU threshold $\delta$ (\ie{} $\forall \delta < 1$), we achieve better results than with the original SCE loss (\ie{} $\delta = 1$). Notably, the best results are achieved with $\delta = 0.5$ (+1.7 p.p.). 
\rev{We also compare the two different contrastive objectives considered for pretraining in \cref{tab:abl_loss_iou}. We can see that using the InfoNCE loss leads to slightly better results (+0.3 p.p.). However, when introducing the location information, SCE benefits much more than InfoNCE (+1.7 p.p. compared to +0.6 p.p.). This might be that the selection of positives from the location helps to introduce easy positive examples, and thanks to this, the relational aspect of SCE can focus on the more difficult positives. In the end, LocSCE achieves stronger results than LocNCE (+0.8 p.p.).}

\paragraph{Hyperparameters} The \cref{tab:abl_hp} presents an ablation study on different important hyperparameters of our approach.
We experimented first with the same batch size applied in \cite{bar2022detreg} (\emph{Abl. Batch}), but found that using a smaller batch size (\emph{Base}) leads to improved results (+0.2 p.p.). 
We evaluated different image scales as a parameter of the weak data augmentations distribution in \emph{Abl. Scales}. \emph{Mid-scale} corresponds to a resizing of the images such that the shortest edge is between 320 and 480 pixels, as used in previous work \cite{dai2021up, bar2022detreg}, and \emph{Large-scale} to a resize between 480 and 800 pixels, used for supervised learning on COCO. Exact values for these parameters can be found in \cref{tab:augmentations_self}. We found that increasing the size of the images during pretraining is important to have more meaningful information in the boxes, and a more precise localization of the boxes (+0.7 p.p.).
Following the best results from \cite{denize2021similarity}, we evaluated the performance for $\tau_t \in \{0.05, 0.07\}$. We found that $\tau_t = 0.07$ leads to the best performance (+0.3 p.p.).
We considered several EMA keep rate parameter values following previous work \cite{he2020momentum, wei2021aligning, denize2021similarity}, and found that 0.999 achieves the best results (+0.1 p.p.).

\begin{table}[ht]
    \centering
    \begin{subtable}{0.59\linewidth}
    \centering
    \begin{tabular}{@{}llc@{}}
    \toprule
    Pretraining & Dataset & mAP \\
    \midrule
        ProSeCo w/ SwAV  & COCO & 27.4 \\
        ProSeCo w/ SwAV  & IN & 27.8 \\
        ProSeCo w/ SCRL  & IN & \textbf{28.8} \\
    \bottomrule
    \end{tabular}%
    \subcaption{ }
    \label{tab:abl_bb}
    \end{subtable}
    \begin{subtable}{0.35\linewidth}
    \centering
    \begin{tabular}{@{}lcc@{}}
        \toprule
        Loss & $\delta$ & mAP \\
        \midrule
            InfoNCE & 1.0 & 26.4 \\
            \textit{LocNCE} & 0.5 & \textbf{27.0} \\
        \midrule
            SCE & 1.0 & 26.1 \\
            \textit{LocSCE (Ours)} & 0.2 & 27.0 \\
            \textit{LocSCE (Ours)} & 0.7 & 27.1 \\
            \textit{LocSCE (Ours)} & 0.5 & \textbf{27.8} \\
        \bottomrule
        \end{tabular}
        \subcaption{ }
        \label{tab:abl_loss_iou}
    \end{subtable}
    \begin{center}
        \caption{\textbf{(a)} Comparison after finetuning when using different pretrained backbone and/or pretraining datasets for ProSeCo. \textbf{(b)} Comparison of the effect of the location information using different IoU threshold $\delta$ and for the different constrastive loss considered. All performance (mAP in \%) are measured on Mini-COCO 5\%.}
    \end{center}
\end{table}

\begin{table}[ht]
    \centering
    \resizebox{0.98\linewidth}{!}{%
    \begin{tabular}{@{}lcccccccccc@{}}
    \toprule
    \multirow{2}{*}{Ablative Variant} & \multicolumn{2}{c}{Batch size} & \multicolumn{2}{c}{Images scale} & \multicolumn{2}{c}{Temperature $\tau_t$} & \multicolumn{3}{c}{EMA} & \multirow{2}{*}{mAP} \\
    \cmidrule(lr){2-3}
    \cmidrule(lr){4-5}
    \cmidrule(lr){6-7}
    \cmidrule(lr){8-10}
    & \textcolor{red}{192} & \textcolor{ForestGreen}{\textbf{64}} & \textcolor{red}{Mid} & \textcolor{ForestGreen}{\textbf{Large}} & \textcolor{red}{0.05} & \textcolor{ForestGreen}{\textbf{0.07}} & \textcolor{red}{0.99} & \textcolor{red}{0.996} & \textcolor{ForestGreen}{\textbf{0.999}} & \\
    \midrule
    Base & & \checkmark & \checkmark & & \checkmark & & & \checkmark & & 26.7 \\
    \midrule
    Abl. Batch & \checkmark & & \checkmark & & \checkmark & & & \checkmark & & 26.5 \\
    \midrule
    Abl. Scale &  & \checkmark & & \checkmark & \checkmark & & & \checkmark & & 27.4 \\
    \midrule
    Abl. Temp. & & \checkmark & \checkmark & & & \checkmark & & \checkmark & & 27 \\
    \midrule
    \multirow{2}{*}{Abl. EMA} & & \checkmark & \checkmark & & \checkmark & & \checkmark & & & 26.3 \\
    & & \checkmark & \checkmark & & \checkmark & & & & \checkmark & 26.8 \\
    \midrule
    \textbf{Best} & & \checkmark & & \checkmark & & \checkmark & & & \checkmark & \textbf{27.8} \\
    \bottomrule
    \end{tabular}%
    }
    \begin{center}
        \caption{Ablation studies on different hyperparameters for the proposed method. The performance (mAP in \%) are measured on Mini-COCO 5\%. \textcolor{ForestGreen}{\textbf{Green and bold columns names}} indicate a \emph{positive} effect on the performance, and \textcolor{red}{red columns} a \emph{negative} effect.}
        \label{tab:abl_hp}
    \end{center}
\end{table}

\paragraph{\rev{Increasing the number of queries}}

\rev{In \cref{tab:abl_queries}, we provide an ablation on the number of object proposals (queries) $N$ in \ddetr, when pretraining with ProSeCo and finetuning afterwards. A higher $N$ leads to more parameters in the model and longer computing time, but we can see that the results of \ddetr are relatively stable w.r.t. to the number of queries. On the other hand, ProSeCo benefits from increasing the number of queries since it means a higher effective batch size during contrastive learning. However, the default value of $N = 300$ leads to the best results both with and without pretraining.}

\begin{table}[ht]
    \centering
    \begin{tabular}{@{}lcc@{}}
    \toprule
    Method & $N$ & Performance \\
    \midrule
        \multirow{4}{*}{Supervised} & 100 & 23.1 \\
        & 200 & 23.0 \\ 
        & 300 & \textbf{23.6} \\
         & 500 & 23.3 \\
     \midrule
        \multirow{4}{*}{\textit{ProSeCo (Ours)}} & 100 & 25.7 \\
        & 200 & 26.5 \\ 
        & 300 & \textbf{27.8} \\
         & 500 & 27.2 \\
    \bottomrule
    \end{tabular}
    \begin{center}
        \caption{Performance (mAP in \%) comparison on Mini-COCO 5\% when changing the number of object proposals in \ddetr.}
        \label{tab:abl_queries}
    \end{center}
\end{table}

\paragraph{\rev{Finetuning with a lot of data}}\label{sec:ft_full_coco}

\rev{In \cref{tab:full_coco}, we present results when finetuning the models on the full COCO dataset (118k training images) under the $1\times$ training schedule \cite{wei2021aligning, li2022dn}, \ie{} 12 training epochs and decaying the learning rate in the last epoch. The improvements in the large-scale annotated data regime are limited, which can be observed also in previous work \cite{dai2021up, bar2022detreg}. As we can see, after pretraining on IN, our ProSeCo reaches similar results than DETReg \cite{bar2022detreg}. We believe that this limitation comes from the pretrained backbone that stays fixed during pretraining, and from the extensive supervision during fine-tuning. However, as we can see in \cref{tab:voc_fsod}, we outperform DETReg on our \emph{FSOD-train} benchmark, which represents a setting with mid-scale annotated data (42k training images).}

\begin{table}[ht]
    \centering
    \begin{tabular}{@{}lccc@{}}
    \toprule
    \multirow{2}{*}{Method} & \multicolumn{3}{c}{COCO}\\
    \cmidrule(lr){2-4}
    & mAP & $\text{AP}_{50}$ & $\text{AP}_{75}$ \\
    \midrule
        Supervised & 37.4 & 55.5 & 40.5 \\
        DETReg \cite{bar2022detreg} & 38.9 & 56.6 & 42.3 \\
        \textit{ProSeCo (Ours)} & 38.9 & 56.2 & 42.4 \\
    \bottomrule
    \end{tabular}
    \begin{center}
        \caption{\rev{Performance (mAP, $\text{AP}_{50}$ and $\text{AP}_{75}$ in \%) comparison on the full COCO dataset with the $1\times$ training schedule.}}
        \label{tab:full_coco}
    \end{center}
\end{table}

\subsection{\rev{Pretraining cost}}\label{sec:cost}

\rev{
We compare the cost of pretraining \emph{in terms of memory and hardware used} to SoCo \cite{wei2021aligning} in \cref{tab:pt_cost} since it is the closest in terms of pretraining pipeline. The information are derived from their paper and official github repository.}

\begin{table}[ht]
    \centering
    \resizebox{0.98\linewidth}{!}{%
    \begin{tabular}{@{}lcccccc@{}}
    \toprule
    Method & IN epochs & Batch Size & Iterations & Time & It. / sec. & Hardware \\
    \midrule
    SoCo \cite{wei2021aligning} & 400 & 2048 & 240k & 140h & 0.5 & 16 V100 32G \\
    ProSeCo (Ours) & 10 & 64 & 187k & 40h & 1.4 & 8 A100 40G \\
    \bottomrule
    \end{tabular}%
    }
    \begin{center}
        \caption{Comparison of pretraining cost between overall pretraining methods. We compare the number of pretraining epochs on IN, the total batch size, the total number of iterations, the total training time (in hours), the number of iterations per seconds (It. / sec.), and the hardware used (number and type of GPUs). We can see that our pretraining is globally less costly than SoCo. }
        \label{tab:pt_cost}
    \end{center}
\end{table}


\rev{Even though A100 GPUs are faster than V100 GPUs, we are \emph{training much faster} which is partly explained by the fact that they learn the backbone along with the detection heads during pretraining, leading to more parameters to learn and more computations.
Furthermore, our ProSeCo requires a smaller batch size leading to less memory and thus less GPUs needed.}

\section{Conclusion}\label{sec:conclusion_self}

In this \rev{chapter}, we aim to use large unlabeled datasets for an unsupervised pretraining of the overall detection model to improve performance when having access to fewer labeled data. In this end, we propose \emph{Proposal Selection Contrast (ProSeCo)}, a novel pretraining approach for Object Detection. Our method leverages the large number of object proposals generated by transformer-based detectors for contrastive learning, reducing the necessity of a large batch size, and introducing the location information of the objects for the selection of positive examples to contrast. We show from various experiments on standard and novel benchmarks in learning with few training data that ProSeCo outperforms previous pretraining methods. Throughout this \rev{chapter}, we advocate for consistency in the level of information encoded in the features when pretraining. Indeed, learning object-level features during pretraining is more important than image-level when applied to a dense downstream task such as Object Detection. 

\rev{Unsupervised pretraining allows to reduce the total amount of labels used. It improves the performance when fine-tuning on limited data by providing a better initialization of the weights \emph{before the fine-tuning phase}. In the following chapter, we are interested in improving the performance \emph{after the fine-tuning phase} through semi-supervision.}

%% file: chapters/ssod.tex
\chapter{Few Annotation Learning for Semi-Supervised Object Detection}\label{chap:ssod}

\minitoc

\begin{abstract}
    \textit{
        In this chapter, we focus on the post-finetuning stage. We recall the positioning in the general training pipeline in \cref{fig:position_wacv_chap}. We want to improve object detectors based on Transformers and trained on few annotated data by leveraging unlabeled data through semi-supervision. 
        The present chapter is organized as follows. In \cref{sec:related_works_semi}, we review previous work in the topic of object detection and semi-supervised learning and discuss their relations with our proposed method. Then, in \cref{sec:mt_detr}, we introduce our proposed MT-DETR and the ideas it is built upon. In \cref{sec:experiments}, we present the results on different FAL benchmarks and an ablation study of the effect of the different architecture choices. Finally, we conclude the chapter in \cref{sec:ccl_semi}.
    }
\end{abstract}

\begin{figure}
    \centering
    \includegraphics[width=0.8\linewidth]{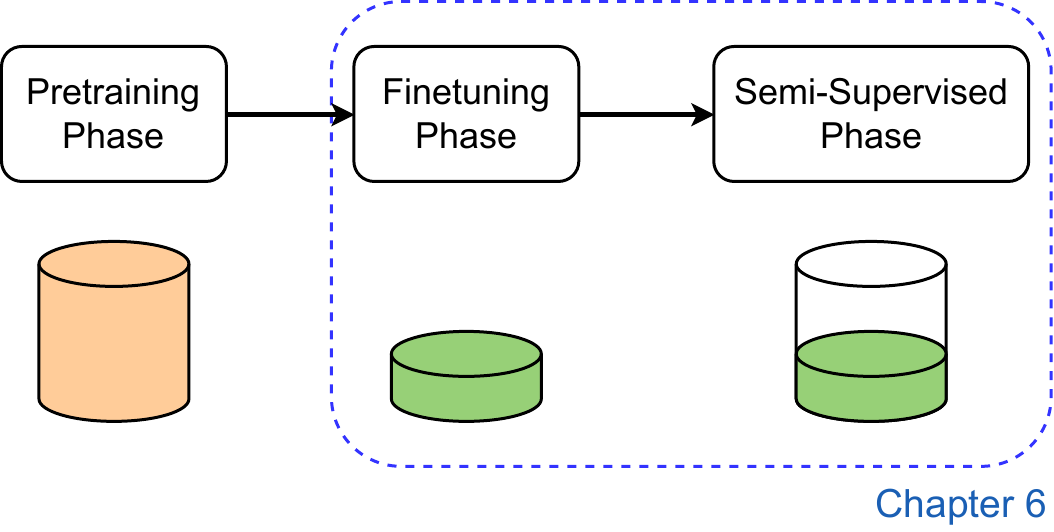}
    \caption{Illustration of the position of our contributions from this chapter in the full training pipeline presented in \cref{chap:intro}. We are interested on Semi-Supervised learning for Object Detection.}
    \label{fig:position_wacv_chap}
\end{figure}

\section{Introduction}

\rev{In the previous chapter, we studied how unsupervised pre-training can be beneficial for object detection task. In this new chapter, we consider a complementary setting where we can guide the learning using only a handful of labeled examples while simultaneously leveraging a large amount of unlabeled data. This corresponds to a particular case of \emph{semi-supervised learning} (SSL) called \emph{few-annotation learning} (FAL) hereafter.}

    
    For the task of \emph{Object Detection} (OD), 
    methods in the literature that tackle this setting~\cite{jeong2019consistency,sohn2020simple,liu2021unbiased,xu2021end,zhou2021instant,tang2021humble} have all considered object detectors based on traditional convolutional networks~\cite{ren2015faster} with a set of specific post-processing heuristics required for them to work~\cite{hosang2017learning,zhang2020bridging}. More recent object detectors are based on an encoder-decoder architecture using transformers \cite{vaswani2017attention} that allows for end-to-end OD without depending on this hand-crafted pipeline~\cite{carion2020end,zhu2020deformable}. However, they have not yet been tested in the SSL context. 
    
    The starting point of this \rev{contribution} is the observation that current state-of-the-art transformer-based architecture~\cite{zhu2020deformable} performs much better than traditional object detectors in a data-scarce fully supervised learning setting, also called \emph{Few-Shot Learning} (FSL), for an equal number of parameters.
    However, when plugging it into a state-of-the-art \emph{Semi-Supervised Object Detection} (SSOD) method~\cite{liu2021unbiased}, we observe that the model fails to converge, meaning that when used as is (\cref{fig:fal}), applying SSL methods from literature to transformer-based object detectors does not guarantee good results. 
    Thus, we propose a novel SSL method tailored for transformer-based architectures in order to take advantage of the effectiveness of transformers in FSL, and upscale these methods for FAL. Our proposed method achieves state-of-the-art in several FAL benchmarks. 
    
    More precisely, our contributions are summarized as: 
    \begin{itemize}
        \item After showcasing the strong performance of transformer-based detectors using few labeled data, we propose \emph{Momentum Teaching DETR} (MT-DETR), an approach for SSOD that leverages the specificities of transformer-based architectures and outperforms previous semi-supervised approaches in FAL settings.
        \item Contrary to convolution-based OD methods, our approach does not rely on heuristics and post-processing for constructing pseudo-labels. Thus, it eliminates sensitive hyperparameters.
    \end{itemize}


\rev{Before presenting these contributions in details, we provide an overview of related work close to the setting considered in this chapter.}

\section{Related Work}\label{sec:related_works_semi}


\subsection{Fully Supervised Object Detection}

\rev{We have seen in previous \cref{chap:selfod,chap:learning} that OD combines the tasks of object localization and classification. }
The most popular OD models have been based on fully convolutional neural networks~\cite{girshick2014rich, ren2015faster, redmon2016you}. These methods can be separated into \emph{two-stage}~\cite{girshick2014rich,girshick2015fast,ren2015faster,lin2017feature} or \emph{one-stage}~\cite{redmon2016you, liu2016ssd, tian2019fcos} detectors. The former methods make predictions of boxes and their class labels based on region proposals, \eg{} from a Region Proposal Network (RPN)~\cite{ren2015faster}, while the latter make predictions \wrt{} to anchors~\cite{lin2017focal} or a grid of possible reference points~\cite{redmon2016you, zhou2019objects, tian2019fcos}.
Both categories depend heavily on hand-designed heuristics, with the most prominent example being the Non-Maximal Suppression (NMS) post-processing, widely used in state-of-the-art OD methods~\cite{hosang2017learning, bodla2017soft}. 
\rev{Similarly to previous \cref{chap:selfod}, we remain interested in this chapter to the recent transformer-based architectures. }
We found in our experiments that \ddetr is a stronger baseline for FSL than the more popular \faster~\cite{ren2015faster} widely used in previous work, which motivated the focus on the transformer-based architectures.



\begin{table}[ht]
\centering
\resizebox{0.95\linewidth}{!}{%
\begin{tabular}{@{}llcccccc@{}}
\toprule
\multirow{2}{*}{Method} & \multirow{2}{*}{Params.} & \multicolumn{4}{c}{COCO} & \multicolumn{2}{c}{VOC07} \\
\cmidrule(lr){3-6}
\cmidrule(lr){7-8}
 & & 0.5\% (590) & 1\% (1180) & 5\% (5900) & 10\% (11800)  & 5\% (250) & 10\% (500) \\
\midrule
    FRCNN + FPN$^\dagger$ & 42M & $6.83 \pm 0.15$ & $9.05 \pm 0.16$ & $18.47 \pm 0.22$ & $23.86 \pm 0.81$ & $18.47 \pm 0.39$ & $25.23 \pm 0.22$ \\
    \ddetr & 40M & $\textbf{8.95} \pm \textbf{0.51}$ & $\textbf{12.96} \pm \textbf{0.08}$ & $\textbf{23.59} \pm \textbf{0.21}$ & $\textbf{28.55} \pm \textbf{0.08}$ & $\textbf{22.87} \pm \textbf{0.38}$ & $\textbf{29.03} \pm \textbf{0.46}$ \\
    $\Delta$ & & \textcolor{ForestGreen}{$+2.12$} & \textcolor{ForestGreen}{$+3.91$} & \textcolor{ForestGreen}{$+5.12$} & \textcolor{ForestGreen}{$+4.69$} & \textcolor{ForestGreen}{$+4.40$} & \textcolor{ForestGreen}{$+3.80$} \\
\bottomrule
\end{tabular}%
}
\begin{center}
    \caption{Performance (mAP in \%) comparison between \faster (FRCNN)~\cite{ren2015faster} with Feature Pyramid Network (FPN)~\cite{lin2017feature}, a two-stage detector commonly used in SSOD methods, and Deformable DETR (\ddetr)~\cite{zhu2020deformable}, a state-of-the-art transformer-based object detector, with the same ResNet-50 backbone model. 
    The performances are reported for different percentages (and the corresponding number of images) of COCO and VOC07 labeled training data. See \cref{sec:exp_details_semi} for more details on the experiments.
    \ddetr performs better than FRCNN + FPN with fewer labeled data for a similar amount of parameters.
    $^\dagger$: Results from \cite{liu2021unbiased} if available, from our reproduction otherwise.}
    \label{tab:comp_ddetr_frcnn}
\end{center}
\end{table}

\subsection{Semi-supervised Object Detection for Few-Annotation Learning}
The goal of semi-supervised learning is to take advantage of unlabeled data along with labeled data during training. In the more specific case of FAL, it allows reducing the need of a large amount of labeled data by leveraging the use of unlabeled data. \rev{The problem of SSL in computer vision was historically tackled first for the image classification task as presented \cref{chap:sota}. Our work takes inspiration from recent \emph{hybrid methods} \cite{sohn2020fixmatch, chen2020big} adapted to OD, by training a student model to match the predicted \emph{probability distributions} of proposals made by a teacher model.}


Methods in the literature for SSOD are mainly relying on pseudo-labels provided by a teacher model after applying strong data augmentations on unlabeled data~\cite{jeong2019consistency,sohn2020simple,liu2021unbiased,xu2021end,zhou2021instant,tang2021humble}. 
The use of geometric transformations in these strong augmentations is particularly important for OD~\cite{sohn2020simple}, due to the localization task intrinsic to the problem.  
The most recent and best performing ones~\cite{liu2021unbiased,xu2021end,tang2021humble} are also updating the teacher through Exponential Moving Average (EMA)~\cite{lillicrap2015continuous} of the student's weights to continuously improve the teacher and, thus, the pseudo-labels given to the student.
Although the use of EMA has improved the performance of the models, we propose in our work to stabilize the teacher, by applying an updating strategy throughout training, inspired by recent advances in self-supervised learning~\cite{grill2020bootstrap,caron2021emerging}.
Pseudo-labels are obtained, either by using a \emph{hard labeling}~\cite{jeong2019consistency,sohn2020simple,liu2021unbiased,xu2021end,zhou2021instant} approach, which consists in applying an $\arg \max$ to the predictions, or a \emph{soft labeling}~\cite{tang2021humble} approach, by fully using the predicted distribution. All the previous methods are relying on NMS and thresholding the \emph{confidence scores}, \ie{} the $\softmax$ of the predictions, given by the teacher model. However, the above-mentioned post-processing steps are sensitive to hyperparameters and introduce a bias into the model incentivizing it to be highly confident in its predictions, which may be suboptimal, particularly when few labeled data are available. Therefore, we aim to remove all these post-processing steps in this work.
Furthermore, SSOD methods in the literature have been exclusively built and evaluated using two-stage OD architectures, and we found that they do not work as is for the more recent detection models based on transformers. 


In this \rev{chapter}, we investigate SSOD through the lens of FAL, and we focus our experiments in this setting, in contrast to previous work that address FAL with only a limited number of experiments.

\section[A semi-supervised learning approach for transformer-based object detection]{A semi-supervised learning approach for \\ transformer-based object detection}\label{sec:mt_detr}

In this section, we first motivate our main idea to use a recent state-of-the-art transformer-based OD method in an SSL context by providing several results on both FSL and FAL settings. Then, we present Momentum Teaching DETR (MT-DETR), our transformer-based SSOD method more adapted to FAL and illustrated in \cref{fig:method}. More specifically, we describe the construction of the pseudo-labels for unlabeled data, and the update scheduling for the teacher model.

\subsection{How do object detectors handle data scarcity ?}\label{sec:fal_comp}

\begin{figure}
    \centering
    \includegraphics[width=0.9\linewidth]{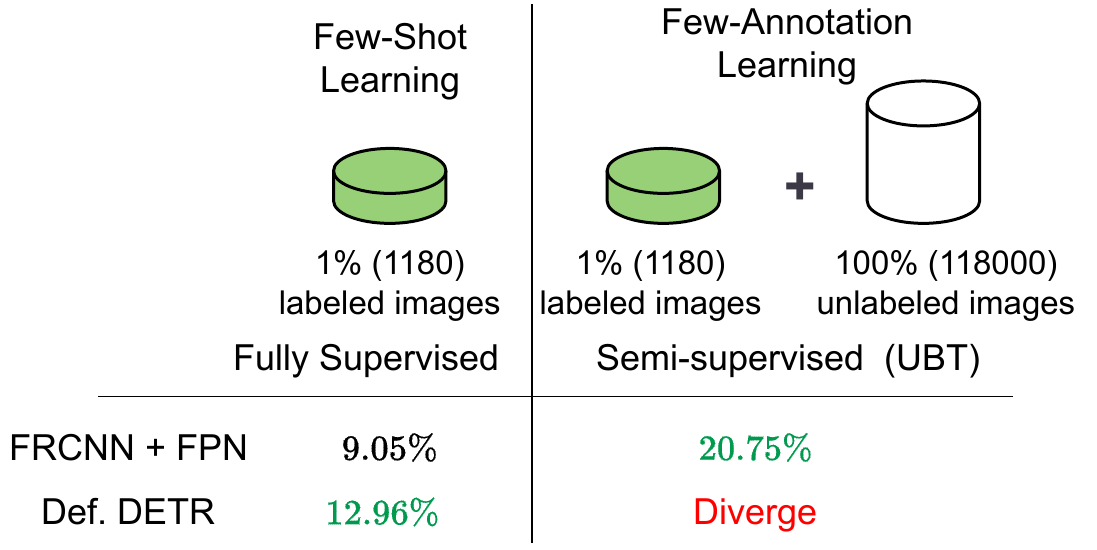}
    \caption{Comparison of mean final performance (mAP in \%) between \faster (FRCNN)~\cite{ren2015faster} with Feature Pyramid Network (FPN) \cite{lin2017feature} and Deformable DETR (\ddetr)~\cite{zhu2020deformable} in the Few-Shot and Few-Annotation Learning settings, using only 1\% of labeled data on COCO (about 1180 images). See \cref{sec:exp_details_semi} for experimental details. 
    In the fully supervised case, \ddetr achieves better results than FRCNN. However, in the semi-supervised case implemented in Unbiased Teacher (UBT)~\cite{liu2021unbiased}, \ddetr cannot converge. 
    }
    \label{fig:fal}
\end{figure}

From the results presented in \cref{tab:comp_ddetr_frcnn}, we can see that Deformable DETR (\ddetr) \cite{zhu2020deformable}, a recent state-of-the-art detection model based on transformers, achieves consistently better performance than the most popular two-stage method in FSL. We refer the reader to \cref{sec:exp_details_semi} for all the implementation details. 

These results motivated us to implement \ddetr in a state-of-the-art SSOD method to see how it performs in FAL settings. We opted for the recent Unbiased Teacher (UBT)~\cite{liu2021unbiased}, as its strong results in FAL were easily reproducible with the provided code.
Surprisingly, we observed that with \ddetr detector, the model does not converge in all the FAL settings tested: 1\% of COCO as labeled data (\ie{} about 1180 labeled images), 5\% and 10\% of VOC 07 (\ie{} 250 and 500 labeled images respectively). Even though it passes by an early best (about 17\% mAP on 1\% of COCO) at the beginning of training, the model collapses soon after. This diverging behavior is not satisfying in practice, even more so that the same method used with a \faster~\cite{ren2015faster} architecture converges (it achieves about 20\% final mAP on 1\% of COCO) in similar settings (\cf{} \cref{fig:fal}).
All of this shows that current state-of-the-art SSOD methods are not adapted to more recent transformer-based architectures. 

Inspired by these results, we propose an SSL method tailored for transformer-based OD called \emph{Momentum Teaching DETR (MT-DETR)}.


\subsection{Overview of our approach}

\begin{figure*}[!t]
    \centering
    \includegraphics[width=1.\linewidth]{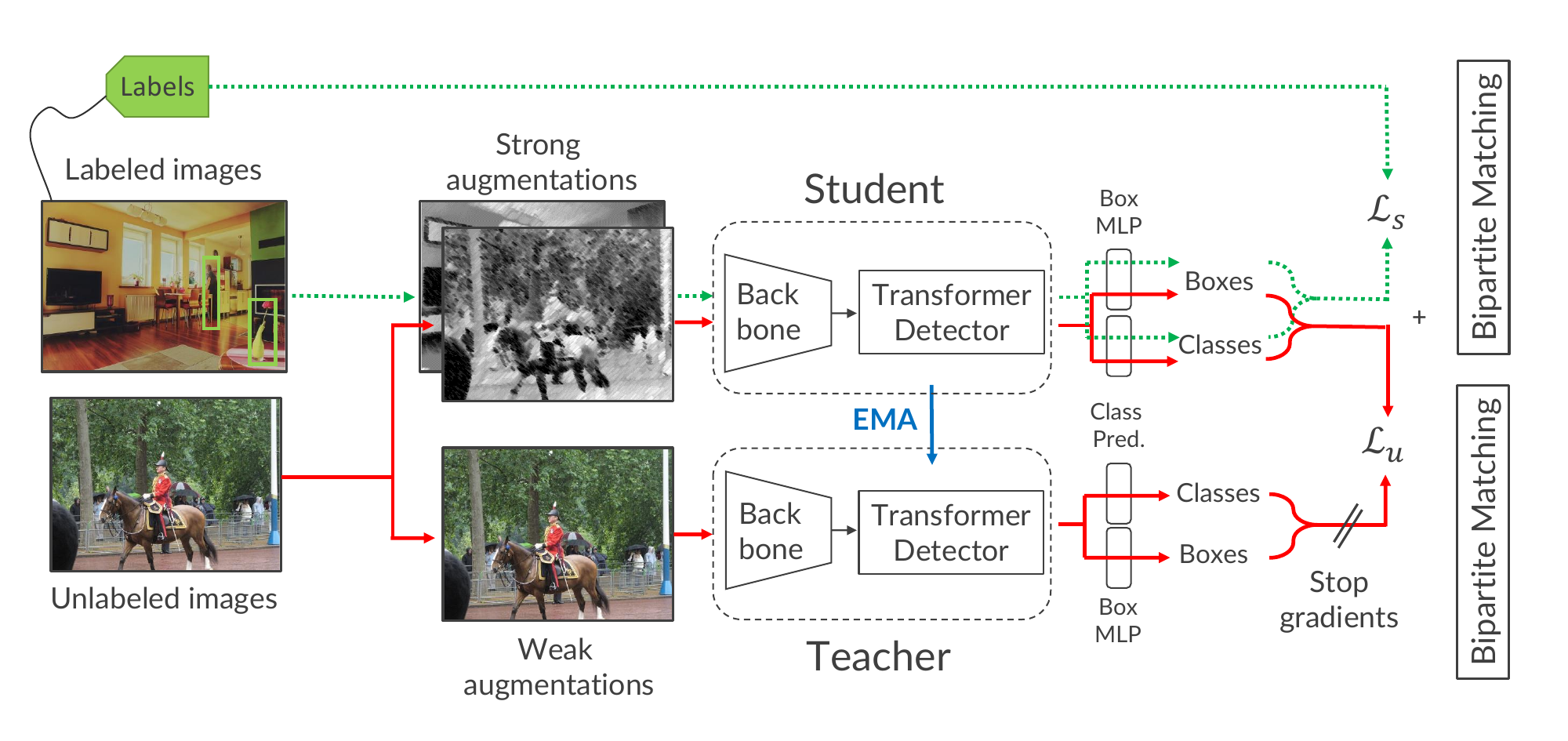}
    \caption{Overview of our Momentum Teaching DETR (MT-DETR) approach for SSOD. The method follows a student-teacher architecture, with the teacher updated through an Exponential Moving Average (EMA) of the student. The keep rate parameter for the EMA follows a \emph{cosine scheduling}. In the supervised branch (\emph{in dotted and \textcolor{ForestGreen}{green}}), the supervised loss $\Lcal_s$ is computed with the predictions of the student on the labeled images. In the unsupervised branch (\emph{in straight and \textcolor{red}{red}}), the \emph{raw}, \ie{} \emph{unprocessed}, outputs of the teacher model for the weakly augmented unlabeled images are used as \emph{soft} pseudo-labels without applying any heuristic like NMS or confidence thresholding. After finding the best corresponding detection proposals with bipartite matching, the student model learns from the strongly augmented images to match the distribution of class probabilities and the bounding boxes in these pseudo-labels through the unsupervised loss $\Lcal_u$.}
    \label{fig:method}
\end{figure*}
As shown in \cref{fig:method}, our approach is composed of a \emph{student-teacher architecture}, which is common for semi-supervised learning~\cite{tarvainen2017mean,sohn2020fixmatch}.
\rev{Even though the general structure of the approach bears similarities with our ProSeCo presented in the previous \cref{chap:selfod}, such as the student-teacher architecture, the strong and weak augmentations, and the dual bipartite matching, there are some important differences coming from the practical data conditions considered. With ProSeCo, the input data being fully unlabeled, both Hungarian algorithms are unsupervised, and the localization task has to be learned partly from region proposals given by an unsupervised algorithm \cite{uijlings2013selective}. This is important to avoid a collapse of the model if the student relies only on the teacher. In the FAL settings that we consider here, the supervision from the labels, even with few of them, stabilizes the training and avoids a collapse of the model. Furthermore, the model has also access to class information from the supervision, such as the total number of class and also reusing classification layers for pseudo-labeling. We describe in this section the full semi-supervised training process.}

Both student and teacher models are initialized from a fully supervised model trained on the \emph{few labeled data} available. Then, during the semi-supervised training, the method takes as inputs a batch of labeled images $\mathcal{B}^l = \{(\rvx^l_i, \rvy^l_i)\}_{i=1}^{N^l}$ and a batch of unlabeled images $\mathcal{B}^u = \{\rvx^u_i\}_{i=1}^{N^u}$. We define $\rvx^l_i$ and $\rvx^u_i$ as the $i$\textsuperscript{th} labeled and unlabeled image respectively, $\rvy^l_i = \{\rvy^l_{(i,j)} \}_{j=1}^{k_i} = \{ (c^l_{(i,j)}, \rvb^l_{(i,j)}) \}_{j=1}^{k_i} \in \{ \{1,2,\dots,C\} \times \mathbb{R}^4 \}_{j=1}^{k_i}$ as the corresponding $k_i$ ground truth class labels and box coordinates, and finally, $N^l$ and $N^u$ are respectively the labeled and unlabeled batch sizes.
The student model is updated by a weighted combination of a supervised loss $\Lcal_s$ and an unsupervised loss $\Lcal_u$ with weight $\lambda_u \in \mathbb{R}^+$:
\begin{equation}
\Lcal(\mathcal{B}^l, \mathcal{B}^u) = \frac{1}{N^l} \Lcal_s(\mathcal{B}^l) + \frac{\lambda_u}{N^u} \Lcal_u(\mathcal{B}^u).
\end{equation}

\noindent Below, we first describe the \emph{supervised branch}, which computes the \emph{supervised loss} using the batch of labeled data $\mathcal{B}^l$. Then, we detail the \emph{unsupervised branch}, which computes the \emph{unsupervised loss} with the batch of unlabeled data $\mathcal{B}^u$. 

\textbf{Supervised branch}
To compute the supervised loss, the supervised branch follows the supervised learning of \ddetr~\cite{zhu2020deformable}, which is an improved version of  DETR~\cite{carion2020end}.
For each image $\rvx^l_i$, the student model infers $N$ predictions $\hat{\rvy}^l_i = \{\hat{\rvy}^l_{(i,j)}\}_{j=1}^N = \{ (\hat{\rvc}^l_{(i,j)},\hat{\rvb}^l_{(i,j)}) \}_{j=1}^N$ of boxes $\hat{\rvb}^l_{(i,j)}$ 
and their associated predicted labels \emph{logits} $\hat{\rvc}^l_{(i,j)} \in \mathbb{R}^{C+1}$, with the ($C+1$)\textsuperscript{th} logit representing the \emph{no object} ($\varnothing$) class. 
Then, the Hungarian algorithm \cite{munkres1957algorithms} finds from all the permutations of $N$ elements $\mathfrak{S}_N$, the optimal bipartite matching $\hat{\sigma}^l_i$ between the predictions $\hat{\rvy}^l_i$ of the student model and the ground truth labels $\rvy^l_i$:
$\hat{\sigma}^l_i = \arg \min_{\sigma \in \mathfrak{S}_N}  \sum_j^N \Lcal_{\text{match}}(\rvy^l_{(i,j)}, \hat{\rvy}^l_{(i,\sigma(j))}).$
Thus, for each labeled image $\rvx^l_i$, the $j$\textsuperscript{th} ground truth $\rvy^l_{(i,j)}$ is associated to $\hat{\sigma}^l_i(j)$.
Similarly to the loss used in object detectors, the matching cost $\Lcal_{\text{match}}$ used in the Hungarian algorithm takes into account both class and bounding box predictions through a linear combination of the \emph{Focal loss}~\cite{lin2017focal} $\Lcal_{\text{focal}}$, the $\ell_1$ loss of the box coordinates, and the generalized IoU loss~\cite{rezatofighi2019generalized} $\Lcal_\text{giou}$, respectively. These loss functions are then used to compute the supervised loss $\Lcal_s$ as well:
\begin{align}
\begin{split}
& \Lcal_\text{match}(\rvy^l_{(i,j)}, \hat{\rvy}^l_{(i,\sigma(j))}) = \1_{\{ \hat{\rvc}^l_{(i,\sigma(j))} \neq \varnothing \}} \Bigl[ \lambda_\text{class} \Lcal_\text{focal}\left(c^l_{(i,j)}, \hat{\rvc}^l_{(i,\sigma(j))}\right) \\ 
& \qquad + \lambda_{\ell_1} \|\rvb^l_{(i,j)} - \hat{\rvb}^l_{(i,\sigma(j))}\|_1 + \lambda_\text{giou} \Lcal_\text{giou}\left(\rvb^l_{(i,j)}, \hat{\rvb}^l_{(i,\sigma(j))}\right)\Bigr], \\
\end{split} \\
\begin{split}
& \Lcal_s(\mathcal{B}^l) = \sum_{i=1}^{N^l} \sum_{j=1}^N \Bigl[ \lambda_\text{class} \Lcal_{\text{focal}}\left(c^l_{(i,j)}, \hat{\rvc}^l_{(i,\hat{\sigma}^l_i(j))}\right) \\
&\qquad + \1_{\{ \hat{\rvc}^l_{(i,\hat{\sigma}^l_i(j))} \neq \varnothing \}} \lambda_{\ell_1} \|\rvb^l_{(i,j)} - \hat{\rvb}^l_{(i,\hat{\sigma}^l_i(j))}\|_1 \\
&\qquad + \1_{\{ \hat{\rvc}^l_{(i,\hat{\sigma}^l_i(j))} \neq \varnothing \}} \lambda_\text{giou} \Lcal_{\text{giou}}\left(\rvb^l_{(i,j)}, \hat{\rvb}^l_{(i,\hat{\sigma}^l_i(j))}\right) \Bigr]. 
\end{split}
\end{align}%
\noindent In the above equations, we define $\lambda_\text{class}, \lambda_{\ell_1}, \lambda_\text{giou} \in \mathbb{R}^+$ as the coefficients in the matching cost and $\1_{\mathcal{X}}$ the \emph{indicator function}, such that $\forall x, \1_{\mathcal{X}}(x) = 1$ iff $x \in \mathcal{X}$.


\textbf{Unsupervised branch}
Our main contribution is the unsupervised loss for transformer-based OD. In the unsupervised branch, we produce two different views for each unlabeled image $\rvx^u_i$: a \emph{weakly augmented view} ${\rvx^u_i}^\prime$ and a \emph{strongly augmented view} ${\rvx^u_i}''$\footnote{The weak and strong augmentations are described in \cref{sec:exp_details_semi}.}. Then, the teacher model provides \emph{soft pseudo-labels} $\rvy^u_i = \{\rvy^u_{(i,j)}\}_{j=1}^N = \{ (\rvc^u_{(i,j)},\rvb^u_{(i,j)}) \}_{j=1}^N$, with $\rvc^u_{(i,j)}$ the predicted \emph{logits}, for each \emph{weakly augmented} unlabeled image ${\rvx^u_i}^\prime$, and the student model infers predictions $\hat{\rvy}^u_i = \{\hat{\rvy}^u_{(i,j)}\}_{j=1}^N = \{ (\hat{\rvc}^u_{(i,j)},\hat{\rvb}^u_{(i,j)}) \}_{j=1}^N$  from the corresponding \emph{strongly augmented} unlabeled view ${\rvx^u_i}''$.

We apply the same Hungarian algorithm with the same matching cost $\Lcal_{\text{match}}$ to obtain the best permutation $\hat{\sigma}^u_i = \arg \min_{\sigma \in \mathfrak{S}_N}  \sum_j^N \Lcal_{\text{match}}(\rvy^u_{(i,j)}, \hat{\rvy}^u_{(i,\sigma(j))})$, that matches the predictions of the student with the closest pseudo-label.
In the unsupervised loss $\Lcal_u$, we follow the consistency regularization paradigm~\cite{bucilua2006model,chen2020big,caron2021emerging}. We train the student network to match the probability distributions of the classes predicted by the student with the soft pseudo-labels proposed by the teacher. We learn to match these distributions by minimizing the cross-entropy between the two class distribution outputs normalized by a $\softmax$ function. We define respectively:
\begin{equation}
{p^s_{(i,j)}}^{(k)} = \softmax(\rvc^u_{(i,j)})^{(k)} = \frac{\exp({c^u_{(i,j)}}^{(k)})}{\sum_{n=1}^{C+1}\exp({c^u_{(i,j)}}^{(n)})},
\end{equation}

\noindent and ${p^t_{(i,j)}}^{(k)} = \text{softmax}(\hat{\rvc}^u_{(i,j)})^{(k)}$, the student and teacher class distribution outputs, where $c^{(k)}$ is the $k$\textsuperscript{th} element of $\rvc, \forall \rvc \in \mathbb{R}^{C+1}$. Then the cross-entropy loss is defined as:
\begin{equation}
\Lcal_{\text{CE}}(\rvc^u_{(i,j)},\hat{\rvc}^u_{(i,j)})  = - \sum_{k=1}^{C+1} {p^s_{(i,j)}}^{(k)} \log {p^t_{(i,j)}}^{(k)},
\end{equation}
and finally, we compute the unsupervised loss $\Lcal_u$ as:
\begin{equation}
\begin{split}
\Lcal_u(\mathcal{B}^u) = & \sum_{i=1}^{N^u} \sum_{j=1}^N \Bigl[ \lambda_\text{class} \Lcal_{\text{CE}}\left(\rvc^u_{(i,j)}, \hat{\rvc}^u_{(i,\hat{\sigma}^u_i(j))}\right) \\
&+ \1_{\{ \hat{\rvc}^u_{(i,\hat{\sigma}^u_i(j))} \neq \varnothing \}} \lambda_{\ell_1} \| \rvb^u_{(i,j)} - \hat{\rvb}^u_{(i,\hat{\sigma}^u_i(j))} \|_1 \\
&+ \1_{\{ \hat{\rvc}^u_{(i,\hat{\sigma}^u_i(j))} \neq \varnothing \}} \lambda_\text{giou} \Lcal_{\text{giou}}\left(\rvb^u_{(i,j)}, \hat{\rvb}^u_{(i,\hat{\sigma}^u_i(j))}\right) \Bigr].
\end{split}
\end{equation}

For FAL, we have little information from the labeled data. Therefore, the quality of the pseudo-labels and their contained information play an important part in the training.

\subsection{Construction of the pseudo-labels}

As mentioned above, the unsupervised loss $\Lcal_u$ takes into account the class predictions through a cross-entropy between the outputs of the student model and the matched outputs of the teacher model. We use the $\softmax$ of the outputs of the teacher model as \emph{soft pseudo-labels} for the cross-entropy, as opposed to \emph{hard pseudo-labels} obtained after taking the $\arg \max$.  

Following the DETR philosophy~\cite{carion2020end}, we give to the students the \emph{raw soft pseudo-labels} obtained from the teacher, \ie{} we remove all handmade heuristics to process the teacher outputs, namely, the NMS and confidence thresholding. Both of these post-processing steps are sensitive to hyperparameters and restrict the diversity in the pseudo-labels. By introducing a bias to keep the most confident proposals, they have the unwanted effect of encouraging the models to always be highly confident in their predictions. 
In the case of FAL, where we have access to only a few labeled examples for each class, the model might not be confident for some classes, leading them to be discarded early by the post-processing. Relying on the model's confidence in certain predictions can be tricky. 
Using the full distributions makes the model less prone to focus on being highly confident in their predictions, and forces the model to take into account the relations between classes. 
Furthermore, the Hungarian algorithm used in transformer-based OD methods leverages the diversity of proposals given by the model and benefits from the fact that the model is not overconfident on a single class thanks to the matching loss. 
Indeed, the bipartite matching can favor proposals with better localizations even if the model is less confident in its class predictions, making the use of raw soft pseudo-labels more suitable for transformer-based detectors.

To obtain strong and insightful pseudo-labels helping the student, the teacher must be updated throughout training. We describe the update process in the following section.

\subsection{Updating the Teacher model}

To avoid a poor supervision from the teacher, its weights $\theta_t$ are updated by an Exponential Moving Average (EMA) from the student's weights $\theta_s$ using a keep rate $\alpha \in [0,1]$, \rev{which we recall here from \cref{chap:selfod}}:
\begin{equation}
\theta_t \leftarrow \alpha \theta_t + (1-\alpha) \theta_s.
\end{equation}
For $\alpha=1$, the teacher is constant and for $\alpha=0$ its weights are equal to the student's. Therefore, there is a trade-off between a too high and too low keep rate parameter. 
Inspired by the Self-supervised learning literature~\cite{grill2020bootstrap, caron2021emerging}, we update $\alpha$ following a \emph{cosine scheduling} from $\alpha_{\text{start}}$ to $\alpha_{\text{end}}$:
\begin{equation}
\alpha \triangleq \alpha_\text{end} - (\alpha_\text{end} - \alpha_{\text{start}}) \cdot (\cos(\pi k /K)+1)/2, 
\end{equation}
with $k$ the current \emph{epoch} and $K$ the maximum number of \emph{epochs}. 
This scheduling stabilizes the teacher model, especially in the last training iterations, to make it converge at the end of training.

\begin{table}
\centering
\resizebox{0.95\linewidth}{!}{%
\begin{tabular}{@{}lccccc@{}}
\toprule
\multirow{2}{*}{Augmentations} & \multirow{2}{*}{Probability} & \multirow{2}{*}{Parameters} & \multirow{2}{*}{Supervised branch} & \multicolumn{2}{c}{Unsupervised branch} \\
\cmidrule(lr){5-6}
& & & & Weak & Strong \\
\midrule
Horizontal Flip & 0.5 & -- & \checkmark & \checkmark & \checkmark \\
\midrule
Resize & 1.0 & short edge $\in$ range(480,801,32) & \checkmark & \checkmark & \checkmark \\
\midrule
\multirow{2}{*}{Color Jitter} & \multirow{2}{*}{0.8} & (brightness, contrast, saturation, hue) & \multirow{2}{*}{\checkmark} & & \multirow{2}{*}{\checkmark} \\
 & & = (0.4, 0.4, 0.4, 0.1) & & & \\
\midrule
Grayscale & 0.2 & -- & \checkmark & & \checkmark \\
\midrule
Gaussian Blur & 0.5 & $\sigma \in [0.1, 2.0]$ & \checkmark & & \checkmark \\
\midrule
\multirow{3}{*}{CutOut} & 0.7 & scale $\in [0.05, 0.2]$, ratio $\in [0.3, 3.3]$ & \checkmark & & \checkmark \\
 & 0.5 & scale $\in [0.02, 0.2]$, ratio $\in [0.1, 6]$ & \checkmark & & \checkmark  \\
 & 0.3 & scale $\in [0.02, 0.2]$, ratio $\in [0.05, 8]$ & \checkmark & & \checkmark \\
\midrule
Rotate & 0.3 & degrees $\in [-30,30]$ & & & \checkmark \\
\midrule
Shear & 0.3 & $\text{shear}_x \in [-30,30]$, $\text{shear}_y \in [-30,30]$ & & & \checkmark \\
\midrule
Rescale + Pad & \multirow{2}{*}{0.5} & $\text{translate}_x \in [0,0.25], \text{translate}_y \in [0,0.25]$ & & & \multirow{2}{*}{\checkmark} \\
+ Translation & & $\text{scale}_x \in [0.25,0.75], \text{scale}_y \in [0.25,0.75]$ & & & \\
\bottomrule
\end{tabular}%
}
\begin{center}
    \caption{The different sets of augmentations used during SSL for each branch. The Horizontal Flip and Resize augmentations follow standard supervised training~\cite{carion2020end,zhu2020deformable}. The Color Jitter, Grayscale, Gaussian Blur and CutOut augmentations follow Unbiased Teacher~\cite{liu2021unbiased} training, and the geometric augmentations (Rotate, Shear, Rescale, Pad and Translation) follow Soft Teacher~\cite{xu2021end} training.}
\label{tab:augmentations_semi}
\end{center}
\end{table}



\section{Experimental Results}\label{sec:experiments}

\begin{table}
\centering
\resizebox{0.99\linewidth}{!}{%
\begin{tabular}{@{}lllcccc@{}}
\toprule
\multirow{2}{*}{Method} & \multirow{2}{*}{Pretrain.} & \multirow{2}{*}{Arch.} & \multicolumn{4}{c}{FAL-COCO} \\
\cmidrule(lr){4-7}
& & & 0.5\% (590) & 1\% (1180) & 5\% (5900) & 10\% (11800) \\
\midrule
    Supervised & Sup. & 2S & $6.83 \pm 0.15$ & $9.05 \pm 0.16$ & $18.47 \pm 0.22$ & $23.86 \pm 0.81$ \\
    STAC~\cite{sohn2020simple} & Sup. & 2S & $9.78 \pm 0.53$ & $13.97 \pm 0.35$ & $24.38 \pm 0.12$ & $28.64 \pm 0.21$ \\
    Instant-Teaching~\cite{zhou2021instant} & Sup. & 2S & -- & $18.05 \pm 0.15$ & $26.75 \pm 0.05$ & $30.40 \pm 0.05$ \\
    Humble Teacher~\cite{tang2021humble} & Sup. & 2S & -- & $16.96 \pm 0.38$ & $27.70 \pm 0.15$ & $31.61 \pm 0.28$ \\
    Unbiased Teacher~\cite{liu2021unbiased} & Sup. & 2S & $16.94 \pm 0.23$ & $20.75 \pm 0.12$ & $28.27 \pm 0.11$ & $31.50 \pm 0.10$ \\
    Soft Teacher~\cite{xu2021end} & Sup. & 2S & -- & $20.46 \pm 0.39$ & $30.74 \pm 0.08$ & $34.04 \pm 0.14$ \\
\midrule
    \rev{Supervised} & \rev{Sup.} & \rev{T} & \rev{$8.95 \pm 0.51$} & \rev{$12.96 \pm 0.08$} & \rev{$23.59 \pm 0.21$} & \rev{$28.55 \pm 0.08$} \\ 
    \rev{Supervised} & \rev{ProSeCo} & \rev{T} & \rev{$11.37 \pm 0.40$} & \rev{$17.9 \pm 0.08$} & \rev{$28.33 \pm 0.33$} & \rev{$32.60 \pm 0.28$} \\ 
    MT-DETR \emph{(Ours)} & Sup. & T & $\mathbf{17.84} \pm 0.54$ & $\mathbf{22.03} \pm 0.17$ & $\mathbf{31.00} \pm 0.11$ & $\mathbf{34.52} \pm 0.07$ \\
    \rev{MT-DETR \emph{(Ours)}} & \rev{ProSeCo} & \rev{T} & \rev{$14.33 \pm 0.17$} & \rev{$21.73 \pm 0.12$} & \rev{$\mathbf{32.00} \pm 0.16$} & \rev{$\mathbf{35.83} \pm 0.17$} \\
\bottomrule
\end{tabular}
}%
\begin{center}
    \caption{Performance (mAP in \%) of our proposed approach on FAL-COCO, using different percentage of labeled data (with the corresponding number of images reported) and 100\% of the dataset as unlabeled data. \rev{We also report the \emph{pretrained (Pretrain.)} backbone used, the \emph{architecture (Arch.) } of the underlying object detector which is either \emph{Two-Stage (2S)} or \emph{Transformer-based (T)}.} }
\label{tab:perf_coco}
\end{center}
\end{table}

\begin{table}
\centering
\resizebox{0.95\linewidth}{!}{%
\begin{tabular}{@{}llccc@{}}
\toprule
\multirow{2}{*}{Method} & \multirow{2}{*}{Arch.} & \multicolumn{3}{c}{FAL-VOC 07-12} \\
\cmidrule(lr){3-5}
& & 5\% (250) & 10\% (500) & 100\% (5000) \\
\midrule
    STAC~\cite{sohn2020simple} & 2S & -- & -- & 44.64 \\
    Instant-Teaching~\cite{zhou2021instant} & 2S & -- & -- & 50.00 \\
    Humble Teacher~\cite{tang2021humble} & 2S & -- & -- & 53.04 \\
    Unbiased Teacher [GitHub] & 2S & -- & -- & 54.48 \\
    Unbiased Teacher$^*$ & 2S & $35.98 \pm 0.71$ & $40.34 \pm 0.95$ & 54.61 \\
\midrule
    MT-DETR \emph{(Ours)} & T & $\mathbf{36.95} \pm 0.53$ & $\mathbf{43.15} \pm 1.10$ & \textbf{56.2} \\ 
\bottomrule
\end{tabular}%
}
\begin{center}
    \caption{Performance (mAP in \%) of our proposed approach on VOC with fully labeled VOC07 and unlabeled VOC12 to compare with previous work, and in the novel FAL-VOC 07-12 settings. Different percentage of VOC07 are used as labeled data (5\%, 10\% or 100\%, with the corresponding number of images reported), and the full VOC12 dataset is used as unlabeled data. \rev{We also report the \emph{architecture (Arch.) } of the underlying object detector which is either \emph{Two-Stage (2S)} or \emph{Transformer-based (T)}}. $^*$ indicates our implementation of Unbiased Teacher~\cite{liu2021unbiased} in this novel setting to compare with our approach. [GitHub] : updated results after publication~\cite{liu2021unbiased} taken from their official code released.\cref{fn:ubt_code}}
\label{tab:perf_voc}
\end{center}
\end{table}

In this section, we present a comparative study of the results of our method to the state-of-the-art on FAL benchmarks, as well as an ablative study on the most relevant parts. Before that, we detail the datasets, the evaluation and training settings used for the different experiments. 

\subsection{Datasets, evaluation and training details}\label{sec:exp_details_semi}

\textbf{Datasets and evaluation protocol} 
\rev{We briefly recall the datasets introduced in \cref{chap:sota} that we used here.}
To evaluate our proposed method, we use the MS-COCO (COCO)~\cite{lin2014microsoft} and PASCAL VOC (VOC)~\cite{everingham2010pascal} datasets which are standard for object detection, following the settings of existing works~\cite{jeong2019consistency,sohn2020simple,liu2021unbiased,xu2021end,zhou2021instant,tang2021humble}. COCO is a dataset with 80 classes, and VOC contains 20 classes.
We are specifically interested in two \emph{Few Annotation Learning} (FAL) settings: \\
On \emph{FAL-COCO}, we randomly sample 0.5, 1, 5 or 10\% (respectively about 590, 1180, 5900 and 11800 images) of the training set (\emph{train2017}) used as the labeled set and use the full training set for the unlabeled set (about 118k images). Performance is evaluated on \emph{val2017}. \\
On \emph{FAL-VOC 07-12}, we restrict the labeled training set (VOC07 \emph{trainval}) to a random sample of 5 or 10\% (respectively 250 and 500 labeled images), and use the full VOC12 \emph{trainval} (about 11k images) as unlabeled training set. We introduce this novel setting to evaluate our approach in a FAL setting on VOC. We also compare the results with previous SSOD methods using the full VOC07 \emph{trainval} labeled training set (5k labeled images) and VOC12 \emph{trainval} as unlabeled training set. Performance is evaluated on VOC07 \emph{test} set. 

In all settings, performance is reported and compared using the $AP_{50:95}$ (mAP, in \%) evaluation metric using the official COCO and VOC evaluation codes, respectively.

\textbf{Training}
For a fair comparison, a fully supervised ResNet-50~\cite{he2016deep} pretrained on ImageNet~\cite{russakovskyImageNetLargeScale2015} is used as a backbone for all the methods. 
For fine-tuning \ddetr~\cite{zhu2020deformable} on the few labeled data, we train the model with a batch size of 32 images on 8 GPUs until the validation performance stops increasing, \ie{} for COCO, up to 2000 epochs for 1\%, 500 epochs for 5\%, 400 epochs for 10\%, and for VOC, up to 2000 epochs for both 5\% and 10\%. 
For semi-supervised learning, we train MT-DETR for 50 (respectively, 250) epochs of the unlabeled data on COCO (respectively, VOC) with a batch size of 48 labeled images and 48 unlabeled images (respectively, 24 and 24) on 8 GPUs. All experiments with less than 100\% of labeled data are reproduced on 3 different random subsets\footnote{\:\url{https://github.com/CEA-LIST/MT-DETR}}.
The training hyperparameters, are defined as in \ddetr~\cite{zhu2020deformable}. 
The coefficients for the losses are set as $\lambda_\text{class}=2, \lambda_{\ell_1}=5, \lambda_\text{giou}=2$, and $\lambda_u = 4$. Following the training schedule of \ddetr, we always decay the learning rates by a factor of 0.1 after about 80\% of training. The keep rate parameter $\alpha$ follows a \emph{cosine scheduling} from $\alpha_\text{start} = 0.9996$ to $\alpha_\text{end} = 1$, with the value of $\alpha_\text{start}$ chosen according to previous work~\cite{liu2021unbiased}. \\
When using Unbiased Teacher~\cite{liu2021unbiased}, we follow the official implementation\footnote{\label{fn:ubt_code}\noindent \:\url{https://github.com/facebookresearch/unbiased-teacher}} and the hyperparameters provided.

\textbf{Augmentations}
For strong and weak data augmentations, we follow the common data augmentations used in previous works~\cite{sohn2020simple,liu2021unbiased,xu2021end}. We apply a random resizing and random horizontal flip for weak augmentations. We randomly add color jittering, grayscale, Gaussian blur, CutOut patches for strong augmentations and also randomly add rescaling, translation with padding, shearing and rotating as geometric transformations~\cite{sohn2020simple} in strong augmentations. 
In the supervised branch, images are also randomly augmented using weak and strong augmentations without any geometric transformations following Soft Teacher~\cite{xu2021end} practices. It helps the student model to be augmentation-agnostic, to better predict pseudo-labels coming from non-augmented images in the unsupervised branch. We remove the CutOut augmentation in the supervised branch in the most difficult settings of FAL-COCO 0.5\% and 1\%, since it can cover the only labeled small boxes available and is counterproductive.  
All the parameters for the different augmentations can be found in \cref{tab:augmentations_semi}.

\begin{table}
\centering
\resizebox{1.\linewidth}{!}{%
\begin{tabular}{@{}lcccccccccc@{}}
\toprule
\multirow{2}{*}{Ablative Variant} & \multicolumn{2}{c}{EMA Scheduling} & \multicolumn{2}{c}{Initialization} & \multirow{2}{*}{\textcolor{red}{NMS}} & \multicolumn{4}{c}{Confidence Thresholding} & \multirow{2}{*}{mAP (in \%)} \\
\cmidrule(lr){2-3}
\cmidrule(lr){4-5}
\cmidrule(lr){7-10}
& \textcolor{ForestGreen}{\textbf{Cosine}} & \textcolor{red}{Constant} & \textcolor{ForestGreen}{\textbf{After FT}} & \textcolor{red}{From scratch} & & \textcolor{ForestGreen}{\textbf{\o}} & \textcolor{red}{0.5} & \textcolor{red}{0.7} & \textcolor{red}{0.9} & \\
\midrule
Best & \checkmark & & \checkmark & & & \checkmark & & & & \textbf{22.25} \\
\midrule
Abl\onedot Sched\onedot &  & \checkmark & \checkmark & & & \checkmark & & & & 21.48 \\
\midrule
Abl\onedot Init\onedot & \checkmark &  & & \checkmark & & \checkmark & & & & 16.51 \\
\midrule
Abl. NMS & \checkmark &  & \checkmark & & \checkmark & \checkmark & & & & 19.85 \\
\midrule
\multirow{3}{*}{Abl\onedot Thresh\onedot} & \checkmark &  & \checkmark & & & & \checkmark & & & 10.26 \\
 & \checkmark &  & \checkmark & & & & & \checkmark & & 17.34 \\
 & \checkmark &  & \checkmark & & & & & & \checkmark & 12.37 \\
\bottomrule
\end{tabular}%
}
\begin{center}
    \caption{Ablation studies of the different parts of our method. \textcolor{ForestGreen}{\textbf{Green and bold columns names}} indicate a \emph{positive} effect on the performance and \textcolor{red}{red columns} a \emph{negative} effect. The use of \emph{cosine scheduling}, \emph{an initialization after fine-tuning (FT)} and \emph{raw soft pseudo labels} corresponds to the best combination found.}
    \label{tab:abl_studies}
\end{center}
\end{table}


\subsection{Results of FAL on COCO and VOC}

\cref{tab:perf_coco,tab:perf_voc} present the results (mAP in \%) obtained by our method compared to previous methods in the literature on the FAL-COCO and FAL-VOC 07-12 benchmarks. As can be seen in both tables, our approach is the only one to consider a transformer-based OD architecture (\ddetr), as opposed to the commonly used two-stage architecture (FRCNN + FPN). When we implemented \ddetr into Unbiased Teacher~\cite{liu2021unbiased} (UBT), we found that the model cannot converge in FAL settings (\cf{} \cref{fig:fal}). 

First, we can see from both tables that our method always improve performance over the corresponding fully supervised FSL baseline (\cf{} \cref{tab:comp_ddetr_frcnn}). 
With our method, we outperform state-of-the-art results on all labeled fractions of the dataset, and obtain even more strong results specifically when the annotations are scarce: globally about +1 performance point (p.p\onedot) when using 1k or less labeled images, which is even more significant when the overall performance is low. For FAL-COCO with 1\% of labeled images, our method achieves a mean of 22.03 mAP, which is about 1.2 p.p\onedot, or 6\% of improvement over the state-of-the-art, UBT. Notably, on FAL-VOC with 10\% of labeled images, we obtain mean performance of 43.15 mAP, corresponding to 2.81 p.p\onedot or 7\% of improvement over UBT. We note that our method also outperform the state-of-the-art when using more labeled data, such as with the 100\% labeled VOC07 setting, where we improve of about 1.5 p.p\onedot over UBT.

\subsection{Ablation studies}
In \cref{tab:abl_studies}, we present an ablation study on the main parts of our approach. We review each ablation below.

\textbf{EMA scheduling}
The effect of the EMA scheduling is compared between the \emph{Best} and \emph{Abl\onedot Sched\onedot} rows. We can see that using a \emph{cosine scheduling} to gradually reduce the EMA keep rate parameter $\alpha$ leads to an improvement of about 0.7 p.p., as opposed to using a \emph{constant} value for $\alpha$ as done in other SSL approaches~\cite{liu2021unbiased,xu2021end,tang2021humble}.

\textbf{Initialization}
In this ablation, we study the effect of \emph{end-to-end semi-supervised learning}~\cite{xu2021end} in the row \emph{Abl\onedot Init\onedot} which consists in starting the semi-supervised training \emph{from scratch} compared to an initialization \emph{after Fine-Tuning (FT)} in the row \emph{Best}, in which we initialize both student and teacher models from the weights of the fine-tuned model on the few labeled data. 
As can be seen in \cref{tab:abl_studies} and contrary to Soft Teacher~\cite{xu2021end}, starting the semi-supervised training from fine-tuned weights is much more effective (about 5.7 p.p\onedot better) than starting from randomly initialized weights, since the teacher model will give useful pseudo-labels to the student from the start of training.  

\textbf{NMS}
The importance of removing NMS to avoid filtering interesting pseudo-labels and introducing bias is showcased between the rows \emph{Best} and \emph{Abl\onedot NMS}. We can see that, contrary to the common practice when using other detectors~\cite{liu2021unbiased,xu2021end,tang2021humble}, the introduction of NMS leads to a performance drop of about 2.5 p.p. This is why we used \emph{raw pseudo labels}, \ie{} without any post-processing.

\textbf{Confidence Thresholding}
The effect of introducing a threshold to filter out the pseudo-labels given by the teacher with poor confidence is shown in the rows \emph{Best} and \emph{Abl\onedot Thresh\onedot}. We test the results using several common values in the literature (0.5, 0.7 and 0.9)~\cite{sohn2020fixmatch,liu2021unbiased,xu2021end}. A value of 0.7 seems to give the best final results (17.34 mAP) between the thresholding variants, but we can see that choosing the best threshold to apply is extremely sensitive. Similarly to Humble Teacher~\cite{tang2021humble}, we also found that removing the confidence threshold to use all the \emph{soft pseudo-labels}, which corresponds to the column with~\o, leads to stronger results (22.24 mAP), less sensitivity and fewer hyperparameters. 


\textbf{\rev{Data augmentation}} 
\rev{In this ablation study, we are interested in the effect of the different sets of data augmentation to generate the \emph{strong view} in the unsupervised branch. In \cref{tab:group_augm_name}, we group similar augmentations together and then study the effects of each group in \cref{tab:augm_abl}. First, we can see that the model does not converge when using the same setting as Unbiased Teacher \cite{liu2021unbiased}, confirming our observation in \cref{fig:fal}, but achieves strong result when using \emph{raw soft pseudo-labels}. Furthermore, using more augmentations to generate the \emph{strong view} leads to better results.}

\subsection{\rev{Self-supervised Backbone}}

\rev{
In \cref{tab:perf_coco}, we also report results when initializing with a \emph{Self-supervised} pretraining instead of a \emph{supervised} one before the fine-tuning phase. As is common practice in Object Detection, detectors are usually initialized with the weights of the \emph{backbone} obtained after a supervised training phase on ImageNet \cite{ren2015faster,He_2019_ICCV}. In these experiments, we use instead weights from our fully unsupervised \emph{ProSeCo} pretraining on ImageNet, presented in \cref{chap:selfod}. This reduces even more the total amount of labels used in the overall training.
}

\rev{We can see first that initializing with ProSeCo improves the fine-tuning performance for all labeled fractions of COCO considered, confirming the results from \cref{chap:selfod}. Then, with semi-supervision with MT-DETR, we observe that the unsupervised pretraining improves over the supervised one on the 5\% and 10\% fractions, but is less effective on the scarcer 0.5\% and 1\% fractions. This might be due to an overlapping between the classes contained in both datasets. The model benefits from having seen more examples of some class of object during the supervised pretraining for the most label-scarce benchmarks, since they contain very few examples for each class.}

\begin{table}
    \begin{subtable}{0.35\linewidth}
    \centering
    \begin{tabular}{@{}ll@{}}
    \toprule
    Name & Augmentations \\
    \midrule
    \multirow{2}{*}{Basic} & Horizontal Flip \\
    & Resize \\
    \midrule
    \multirow{3}{*}{Photo.} & Color Jitter \\
     & Grayscale \\
     & Gaussian Blur \\
    \midrule
    CutOut & CutOut \\
    \midrule
    \multirow{3}{*}{Geom.} & Rotate \\
    & Shear \\
    & Rescale + Pad \\
    \bottomrule
    \end{tabular}
    \subcaption{ }
    \label{tab:group_augm_name}%
    \end{subtable}
    \begin{subtable}{0.65\linewidth}
    \centering
    \begin{tabular}{@{}ll@{}}
    \toprule
    Augmentations used & mAP (in \%) \\
    \midrule
    Basic + Photo. & 17.8 \\
    Basic + Photo. + CutOut & \\
    \quad \textbar{} w/ NMS + Hard PL \cite{liu2021unbiased} & \textcolor{red}{Div.} \\
    \quad \textbar{} w/o NMS + Soft PL \textit{(Ours)} & \textcolor{ForestGreen}{21.1} \\
    Basic + Photo. + CutOut + Geom. & \textbf{21.6} \\
    \midrule
    Basic + Photo. + CutOut + Geom. & \multirow{2}{*}{\textbf{22.3}} \\
     + Augmentations in Supervised branch & \\
    \bottomrule
    \end{tabular}%
    \subcaption{ }
    \label{tab:augm_abl}
    \end{subtable}
    \begin{center}
        \caption{\rev{(a) The different sets of augmentations considered during SSL. The \emph{Basic} augmentations follow standard supervised training~\cite{carion2020end,zhu2020deformable}. The \emph{Photometric} augmentations are used in Self-supervised learning and in Unbiased Teacher~\cite{liu2021unbiased} along with \emph{CutOut} augmentations. The \emph{Geometric} augmentations follow Soft Teacher~\cite{xu2021end} training.} \\
        \rev{(b) Performance comparison of different combination of data augmentations on FAL-COCO 1\%. Using all data augmentations leads to the best results (\textbf{in bold}). Furthermore, we can see that the model do not converge when using NMS and confidence thresholding along with Basic, Photometric and CutOut augmentations, which corresponds to the same setting as Unbiased Teacher \cite{liu2021unbiased} (\textcolor{red}{in red}). Using \emph{raw soft pseudo-labels (PL)} stabilizes the training and makes it converge to a strong solution (\textcolor{ForestGreen}{in green}).}
        }
    \end{center}
    \end{table}

\section{Conclusion}\label{sec:ccl_semi}

In this work, we experimented in different data scarce settings with the state-of-the-art transformer-based object detector \ddetr~\cite{zhu2020deformable} and showed that it performs much better than the most popular two-stage detector \faster~\cite{ren2015faster} with FPN~\cite{lin2017feature}. Surprisingly, we found that Unbiased Teacher~\cite{liu2021unbiased}, a state-of-the-art SSOD method, did not converge when applied with \ddetr{}. 

To address this issue, we propose Momentum Teaching DETR (MT-DETR), an SSL approach tailored for OD based on transformers, in order to leverage their good results with few labeled data.
Our method is based on a student-teacher architecture and, contrary to common practice, discards all previously used handcrafted heuristics to process pseudo-labels generated by the teacher. 
These processing steps are sensitive to hyperparameters, and introduce biases with the unwanted effect of forcing the models to be overconfident in their predictions.
We show that our proposed MT-DETR outperforms state-of-the-art methods, especially in FAL settings. 
\rev{Then, we combined our unsupervised pretraining with our semi-supervised method to reduce even more the number of labels used.}

\rev{This concludes our contributions in this thesis. In the last chapter we give a summary of the different contributions presented and outline future perspectives for learning with limited labels.}

%% file: chapters/ccl.tex
\chapter{Conclusions and Perspectives}\label{chap:ccl}

\minitoc

\section{Summary of the contributions}

Although Machine Learning (ML) and Deep Learning (DL) systems are becoming increasingly used in research and industry, they require a huge amount of labeled data to be effective. Major breakthroughs have been possible thanks to large-scale labeled datasets such as ImageNet \cite{russakovskyImageNetLargeScale2015}, containing 14M labeled images divided into 21k classes, or MS COCO \cite{lin2014microsoft}, with 2.5M labeled instances of 91 classes in 328k images. While images in these datasets represent common objects in natural scenes and considerable performance improvements have been achieved by iterating on them, they can be really semantically far from more specialized tasks which would require specific annotations. 
Annotating such large-scale datasets are incredibly time-consuming and costly, which represents a high entry cost to apply DL algorithms in real-world settings where there may not be a large amount of labeled data available. This explains why in this thesis, we are working towards \emph{Few-Annotation Learning} (FAL), \ie{} learning efficient models with limited labels while having access to unlabeled data, such that they can still perform well on new data, even if they have only seen a few \emph{labeled} examples of that data during training. 
Our contributions are organized in three parts, each detailed in the previous chapters that we summarize below.

In \cref{chap:contrib_eccv}, we investigate the \emph{Meta-Learning} paradigm that is increasingly popular for \emph{Few-Shot Learning} (FSL). We are specifically interested in theoretical justifications of the good empirical results, observed through the lens of \emph{Multi-task Representation Learning}. We provide proofs and a better understanding of the behavior of meta-learning algorithms as well as practical ways to force them to follow the best learning settings. To do so, we propose spectral-based regularization terms and normalization schemes to enforce important theoretical assumptions. This leads to more efficient meta-learning as supported by better empirical results.

Then, in \cref{chap:selfod}, we develop a novel approach for pretraining Transformer-based Object Detectors, Proposal Selection Contrast (ProSeCo), such that every part of the detector can be properly initialized. The pretraining is based on a Proposal-Contrastive Learning framework. Object proposals given by a teacher model are contrasted with predictions of a student model. To introduce the location information that is important in Object Detection, we compute the \emph{Intersection over Union} between proposals and consider the overlapping ones as positives during contrastive learning. This allows for more sample-efficient learning, as pretrained detectors achieve better performance for fewer annotated samples.

Finally, in \cref{chap:ssod}, we focus on Semi-Supervised Learning for Object Detection to leverage large-scale unlabeled data along with few annotated samples. We propose the first method specifically designed for Transformer-based detectors, since training with previous approaches does not converge when using few annotated data. We observe that previous semi-supervised methods do not converge in FAL when applied with these detectors. To fix this converging issue, we investigate into different parts of the model and improve methodological aspects such as data augmentation, soft pseudo-labels without counterproductive post-processing. When comparing performance on different few annotation learning benchmarks, our proposed method outperforms state of the art with a more significant gap when labels are scarce.


\section{Perspectives and discussions}

As mentioned in \cref{chap:sota}, research in learning with limited labels can be tackled in various ways. There is still a large room for improvements, and open problems for future works. We discuss potential research directions in this section.

\subsection{Meta-Learning}

While the work presented in \cref{chap:contrib_eccv} proposes an initial approach to bridging the gap between theory  and practice for Meta-Learning and Multi-Task Representation Learning, some questions remain open on the inner workings of these algorithms. In particular, being able to take better advantage of the particularities of the training tasks during meta-training could help improve the effectiveness of these approaches. 
Furthermore, the similarity between source and test tasks was not taken into account in this work, which is an additional assumption in the theory developed in \cite{duFewShotLearningLearning2020}. 
Even though we provide a preliminary experimental study using different datasets between meta-training and meta-testing to foster future work on this topic, the theoretical learning bounds have to be updated to take into account the difference in the assumptions.

However, in light of recent works challenging the effectiveness of meta-learning compared to fine-tuning approaches \cite{chenCLOSERLOOKFEWSHOT2019,tian2020rethinking,DBLP:conf/icml/WangH0DY20}, which require less computation, the benefit of meta-learning and episodic training for few-shot learning remains to be proven. While meta-learning can be interesting for specific problems such as Reinforcement Learning \cite{finnModelAgnosticMetaLearningFast2017} or Continual Learning \cite{gupta2020look}, it is still unsure if \rev{it is more advantageous for Few-Shot Learning}.



\subsection{Unsupervised pretraining}

Future work for pretraining transformer-based object detectors, as presented in \cref{chap:selfod}, could update the backbone during pretraining to further improve the consistency between the backbone and the detection heads. Even though it would likely be more costly to do so, it might help to improve performance in the large-scale annotated data settings.
However, the benefits of pretrained backbone for downstream tasks are also contested. While they are common practice in object detection \cite{ren2015faster} and allows achieving good performance with lower supervised training iterations \cite{He_2019_ICCV}, similar performance can still be achieved with random initialization and more supervised training time \cite{He_2019_ICCV,DBLP:conf/cvpr/NewellD20}. This implies that pretrained backbone do not lead to intrinsically better representations. Moreover, the best backbones are currently selected according to ImageNet classification performance, which might not be suited for all downstream problems \cite{DBLP:conf/cvpr/EricssonGH21}. 

Even though unsupervised pretraining does not require labeled data, it still needs large-scale datasets to be effective. Pretraining on ImageNet currently remains better than pretraining on the target dataset, thanks to the huge amount of diverse data available in ImageNet \cite{wei2021aligning}. However, pretraining on different datasets one after the other can help for target tasks with smaller-scale data \cite{reed2022self}.


\subsection{Semi-Supervised Object Detection}

In SSOD, future works could push the data scarcity in OD even further to consider very few labeled examples for each class, and better understand how to match the performance of SSL methods for image classification in this setting~\cite{sohn2020fixmatch}. Better taking into account the uncertainty in the localization part might be an interesting step forward \cite{DBLP:conf/cvpr/LiuMK22}.
Regarding transformer-based detectors, we found that improvements from semi-supervision are less significant than with convolutional methods, which would require more understanding. For instance in \cref{tab:perf_coco}, we achieve an improvement of about 9 p.p\onedot{} with our MT-DETR compared to the supervised baseline in the FAL-COCO 1\% setting, whereas the best convolutional methods achieve 11 p.p\onedot{} of improvements. 
Furthermore, we also found in \cref{tab:perf_coco} that the fewer the annotated data, the fewer the benefits of unsupervised pretraining in semi-supervision with transformer-based detectors. More experiments are needed to fully understand how to leverage unsupervised pretraining for semi-supervised learning in FAL, as this can reduce the total amount of label used. 


\section{Broader Impacts}


\rev{We propose to conclude this thesis by discussing the potential benefits and implications of research towards Few-Annotation Learning. Three themes are thus addressed in this section covering the accessibility of Deep Learning, the impact on computational costs and on environmental footprints which represent major issues nowadays.}

\subsection{Accessibility of Deep Learning}

Improving the ability of DL methods to learn with fewer labeled data would make DL accessible to a wider audience, since it can be currently a technical barrier for applying DL methods to specialized problems such as medical or geological imaging. These tasks might not have enough data to begin with, since the gathering of data itself can be a challenge because of the cost of the procedures. Furthermore, annotating these data requires expertise and thus, qualified workers, which is expensive and time-consuming. These practical problems have motivated research in learning with limited data and labels, but the means found to alleviate the necessity of labels comes with an additional computational cost which can also restrict accessibility of such techniques.

\subsection{Computational costs}

Unfortunately, few annotation learning does not imply few computations, and it's often the opposite. To compensate for the scarcity in labels, one must rely on costly unsupervised trainings, along or before supervised training. Compared to supervised pretraining, unsupervised pretraining require bigger batch size and longer training time \cite{chen2020big}. Whereas supervised training on the fully labeled ImageNet is done with a batch size of 256 and 90 epochs \cite{DBLP:journals/corr/abs-1711-04325}, recent self-supervised methods train for about 800 epochs with a batch size of 4096 \cite{chen2020big}. While self-supervised backbones achieve the best performance faster in terms of training time on the target dataset \cite{He_2019_ICCV,DBLP:conf/cvpr/NewellD20} compared to supervised backbones, the full training might not be globally faster if we also take into account the time of the pretraining phase. Thus, using off-the-shelf weights pretrained on large-scale common datasets can save a lot of training time, but in general, pretraining on novel datasets might not be interesting in practice. The information contained in the labels makes supervised training very efficient.

Even in few-shot learning, \ie{} without access to large-scale unlabeled data, the popular Meta-learning framework, and in particular \emph{episodic training}, requires to split the training dataset into small, independent, but overlapping tasks and to train on a large amount of such tasks \cite{finnModelAgnosticMetaLearningFast2017}. Recent methods \cite{ye2020fewshot,yeHowTrainYour2021a} even initialize the episodic training phase with fully supervised weights, making the full meta-learning training much more computationally expensive than simpler fine-tuning approaches. 

These computational perspectives have to be taken into account when considering FSL and FAL methods. However, with a high computational cost comes an environmental cost, but so does the process of annotating a dataset. 

\subsection{Environmental footprints}

Looking through the lens of CO2 emissions, even though unsupervised trainings are obviously expensive because of hardware resources and computing time, the process of annotating a large-scale dataset is not green either. We generally only measure the cost of the training phase \cite{DBLP:journals/corr/abs-2211-02001}, but rarely look at the annotation part. All numerical details from the following discussion can be found in \cref{ax:comp_co2}.

Let us consider the ILSVRC-2012 \cite{russakovskyImageNetLargeScale2015} dataset from ImageNet, having 1.2M labeled images and widely used for supervised pretraining of backbones. The annotation tasks were given to Amazon Mechanical Turk (AMT) to parallelize the process, using about 5 independent annotators to verify each image \cite{russakovskyImageNetLargeScale2015} which results in about \emph{2000 hours of annotation}. From the carbon intensity of the resulting electrical consumption, detailed in \cref{ax:comp_co2}, we obtain an estimate of the total carbon footprint of this annotation of \emph{286.4 kgCO2eq}. 

To put this footprint in perspective, it is equivalent to about 55 days of compute of a node with 8 GPUs A100 used at 100\% in a cluster based in France, or 10 days of compute if the cluster is based in the USA (\rev{\cf{} \cref{ax:comp_co2} for more details}). Considering that a full self-supervised pretraining of a regular sized CNN (\eg{} a ResNet-50) on the same dataset takes about 2 days with the same computational budget, this might represent a lot of possible unsupervised experiments depending on the country.

While these numbers might seem low, we considered here an image classification dataset. The annotation process is simple and can be done quickly since there is a single label per image. For more complicated annotations like in Object Detection tasks, each image can contain multiple instance of objects, and each one of them have to be segmented to provide precise location information. If we consider the COCO dataset \cite{lin2014microsoft}, the full annotation process took about \emph{85 000 hours}. This represents a total carbon footprint of about \emph{12 tons of CO2eq}, or about \emph{12 transatlantic round trips by plane} (from France to New-York), which is on another scale than ImageNet. This footprint is equivalent to \emph{51 years} of computation of a node with 8 A100 GPUs running at 100\% of their capacity located in France, and about \emph{9 years} if the node is located in the USA. Once again, the numerical details can be found in \cref{ax:comp_co2}.

One might argue that the annotation process of these two popular large-scale datasets has since then been largely profitable both to the community and from an ecological point of view, considering the quantity of experiments using either these labels or pretrained weights resulting from them. Nevertheless, this might not be the case for \emph{private} or single-use datasets, on which a small-scale annotation process along with Few-Annotation Learning methods would be more beneficial, both ecologically and economically. In the end, the amount of data to annotate must take into account the annotation cost and time, the complexity of the target task, the distance of the data to other already available datasets, and the number of experiments planned both in supervised and unsupervised settings.


%% file: chapters/appendix.tex
\chapter{Appendix}

\minitoc

\begin{abstract}
    \rev{The Appendix is organized as follows. \cref{ax:MTR} provides an additional review on Multi-task Representation Learning theory. \cref{ax:intro_imp_mc} details two more advanced meta-learning algorithms used in experiments. \cref{ax:meta_mtr_proofs} provides the full proofs of the theoretical results presented in \cref{chap:contrib_eccv}. \cref{ax:detailed_perf} gives the detailed performance results of experiments from \cref{sec:expe}. \cref{ax:comp_co2} details the computation of carbon footprint resulting from labelization of popular dataset used in the literature.}
\end{abstract}

\section{Review of Multi-task Representation Learning Theory} \label{ax:MTR}
We formulate the main results of the three main theoretical analyses of Multi-task Representation (MTR) Learning Theory provided in \cite{maurerBenefitMultitaskRepresentation2016,duFewShotLearningLearning2020,tripuraneniProvableMetaLearningLinear2020} in \cref{sota:meta_learning_theory} to give additional details for \cref{sec:theoretical_results,sec:ood}.

\renewcommand*{\arraystretch}{1.5}
\begin{table}[tb]
    \centering
    \resizebox{\textwidth}{!}{\begin{tabular}{l|l|c|c}
        \textbf{Paper} & \multicolumn{1}{c}{\textbf{Assumptions}} & $\Phi$ & \textbf{Bound}\\
        \cite{maurerBenefitMultitaskRepresentation2016} & \textbf{A1}. $\forall t \in [[T+1]],\ \mu_t \sim \eta$ & -- & $O\left(\frac{1}{\sqrt{n_1}}+\frac{1}{\sqrt{T}}\right)$\\\hline
        \multirow{6}{*}{\cite{duFewShotLearningLearning2020}} &
        \textbf{A2.0}. $\forall t$, $\|\rvw_t^*\| = \Theta(1)$ &  &  \\
        & \textbf{A2.1}. $\forall t$, $\Bar{\rvx}$ is $\rho^2$-subgaussian &  &  \\
        & \textbf{A2.2}. $\forall t \in [[T]], \exists c>0: \Sigma_t \succeq c\Sigma_{T+1}$ & \textbf{A2.1-2.4}, linear, $k\ll d$ & $O\left(\frac{kd}{cn_1T}+\frac{k}{n_2}\right)$\\
        & \textbf{A2.3}. $\frac{\sigma_1(\rmW^*)}{\sigma_k(\rmW^*)} = O(1)$ & \textbf{A2.3-2.5}, general, $k\ll d$ & $O\left(\frac{\gC(\Phi)}{n_1T}+\frac{k}{n_2}\right)$\\
         & \textbf{A2.4}. $\rvw_{T+1}^* \sim \mu_\rvw: ||\mathop{\E}_{\rvw\sim\mu_\rvw}[\rvw\rvw^T]||\leq O(\frac{1}{k})$ & \textbf{A2.1,2.5,2.6}, linear + $\ell_2$ regul., $k\gg d$ & $\sigma\Bar{R}\tilde{O}\left(\frac{\sqrt{\text{Tr}(\Sigma)}}{\sqrt{n_1T}}+\frac{\sqrt{||\Sigma||_2}}{\sqrt{n_2}}\right)$\\
        & \textbf{A2.5}. $\forall t$, $p_t = p, \Sigma_t = \Sigma$ & \textbf{A2.1,2.5,2.6,2.7}, two-layer NN (ReLUs+ $\ell_2$ regul.) & $\sigma\Bar{R}\tilde{O}\left(\frac{\sqrt{\text{Tr}(\Sigma)}}{\sqrt{n_1T}}+\frac{\sqrt{||\Sigma||_2}}{\sqrt{n_2}}\right)$\\
        & \textbf{A2.6}. Point-wise+unif. cov. convergence &  & \\
        & \textbf{A2.7}. Teacher network &  & \\
        \hline
        \multirow{4}{*}{\cite{tripuraneniProvableMetaLearningLinear2020}} & \textbf{A3.1}. $\forall t$, $\rvx\sim \mu_{\sX_t}$ is $\rho^2$-subgaussian & \multirow{4}{*}{\textbf{A1-4}, linear, $k\ll d$} & \multirow{4}{*}{$\tilde{O}\left(\frac{kd}{n_1T}+\frac{k}{n_2}\right)$} \\
        & \textbf{A3.2}. $\frac{\sigma_1(\rmW^*)}{\sigma_k(\rmW^*)} = O(1)$ and $\forall t, \|\rvw_t\| = \Theta(1)$ &  &  \\
        & \textbf{A3.3}. $\widehat{\rmW}$ learned using the Method of Moments &  &  \\
        & \textbf{A3.4}. $\rvw_{T+1}^*$ is learned using Linear Regression & & \\
        \hline
    \end{tabular}}
    \caption{Overview of main theoretical contributions related to MTR learning with their assumptions, considered classes of representations and the obtained bounds on the excess risk. Here $\tilde{O}(\cdot)$ hides logarithmic factors.}
    \label{sota:meta_learning_theory}
\end{table}

One may note that all the assumptions presented in this table can be roughly categorized into two groups. First one consists of the assumptions related to the data generating process (A1, A2.1, A2.4-7 and A3.1), technical assumptions required for the manipulated empirical quantities to be well-defined (A2.6) and assumptions specifying the learning setting (A3.3-4). We put them together as they are not directly linked to the quantities that we optimize over in order to solve the meta-learning problem. The second group of assumptions includes A2.2 and A3.2: both defined as a measure of diversity between source tasks' predictors that are expected to cover all the directions of $\sR^k$ evenly. These assumptions are of primary interest as it involves the matrix of predictors optimized in \cref{eq:mtr_pb} as thus one can attempt to force it in order for $\widehat{\rmW}$ to have the desired properties. 

Finally, we note that assumption A2.2 related to the covariance dominance can be seen as being at the intersection between the two groups. On the one hand, this assumption is related to the population covariance and thus is related to the data generating process that is supposed to be fixed. On the other hand, we can think about a pre-processing step that precedes the meta-train step of the algorithm and transforms the source and target tasks' data so that their sample covariance matrices satisfy A2.2. While presenting a potentially interesting research direction, it is not clear how this can be done in practice especially under a constraint of the largest value of $c$ required to minimize the bound. \cite{duFewShotLearningLearning2020} circumvent this problem by adding A2.5, stating that the task data marginal distributions are similar. 



\section{\rev{Introduction to IMP and MC algorithms}}\label{ax:intro_imp_mc}

\begin{figure}
    \centering
    \includegraphics[width=\linewidth]{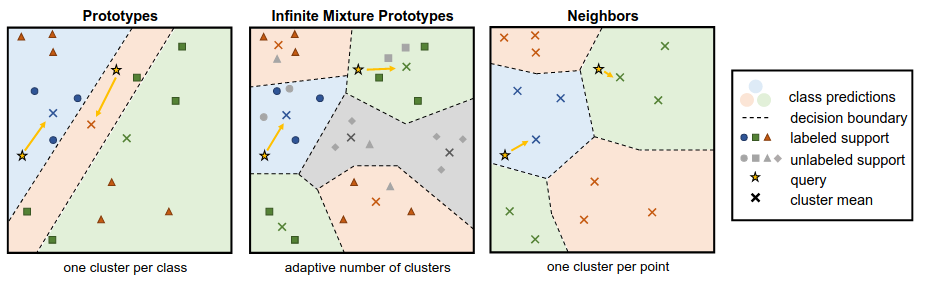}
    \caption{\rev{Illustration of the IMP methods \cite{pmlr-v97-allen19b}. IMP represents each class by a set of clusters, and infers the number of clusters from the data to adjust its modeling capacity.}}
    \label{fig:imp_fig}
\end{figure}

\begin{figure}
    \centering
    \includegraphics[width=0.6\linewidth]{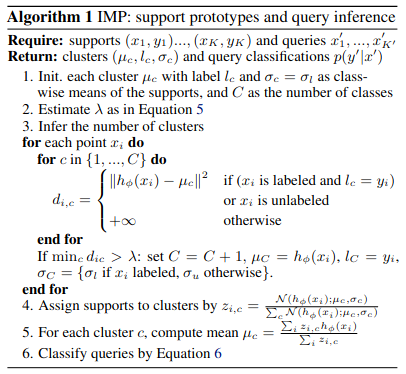}
    \caption{\rev{Pseudocode algorithm of IMP from \cite{pmlr-v97-allen19b}.}}
    \label{fig:imp_algo}
\end{figure}

\subsection{Infinite Mixture Prototypes}
\rev{The \emph{Infinite Mixture Prototypes (IMP)} \cite{pmlr-v97-allen19b} algorithm represents each class by a set of clusters, instead of a single cluster like \Proto{} \cite{snellPrototypicalNetworksFewshot2017}. By inferring the number of clusters, IMP interpolates between nearest neighbor and prototypical representations. IMP can adapt its capacity to avoid underfitting by learning the cluster variance and by multi-modal clustering. \cref{fig:imp_fig} gives a schematic view of the multi-modal representation.}

\rev{\cref{fig:imp_algo} describes the IMP algorithm in pseudocode. Suppose we are given episodes with \emph{support set} $\Scal = \{ (\rvx_1, y_1), \dots, (\rvx_K, y_K)\} \in \sX^K \times \R^K$ of $K$ labeled points and a \emph{query set} $\mathcal{Q} = \{ (\rvx^\prime_1, y^\prime_1), \dots, (\rvx^\prime_{K^\prime}, y^\prime_{K^\prime})\} \in \sX^{K^\prime} \times \R^{K^\prime}$ of ${K^\prime}$ labeled testing points. Within an episode, the cluster are initialized with class-wise means. Then, during inference, the distance to all existing clusters are computed for all support points. If the distance exceeds a threshold $\lambda$, a new cluster is made with mean equal to that point. \IMP{} then updates soft cluster assignments $z_{i,c}$ as the normalized Gaussian density for cluster membership. Finally, cluster means $\mu_c$ are computed by the weighted mean of their members. Since each class can have multiple clusters, query points $\rvx^\prime$ are classified by the softmax over distances to the closes cluster in each class $n$: }

\begin{equation}
    P_\phi (y^\prime = n | \rvx^\prime) = \frac{\exp (-d(h_\phi (\rvx^\prime), \mu_{c_n^*}))}{\sum_{n^\prime} \exp (-d (h_\phi(\rvx^\prime),\mu_{c_{n^\prime}^*}))},
\end{equation}

\noindent \rev{
    with $c^*_n = \argmin_{c:l_c = n} d(h_\phi(\rvx^\prime),\mu_c)$ indexing the clusters, where each cluster $c$ has label $l_c$, and $d(\cdot, \cdot)$ is the distance function of choice. IMP optimizes the embedding function $h_\phi$ with parameters $\phi$ and cluster variances $\sigma$ by stochastic gradient descent across episodes.
}
\subsection{Meta-Curvature} 

\begin{figure}
    \centering
    \includegraphics[width=\linewidth]{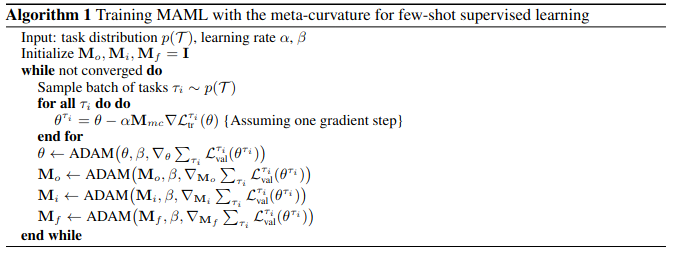}
    \caption{\rev{Pseudocode algorithm of MC from \cite{ParkO19}. The model considered here has only one layer, it is straightforward to extend to multiple layers.}}
    \label{fig:mc_algo}
\end{figure}

\begin{figure}
    \centering
    \includegraphics[width=\linewidth]{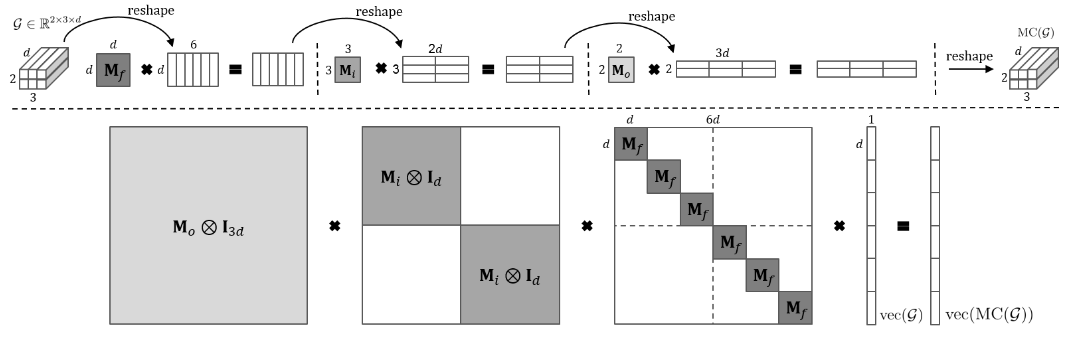}
    \caption{\rev{Illustration from \cite{ParkO19} of meta-curvature computation with $\etG \in \R^{2 \times 3 \times d}$. \emph{(Top)} Tensor algebra view. \emph{(Bottom)} Matrix-vector product view.}}
    \label{fig:mc_fig}
\end{figure}

\rev{
    The \emph{Meta-Curvature (MC)} \cite{ParkO19} method learns a curvature for better generalization and faster model adaptation in the meta-learning framework. The curvature is learned along with the model’s initial parameters simultaneously via the meta-learning process. The goal is that the meta-learned curvature works collaboratively with the meta-learned model’s initial parameters to produce good generalization performance on new tasks with fewer gradient steps. The meta-curvature is computed efficiently through tensor algebra.}

\rev{
    If we consider convolutional neural networks as our model, we can formally define meta-curvature matrices $\rmM_o \in \R^{C_{out} \times C_{out}}$, $\rmM_i \in \R^{C_{in} \times C_{in}}$, and $\rmM_f \in \R^{d \times d}$, where $C_{out}$, $C_{in}$ and $d$ are the number of output channels, the number of input channels and the filter size respectively. Then the meta-curvature function for a multidimensional tensor $\etG$ is computed through $n$-mode products and has all matrices as learnable parameters: }
     
\begin{equation}
    \text{MC}(\etG) = \etG \times_3 \rmM_f \times_2 \rmM_i \times_1 \rmM_o.
\end{equation}

\noindent \rev{
    Then, we can expand the matrices as follows:
}

\begin{equation}
    \widehat{\rmM}_o = \rmM_o \otimes \rmI_{C_{in}} \otimes \rmI_d, \quad \widehat{\rmM}_i = \rmI_{C_{out}} \otimes \rmM_i \otimes \rmI_d, \quad \widehat{\rmM}_f = \rmI_{C_{out}} \otimes \rmI_{C_{in}} \otimes \rmM_f,
\end{equation}

\noindent \rev{
    where $\otimes$ is the Kronecker product, $\rmI_k$ is the $k$ dimensional identity matrix, and the three expanded matrices are all the same size $\widehat{\rmM}_o, \widehat{\rmM}_i, \widehat{\rmM}_f \in \R^{C_{out}C_{in}d \times C_{out}C_{in}d}$. Finally, the gradients can be transformed with the meta-curvature learned as:
}
    
\begin{equation}
    \text{vec}(\text{MC}(\etG)) = \rmM_{mc} \text{vec}(\etG),
\end{equation}

\noindent \rev{
    where $\rmM_{mc} = \widehat{\rmM}_o \widehat{\rmM}_i \widehat{\rmM}_f$. \cref{fig:mc_fig} shows an example of computation with $\etG \in \R^{2 \times 3 \times d}$, and \cref{fig:mc_algo} shows the details of the algorithm to train meta-curvature matrices and the initial model parameters.
}

\section{Full proofs from Chapter 4}
\label{ax:meta_mtr_proofs}

\subsection{Proof of \cref{th:norm_proto}}

\textbf{Prototypical Loss}
We start by recalling the prototypical loss $\Lcal_{proto}$ used during training of Prototypical Networks for a single episode with support set $S$ and query set $Q$:
\begin{align*}
\Lcal_{proto}(S,Q, \phi) &= \E_{(\rvq,i) \sim Q} \left[ - \log \frac{\exp(-d(\phi(\rvq), \rvc_{i}))}{\sum_{j} \exp{(- d(\phi(\rvq), \rvc_j))}} \right] \\
 &=  \underbrace{ \E_{(\rvq, i) \sim Q} \left[ d(\phi(\rvq), \rvc_{i}) \right] }_{(1)} \\
 & \quad + \underbrace{\E_{\rvq \sim Q} \log \sum_{j = 1}^{n} \exp{(-d(\phi(\rvq), \rvc_j))} }_{(2)}
\end{align*}
with $\rvc_i = \frac{1}{k} \sum_{\rvs \in S_i} \phi(\rvs)$ the prototype for class $i$, $S_i\subseteq S$ being the subset containing instances of $S$ labeled with class $i$.

\noindent
\textbf{Distance} 
For \Proto, we consider the Euclidean distance between the representation of a query example $\phi(\rvq)$ and the prototype of a class $i$ $\rvc_i$:
\begin{align}
-d(\phi(\rvq),\rvc_i) &= -\| \phi(\rvq) - \rvc_i \|^2_2 \\
 &= -\phi(\rvq)^\top \phi(\rvq) + 2 \rvc_i^\top \phi(\rvq) - \rvc^\top_i \rvc_i.
\end{align}

Then, with respect to class $i$, the first term is constant and do not affect the softmax probabilities. The remaining terms are:
\begin{align}
-d(\phi(\rvq), \rvc_i) &= 2 \rvc_i^\top \phi(\rvq) - \|\rvc_i\|^2_2 \\
 &= \frac{2}{|S_i|} \sum_{\rvs \in S_i } \phi(\rvs)^\top \phi(\rvq) - \|\rvc_i\|^2_2.
\end{align}





\noindent \rev{Now we can recall and prove \cref{th:norm_proto}:}
\begin{theorem-non}[\ref{th:norm_proto}]\emph{(Normalized \Proto{})} \\
    If $\forall i\ \|\rvc_i\| = 1$, then $\forall \hat{\phi} \in \argmin_{\phi} \Lcal_{proto}(S,Q,\phi)$, the matrix of the optimal prototypes $\rmW^*$ is well-conditioned, \ie{} $\kappa(\rmW^*) = O(1)$.
\end{theorem-non}

\begin{proof}

We can rewrite the first term in $\Lcal_{proto}$ as
\begin{align}
    &\E_{(\rvq,i) \sim Q} \left[ d(\phi(\rvq), \rvc_{i}) \right ] \\
    &= - \E_{(\rvq,i) \sim Q} \left[ \frac{2}{|S_i|} \sum_{\rvs \in S_i} \phi(\rvs)^\top \phi(\rvq) - \|\rvc_{i}\|^2_2 \right ] \\
    &= - \E_{(\rvq,i) \sim Q} \left[ \frac{2}{|S_i|} \sum_{\rvs \in S_i} \phi(\rvs)^\top \phi(\rvq) \right ] \\
    &\quad + \E_{(\rvq,i) \sim Q} \left[ \|\rvc_{i}\|^2_2 \right],
\end{align}

and the second term as
\begin{align}
    & \E_{\rvq \sim Q} \left [ \log \sum_{j = 1}^{n} \exp{(-d(\phi(\rvq), \rvc_j))} \right ] \\
    &= \E_{\rvq \sim Q} \left [ \log \sum_{j = 1}^{n} \exp{(\frac{2}{|S_j|} \sum_{\rvs \in S_j} \phi(\rvs)^\top \phi(\rvq) - \|\rvc_j\|^2_2)} \right ] \\
    &= \E_{\rvq \sim Q} \left [ \log \sum_{j = 1}^{n} \exp{(2 \rvc_j^\top \phi(\rvq)-\|\rvc_j\|^2_2)} \right ] \\
    &= \E_{\rvq \sim Q} \left [ \log \left ( n \sum_{j = 1}^{n} \frac{1}{n} \left [ \exp{(2 \rvc_j^\top \phi(\rvq)-\|\rvc_j\|^2_2)} \right ] \right ) \right ] \\
    &= \E_{\rvq \sim Q} \left [ \log \sum_{j = 1}^{n} \frac{1}{n} \left [ \exp{(2 \rvc_j^\top \phi(\rvq)-\|\rvc_j\|^2_2)} \right ] + \log n \right ]. \\
\end{align}

By dropping the constant part in the loss, we obtain:

\begin{align}
    \Lcal_{proto}(S,Q,\phi) &= - \E_{(\rvq,i) \sim Q} \left[ \frac{2}{|S_i|} \sum_{\rvs \in S_i} \phi(\rvs)^\top \phi(\rvq) \right ] \\
    & + \E_{\rvq \sim Q} \left [ \log \sum_{j = 1}^{n} \frac{1}{n} \left [ \exp{(2 \rvc_j^\top \phi(\rvq))} \right ] \right ].
\end{align}
Let us note $\Scal^d$ the hypersphere of dimension $d$, and $\Mcal(\Scal^d)$ the set of all possible Borel probability measures on $\Scal^d$. $\forall \mu \in \Mcal(\Scal^d), u \in \Scal^d$, we further define the continuous and Borel measurable function:
$$
U_\mu (u) := \int_{\Scal^d} \exp(2u^\top v) d\mu(v).
$$
Then, we can write the second term as 
\begin{align}
    & \E_{\rvq \sim Q} \left [ \log \E_{\rvc \sim C \circ \phi^{-1}} \left [ \exp{(2 \phi(\rvc)^\top \phi(\rvq))} \right ] \right ] \\
    & = \E_{\rvq \sim Q} \left [ \log U_{C \circ \phi^{-1}} (\phi(\rvq)) \right ],
\end{align}

where $C$ is the distribution of prototypes of $S$, \ie each data point in $C$ is the mean of all the points in $S$ that share the same label, and $C \circ \phi^{-1}$ is the probability measure of prototypes, \ie the pushforward measure of $C$ via $\phi$.

We now consider the following problem:
\begin{align}
    \min_{\mu \in \Mcal(\Scal^d)} \int_{\Scal^d} \log U_\mu(u) d\mu(u).
    \label{eq:wang_isola}
\end{align}
The unique minimizer of \cref{eq:wang_isola} is the \emph{uniform distribution on $\Scal^d$}, as shown in~\cite{pmlr-v119-wang20k}. 
This means that learning with $\Lcal_{proto}$ leads to prototypes uniformly distributed in the embedding space. \rev{By considering $\rmW^*$ the matrix of the optimal prototypes for each task then $\rmW^*$ is \emph{well-conditioned}, \ie $\kappa(\rmW^*) = O(1)$.}





\end{proof}

\subsection{Proof of \cref{prop:maml_kappa}}

\rev{Let us first recall the learning model of interest:}
\begin{align}
\hat{y}_t = \langle \rvw_t, \rvx_t\rangle, \quad \ell_t = \mathbb{E}_{p(\rvx_t,y_t|\boldsymbol{\theta}_t)} (y_t-\langle \rvw_t, \rvx_t\rangle)^2.   
\label{eq:setup_maml_ax}
\end{align}
\rev{We can now recall \cref{prop:maml_kappa}:}

\begin{proposition-non}[\ref{prop:maml_kappa}]
    Let $\forall t \in [[T]]$, $\boldsymbol{\theta}_t \sim \mathcal{N}(\bm{0}_d,\bm{I}_d)$, $\rvx_t \sim \mathcal{N}(\bm{0}_d,\bm{I}_d)$ and $y_t \sim \mathcal{N}(\langle \boldsymbol{\theta}_t, \rvx_t\rangle, 1)$. Consider the learning model from \cref{eq:setup_maml_ax}, let $\boldsymbol{\Theta}_i := [\boldsymbol{\theta}_i, \boldsymbol{\theta}_{i+1}]^T$, and denote by $\widehat{\rmW}_2^i$ the matrix of last two predictors learned by \Maml{}\ at iteration $i$ starting from $\widehat{\rvw}_0 = \bm{0}_d$. Then, we have that: \\
    \begin{equation}
        \forall i, \quad \kappa(\widehat{\rmW}_2^{i+1})\geq \kappa(\widehat{\rmW}_2^i),\ \text{ if } \sigma_\text{min}(\boldsymbol{\Theta}_i) = 0.
    \end{equation}
\end{proposition-non}

\begin{proof}
We follow \cite{sha_maml_2021} and note that in the considered setup the gradient of the loss for each task is given by 
$$\frac{\partial \ell_t(\widehat{\rvw} - \alpha \nabla \ell_t(\boldsymbol{\theta}))}{\partial \widehat{\rvw}} \propto (1-\alpha)^2 (\widehat{\rvw}_t - \boldsymbol{\theta}_{t})$$
so that the meta-training update for a single gradient step becomes:
$$\widehat{\rvw}_{t} \leftarrow  \widehat{\rvw}_{t-1} - \beta (1-\alpha)^2 (\widehat{\rvw}_{t-1} - \boldsymbol{\theta}_{t}),$$
where $\beta$ is the meta-training update learning rate. 
Starting at $\widehat{\rvw}_0 = \bm{0}_d$, we have that 
\begin{align*}
    &\widehat{\rvw}_{1} = c\boldsymbol{\theta}_{1},\\
    &\widehat{\rvw}_{2} = c((c-1)\boldsymbol{\theta}_{1} +\boldsymbol{\theta}_{2}),\\
    & \dots\\
    &\widehat{\rvw}_n = c\sum_{i=1}^n \boldsymbol{\theta}_i(c-1)^{n-i},
\end{align*}
where $c :=\beta (1-\alpha)^2$. We can now define matrices $\widehat{\rmW}_2^{i}$ as follows:
\begin{align*}
    &\widehat{\rmW}_2^{1} = \begin{pmatrix}
                            c\boldsymbol{\theta}_{1},\\
                            c((c-1)\boldsymbol{\theta}_{1} +\boldsymbol{\theta}_{2})
                            \end{pmatrix}, \\
    &\widehat{\rmW}_2^{2} = \begin{pmatrix}
                            c((c-1)\boldsymbol{\theta}_{1} +\boldsymbol{\theta}_{2}),\\
                            c((c-1)^2\boldsymbol{\theta}_{1}+(c-1)\boldsymbol{\theta}_{2}+\boldsymbol{\theta}_3)
                            \end{pmatrix},\\
    & \dots \\
    &\widehat{\rmW}_2^{n} = \begin{pmatrix}
                            c\sum_{i=1}^{n} \boldsymbol{\theta}_i(c-1)^{n-i},\\
                            c\sum_{i=1}^{n+1} \boldsymbol{\theta}_i(c-1)^{n-i}
                            \end{pmatrix}.       
\end{align*}
We can note that for all $i>1$:
$$\widehat{\rmW}_2^{i+1} = (c-1)\widehat{\rmW}_2^{i} + c \boldsymbol{\Theta}_i.$$
Now, we can write:
\begin{align*}
    \kappa(\widehat{\rmW}_2^{i+1}) = \frac{\sigma_1(\widehat{\rmW}_2^{i+1})}{\sigma_2(\widehat{\rmW}_2^{i+1})} = \frac{\sigma_1((c-1)\widehat{\rmW}_2^{i} + c\boldsymbol{\Theta}_i)}{\sigma_2((c-1)\widehat{\rmW}_2^{i} + c\boldsymbol{\Theta}_i)}\\
    \geq \frac{\sigma_1((c-1)\widehat{\rmW}_2^{i}) - \sigma_2(c\boldsymbol{\Theta}_i)}{\sigma_2((c-1)\widehat{\rmW}_2^{i} + c\boldsymbol{\Theta}_i)} \\
    \geq \frac{\sigma_1((c-1)\widehat{\rmW}_2^{i}) - \sigma_2(c\boldsymbol{\Theta}_i)}{\sigma_2((c-1)\widehat{\rmW}_2^{i}) + \sigma_2(c\boldsymbol{\Theta}_i)}\\
    \geq \kappa(\widehat{\rmW}_2^{i}).
\end{align*}
where the second and third lines follow from the inequalities for singular values $\sigma_1(A+B) \leq \sigma_1(A) + \sigma_2(B)$ and $\sigma_i(A+B) \geq \sigma_i(A) - \sigma_{\text{min}}(B)$ and the desired result is obtained by setting $\sigma_\text{min}(\boldsymbol{\theta}_i) = 0.$
\end{proof}

\subsection{Proof of \cref{prop:new_bound}}

\rev{Let us first recall \cref{prop:new_bound}:}

\rev{
\begin{proposition-non}[\ref{prop:new_bound}]
    If $\forall t \in [[T]], \|\rvw_t^*\| = O(1)$ and $\kappa(\rmW^*) = O(1)$, and $\rvw_{T+1}$ follows a distribution $\nu$ such that $\| \E_{\rvw \sim \nu}[\rvw\rvw^\top]\| \leq O \left(\frac{1}{k} \right)$, then
    \begin{equation*}
        \text{ER}(\hat{\phi}, \hat{\rvw}_{T+1}) \leq O\left(\frac{C(\Phi)}{n_1T}\cdot \kappa(\rmW^*) +\frac{k}{n_2}\right).
    \end{equation*}
\end{proposition-non}
}

\rev{
\begin{proof}
Du et al. \cite{duFewShotLearningLearning2020} assume that $\sigma_k(\rmW^*) \gtrsim \frac{T}{k}$ (Assumption 4.3 in their work). However, since we also have $\|\rvw_t^*\| = O(1)$, it is equivalent to $\frac{\sigma_1(\rmW^*)}{\sigma_k(\rmW^*)} = O(1)$. \\
We have $\sigma_1(\rmW^*) \gtrsim \sigma_k(\rmW^*) \gtrsim \frac{T}{k}$ and then $\frac{\sigma_1(\rmW^*)}{T \cdot \sigma_k(\rmW^*)} = \frac{1}{T} \cdot \kappa(\rmW^*) \gtrsim \frac{1}{k \cdot \sigma_k(\rmW^*)}$ which we use in their proof of Theorem 5.1 instead of $\frac{1}{T} \gtrsim \frac{1}{k \cdot \sigma_k(\rmW^*)}$ to obtain the desired result.
\end{proof}
}

\subsection{Proof of \cref{prop:example}}

Let us recall the data generating process and \cref{prop:example}:
\begin{equation}
    \forall t \in [[T+1]] \text{  and  } (\rvx,y) \sim \mu_t, \quad y = \langle \rvw_t^*, \phi^*(\rvx)\rangle + \varepsilon, \quad \varepsilon \sim \gN(0,\sigma^2).
\label{eq:data_gen_model_ax}
\end{equation}


\begin{proposition-non}[\ref{prop:example}]
    Let $T=2$, $\sX \subseteq \sR^d$ be the input space and $\sY = \{-1,1\}$ be the output space. Then, there exist distributions $\mu_1$ and $\mu_2$ over $\sX \times \sY$, representations $\widehat{\phi}\neq \phi^*$ and matrices of predictors $\widehat{\rmW} \neq \rmW^*$ that satisfy the data generating model (\cref{eq:data_gen_model_ax}) with $\kappa(\widehat{\rmW})\approx 1$ and $\kappa(\rmW^*) \gg 1$.

\end{proposition-non}
\begin{proof}
Let us define two uniform distributions $\mu_1$ and $\mu_2$ parametrized by a scalar $\varepsilon>0$ satisfying the data generating process from \cref{eq:data_gen_model_ax}:
\begin{enumerate}
    \item $\mu_1$ is uniform over $\{1-k\varepsilon,k,1, \underbrace{\dots}_{d-3}\}\times\{1\} \cup \{1+k\varepsilon,k,-1, \underbrace{\dots}_{d-3}\}\times\{-1\}$;
    \item $\mu_2$ is uniform over $\{1+k\varepsilon,k,\frac{k-1}{\varepsilon},  \underbrace{\dots}_{d-3}\}\times\{1\} \cup \{-1+k\varepsilon,k,\frac{1+k}{\varepsilon}, \underbrace{\dots}_{d-3}\}\times\{-1\}$.
\end{enumerate}
where last $d-3$ coordinates of the generated instances are arbitrary numbers.
We now define the optimal representation and two optimal predictors for each distribution as the solution to the MTR problem over the two data generating distributions and $\Phi = \{\phi |\ \phi(\rvx) = \boldsymbol{\Phi}^T\rvx,\ \boldsymbol{\Phi}\in \sR^{d\times 2}\}$:
\begin{align}
    \phi^*, \rmW^* = \argmin_{\phi \in \Phi, \rmW \in \sR^{2\times 2}}\sum_{i=1}^2\underset{(\rvx,y)\sim\mu_i}{\E} \ell(y,\langle \rvw_i, \phi(\rvx)\rangle),
    \label{eq:meta_train_true}
\end{align}
One solution to this problem can be given as follows:
\[\boldsymbol{\Phi}^* = \begin{pmatrix} 1 & 0 & 0 & \dots & 0\\0 & 1 & 0 & \dots & 0\end{pmatrix}^T, \quad \rmW^* = \begin{pmatrix} 1 & \varepsilon\\1 & -\varepsilon\end{pmatrix},
\]
where $\boldsymbol{\Phi}^*$ projects the data generated by $\mu_i$ to a two-dimensional space by discarding its $d-2$ last dimensions and the linear predictors satisfy the data generating process from \cref{eq:data_gen_model} with $\varepsilon=0$. One can verify that in this case $\rmW^*$ have singular values equal to $\sqrt{2}$ and $\sqrt{2}\varepsilon$, and $\kappa(\rmW^*) = \frac{1}{\varepsilon}$. When $\varepsilon \rightarrow 0$, the optimal predictors make the ratio arbitrary large thus violating Assumption 1.

Let us now consider a different problem where we want to solve \cref{eq:meta_train_true} with constraints that force linear predictors to satisfy both assumptions:
\begin{align}
    \begin{split}
    \widehat{\phi}, \widehat{\rmW} &= \argmin_{\phi \in \Phi, \rmW \in \sR^{2\times 2}}\sum_{i=1}^2\underset{(\rvx,y)\sim\mu_i}{\E} \ell(y,\langle \rvw_i, \phi(\rvx)\rangle), \\
    &\quad \text{s.t.}\ \kappa(\rmW) \approx 1 \quad \text{ and } \quad \forall i, \quad \|\rvw_i\| \approx 1.
    \end{split}
\end{align}
Its solution is different and is given by  
\[\widehat{\boldsymbol{\Phi}} = \begin{pmatrix} 0 & 1 & 0 & \dots & 0\\0 & 0 & 1 & \dots & 0\end{pmatrix}^T, \quad \widehat{\rmW} = \begin{pmatrix} 0 & 1 \\1 & -\varepsilon\end{pmatrix}.
\]
Similarly to $\boldsymbol{\Phi}^*$, $\widehat{\boldsymbol{\Phi}}$ projects to a two-dimensional space by discarding the first and last $d-3$ dimensions of the data generated by $\mu_i$. The learned predictors in this case also satisfy \cref{eq:data_gen_model} with $\varepsilon=0$, but contrary to $\rmW^*$, $\kappa(\widehat{\rmW}) = \sqrt{ \frac{2 + \varepsilon^2 + \varepsilon \sqrt{\varepsilon^2 + 4}}{2 + \varepsilon^2 - \varepsilon \sqrt{ \varepsilon^2 + 4 }} }$ tends to 1 when $\varepsilon \rightarrow 0$.
\end{proof}

\section{Detailed Performance Comparisons from Section 4.4}\label{ax:detailed_perf}

\begin{figure}
    \centering
    \includegraphics[width=0.8\linewidth]{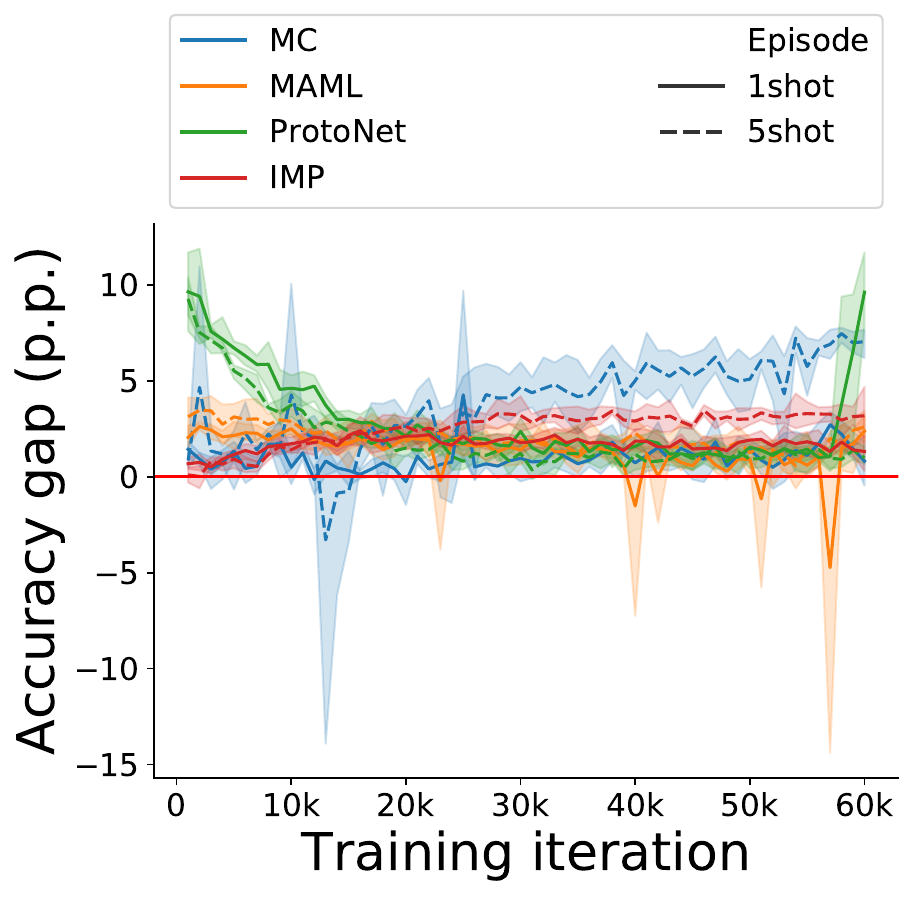}
    \caption{Performance gap (in p.p.) when applying regularization for gradient-based and normalization for metric-based methods throughout the training process on 5-way 1-shot and 5-shot episodes on miniImageNet (\emph{better viewed in color}). Each data point is averaged over 2400 validation episodes and 4 different seeds and shaded areas report $95\%$ confidence interval. We can see that the gap is globally positive throughout training and generally higher at the beginning of training. The increase in the gap at the end of training is linked to a lower overfitting.}
    \label{fig:gap_curves}
\end{figure}

\cref{tab:perf} provides the detailed performance of our reproduced methods with and without our regularization for gradient-based methods in \cref{tab:exact_perf_grad}, or normalization for metric-based methods in \cref{tab:exact_perf_metric}.
\cref{fig:gap_curves} shows the performance gap throughout training for all methods on miniImageNet. 
We can see on both \cref{tab:perf} and \cref{fig:gap_curves} that the gap is globally positive, which shows the increased generalization capabilities of enforcing the assumptions.

\begin{table}[ht]
    \begin{subtable}{\linewidth}
        \centering
        \begin{tabular}{@{}lllll@{}} \toprule
            Method & Dataset & Episodes & without Reg./Norm. & with Reg./Norm. \\
            \midrule
            
            \multirow{7}{*}{\Proto} & \multirow{2}{*}{Omniglot} & 1-shot & $95.56 \pm 0.10\%$ & $\mathbf{95.89 \pm 0.10\%}$ \\
             & & 5-shot & $\mathbf{98.80 \pm 0.04\%}$ & $\mathbf{98.80 \pm 0.04\%}$ \\
            \cmidrule{2-5}
             & \multirow{2}{*}{miniImageNet} & 1-shot & $49.53 \pm 0.41\%$ & $\mathbf{50.29 \pm 0.41\%}$ \\
             & & 5-shot & $65.10 \pm 0.35\%$ & $\mathbf{67.13 \pm 0.34\%}$ \\
            \cmidrule{2-5}
             & \multirow{2}{*}{tieredImageNet} & 1-shot & $51.95 \pm 0.45\%$ & $\mathbf{54.05 \pm 0.45\%}$ \\
             & & 5-shot & $\mathbf{71.61 \pm 0.38\%}$ & $\mathbf{71.84 \pm 0.38\%}$ \\
             
            \midrule
            
            \multirow{7}{*}{\IMP} & \multirow{2}{*}{Omniglot} & 1-shot & $\mathbf{95.77 \pm 0.20\%}$ & $\mathbf{95.85 \pm 0.20\%}$ \\
             & & 5-shot & $\mathbf{98.77 \pm 0.08\%}$ & $\mathbf{98.83 \pm 0.07\%}$ \\
            \cmidrule{2-5}
             & \multirow{2}{*}{miniImageNet} & 1-shot & $48.85 \pm 0.81\%$ & $\mathbf{50.69 \pm 0.80\%}$ \\
             & & 5-shot & $66.43 \pm 0.71\%$ & $\mathbf{67.29 \pm 0.68\%}$ \\
            \cmidrule{2-5}
             & \multirow{2}{*}{tieredImageNet} & 1-shot & $52.16 \pm 0.89\%$ & $\mathbf{53.46 \pm 0.89\%}$ \\
             & & 5-shot & $\mathbf{71.79 \pm 0.75\%}$ & $\mathbf{72.38 \pm 0.75\%}$ \\
             \bottomrule
        \end{tabular}
        \subcaption{Metric-based methods}
        \label{tab:exact_perf_metric}
    \end{subtable}
\end{table}

    \begin{table}\ContinuedFloat
    \begin{subtable}{\linewidth}
        \centering
        \begin{tabular}{@{}lllll@{}} \toprule
            Method & Dataset & Episodes & without Reg./Norm. & with Reg./Norm. \\
        \midrule
        
        \multirow{7}{*}{\Maml} & \multirow{2}{*}{Omniglot} & 1-shot & $91.72 \pm 0.29\%$ & $\mathbf{95.67 \pm 0.20 \%}$ \\
        &  & 5-shot & $97.07 \pm 0.14\%$ & $\mathbf{98.24 \pm 0.10\%}$ \\
        \cmidrule{2-5}
        & \multirow{2}{*}{miniImageNet} & 1-shot & $47.93 \pm 0.83\%$ & $\mathbf{49.16 \pm 0.85\%}$ \\
        & & 5-shot & $64.47 \pm 0.69\%$ & $\mathbf{66.43 \pm 0.69\%}$ \\
        \cmidrule{2-5}
        & \multirow{2}{*}{tieredImageNet} & 1-shot & $50.08 \pm 0.91\%$ & $\mathbf{51.5 \pm 0.90\%}$ \\
        & & 5-shot & $67.5 \pm 0.79\%$ & $\mathbf{70.16 \pm 0.76\%}$ \\
        
        \midrule
        
        \multirow{7}{*}{\MC} & \multirow{2}{*}{Omniglot} & 1-shot & $\mathbf{96.56 \pm 0.18\%}$ & $95.95 \pm  0.20\%$ \\
        &  & 5-shot & $\mathbf{98.88 \pm 0.08\%}$ & $98.78 \pm 0.08\%$ \\
        \cmidrule{2-5}
        & \multirow{2}{*}{miniImageNet} & 1-shot & $\mathbf{49.28 \pm 0.83\%}$ & $\mathbf{49.64 \pm 0.83\%}$ \\
        & & 5-shot & $63.74 \pm 0.69\%$ & $\mathbf{65.67 \pm 0.70\%}$ \\
        \cmidrule{2-5}
        & \multirow{2}{*}{tieredImageNet} & 1-shot & $55.16 \pm 0.94\%$ & $\mathbf{55.85 \pm 0.94\%}$ \\
        & & 5-shot & $71.95 \pm 0.77\%$ & $\mathbf{73.34 \pm 0.76\%}$ \\

        
         
        
        
        \bottomrule
        \end{tabular}
        \subcaption{Gradient-based methods}
        \label{tab:exact_perf_grad}
    \end{subtable}
    \caption{Exact performance of the meta-learning setting considered in \cref{sec:expe}, without and with our regularization (or normalization in the case of \Proto{} and \IMP{}) to enforce the theoretical assumptions. All accuracy results (in \%) are averaged over 2400 test episodes and 4 different seeds and are reported with $95\%$ confidence interval. Episodes are 20-way classification for Omniglot and 5-way classification for miniImageNet and tieredImageNet.}
    \label{tab:perf}
    \end{table}

\section{Detailed computations of CO2 emissions from popular dataset labelization}\label{ax:comp_co2}

\begin{table}
    \centering
    \begin{tabular}{@{}lcc@{}}
        \toprule
        Country & AMT demography & Carbon intensity of electricity \\
        \midrule
            USA & 47\% & 379 \\
            India & 34\% & 633 \\
            Other & 19\% & 442 \\
            France & -- & 68 \\
        \bottomrule
        \end{tabular}
    \caption{General demography of AMT (in \% of total AMT users) with the corresponding carbon intensity of the electricity grid (in gCO2eq/kWh) of each country. The carbon intensity of the \emph{other} category is taken as the world average, and the value for France is given for reference.}
    \label{tab:co2_amt_elec}
\end{table}

\rev{We measure in this section the carbon footprint resulting from the labelization of two popular datasets widely used in the literature, namely ILSVRC-2012 \cite{russakovskyImageNetLargeScale2015} and COCO \cite{lin2014microsoft}.}

The ILSVRC-2012 \cite{russakovskyImageNetLargeScale2015} dataset, is composed of about 1.2M labeled images. The annotation tasks were given to Amazon Mechanical Turk (AMT) to parallelize the process, using about 5 independent annotators (turkers) to verify each image \cite{russakovskyImageNetLargeScale2015}, leading to about 6M total labels given. Since the average user can identify about 250 images in five minutes\footnote{\url{https://www.nytimes.com/2012/11/20/science/for-web-images-creating-new-technology-to-seek-and-find.html}}, this results in about 2000 worker hours.
Turkers are using their own computers to complete the annotation tasks, and with an average computer requiring about 300 W, we obtain a total electrical consumption for the workers of about 600 kWh, without taking into account the Amazon servers running the services behind.
Then, from the average demography of AMT \cite{ipeirotis2010demographics} and the carbon intensity of the electricity grid for each country\footnote{\url{https://ourworldindata.org/grapher/carbon-intensity-electricity}} given in \cref{tab:co2_amt_elec}, we can compute an estimate of the total electric consumption of about 1200 kWh and a resulting carbon footprint of the annotation process of about \emph{286.4 kgCO2eq}. To compare with an actual training experiment, we suppose that we have a node with 8 A100 GPUs, and we estimate that each GPU require 400 W\footnote{\url{https://www.nvidia.com/en-us/data-center/a100/}}. Thus, from \cref{tab:co2_amt_elec}, the annotation process is equivalent to the emissions from \emph{55 days of computation} if the node is located in France, and about \emph{10 days} if it is located in the USA.

For the COCO dataset \cite{lin2014microsoft}, the authors report the time taken for each stage of the annotation process. The \emph{Category Labeling} stage took about 20 000 worker hours, the \emph{Instance Spotting} stage about 10 000 worker hours, and the \emph{Instance Segmentation} stage about 55 000 worker hours, resulting in a total of \emph{85 000 worker hours}. Using the values given in \cref{tab:co2_amt_elec}, we obtain a total electric consumption of about 25 500 kWh and a carbon footprint of about \emph{12 tons of CO2eq}. This footprint is equivalent to \emph{51 years} of computation of a node with 8 A100 GPUs running at 100\% located in France, and about \emph{9 years} if the node is located in the USA.